\documentclass{article}

\PassOptionsToPackage{numbers, compress}{natbib}
\usepackage[preprint]{neurips_2026}

\usepackage[utf8]{inputenc} 
\usepackage[T1]{fontenc}    
\usepackage{hyperref}       
\usepackage{url}            
\usepackage{booktabs}       
\usepackage{amsfonts}       
\usepackage{nicefrac}       
\usepackage{microtype}      
\usepackage{xcolor}         

\usepackage{amsmath}
\usepackage{graphicx}
\usepackage{caption}
\usepackage{wrapfig}
\usepackage[most]{tcolorbox}
\usepackage{enumitem}   
\title{Delta-Adapter: Scalable Exemplar-Based Image Editing with Single-Pair Supervision}

\author{%
Jiacheng Chen$^{1}$ \quad
Songze Li$^{1}$ \quad
Han Fu$^{1}$ \quad
Baoquan Zhao$^{1}$ \quad
Wei Liu$^{2}$ \\
\textbf{Yanyan Liang}$^{3}$  \quad
\textbf{Qing Li}$^4$ \quad
\textbf{Xudong Mao}$^1$\thanks{Corresponding author (xudong.xdmao@gmail.com).} 
\\
$^1$Sun Yat-sen University \quad  $^2$Video Rebirth 
\quad $^3$Macau University of Science and Technology
\\
$^4$The Hong Kong Polytechnic University
\\
}

\begin{document}

\maketitle

\begin{abstract}

Exemplar-based image editing applies a transformation defined by a source-target image pair to a new query image. Existing methods rely on a pair-of-pairs supervision paradigm, requiring two image pairs sharing the same edit semantics to learn the target transformation. This constraint makes training data difficult to curate at scale and limits generalization across diverse edit types. We propose Delta-Adapter, a method that learns transferable editing semantics under single-pair supervision, requiring no textual guidance. Rather than directly exposing the exemplar pair to the model, we leverage a pre-trained vision encoder to extract a semantic delta that encodes the visual transformation between the two images. This semantic delta is injected into a pre-trained image editing model via a Perceiver-based adapter. Since the target image is never directly visible to the model, it can serve as the prediction target, enabling single-pair supervision without requiring additional exemplar pairs. This formulation allows us to leverage existing large-scale editing datasets for training. To further promote faithful transformation transfer, we introduce a semantic delta consistency loss that aligns the semantic change of the generated output with the ground-truth semantic delta extracted from the exemplar pair. Extensive experiments demonstrate that Delta-Adapter consistently improves both editing accuracy and content consistency over four strong baselines on seen editing tasks, while also generalizing more effectively to unseen editing tasks. Code will be available at \url{https://delta-adapter.github.io}.

\end{abstract}
\begin{figure*}[h]
 \centering
\vspace{-3.55mm}
 \includegraphics[width=1\linewidth]{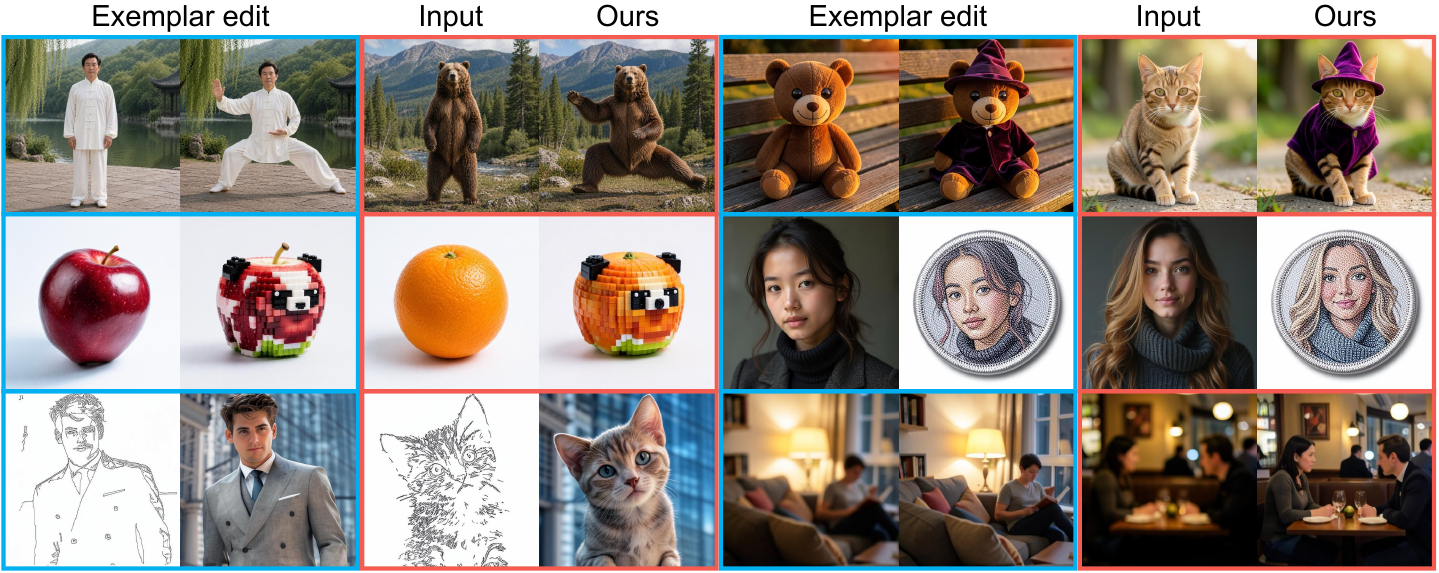}
\caption{Exemplar-based image editing with Delta-Adapter. Our method learns complex transformations from exemplar image pairs and faithfully applies them to new input images.}
\label{fig:teaser}
\end{figure*}

\section{Introduction}
\label{sec:intro}
Instruction-based image editing~\cite{instructpix2pix,Kontext} has demonstrated powerful and flexible image manipulation through natural language. However, certain edits, such as subtle appearance shifts and edit extent, are inherently difficult to articulate precisely in words. This limitation motivates exemplar-based image editing~\cite{visual_prompting,imagebrush}, also known as image analogy~\cite{jacobs2001image}, where a source/target exemplar pair defines the desired transformation, which is then applied analogously to a new query image. Compared to text instructions, exemplar pairs convey editing intent more directly and unambiguously.

Existing exemplar-based editing methods~\cite{imagebrush,gong2025relationadapter,li2025visualcloze} predominantly adopt a \textit{pair-of-pairs supervision} paradigm: given two image pairs $\{a, a'\}$ and $\{b, b'\}$ sharing the same edit semantics, the model learns to predict $b'$ from the tuple $(a, a', b)$ by transferring the transformation observed in $\{a, a'\}$. Despite its effectiveness, this formulation is inherently restrictive. To reliably isolate the intended edit, both pairs must exhibit closely matched transformations, and any uncontrolled discrepancy can introduce ambiguity that undermines learning. This strict alignment requirement makes training data difficult to curate and scale, limiting the diversity of learnable edit types and the model's generalization capacity. Moreover, existing methods often rely on textual guidance at both training and inference time, making performance sensitive to prompt wording and imposing an extra burden on users.

These limitations raise a central question: Can transferable editing semantics be learned under \textit{single-pair supervision}, without textual guidance? The reliance on two pairs in existing methods stems from a specific architectural choice: the model is conditioned on the complete exemplar pair $\{a, a'\}$, directly exposing the edited image $a'$ as input. Because the target appearance is fully observable, a second pair becomes necessary to supervise the prediction of $b'$. Our key insight is to adopt a fundamentally different conditioning strategy. Rather than exposing $a'$ directly, we extract a semantic delta $\Delta_{a \to a'}$ that encodes the visual transformation from $a$ to $a'$, and condition the model solely on the tuple $(a, \Delta_{a \to a'})$. Since $a'$ is never directly visible to the model, it can serve as the prediction target, enabling single-pair supervision without requiring additional exemplar pairs.

We instantiate this idea in Delta-Adapter, a framework for exemplar-based image editing under single-pair supervision that requires no textual guidance. Given a single exemplar pair $\{a, a'\}$, we leverage a pre-trained vision encoder~\cite{siglip} to compute a semantic delta $\Delta_{a \to a'}$ that encodes the visual transformation between the two images. This delta is injected into a pre-trained image editing model via a Perceiver-based adapter~\cite{perceiver}. During training, only the adapter parameters are optimized to reconstruct $a'$ from the tuple $(a, \Delta_{a \to a'})$, while the base editing model remains entirely frozen. To further improve editing fidelity, we introduce a semantic delta consistency loss that encourages the feature-space displacement between the source and generated images to align with the ground-truth semantic delta.

Our proposed single-pair supervision paradigm offers two key practical advantages. First, training requires only individual source/target image pairs, enabling direct use of existing large-scale image editing datasets. This substantially broadens the diversity of edit types seen during training and improves generalization to unseen edits. Second, the single-pair paradigm naturally enables a test-time adaptation strategy: for challenging unseen exemplars, Delta-Adapter can be efficiently fine-tuned on the provided image pair to better capture the intended transformation.

We validate Delta-Adapter through extensive qualitative and quantitative experiments, comparing against four strong baselines across a diverse range of editing tasks. On seen editing tasks, our method achieves superior editing accuracy while better maintaining content consistency. Moreover, Delta-Adapter exhibits better generalization to unseen edits compared to all baselines. When further equipped with the test-time adaptation strategy, performance on unseen tasks improves substantially, reaching levels comparable to those achieved on seen tasks.

\section{Related Work}
\label{sec:related}

\vspace{1pt} \noindent \textbf{Diffusion-based image editing.\ }
Diffusion models have emerged as the dominant paradigm for high-quality image generation and editing~\cite{ddpm,ldm,flux,dit}, and a rich body of work has explored how diverse conditioning signals can guide the editing process. Text-conditioned methods are among the most widely adopted, conveying desired changes through natural language~\cite{sdedit,p2p,diffusionclip,plug_play,zero_shot_i2i,localizing_object,ledits,instructpix2pix,instructdiffusion,emu_edit,hive,guiding_instruction,smartedit,imagic}. While language affords flexible semantic control, it often struggles to precisely capture subtle appearance changes, fine-grained spatial extents, or complex transformations. Mask-conditioned methods address the localization challenge by restricting edits to user-specified regions~\cite{blended,blended_latent,diffedit,inpaint_anything,imagen_editor,task_one_word}, while structure-conditioned methods further enforce spatial faithfulness by incorporating geometric cues such as edges, depth, or pose~\cite{controlnet,T2I_Adapter,Uni_ControlNet}. Reference-guided methods take a complementary approach, transferring appearance, identity, or style from a reference image to the target~\cite{exemplar,anydoor,li2023blipdiffusion,ye2023ip}. Exemplar-based image editing methods~\cite{visual_prompting,visii,imagebrush} condition the model on a before-and-after image pair that jointly defines the desired transformation.

\vspace{1pt} \noindent \textbf{Exemplar-based image editing.\ }
The idea of learning visual transformations from image pairs traces back to the classical image analogy framework~\cite{jacobs2001image}, and has regained significant attention with the rise of large generative models~\cite{visual_prompting,what_makes_good,imagebrush,li2025visualcloze}. Existing diffusion-based approaches can be broadly categorized by how they leverage the exemplar pair at test time. Optimization-based methods adapt learnable parameters to encode the transformation defined by the pair~\cite{visii,pair_customization,lu2025pairedit}. While capable of capturing fine-grained edits, these methods require a costly per-edit optimization process. Training-free methods avoid this overhead by exploiting the in-context reasoning capabilities of pre-trained diffusion models~\cite{Analogist,ReEdit}. Training-based methods instead learn a general editing policy from data, enabling efficient inference without test-time optimization~\cite{imagebrush,image_speak,loc,promptdiffusion,li2025visualcloze,gong2025relationadapter,lorweb}. However, existing training-based approaches rely on pair-of-pairs supervision: two image pairs sharing the same edit semantics are required, where the model observes the transformation in one pair and is trained to predict the target image in the other. This requirement makes training data difficult to curate and scale. Our method addresses this limitation by conditioning on the semantic delta rather than the full exemplar pair, enabling single-pair supervision. Although ReEdit~\cite{ReEdit} also extracts a semantic delta for exemplar-based editing, the two methods differ in fundamental ways. First, ReEdit operates as a training-free method, whereas ours is a trained model. Second, ReEdit conditions on a combination of the semantic delta and the target image representation, while our method conditions solely on the delta, explicitly decoupling the edit operation from image content. Third, ReEdit projects the semantic delta into the textual embedding space and fuses it with a text prompt for conditioning, whereas our model injects the delta directly into the editing backbone via a Perceiver-based adapter.

 \section{Preliminary}
 \label{sec:preliminaries}

\vspace{1pt} \noindent \textbf{Rectified flow.\ }
Our model builds upon FLUX, which formulates image generation as a rectified flow~\cite{flow_matching,rectified_flow} process in the latent space.
Let $z_0$ denote a clean image latent and $z_1 \sim \mathcal{N}(0, I)$ a noise latent.
Rectified flow defines a straight-line interpolation between $z_0$ and $z_1$ as
$z_t = (1-t)\, z_0 + t\, z_1$, where $t \in [0,1]$.
A velocity network $v_\theta$ is trained to predict the constant target velocity $z_1 - z_0$ along this trajectory, conditioned on the noisy latent $z_t$, the timestep $t$, and a text prompt $c$:
\begin{equation}
    \mathcal{L}_{\mathrm{flow}} = \mathbb{E}_{t, z_0, z_1, y, c}
    \left[
        \left\|
            v_\theta(z_t, t, c) - (z_1 - z_0)
        \right\|_2^2
    \right].
\label{eq:flow_matching}
\end{equation}
During training, a coarse estimate of the clean latent can be recovered from the predicted velocity as
\begin{equation}
\hat{z}_0 = z_1 - v_\theta(z_t, t, c).
\label{eq:z0}
\end{equation}
\section{Method}
\label{sec:method}

\subsection{Problem Formulation}
\label{sec:method_formulation}

We address the task of exemplar-based image editing under single-pair supervision. Given a single exemplar pair $\{a, a'\}$, where $a$ is the source image and $a'$ is its edited counterpart, our goal is to learn the visual transformation $a \to a'$ and apply it analogously to an unseen query image $b$, producing an edited output $\hat{b}'$, without any textual guidance at training or inference time.

The key distinction between our formulation and prior work lies in the conditioning input exposed to the model. Existing methods~\cite{gong2025relationadapter,lorweb} condition the editing model on the full exemplar pair $\{a, a'\}$, making the target image $a'$ directly observable. In this setting, supervising the model on the same pair is ill-posed: the desired edited appearance is already present as a conditioning input. Prior methods therefore rely on a second aligned pair $\{b, b'\}$ to supervise whether the edit inferred from $\{a, a'\}$ transfers to $b$. This pair-of-pairs requirement makes training data difficult to curate and fundamentally limits the scalability of model training.

\begin{figure*}[t]
  \centering
  \includegraphics[width=\linewidth]{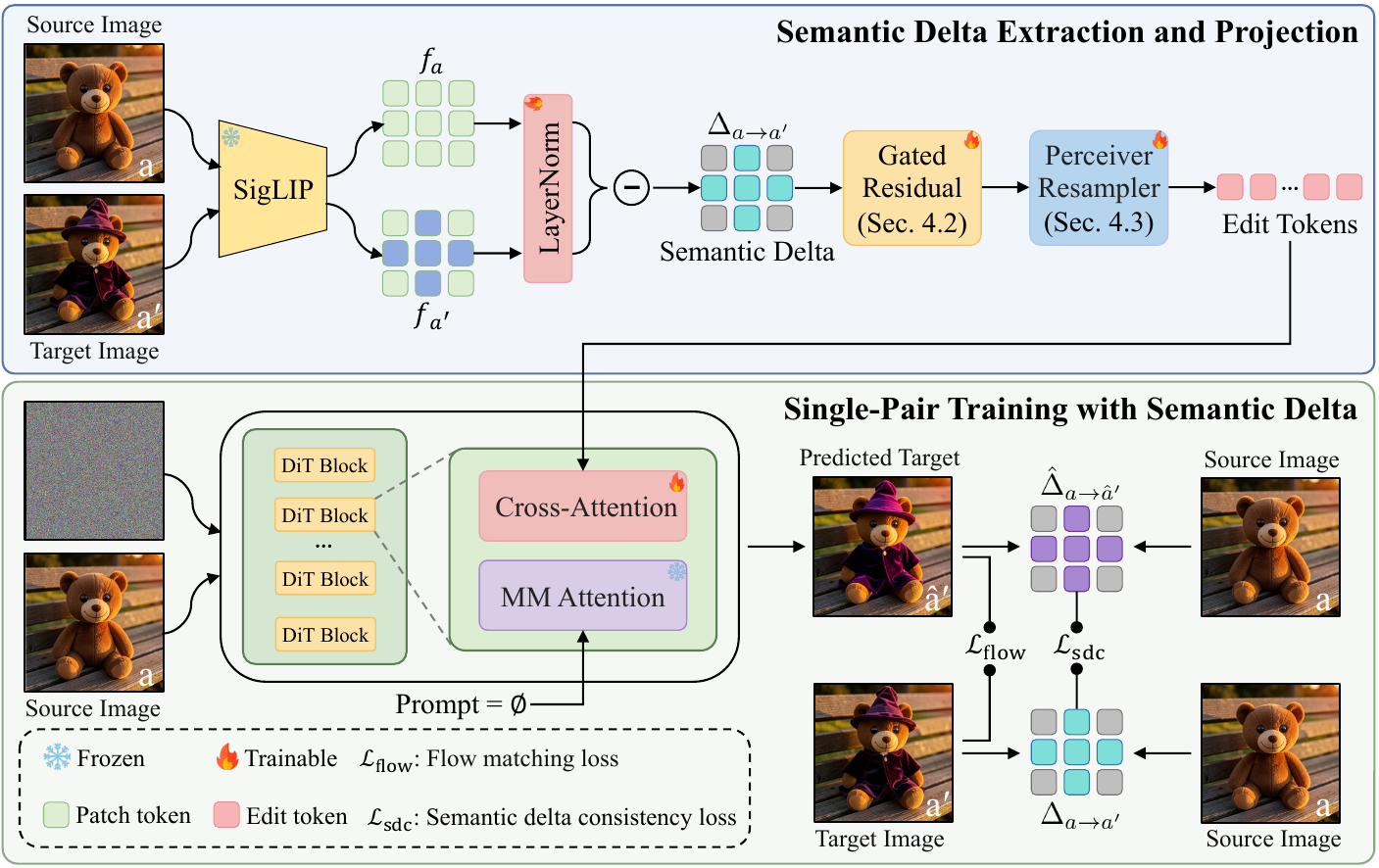}
  \caption{\textbf{Overview of Delta-Adapter.} 
  Given a single exemplar pair $\{a, a'\}$, we first extract patch-level SigLIP features and compute a normalized semantic delta $\Delta_{a\rightarrow a'} = \mathrm{LN}(f_{a'}) - \mathrm{LN}(f_a)$. The delta is refined via a gated residual projection and converted into edit tokens through a Perceiver resampler. These tokens are injected into a frozen DiT-based editing backbone via decoupled cross-attention to reconstruct the target image. Training is supervised by a flow-matching loss combined with a semantic delta consistency loss, which encourages the predicted edit to align with the ground-truth semantic transformation.
  }
  \vspace{-10pt}
  \label{fig:framework}
\end{figure*}

Our key insight is to condition the model on an explicit \emph{semantic delta} $\Delta_{a \to a'}$ that encodes the transformation from $a$ to $a'$, rather than on $a'$ itself. Formally, the model takes the tuple $(a,\, \Delta_{a \to a'})$ as input and is trained to reconstruct $a'$:
\begin{equation}
    \hat{a}' = \mathcal{F}_\theta\!\left(a,\,\Delta_{a \to a'}\right).
\end{equation}
Unlike prior methods that expose the full edited image $a'$ as a condition, our model receives only a semantic displacement $\Delta_{a \to a'}$. This prevents direct copying of the target appearance while retaining the transformation signal necessary for supervision. Consequently, each single image pair can supervise itself, without requiring an additional aligned pair.

As illustrated in Figure~\ref{fig:framework}, our framework operates as follows. Given the exemplar pair $\{a, a'\}$, we first extract a normalized semantic delta $\tilde{\Delta}_{a \to a'}$ (Section~\ref{sec:delta}). This delta is then resampled into a sequence of edit tokens and injected into the pre-trained editing backbone (Section~\ref{sec:injection}). The model is trained to reconstruct $a'$ using a flow matching loss augmented by a semantic delta consistency loss that enforces alignment between the predicted and ground-truth edit directions (Section~\ref{sec:method_loss}).

\subsection{Semantic Delta Extraction}
\label{sec:delta}
The first step is to construct a representation of the visual transformation $a \to a'$. We describe this in two stages: computing a normalized token-level semantic delta, and refining it via a gated residual projection.

\vspace{1pt} \noindent \textbf{Normalized semantic delta.\ }
Given the exemplar pair $\{a, a'\}$, we employ a pre-trained SigLIP~\cite{siglip} encoder to extract dense patch-level features $f_a,\, f_{a'} \in \mathbb{R}^{L \times D_r}$, where $L$ is the number of patch tokens and $D_r$ their dimensionality. Specifically, we extract the last hidden states before the pooling layer, preserving the per-patch spatial structure that is essential for image editing. A natural first attempt is to define the semantic delta as the naive difference $f_{a'} - f_a$. However, this formulation is often dominated by instance-dependent magnitude variations in the raw SigLIP feature space. To address this, we apply token-wise layer normalization~\cite{ba2016layer} before differencing:
\begin{equation}
    \Delta_{a \to a'} = \mathrm{LN}(f_{a'}) - \mathrm{LN}(f_a),
\end{equation}
where $\mathrm{LN}(\cdot)$ normalizes each token independently. This suppresses instance-level magnitude variation while preserving the directional change in feature space.

\vspace{1pt} \noindent \textbf{Gated residual refinement.\ }
Even after normalization, $\Delta_{a \to a'}$ may still contain task-irrelevant variation or imprecisely aligned edit directions. We therefore introduce a gated residual projection to further refine the edit signal:
\begin{equation}
    \tilde{\Delta}_{a \to a'} = \Delta_{a \to a'} + \tanh(g)\,\mathrm{Linear}(\Delta_{a \to a'}),
\end{equation}
where $\mathrm{Linear}(\cdot)$ is a token-wise affine transformation shared across all patch tokens, and $\tanh(g)$ is a bounded learnable scalar gate. The gate is initialized to zero, ensuring the model first learns a stable semantic delta representation before gradually incorporating residual corrections.

\subsection{Semantic Delta Projection and Injection}
\label{sec:injection}
Given the extracted semantic delta $\tilde{\Delta}_{a \to a'}$, we project it into a fixed-length sequence of conditioning tokens and inject them into the DiT-based editing backbone.

\vspace{1pt} \noindent \textbf{Perceiver-based resampling.\ }
Prior IP-Adapter-style methods~\cite{ye2023ip,gong2025relationadapter} map visual encoder features into the generative model via global average pooling followed by an MLP. For exemplar-based editing, however, we find this design generalizes poorly to unseen tasks. We attribute this limitation to the pooling operation: collapsing $\tilde{\Delta}_{a \to a'}$ into a single global vector discards the localized and relational changes that are critical for faithfully representing the intended edit. To address this, we replace the pooling-MLP with a Perceiver resampler~\cite{perceiver}. Specifically, $N$ learnable query tokens cross-attend to the full patch sequence of $\tilde{\Delta}_{a \to a'}$, producing a fixed-length edit representation $R \in \mathbb{R}^{N \times D_r}$. Unlike global average pooling, which treats all patches uniformly, the cross-attention mechanism can exploit the positional information inherent in SigLIP patch tokens when aggregating edit signals.

\vspace{1pt} \noindent \textbf{Per-token projection.\ }
\label{sec:method_token_injection}
A common practice for mapping $R$ into the conditioning space of the DiT blocks~\cite{dit} is to use a shared linear projection for all tokens~\cite{perceiver}. We find this shared mapping overly restrictive for exemplar-based editing: because each token in $R$ is expected to encode a distinct aspect of the edit, a uniform projection suppresses such specialization. We therefore assign each latent token its own affine projection, $e_i = W_i\, r_i + b_i$ for $i = 1, \dots, N$, where $r_i \in \mathbb{R}^{D_r}$ is the $i$-th token of $R$ and $(W_i, b_i)$ are token-specific learnable parameters. The resulting edit tokens $\{e_i\}_{i=1}^{N}$, stacked into $E \in \mathbb{R}^{N \times D_c}$, form the final conditioning representation passed to the editing backbone.

Our Perceiver resampler with per-token projection offers two key advantages. First, as demonstrated in Table~\ref{tab:ablation}, it improves both editing accuracy and content preservation, with particularly pronounced gains on unseen tasks. Second, it is more parameter-efficient, requiring only half the parameters of the pooling-MLP projection employed in~\cite{gong2025relationadapter}.

\vspace{1pt} \noindent \textbf{Decoupled attention injection.\ }
Following~\cite{ye2023ip,gong2025relationadapter}, we inject the edit tokens $E$ into each DiT block via a decoupled cross-attention branch. Specifically, we introduce learnable key and value projections $K_\Delta = E W_k^\Delta$ and $V_\Delta = E W_v^\Delta$, and compute the branch output as $Z_\Delta = \mathrm{Softmax}\!\left({Q K_\Delta^\top}/{\sqrt{d}}\right) V_\Delta$, where $Q$ denotes the query from the original DiT branch. The branch output is then fused with the original attention output via a residual connection: $Z_{\mathrm{new}} = Z + \lambda_{\mathrm{ca}}\, Z_\Delta$, where $\lambda_{\mathrm{ca}}$ is a learnable scalar controlling the injection strength. During training, only $W_k^\Delta$, $W_v^\Delta$, and the preceding projection layers are optimized, while all backbone weights remain frozen.

\subsection{Semantic Delta Consistency Loss}
\label{sec:method_loss}
Our training objective consists of two loss terms. The first applies the flow matching loss (Eq.~\ref{eq:flow_matching}) to reconstruct the target image $a'$. The second is an auxiliary semantic delta consistency loss that provides explicit supervision over the edit semantics.

At each training step, we estimate the denoised latent $\hat{z}_0$ via Eq.~\ref{eq:z0} and decode it through the VAE decoder to obtain the reconstructed image $\hat{a}'$ in pixel space. We then extract patch-level features $f_{\hat{a}'}$ from $\hat{a}'$ using the SigLIP encoder, and compute the predicted semantic delta as $\hat{\Delta}_{a \to \hat{a}'} = \mathrm{LN}(f_{\hat{a}'}) - \mathrm{LN}(f_a)$. Notably, since our backbone model performs denoising in only four steps, the recovered $\hat{z}_0$ is sufficiently sharp to support reliable feature extraction even at the very first denoising step, as illustrated in Figure~\ref{fig:klein_steps}.

Since an edit often affects only a subset of image regions, patches undergoing large semantic shifts should exert a stronger supervisory signal than those that remain nearly unchanged. We therefore assign each patch token $\ell$ a weight $m^{(\ell)} = \|\Delta_{a \to a'}^{(\ell)}\|_2 \,/\, \max_{j}\|\Delta_{a \to a'}^{(j)}\|_2$ proportional to its relative magnitude of change in the ground-truth delta $\Delta_{a \to a'}$. The semantic delta consistency loss then minimizes the patch-weighted cosine distance between the predicted and ground-truth deltas:
\begin{equation}
\mathcal{L}_{\mathrm{sdc}} = 1 -
\frac{1}{L}\sum_{\ell=1}^{L} m^{(\ell)}
\cos\Bigl(
\Delta_{a \to a'}^{(\ell)},\;
\hat{\Delta}_{a \to \hat{a}'}^{(\ell)}
\Bigr),
\label{eq:sdc_loss}
\end{equation}
where $\cos(\cdot,\cdot)$ denotes cosine similarity. This objective encourages the model to produce edits whose semantic deviation from the source image aligns directionally with the intended edit direction.

The full training objective combines both terms:
\begin{equation}
\mathcal{L} = \mathcal{L}_{\mathrm{flow}} +
\lambda_{\mathrm{sdc}}\,\mathcal{L}_{\mathrm{sdc}},
\label{eq:full_objective}
\end{equation}
where $\lambda_{\mathrm{sdc}}$ controls the relative contribution of the semantic consistency term.

\subsection{Test-Time Adaptation}
\label{sec:method_tta}
Despite the generalization benefits of large-scale training under our single-pair supervision paradigm, the model may still struggle to capture fine-grained details on particularly challenging unseen tasks. A key advantage of our paradigm is that it naturally supports test-time adaptation using only the exemplar pair provided at inference. Concretely, we fine-tune Delta-Adapter for a small number of gradient steps (20 in our experiments) using the objective in Eq.~\ref{eq:full_objective}. This stands in contrast to pair-of-pairs methods, which require an additional aligned pair for fine-tuning. As demonstrated in Section~\ref{sec:results}, test-time adaptation yields substantial improvements on challenging unseen tasks. To ensure fair comparison, test-time adaptation is not applied in any comparisons with baselines.

\section{Experiments}
\label{sec:experiments}

\subsection{Implementation and Evaluation Setup}
\label{sec:settings}

\vspace{1pt} \noindent \textbf{Training data.\ }
Since Delta-Adapter requires only single-pair supervision, it can readily leverage existing training datasets designed for instruction-based image editing. Specifically, we train our model on approximately one million image pairs drawn from three sources: Relation~\cite{gong2025relationadapter}, Pico-Banana~\cite{picobanana}, and NHR-Edit~\cite{nhr}. For the evaluation of seen tasks, we train exclusively on 16K image pairs from the Relation dataset to ensure a fair comparison with the baselines.

\vspace{1pt} \noindent \textbf{Implementation details.\ }
Our implementation builds upon the publicly available FLUX.2-klein-4B model, with SigLIP~\cite{siglip} serving as the image encoder.  During training, both $\lambda_{\mathrm{sdc}}$ and $\lambda_{\mathrm{ca}}$ are fixed to 1.0. The model is trained for 100K steps on 4 $\times$ H200 GPUs with a per-GPU batch size of 16, using AdamW with a learning rate of $1\times10^{-4}$ in bfloat16 precision. More implementation details for our method and all baselines are provided in Appendix~\ref{sec:appendix_implementation}.

\begin{figure*}[t]
    \centering
    \renewcommand{\arraystretch}{0.3}
    \setlength{\tabcolsep}{0.6pt}

    {\footnotesize
    \begin{tabular}{c c c @{\hspace{0.05cm}} | @{\hspace{0.05cm}}  c c c c }

        \multicolumn{1}{c}{\normalsize Source ($a$)} &
        \multicolumn{1}{c}{\normalsize Target ($a'$)} &
        \multicolumn{1}{c}{\normalsize Query ($b$)} &
        \multicolumn{1}{c}{\normalsize Edit Transfer} &
        \multicolumn{1}{c}{\normalsize LoRWeB} &
        \multicolumn{1}{c}{\begin{tabular}{c}\normalsize Relation\\\normalsize Adapter\end{tabular}} &
        \multicolumn{1}{c}{\normalsize Ours} \\

        \includegraphics[width=0.14\textwidth]{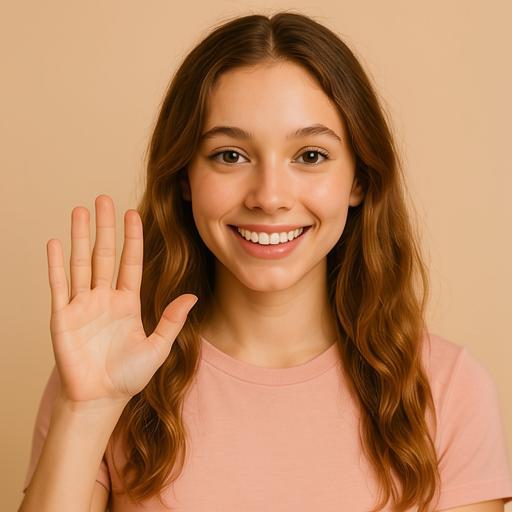} &
        \includegraphics[width=0.14\textwidth]{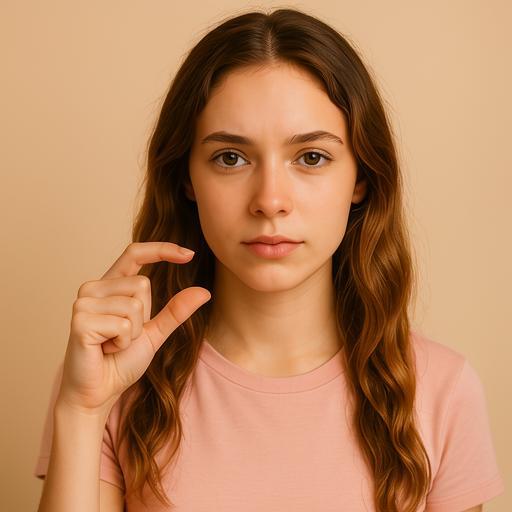} &
        \includegraphics[width=0.14\textwidth]{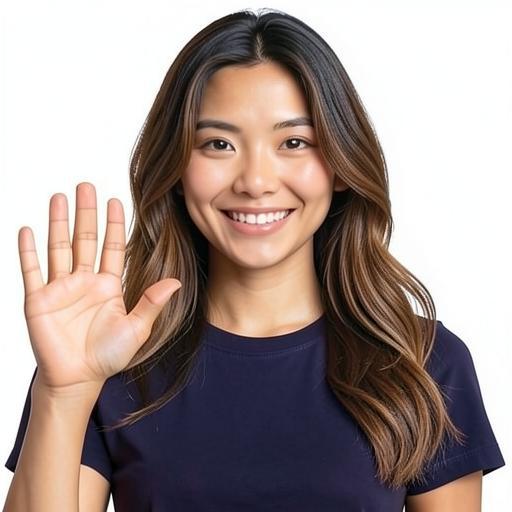} &
        \includegraphics[width=0.14\textwidth]{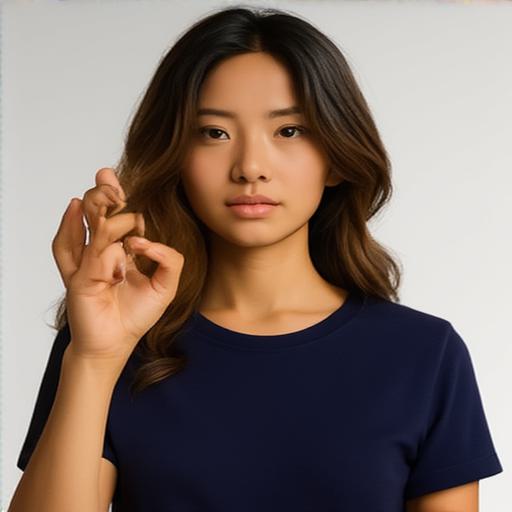} &
        \includegraphics[width=0.14\textwidth]{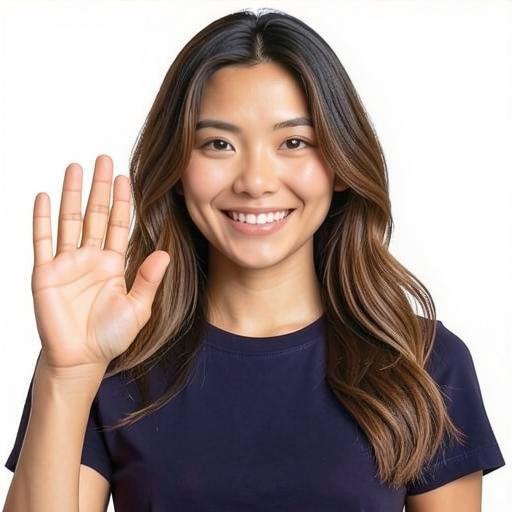} &
        \includegraphics[width=0.14\textwidth]{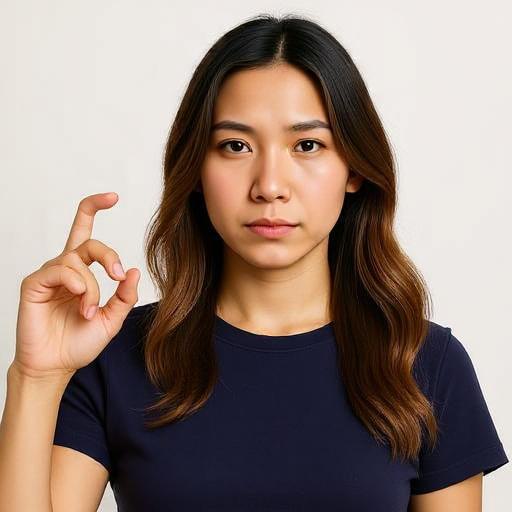} &
        \includegraphics[width=0.14\textwidth]{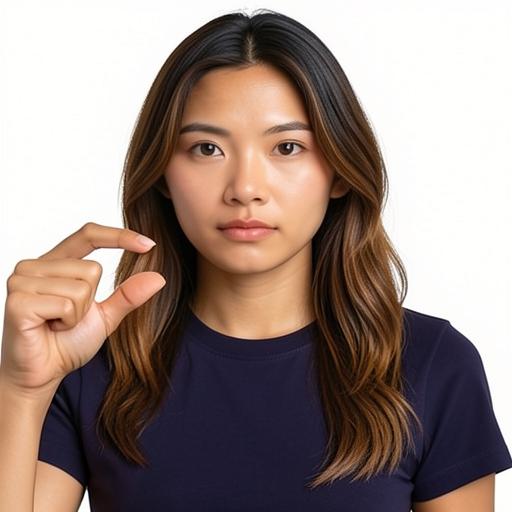} \\

        \includegraphics[width=0.14\textwidth]{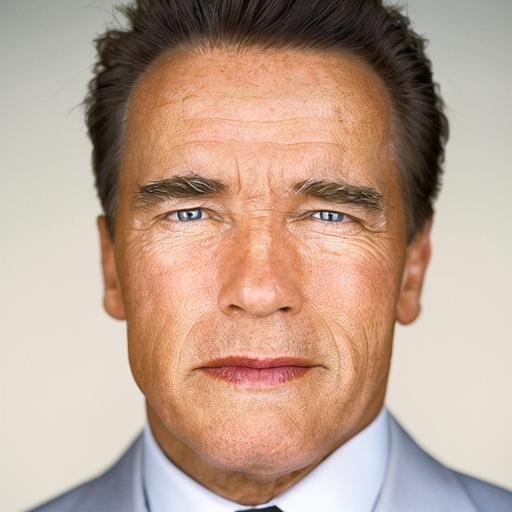} &
        \includegraphics[width=0.14\textwidth]{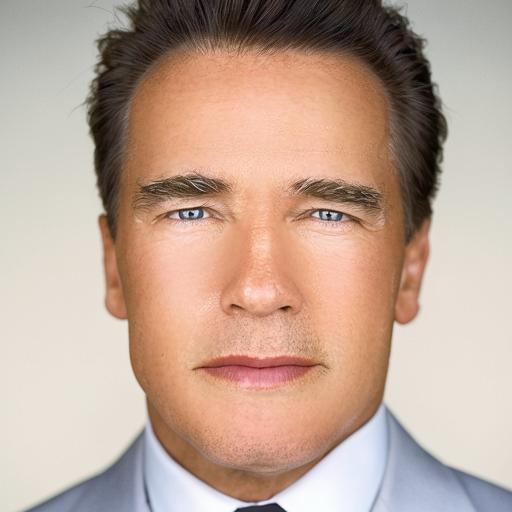} &
        \includegraphics[width=0.14\textwidth]{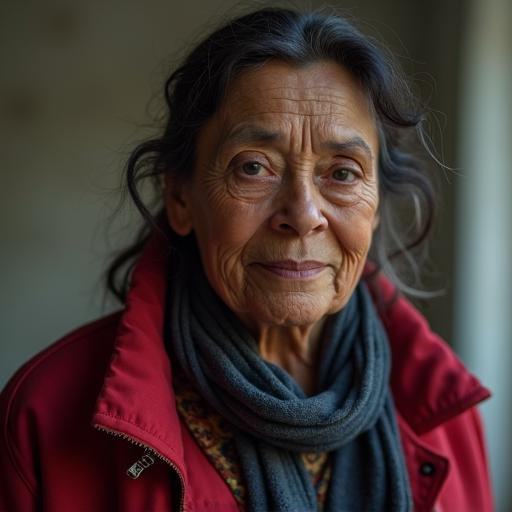} &
        \includegraphics[width=0.14\textwidth]{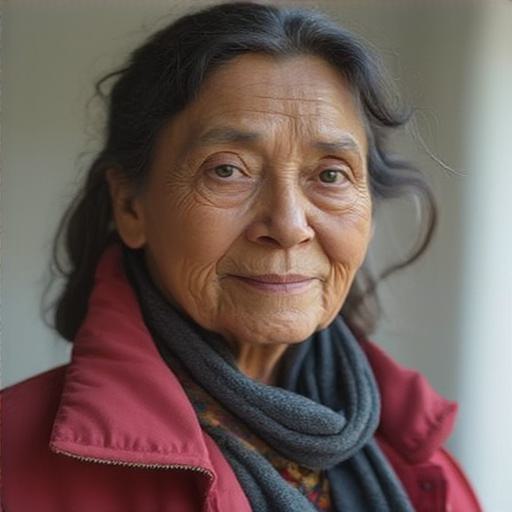} &
        \includegraphics[width=0.14\textwidth]{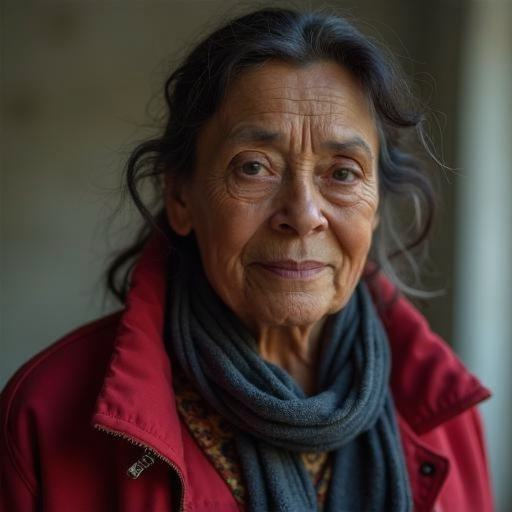} &
        \includegraphics[width=0.14\textwidth]{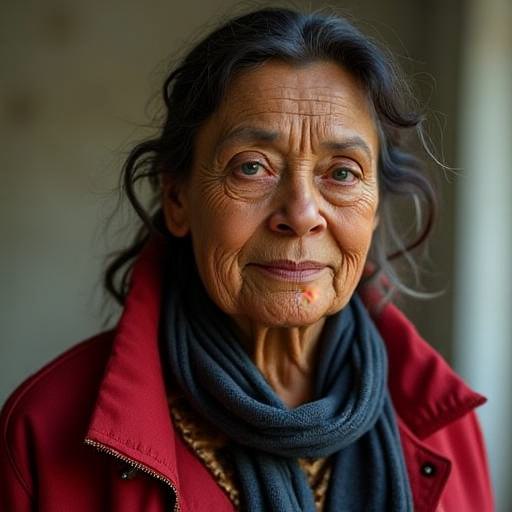} &
        \includegraphics[width=0.14\textwidth]{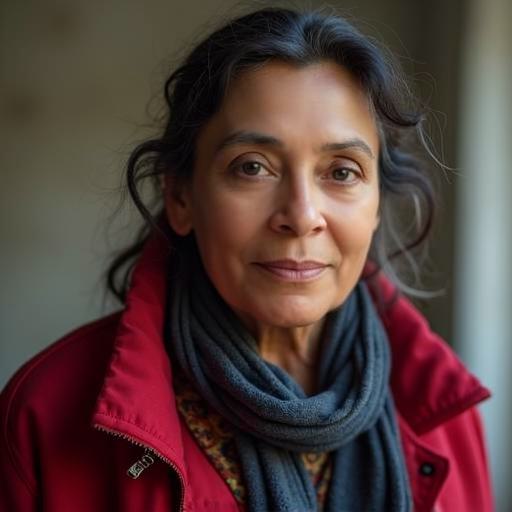} \\

        \includegraphics[width=0.14\textwidth]{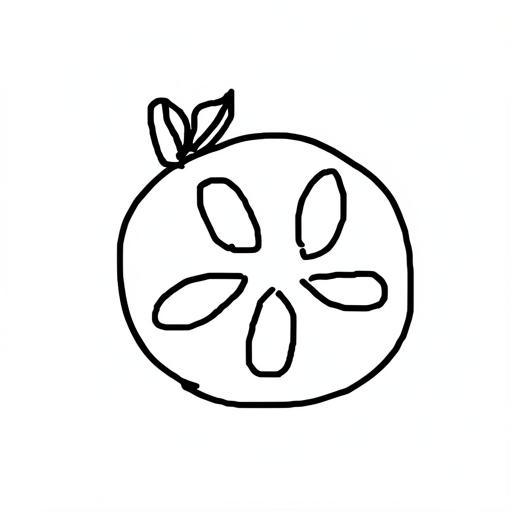} &
        \includegraphics[width=0.14\textwidth]{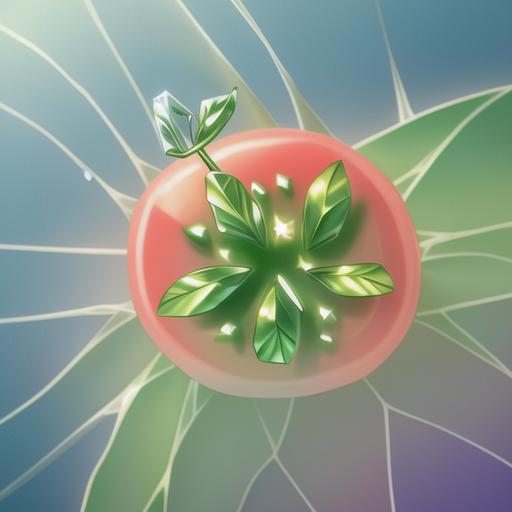} &
        \includegraphics[width=0.14\textwidth]{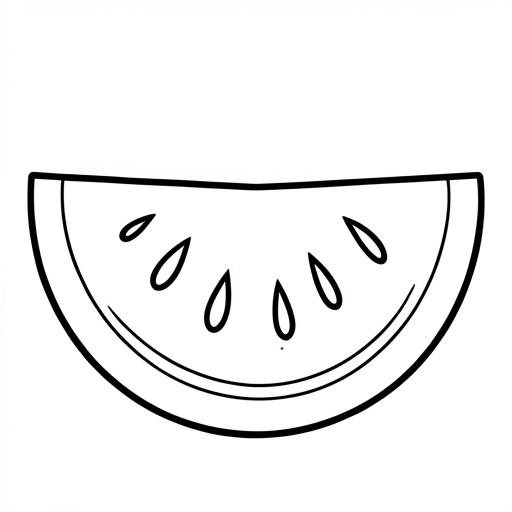} &
        \includegraphics[width=0.14\textwidth]{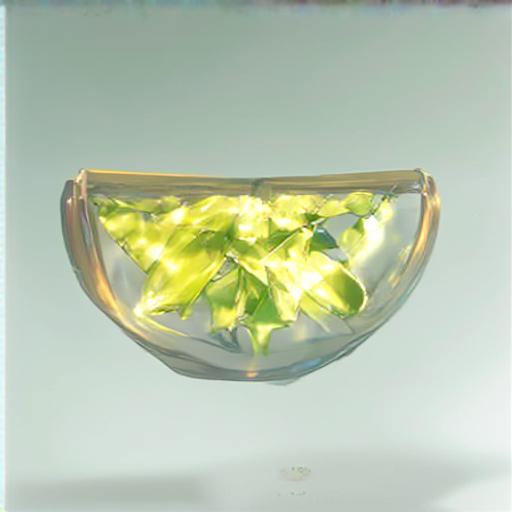} &
        \includegraphics[width=0.14\textwidth]{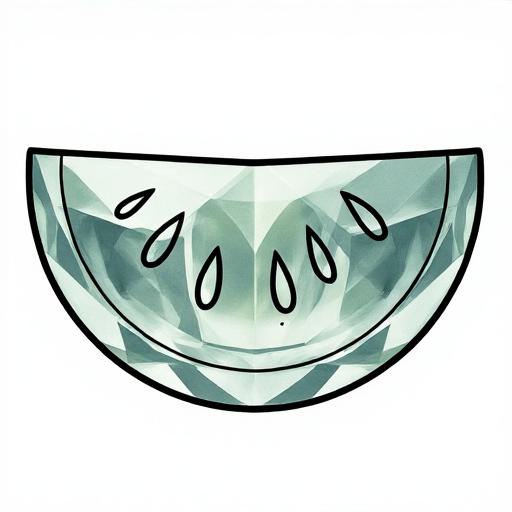} &
        \includegraphics[width=0.14\textwidth]{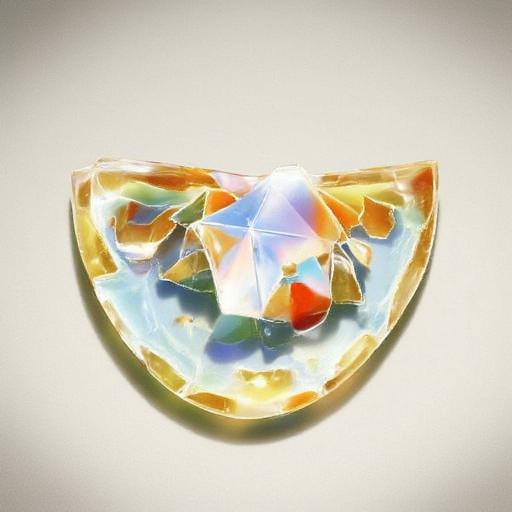} &
        \includegraphics[width=0.14\textwidth]{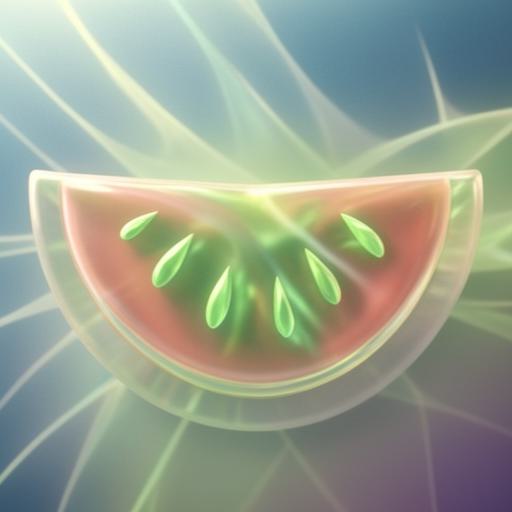} \\

        \includegraphics[width=0.14\textwidth]{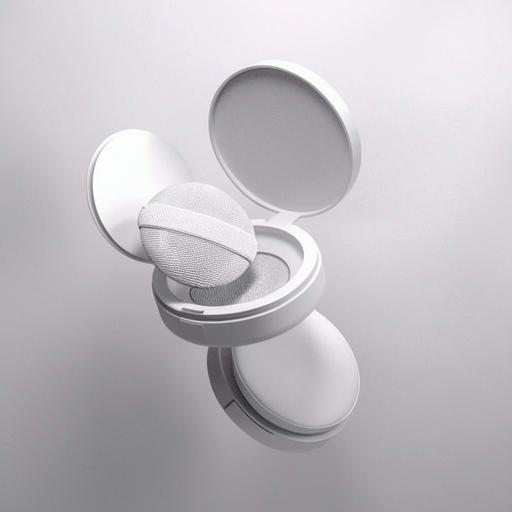} &
        \includegraphics[width=0.14\textwidth]{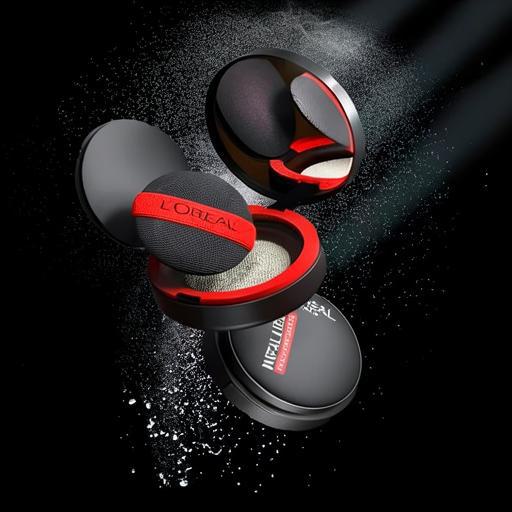} &
        \includegraphics[width=0.14\textwidth]{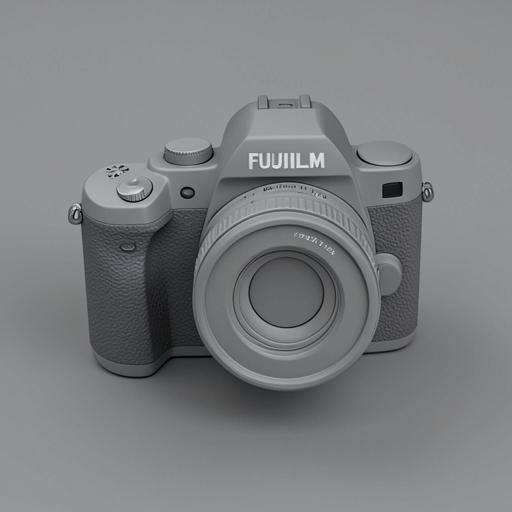} &
        \includegraphics[width=0.14\textwidth]{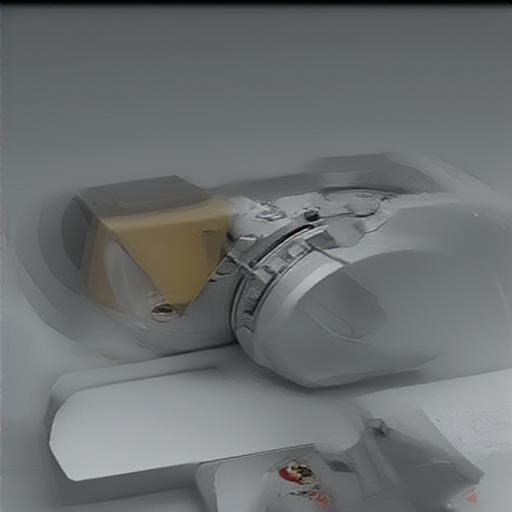} &
        \includegraphics[width=0.14\textwidth]{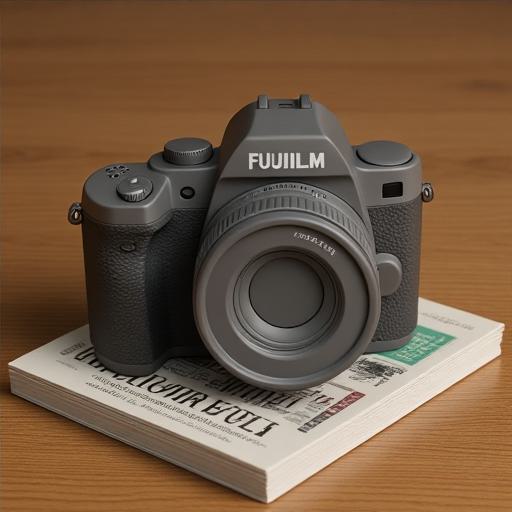} &
        \includegraphics[width=0.14\textwidth]{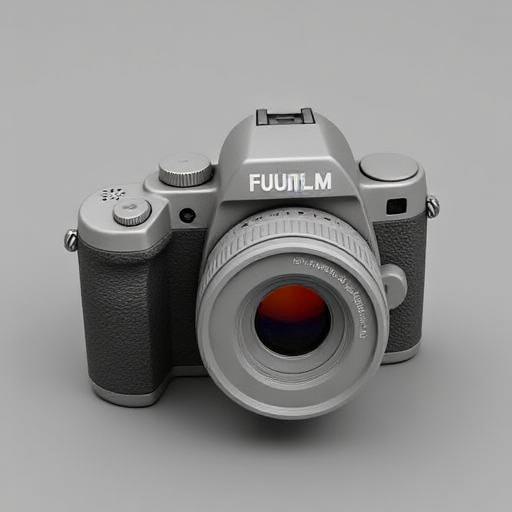} &
        \includegraphics[width=0.14\textwidth]{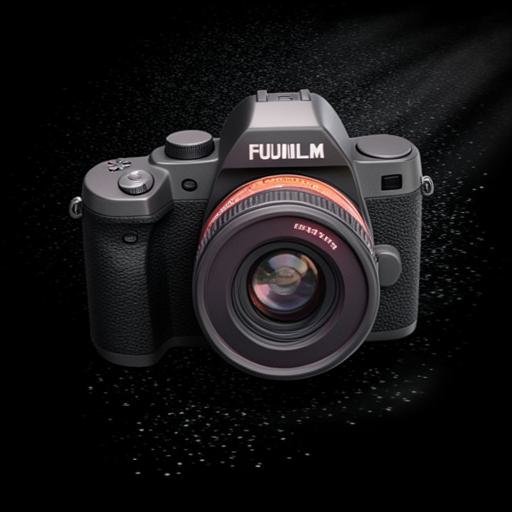} \\

        \includegraphics[width=0.14\textwidth]{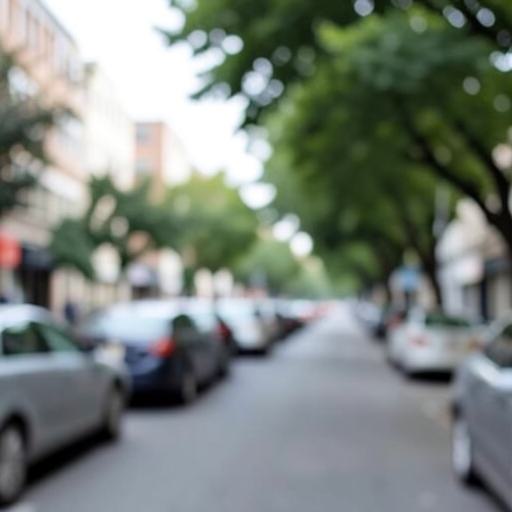} &
        \includegraphics[width=0.14\textwidth]{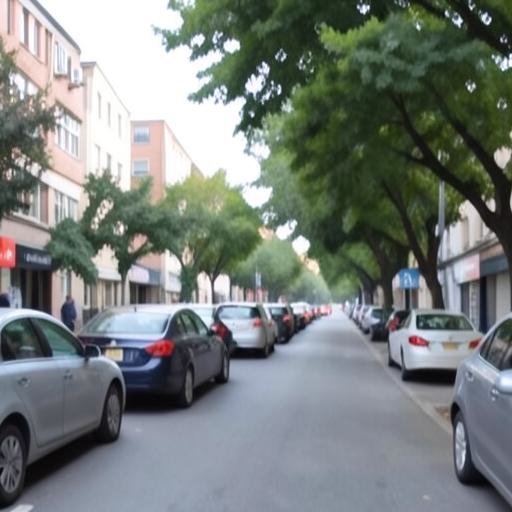} &
        \includegraphics[width=0.14\textwidth]{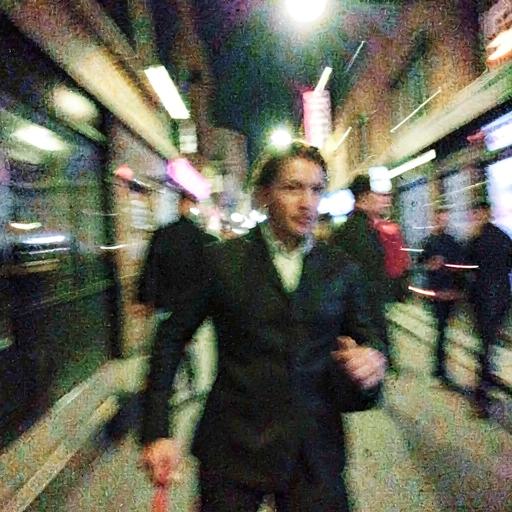} &
        \includegraphics[width=0.14\textwidth]{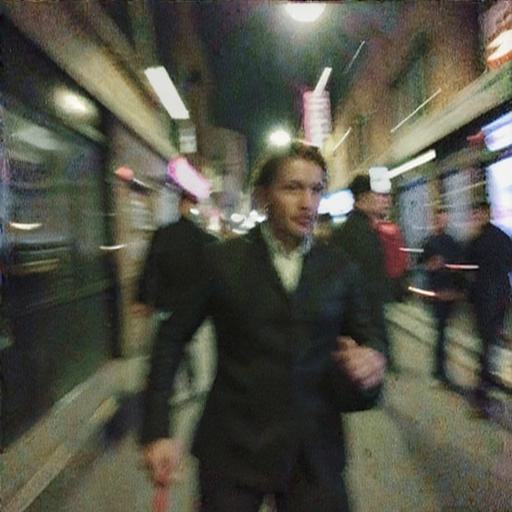} &
        \includegraphics[width=0.14\textwidth]{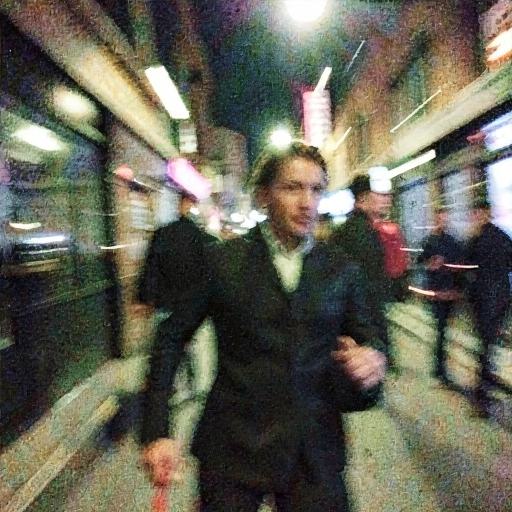} &
        \includegraphics[width=0.14\textwidth]{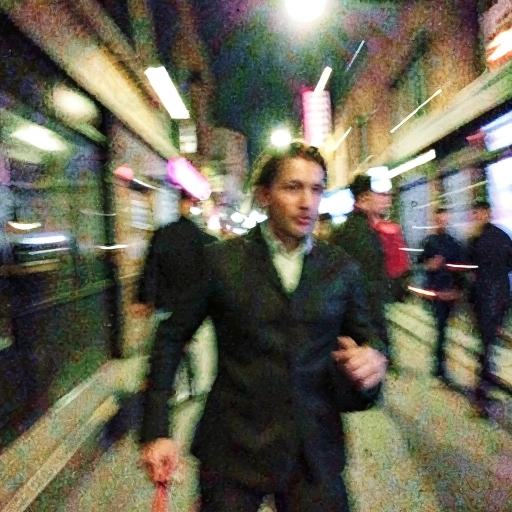} &
        \includegraphics[width=0.14\textwidth]{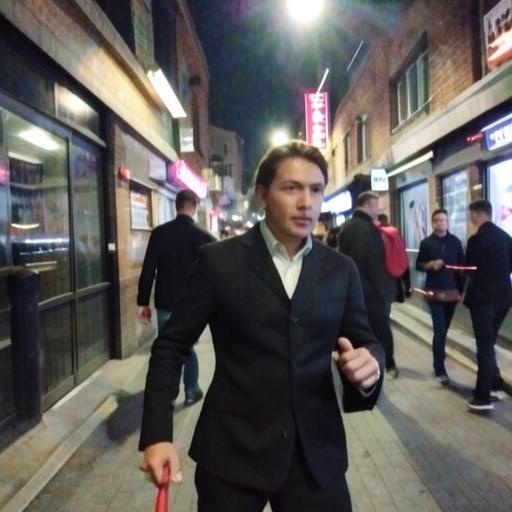} \\

        \includegraphics[width=0.14\textwidth]{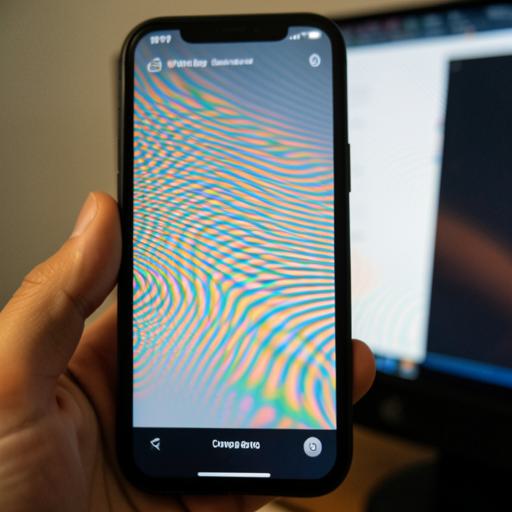} &
        \includegraphics[width=0.14\textwidth]{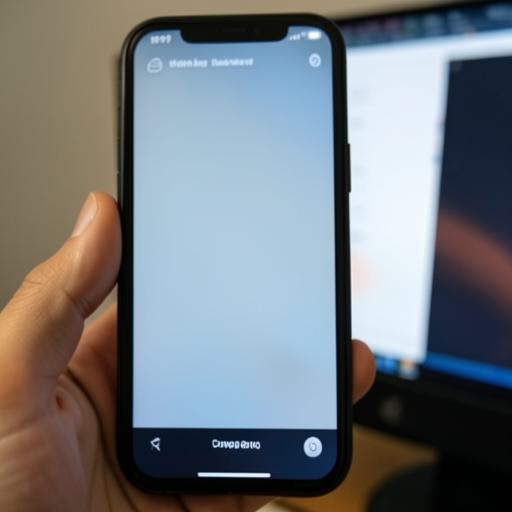} &
        \includegraphics[width=0.14\textwidth]{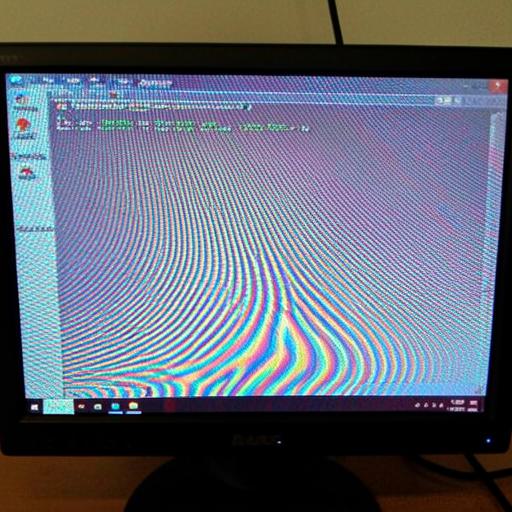} &
        \includegraphics[width=0.14\textwidth]{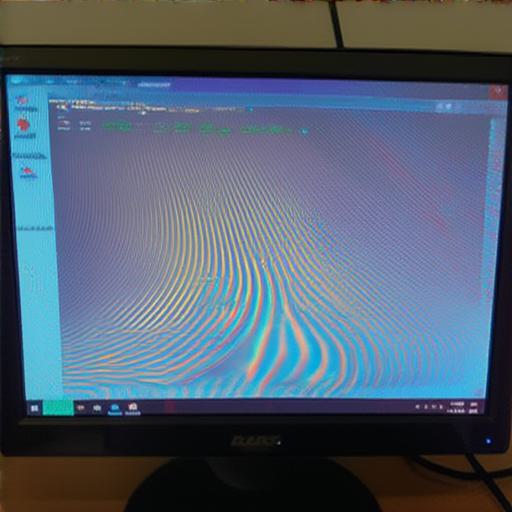} &
        \includegraphics[width=0.14\textwidth]{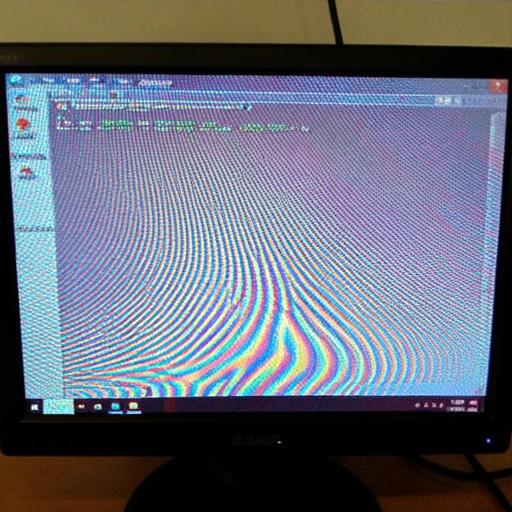} &
        \includegraphics[width=0.14\textwidth]{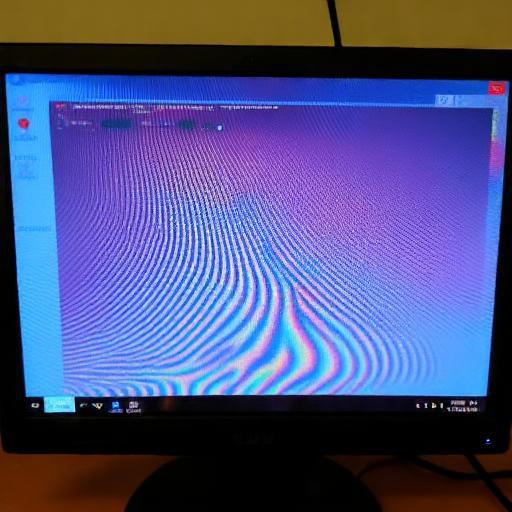} &
        \includegraphics[width=0.14\textwidth]{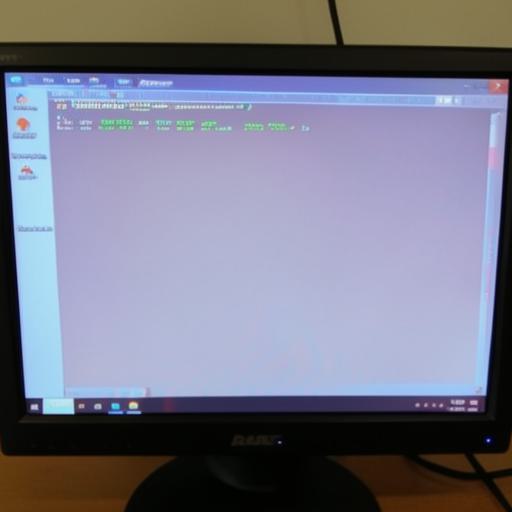} \\

    \end{tabular}
    }
    \caption{\textbf{Qualitative comparison on seen editing tasks.} We compare Delta-Adapter with RelationAdapter~\cite{gong2025relationadapter}, LoRWeB~\cite{lorweb}, and Edit Transfer~\cite{edit_transfer}. Delta-Adapter more faithfully captures the edit semantics implied by the exemplar pair and applies them to the query image, while better preserving its underlying structure and identity.}
  \vspace{-15pt}

\label{fig:qualitative_seen}
\end{figure*}

\vspace{1pt} \noindent \textbf{Baselines.\ }
We compare our method against four representative baselines: RelationAdapter~\cite{gong2025relationadapter}, LoRWeB~\cite{lorweb}, VisualCloze~\cite{li2025visualcloze}, and Edit Transfer~\cite{edit_transfer}. In Appendix~\ref{sec:appendix_more_baselines}, we further include comparisons with two multimodal image generation models, Nano Banana 2~\cite{banana} and GPT-Image-2~\cite{gpt_image}, as well as an optimization-based method, PairEdit~\cite{lu2025pairedit}. 

\vspace{1pt} \noindent \textbf{Evaluation protocol.\ }
We adopt LPIPS~\cite{lpips} and CLIP-I~\cite{clip} to measure perceptual similarity and semantic alignment between the edited and source images, respectively. We further leverage GPT-5.4 to evaluate two aspects of each edited result on a 5-point scale: content consistency of unedited regions (GPT-C) and editing accuracy with respect to the exemplar pair (GPT-A). More details of GPT-based metrics are provided in Appendix~\ref{sec:appendix_gpt_eval}. For seen tasks, we evaluate across all 218 tasks in the Relation dataset with 5 query images each, yielding 1{,}090 generations per method. For unseen tasks, in addition to the unseen validation set from RelationAdapter, which consists of relatively simple tasks, we further construct 50 novel tasks spanning style transfer, attribute change, and object transformation, with 5 query images per task, yielding 250 generations per method.

\subsection{Results}
\label{sec:results}

\begin{figure*}[t]
    \centering
    \renewcommand{\arraystretch}{0.3}
    \setlength{\tabcolsep}{0.6pt}

    {\footnotesize
    \begin{tabular}{c c  c @{\hspace{0.05cm}} | @{\hspace{0.05cm}} c c c c }

        \multicolumn{1}{c}{\normalsize Source ($a$)} &
        \multicolumn{1}{c}{\normalsize Target ($a'$)} &
        \multicolumn{1}{c}{\normalsize Query ($b$)} &
        \multicolumn{1}{c}{\normalsize VisualCloze} &
        \multicolumn{1}{c}{\normalsize LoRWeB} &
        \multicolumn{1}{c}{\begin{tabular}{c}\normalsize Relation\\\normalsize Adapter\end{tabular}} &
        \multicolumn{1}{c}{\normalsize Ours} \\

        \includegraphics[width=0.14\textwidth]{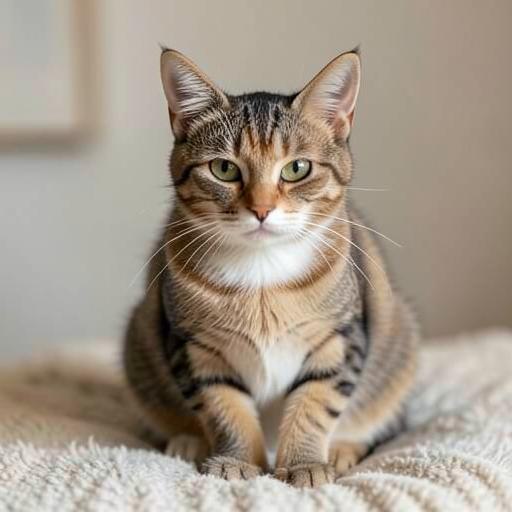} &
        \includegraphics[width=0.14\textwidth]{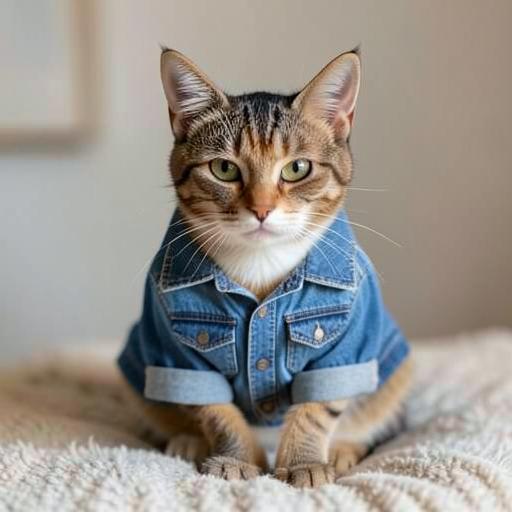} &
        \includegraphics[width=0.14\textwidth]{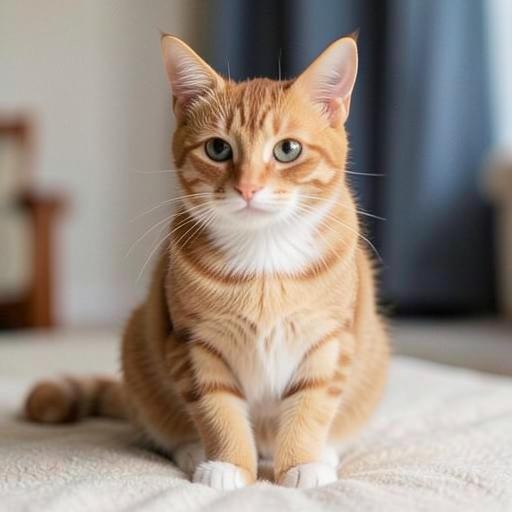} &
        \includegraphics[width=0.14\textwidth]{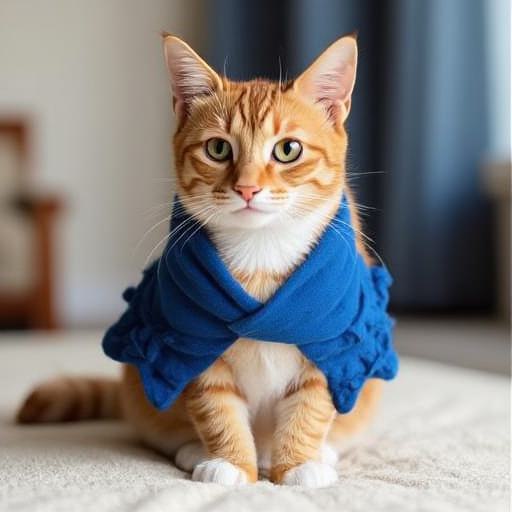} &
        \includegraphics[width=0.14\textwidth]{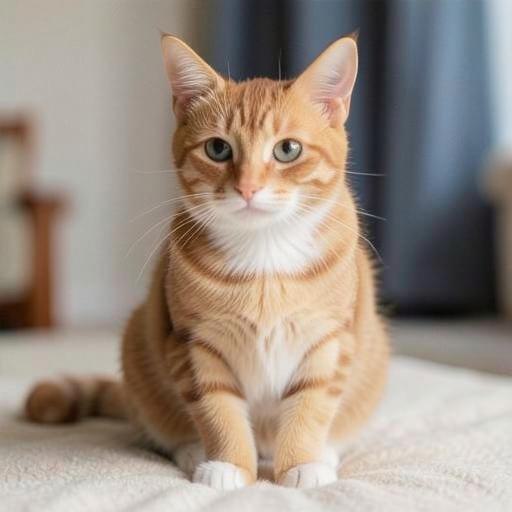} &
        \includegraphics[width=0.14\textwidth]{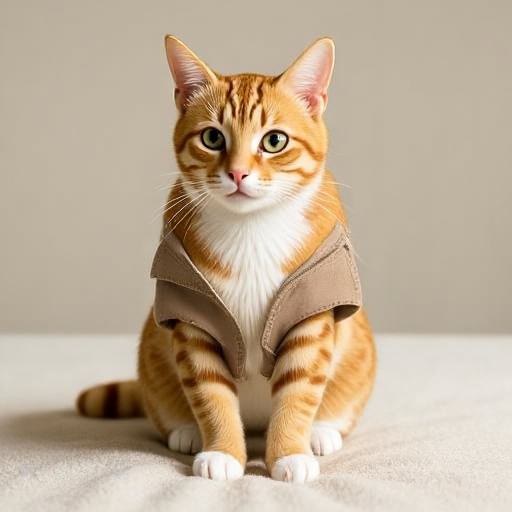} &
        \includegraphics[width=0.14\textwidth]{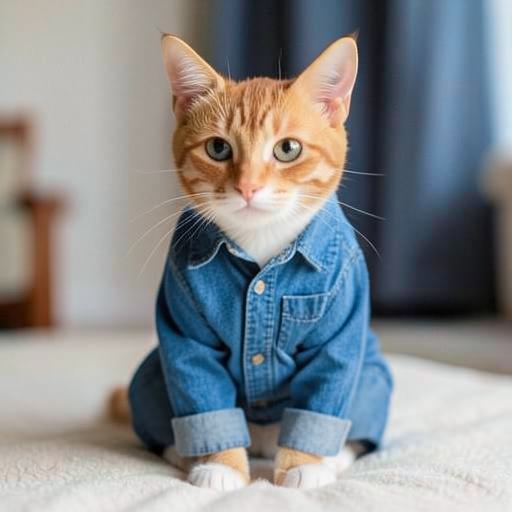} \\

        \includegraphics[width=0.14\textwidth]{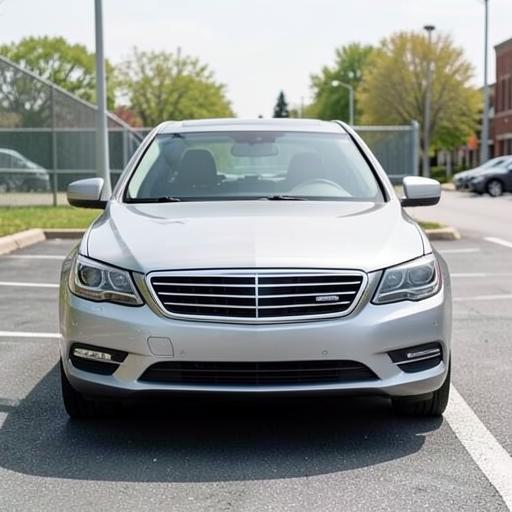} &
        \includegraphics[width=0.14\textwidth]{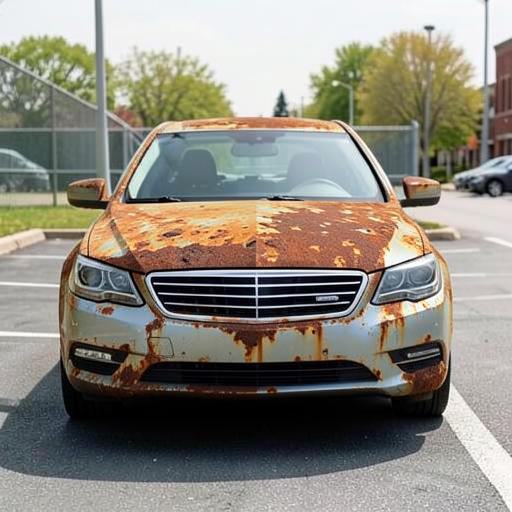} &
        \includegraphics[width=0.14\textwidth]{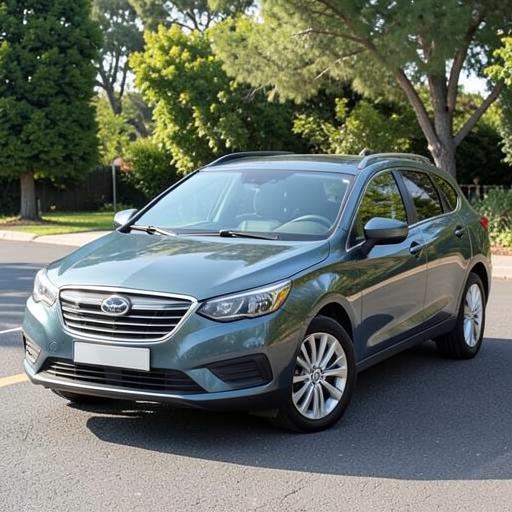} &
        \includegraphics[width=0.14\textwidth]{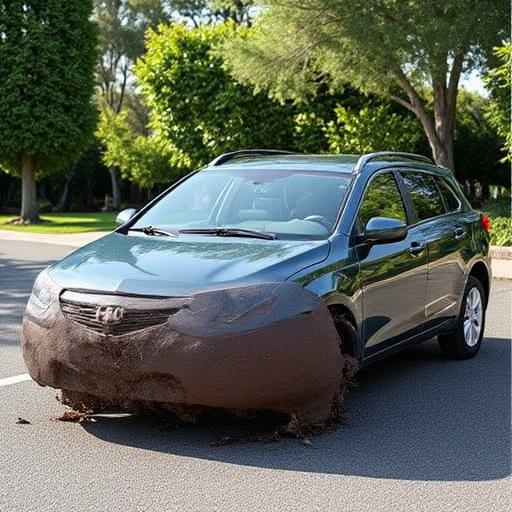} &
        \includegraphics[width=0.14\textwidth]{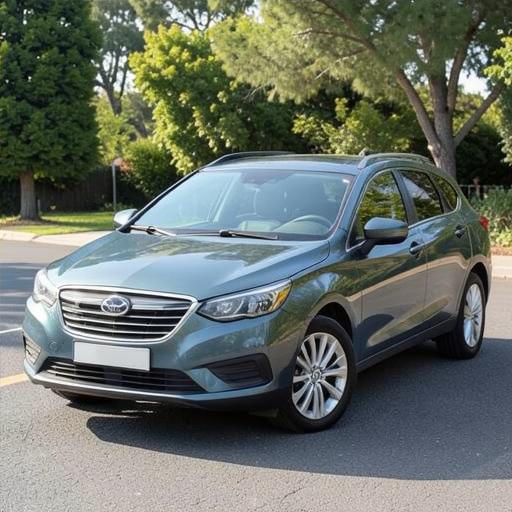} &
        \includegraphics[width=0.14\textwidth]{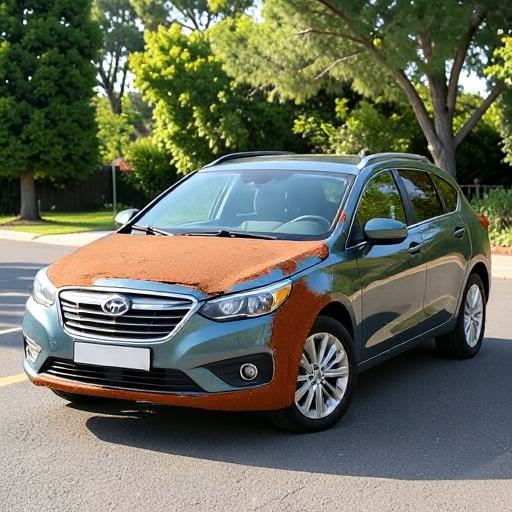} &
        \includegraphics[width=0.14\textwidth]{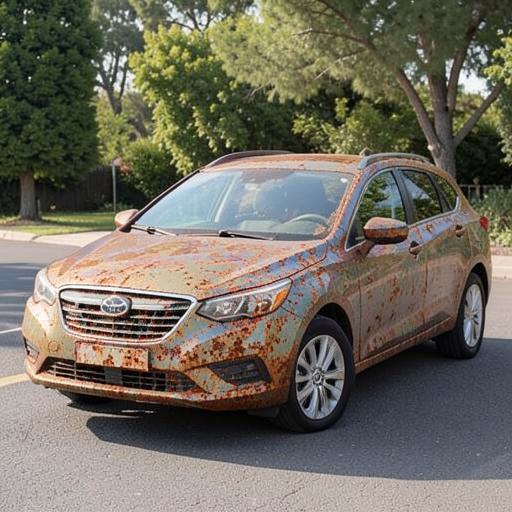} \\

        \includegraphics[width=0.14\textwidth]{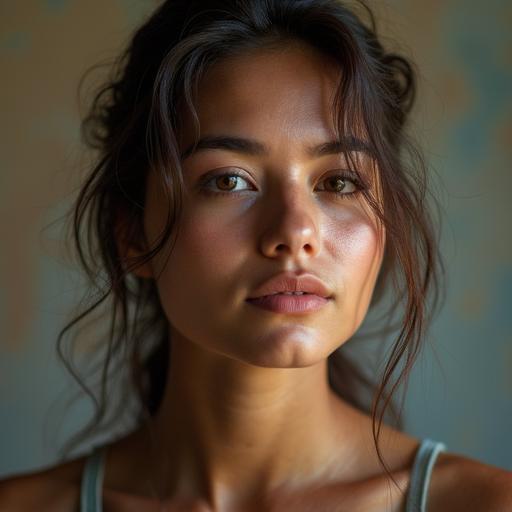} &
        \includegraphics[width=0.14\textwidth]{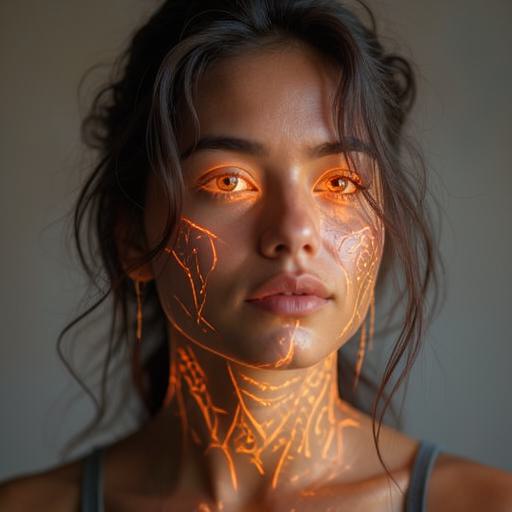} &
        \includegraphics[width=0.14\textwidth]{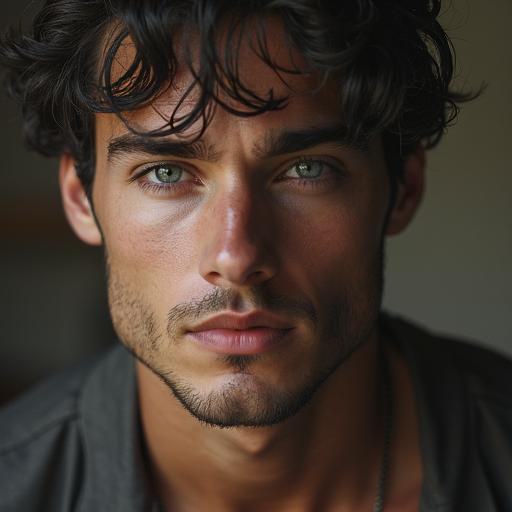} &
        \includegraphics[width=0.14\textwidth]{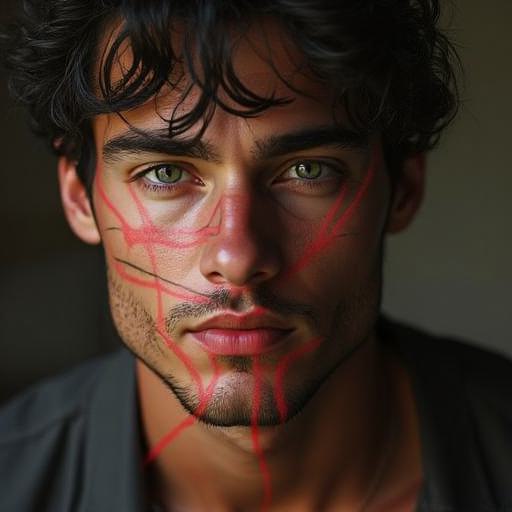} &
        \includegraphics[width=0.14\textwidth]{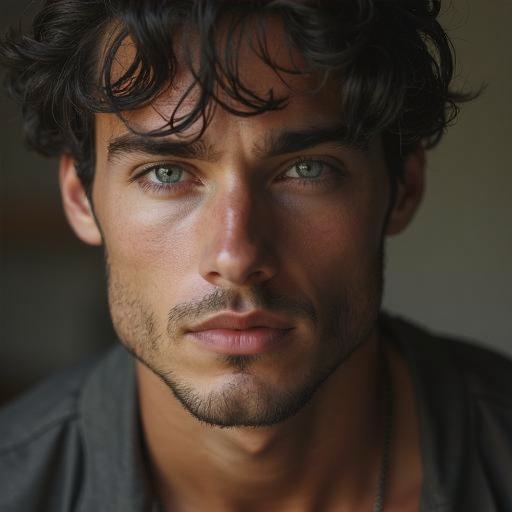} &
        \includegraphics[width=0.14\textwidth]{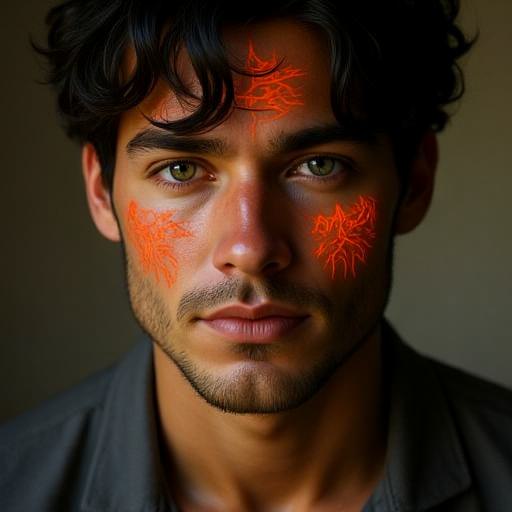} &
        \includegraphics[width=0.14\textwidth]{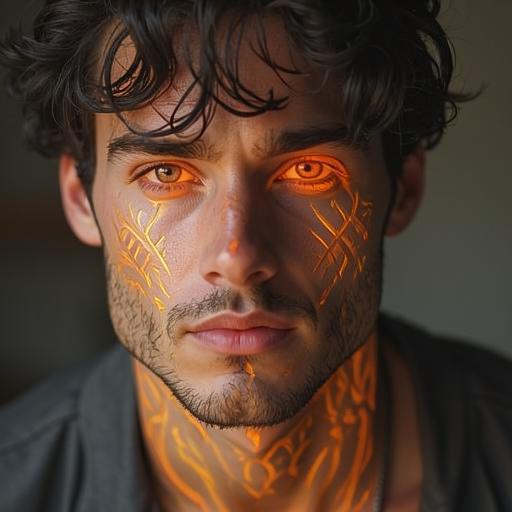} \\

        \includegraphics[width=0.14\textwidth]{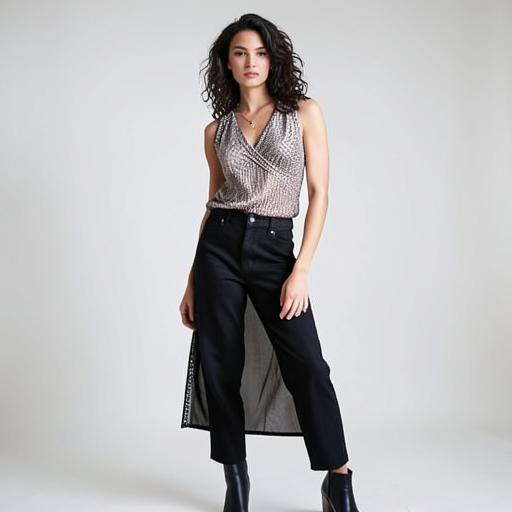} &
        \includegraphics[width=0.14\textwidth]{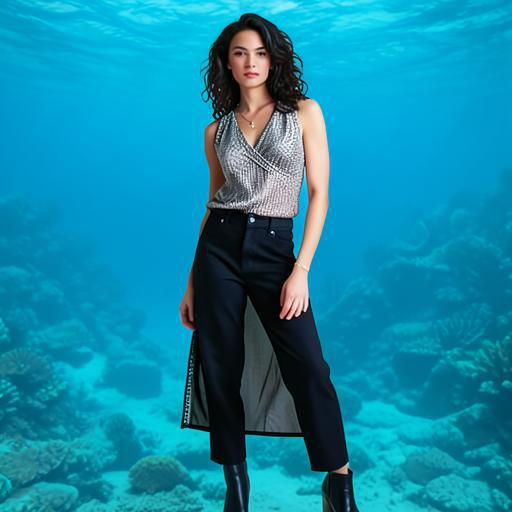} &
        \includegraphics[width=0.14\textwidth]{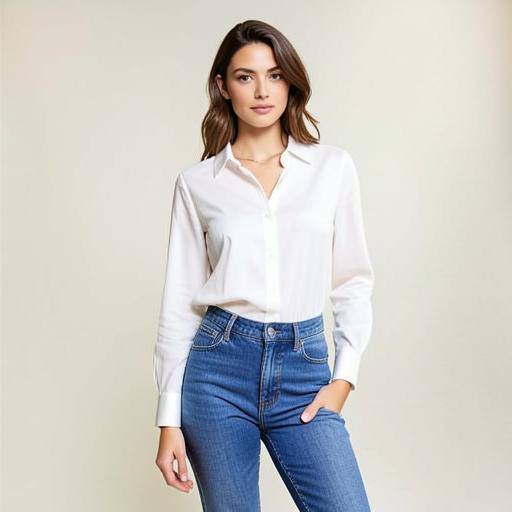} &
        \includegraphics[width=0.14\textwidth]{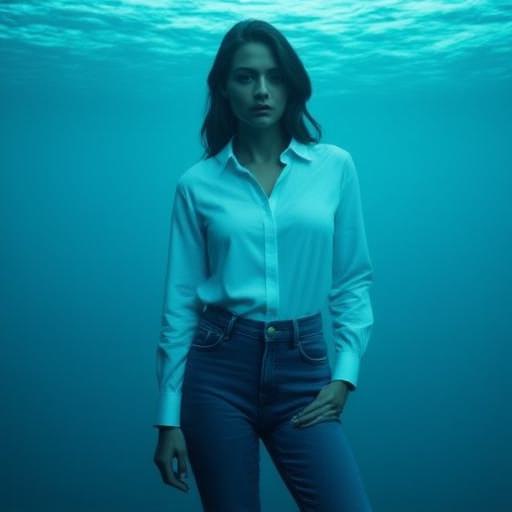} &
        \includegraphics[width=0.14\textwidth]{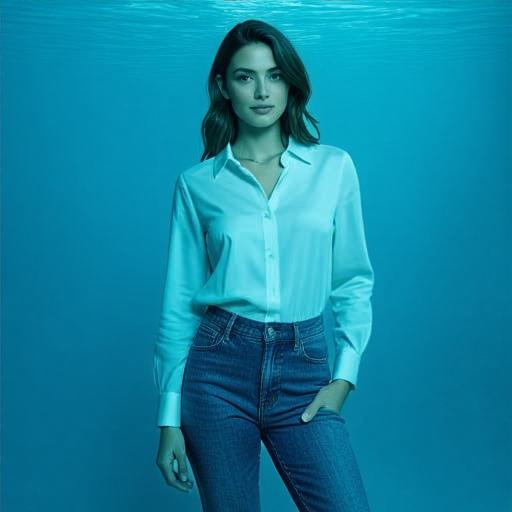} &
        \includegraphics[width=0.14\textwidth]{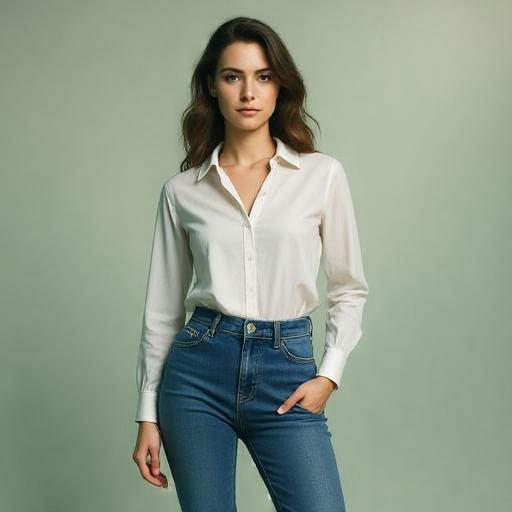} &
        \includegraphics[width=0.14\textwidth]{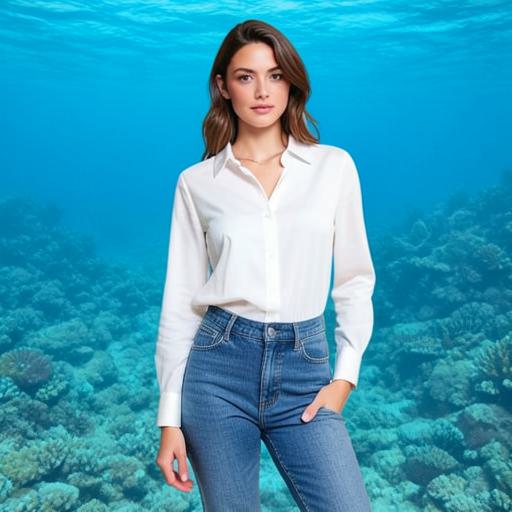} \\

    \end{tabular}
    }
    \caption{\textbf{Qualitative comparison on unseen editing tasks.} We compare Delta-Adapter with RelationAdapter~\cite{gong2025relationadapter}, LoRWeB~\cite{lorweb}, and VisualCloze~\cite{li2025visualcloze}. Across diverse unseen transformations, Delta-Adapter produces outputs that are semantically aligned with the exemplar edit.}
  \vspace{-10pt}

\label{fig:qualitative_unseen}
\end{figure*}

\noindent \textbf{Qualitative evaluation.\ }
Figure~\ref{fig:qualitative_seen} presents qualitative comparisons on seen tasks, where all models are trained on the Relation dataset~\cite{gong2025relationadapter}. LoRWeB~\cite{lorweb} often fails to capture the intended transformation from the exemplar pair, producing outputs that are nearly identical to the query image (rows 1 and 2). Edit Transfer~\cite{edit_transfer} similarly struggles to infer the edit semantics and frequently yields outputs with degraded visual quality. RelationAdapter~\cite{gong2025relationadapter} achieves improved editing fidelity but still falls short on challenging tasks such as motion deblurring (row 5) and image demoiréing (row 6), and struggles to maintain content consistency with the query image (rows 1 and 3). In contrast, Delta-Adapter reliably applies the inferred edit semantics to the query image while maintaining content consistency. We observe that Delta-Adapter successfully learns all editing tasks present in the Relation dataset. As further demonstrated in Figure~\ref{fig:qualitative_unseen}, this advantage extends to unseen tasks, where Delta-Adapter exhibits stronger generalization than all baselines. Notably, all baselines require textual instructions at both training and inference time, whereas our method operates without any textual guidance yet achieves superior performance. 

Beyond the editing tasks in the Relation dataset, we demonstrate results on more complex editing scenarios in Figures~\ref{fig:teaser} and~\ref{fig:appendix_more_ours}. Furthermore, Delta-Adapter supports continuous image editing by adjusting the injection strength $\lambda_{\mathrm{ca}}$ of the decoupled cross-attention, as shown in Figure~\ref{fig:continuous}. This capability arises from the fact that our method injects only the editing signal, whereas methods that inject the full exemplar pair (e.g., RelationAdapter) are unable to achieve such continuous control. Additional qualitative results are provided in Appendix~\ref{sec:appendix_qualitative}.

\begin{table}[t]
\centering
\caption{\textbf{Quantitative comparison on seen and unseen editing tasks.} Ours-16K denotes our model trained on 16K image pairs from the Relation dataset, and Ours-1M is trained on 1M image pairs. VisualCloze is not reported on seen-task evaluation due to training instability on the Relation dataset.}
\label{tab:quantitative}
\setlength{\tabcolsep}{8pt}
\begin{tabular}{@{\hspace{15pt}}lcccc@{\hspace{8pt}}}
\toprule
Method & CLIP-I$\uparrow$ & LPIPS$\downarrow$ & GPT-C$\uparrow$ & GPT-A$\uparrow$ \\
\midrule
\multicolumn{5}{@{\hspace{15pt}}l@{\hspace{8pt}}}{\textit{Seen tasks}} \\
\midrule
Edit Transfer~\cite{edit_transfer} & 0.764 & 0.408 & 2.839 & 2.843 \\
LoRWeB~\cite{lorweb} & \textbf{0.898} & \underline{0.254} & \textbf{4.198} & 3.033 \\
RelationAdapter~\cite{gong2025relationadapter} & 0.858 & 0.327 & 3.201 & \underline{3.417} \\
\textbf{Ours-16K} & \underline{0.896} & \textbf{0.249} & \underline{3.939} & \textbf{4.234} \\
\midrule

\multicolumn{5}{@{\hspace{15pt}}l@{\hspace{8pt}}}{\textit{Unseen tasks}} \\
\midrule
VisualCloze~\cite{li2025visualcloze} & 0.827 & 0.297 & 3.416 & 2.556 \\
LoRWeB~\cite{lorweb} & \textbf{0.945} & \textbf{0.166} & \textbf{4.680} & 2.004 \\
RelationAdapter~\cite{gong2025relationadapter} & 0.862 & 0.370 & 3.804 & 2.884 \\
\textbf{Ours-16K} & 0.882 & \underline{0.248} & 3.924 & \underline{3.332} \\
\textbf{Ours-1M}   & \underline{0.892} & 0.265 & \underline{3.964} & \textbf{4.008} \\
\bottomrule
\end{tabular}
\vspace{-12pt}
\end{table}

\vspace{1pt} \noindent \textbf{Quantitative evaluation.\ }
Table~\ref{tab:quantitative} reports quantitative results on both seen and unseen tasks. Although LoRWeB achieves the highest consistency scores, it suffers from substantially lower editing accuracy. This is primarily due to frequent editing failures in which the output remains nearly identical to the input, consistent with the qualitative observations in Figures~\ref{fig:qualitative_seen} and~\ref{fig:qualitative_unseen}. Excluding this extreme case, our method achieves superior editing accuracy and content consistency over all baselines. On unseen tasks, our method achieves a GPT-A score of 4.008, compared to 2.884 for the strongest baseline, RelationAdapter. Furthermore, scaling training data from 16K to 1M pairs yields substantial gains in editing accuracy on unseen tasks, with GPT-A improving from 3.332 to 4.008, demonstrating the benefit of large-scale single-pair supervision for generalization. Additional quantitative results on the unseen validation set released by RelationAdapter are provided in Appendix~\ref{sec:appendix_quantitative}, where our method also shows consistent improvements over all baselines. We further validate the perceptual quality of our results through a human preference study presented in Appendix~\ref{sec:appendix_user_study}.

\noindent \textbf{Test-time adaptation.\ }
As described in Section~\ref{sec:method_tta}, our single-pair supervision paradigm enables test-time adaptation for challenging unseen exemplar pairs. Empirically, we find that fine-tuning the Delta-Adapter for as few as 20 gradient steps is sufficient. Figure~\ref{fig:tta} provides qualitative comparisons before and after adaptation on several such difficult cases. Without test-time adaptation, the model captures only the coarse semantics of the intended edit. After adaptation, editing fidelity improves substantially, yielding outputs that faithfully reflect the transformation specified by the exemplar.

\begin{wraptable}{r}{0.55\textwidth}
\vspace{-0.41cm}
\centering
        \small
        \caption{Ablation study. All variants are trained on the Relation dataset.}
        \label{tab:ablation}
        \vspace{-7pt}
        \setlength{\tabcolsep}{3pt}
        \renewcommand{\arraystretch}{1}
        \begin{tabular}{lcccc}
            \toprule
            Variant & CLIP-I$\uparrow$ & LPIPS$\downarrow$ & GPT-C$\uparrow$ & GPT-A$\uparrow$ \\
            \midrule
            \multicolumn{5}{l}{\textit{Seen tasks}} \\
            \midrule
            w/o semantic delta               & 0.866 & 0.337 & 3.228 & 3.839 \\
            w/o layernorm                    & 0.875 & 0.276 & 3.772 & 3.865 \\
            w/o gated residual               & 0.887 & 0.273 & 3.824 & 4.091 \\
            w/o perceiver                    & 0.891 & 0.260 & 3.876 & 4.059 \\
            w/o per-token proj.              & 0.873 & 0.323 & 3.568 & 3.624 \\
            w/o $\mathcal{L}_{\mathrm{sdc}}$ & 0.879 & 0.287 & 3.743 & 4.162 \\
            \textbf{Ours}                    & \textbf{0.896} & \textbf{0.249} & \textbf{3.939} & \textbf{4.234} \\
            \midrule
            \multicolumn{5}{l}{\textit{Unseen tasks}} \\
            \midrule
            w/o perceiver   & 0.881 & 0.285 & 3.828 & 2.964 \\
            \textbf{Ours} & \textbf{0.882} & \textbf{0.248} & \textbf{3.924} & \textbf{3.332} \\
            \bottomrule
        \end{tabular}
        \vspace{-8pt}
\end{wraptable}

\subsection{Ablation Study}
\label{sec:ablation}
To validate the contribution of each component in Delta-Adapter, we conduct ablation studies by removing individual components. Specifically, we evaluate six variants: (1) \emph{w/o semantic delta} replaces the semantic delta with the full exemplar pair as conditioning; (2) \emph{w/o layernorm} removes layer normalization applied to the semantic delta; (3) \emph{w/o gated residual} feeds the normalized delta directly into the resampler without gated residual refinement; (4) \emph{w/o perceiver} replaces the Perceiver resampler with a pooling-MLP architecture; (5) \emph{w/o per-token proj.} replaces the per-token projection with a standard shared linear projection; and (6) \emph{w/o $\mathcal{L}_{\mathrm{sdc}}$} removes the semantic delta consistency loss. Quantitative results for all variants are reported in Table~\ref{tab:ablation}. As shown, removing any single component leads to a consistent performance drop, confirming that each design choice contributes meaningfully to the overall framework. Notably, removing the Perceiver resampler (w/o perceiver) causes a substantial degradation on unseen tasks. 

\begin{figure*}[t]
    \centering

    \begin{minipage}[t]{0.49\textwidth}
        \vspace{0pt}
        \centering
        \renewcommand{\arraystretch}{0.25}
        \setlength{\tabcolsep}{0.5pt}

        {\footnotesize
        \begin{tabular}{c @{\hspace{0.04cm}} | @{\hspace{0.04cm}} c c c c}

            \multicolumn{1}{c}{ Query } &
            \multicolumn{4}{c}{ Edit strength + $\xrightarrow{\hspace{2.3cm}}$ } \\

            \includegraphics[width=0.195\linewidth]{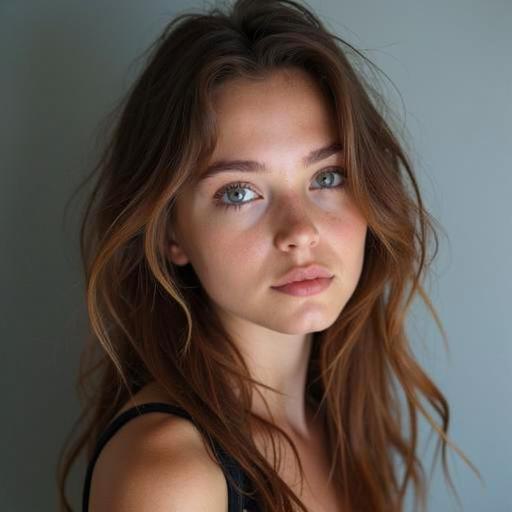} &
            \includegraphics[width=0.195\linewidth]{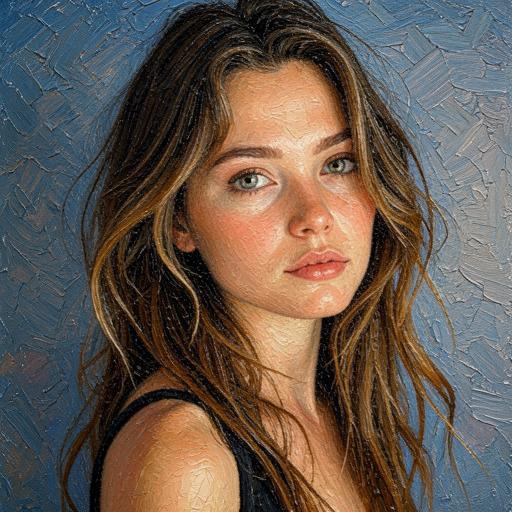} &
            \includegraphics[width=0.195\linewidth]{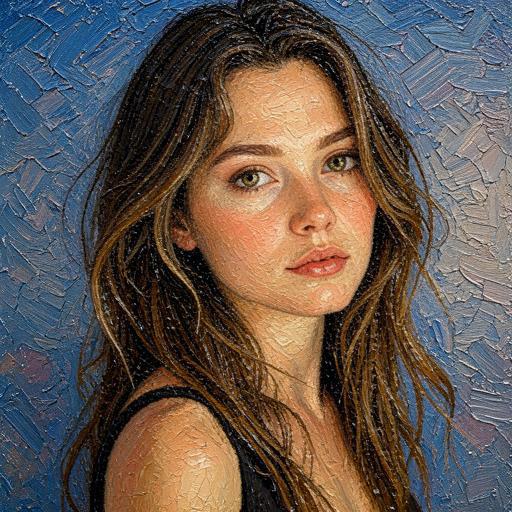} &
            \includegraphics[width=0.195\linewidth]{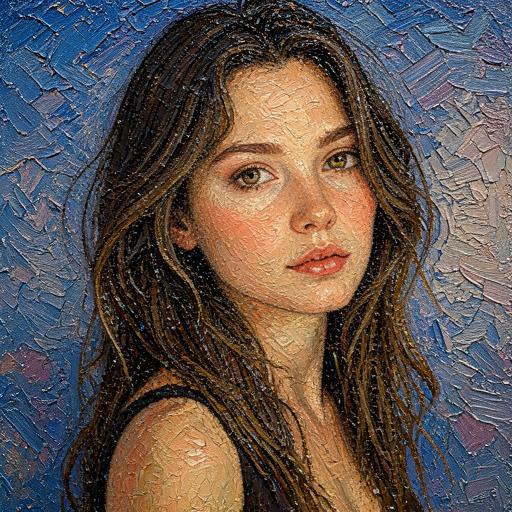} &
            \includegraphics[width=0.195\linewidth]{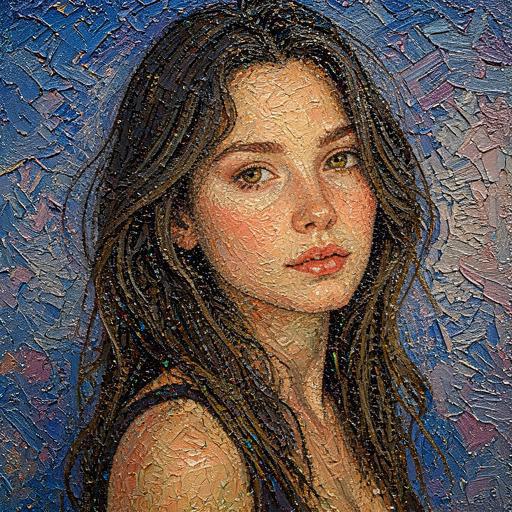} \\

            \includegraphics[width=0.195\linewidth]{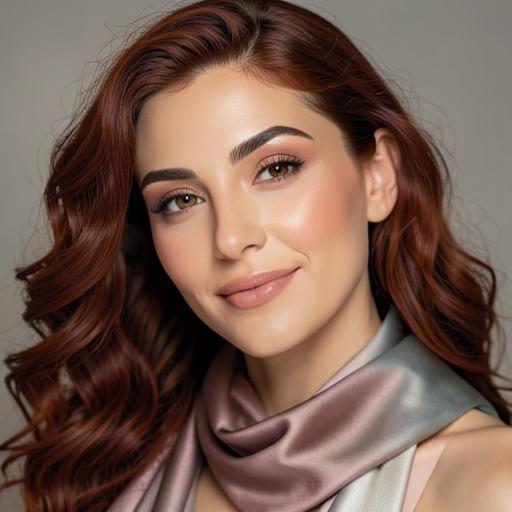} &
            \includegraphics[width=0.195\linewidth]{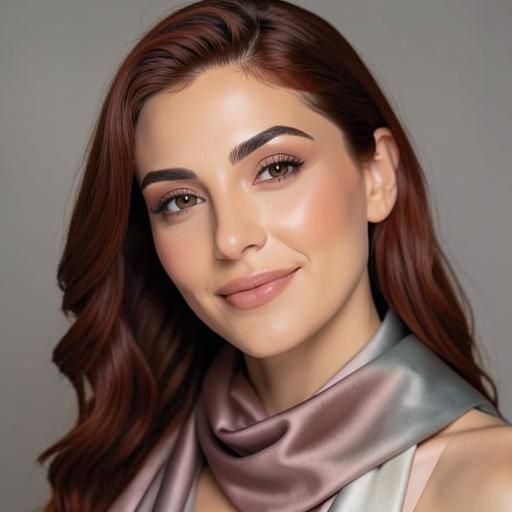} &
            \includegraphics[width=0.195\linewidth]{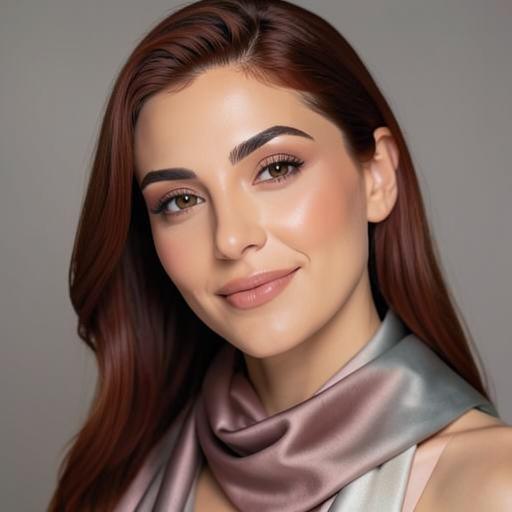} &
            \includegraphics[width=0.195\linewidth]{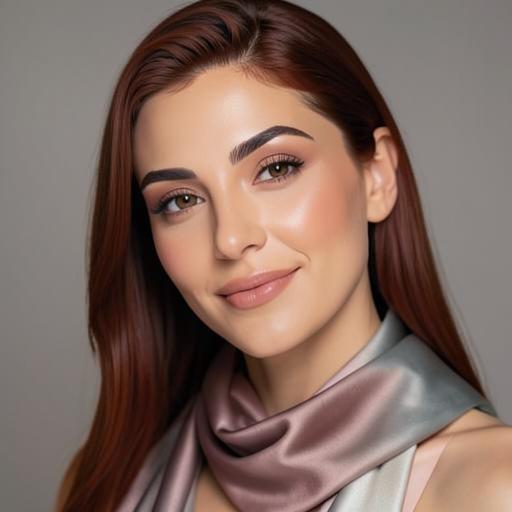} &
            \includegraphics[width=0.195\linewidth]{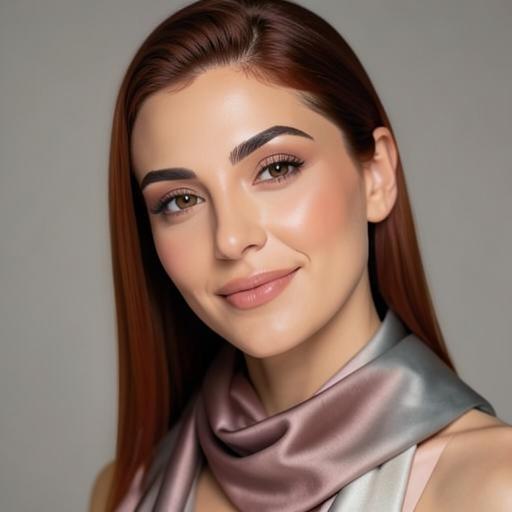} \\

            \includegraphics[width=0.195\linewidth]{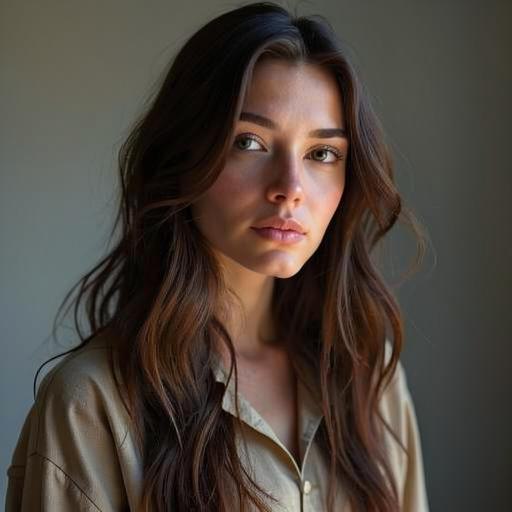} &
            \includegraphics[width=0.195\linewidth]{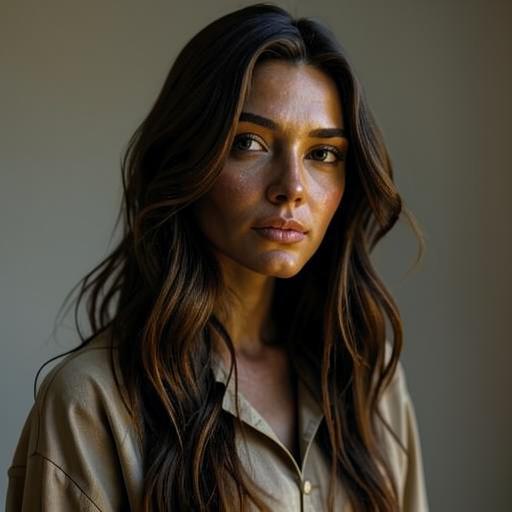} &
            \includegraphics[width=0.195\linewidth]{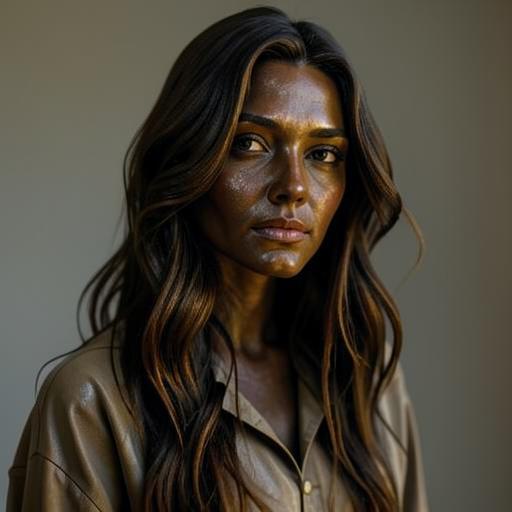} &
            \includegraphics[width=0.195\linewidth]{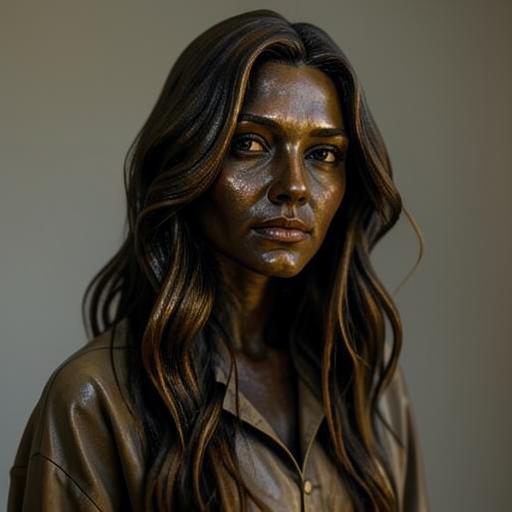} &
            \includegraphics[width=0.195\linewidth]{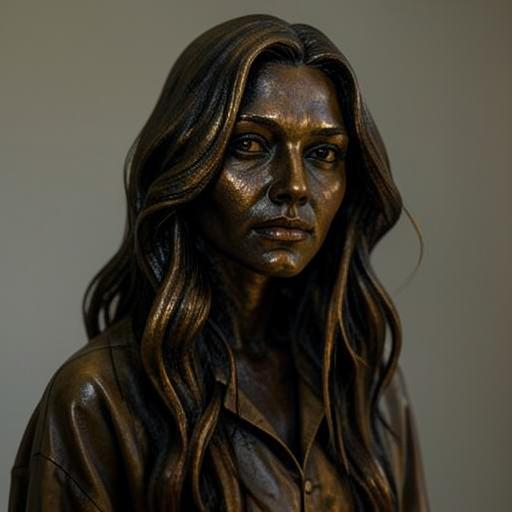} \\

            \includegraphics[width=0.195\linewidth]{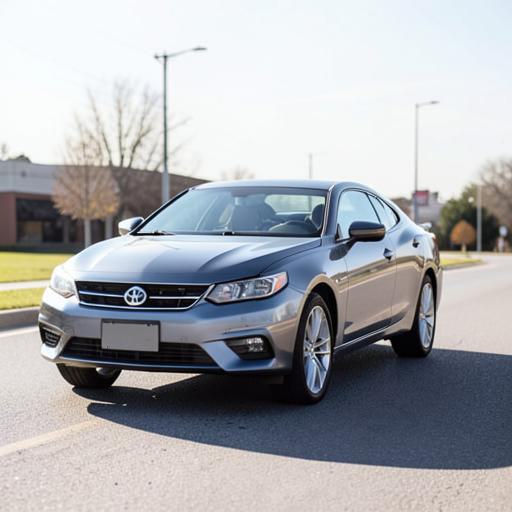} &
            \includegraphics[width=0.195\linewidth]{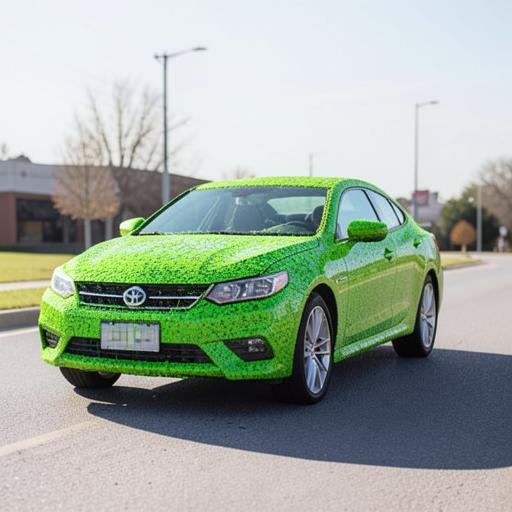} &
            \includegraphics[width=0.195\linewidth]{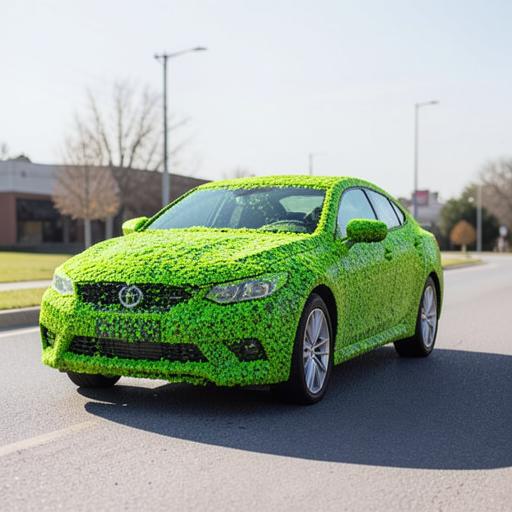} &
            \includegraphics[width=0.195\linewidth]{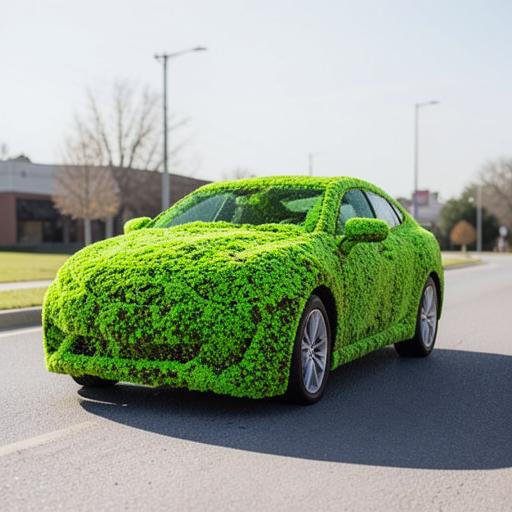} &
            \includegraphics[width=0.195\linewidth]{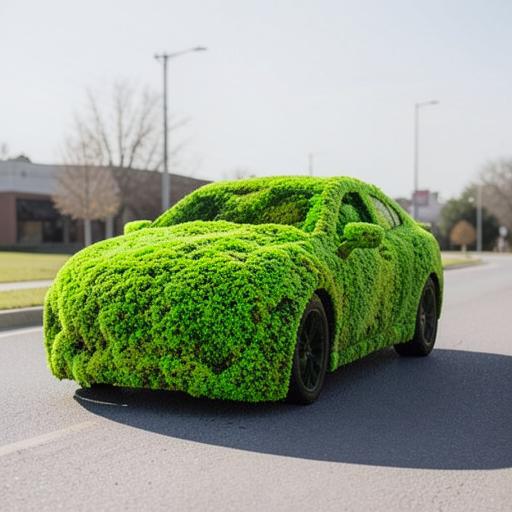} \\

        \end{tabular}
        }

        \vspace{-0.15cm}
        \caption{Continuous image editing with Delta-Adapter.}
        \label{fig:continuous}
    \end{minipage}
    \hfill
    \begin{minipage}[t]{0.49\textwidth}
        \vspace{0pt}
        \centering
        \renewcommand{\arraystretch}{0.25}
        \setlength{\tabcolsep}{0.5pt}

        {\footnotesize
        \begin{tabular}{c c c @{\hspace{0.04cm}} | @{\hspace{0.04cm}} c c}

            \multicolumn{1}{c}{ Source ($a$)} &
            \multicolumn{1}{c}{ Target ($a'$)} &
            \multicolumn{1}{c}{ Query ($b$)} &
            \multicolumn{1}{c}{ w/o TTA} &
            \multicolumn{1}{c}{ w/ TTA} \\

            \includegraphics[width=0.195\linewidth]{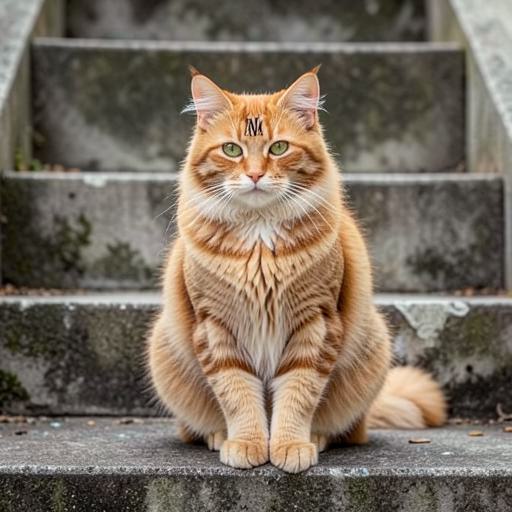} &
            \includegraphics[width=0.195\linewidth]{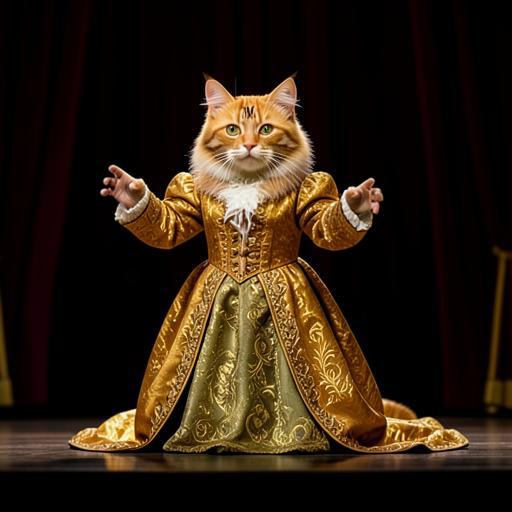} &
            \includegraphics[width=0.195\linewidth]{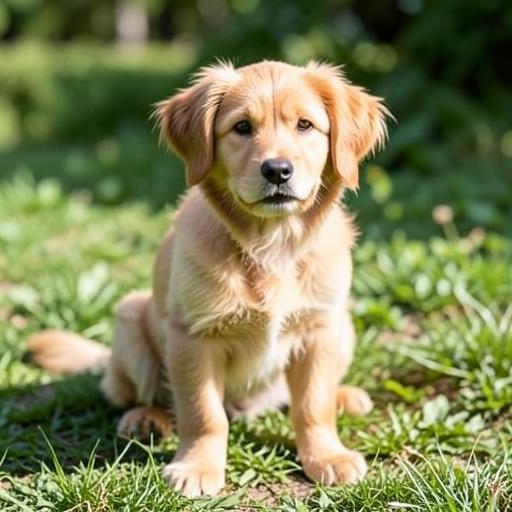} &
            \includegraphics[width=0.195\linewidth]{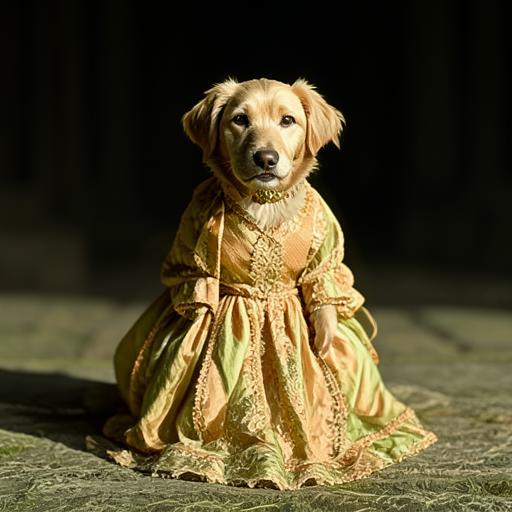} &
            \includegraphics[width=0.195\linewidth]{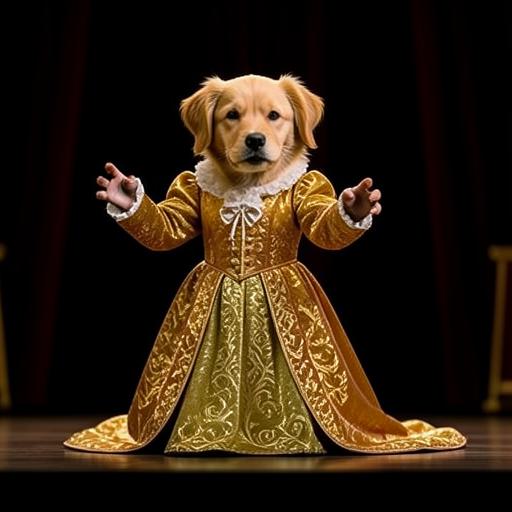} \\

            \includegraphics[width=0.195\linewidth]{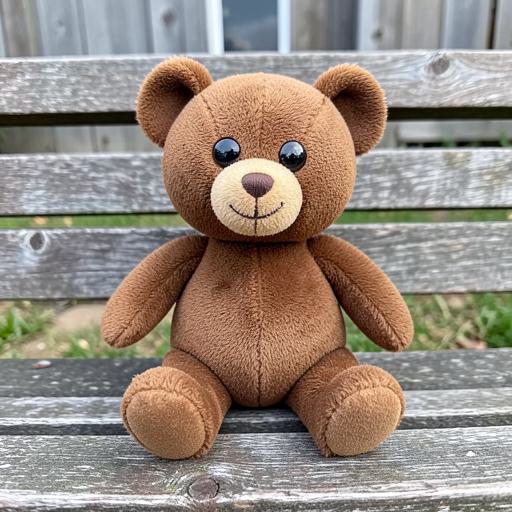} &
            \includegraphics[width=0.195\linewidth]{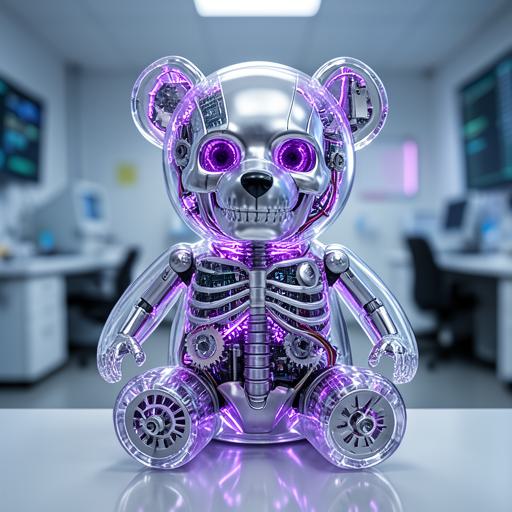} &
            \includegraphics[width=0.195\linewidth]{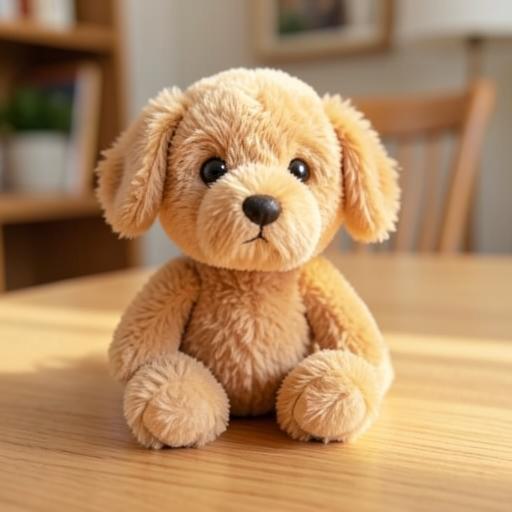} &
            \includegraphics[width=0.195\linewidth]{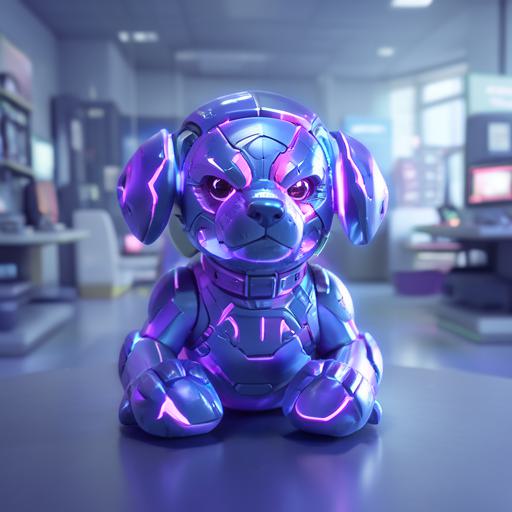} &
            \includegraphics[width=0.195\linewidth]{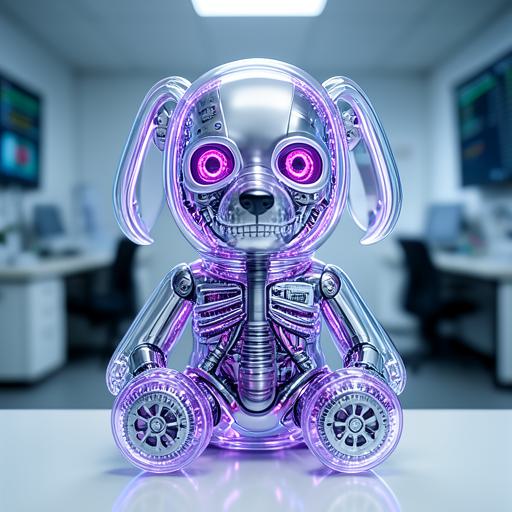} \\

            \includegraphics[width=0.195\linewidth]{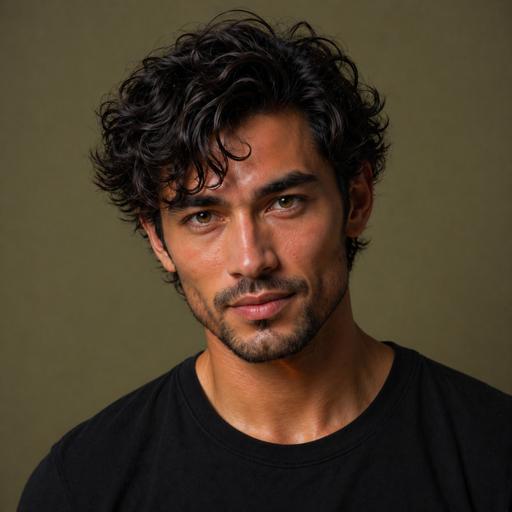} &
            \includegraphics[width=0.195\linewidth]{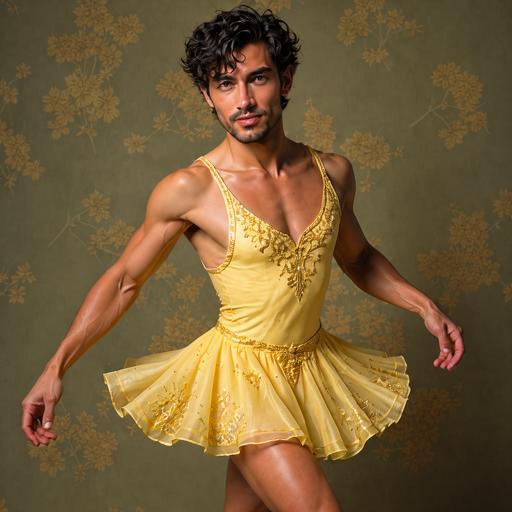} &
            \includegraphics[width=0.195\linewidth]{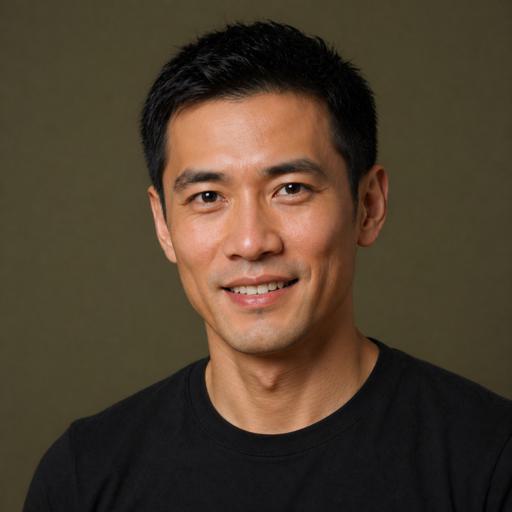} &
            \includegraphics[width=0.195\linewidth]{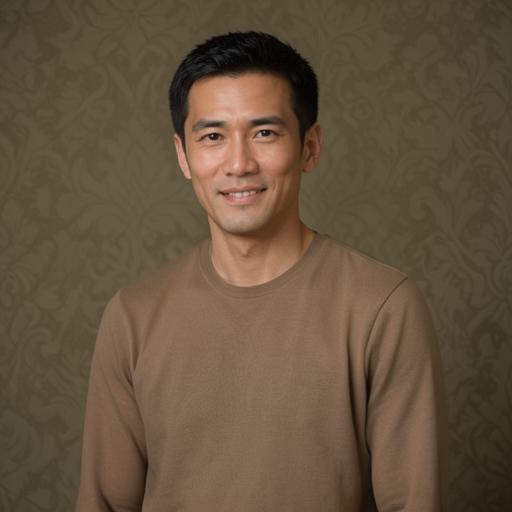} &
            \includegraphics[width=0.195\linewidth]{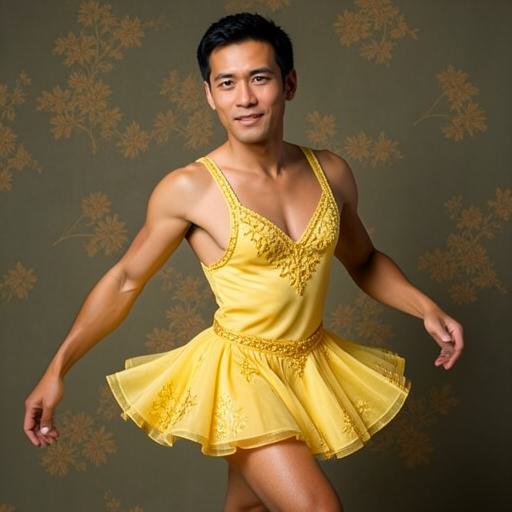} \\

            \includegraphics[width=0.195\linewidth]{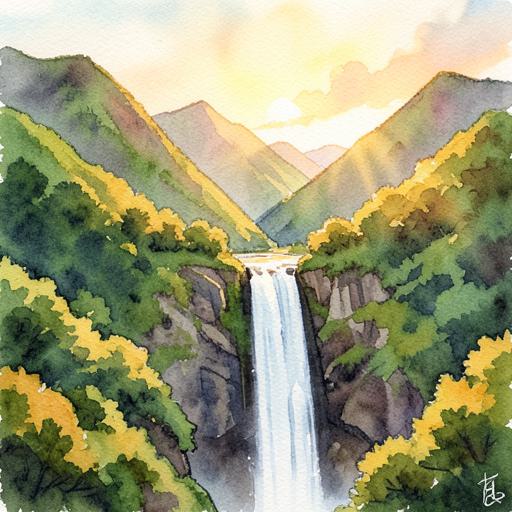} &
            \includegraphics[width=0.195\linewidth]{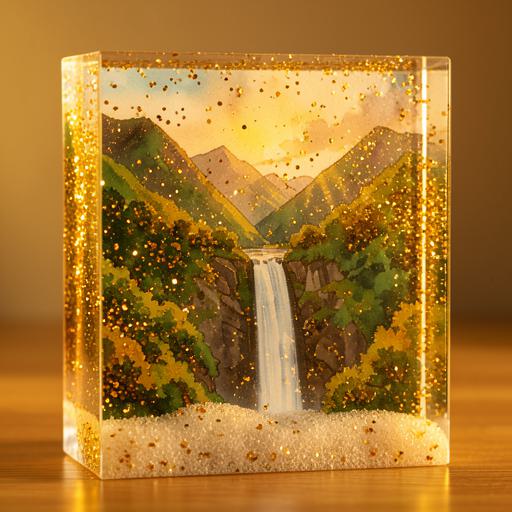} &
            \includegraphics[width=0.195\linewidth]{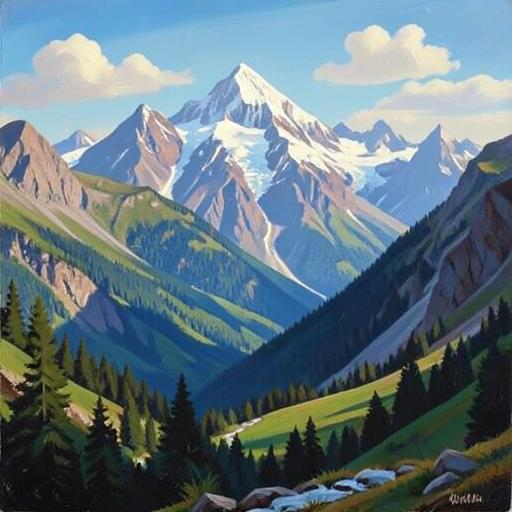} &
            \includegraphics[width=0.195\linewidth]{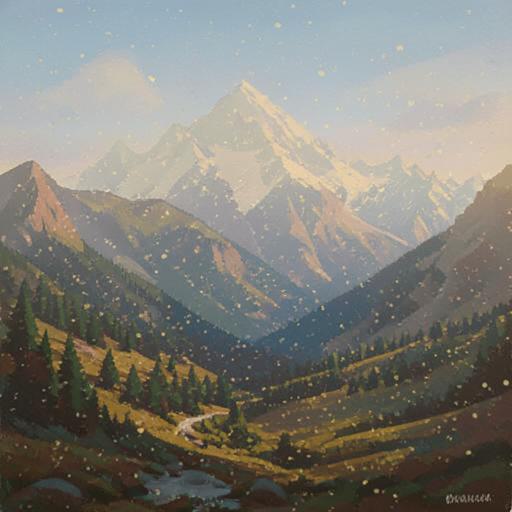} &
            \includegraphics[width=0.195\linewidth]{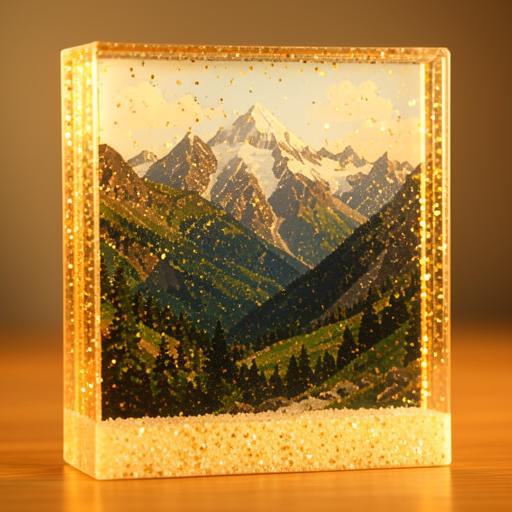} \\

        \end{tabular}
        }

        \vspace{-0.15cm}
        \caption{Test-time adaptation (TTA) on challenging unseen exemplar pairs.}
        \label{fig:tta}
    \end{minipage}
  \vspace{-10pt}
\end{figure*}

\section{Conclusions and Limitations}
\label{sec:conclusion_and_limitation}

We presented Delta-Adapter, a framework for exemplar-based image editing under single-pair supervision without textual guidance. By conditioning on a semantic delta rather than the edited image directly, our method eliminates the need for paired exemplars at training time, enabling scalable data curation and effective test-time adaptation. Experiments demonstrate superior performance over four strong baselines on both seen and unseen editing tasks. One limitation of our approach is that the semantic delta is bounded by the representational capacity of the vision encoder, which operates at a high semantic level and may fail to capture fine-grained visual details such as text rendered within images. Several failure examples are provided in Appendix~\ref{sec:appendix_failure}. Additionally, our model is currently trained on approximately one million image pairs; we anticipate that scaling to larger datasets will further improve editing generalization, which we leave as future work.

\bibliographystyle{plainnat}
\bibliography{ref}


\newpage
\appendix

\section{Implementation Details}
\label{sec:appendix_implementation}

\paragraph{Our method.}
We build Delta-Adapter on top of FLUX.2-klein-4B\footnote{\url{https://huggingface.co/black-forest-labs/FLUX.2-klein-4B}}, using SigLIP-2~\cite{siglip} (\texttt{google/siglip2-base-patch16-224}) as the image encoder. The mapping layers consist of a Perceiver resampler~\cite{perceiver} with 128 learnable queries. We train with the AdamW optimizer ($\text{lr}=1\times10^{-4}$, weight decay $=0.01$, $\beta_1=0.9$, $\beta_2=0.99$) in bfloat16 precision. The full model is trained on approximately one million image pairs for 100K steps, with a per-GPU batch size of 16 across $4\times$ H200 GPUs. For the seen-task comparisons in Figure~\ref{fig:qualitative_seen} and Table~\ref{tab:quantitative}, we additionally train a model exclusively on 16K image pairs from the Relation dataset to ensure a fair comparison with RelationAdapter and LoRWeB, using a batch size of 32 for 10K steps. Throughout training, only the mapping layers and the newly introduced cross-attention layers are optimized, while keeping FLUX.2-klein and SigLIP-2 frozen. For test-time adaptation, we fine-tune Delta-Adapter for 20 gradient steps with a batch size of 1 and a learning rate of $1\times10^{-4}$. Notably, this fine-tuning requires only 33 GB of GPU memory, enabling practical deployment on a single GPU.

\paragraph{Baselines.}
For RelationAdapter~\cite{gong2025relationadapter}, LoRWeB~\cite{lorweb}, and VisualCloze~\cite{li2025visualcloze}, we evaluate using their official pre-trained checkpoints and default inference configurations. For Edit Transfer~\cite{edit_transfer}, we follow the official training protocol and retrain the model on the Relation dataset to ensure a fair comparison. As these baselines additionally require a text instruction alongside the exemplar pair, we generate a concise edit description for each evaluation case using GPT-5.4, conditioned on the exemplar pair $a \rightarrow a'$. All methods are evaluated on identical query images and exemplar pairs under a fixed random seed of 42 to ensure reproducibility.

\section{Additional Qualitative Results}
\label{sec:appendix_qualitative}
Figure~\ref{fig:appendix_more_ours} presents a broader set of challenging image editing results produced by our Delta-Adapter. Figures~\ref{fig:additional_qualitative_seen} and~\ref{fig:additional_qualitative_unseen} provide additional qualitative comparisons with baseline methods on seen and unseen editing tasks, respectively.

\section{Additional Quantitative Results}
\label{sec:appendix_quantitative}
Table~\ref{tab:appendix_quantitative} presents additional quantitative comparisons on the unseen validation set provided by RelationAdapter~\cite{gong2025relationadapter}. For this validation set, we follow RelationAdapter and compute consistency scores (CLIP-I and LPIPS) between the generated and ground-truth images. Our method consistently outperforms all baselines in both editing accuracy and content consistency. Notably, our method achieves a GPT-A score of 4.077, compared to 3.432 for the best-performing baseline, RelationAdapter, representing a relative improvement of 18.8\%.

\section{Comparison with Additional Baselines}
\label{sec:appendix_more_baselines}
In Figure~\ref{fig:qualitative_gpt_nano}, we compare Delta-Adapter against two state-of-the-art multimodal generation models, Nano Banana 2~\cite{banana} and GPT-Image-2~\cite{gpt_image}. As general-purpose generation systems, both models exhibit notable limitations in editing accuracy and content consistency when applied to exemplar-based image editing. Specifically, they often fail to capture the intended edits (rows 1--3) and tend to leak appearance cues from the exemplar images into the output (rows 4--6). In contrast, Delta-Adapter consistently transfers the target edit while faithfully preserving the content of the query image. In Figure~\ref{fig:qualitative_pairedit}, we further compare against PairEdit~\cite{lu2025pairedit}, an optimization-based method that performs test-time optimization independently for each exemplar pair. PairEdit struggles to capture complex edits from the exemplar, whereas Delta-Adapter faithfully transfers the intended transformations to the query image.

\begin{table}[t]
\centering
\caption{\textbf{Additional quantitative comparison on the unseen validation set released by RelationAdapter.} Ours-16K denotes our model trained on 16K image pairs from the Relation dataset, and Ours-1M denotes our model trained on 1M image pairs.}
\label{tab:appendix_quantitative}
\setlength{\tabcolsep}{8pt}
\begin{tabular}{@{\hspace{15pt}}lcccc@{\hspace{8pt}}}
\toprule
Method & CLIP-I$\uparrow$ & LPIPS$\downarrow$ & GPT-C$\uparrow$ & GPT-A$\uparrow$ \\
\midrule
VisualCloze~\cite{li2025visualcloze} & 0.808 & 0.499 & 2.973 & 3.277 \\
LoRWeB~\cite{lorweb} & 0.837 & 0.446 & 3.793 & 3.023 \\
RelationAdapter~\cite{gong2025relationadapter} & 0.851 & 0.423 & 3.847 & 3.432 \\
\textbf{Ours-16K} & \underline{0.853} & \textbf{0.396} & \textbf{4.060} & \underline{3.847} \\
\textbf{Ours-1M}  & \textbf{0.867} & \underline{0.399} & \underline{3.943} & \textbf{4.077} \\
\bottomrule
\end{tabular}
\vspace{-10pt}
\end{table}

\section{Failure Cases}
\label{sec:appendix_failure}
While Delta-Adapter generalizes well across a wide range of editing tasks, it can struggle to preserve fine-grained visual details, particularly textual content. Figure~\ref{fig:failure_cases} illustrates representative failure cases where the model generates characters inconsistent with those in the exemplar pair. This limitation is partly due to the pre-trained vision encoder, which may not faithfully capture such fine-grained visual details.

\section{Details of GPT-Based Evaluation Metrics}
\label{sec:appendix_gpt_eval}

We employ GPT-5.4 as an automated evaluator to assess each edited result along two dimensions: editing accuracy (GPT-A) and content consistency (GPT-C). For each instance, the evaluator receives four images as input: the reference source $a$, the reference target $a'$, the query image $b$, and the candidate edit $\hat{b}'$. It is then prompted to judge whether the intended transformation $a \rightarrow a'$ is faithfully applied to $b$, and whether the non-edited content of the query image is preserved. The complete system prompt is provided in Figure~\ref{fig:gpt_prompt}.

\section{Human Preference Evaluation}
\label{sec:appendix_user_study}
\begin{wraptable}{r}{0.48\textwidth}
\vspace{-1.0em}
\centering
\small
\caption{Pairwise user preference study comparing our method with each baseline.}
\setlength{\tabcolsep}{4pt}
\begin{tabular}{lcc}
\toprule
Baseline & Pref.\ Baseline & Pref.\ Ours \\
\midrule
VisualCloze~\cite{li2025visualcloze}     & 13.54\,\% & \textbf{86.46\,\%} \\
Edit Transfer~\cite{edit_transfer}    & 6.77\,\%  & \textbf{93.23\,\%} \\
LoRWeB~\cite{lorweb}          & 10.42\,\% & \textbf{89.58\,\%} \\
RelationAdapter~\cite{gong2025relationadapter} & 19.27\,\% & \textbf{80.73\,\%} \\
\bottomrule
\end{tabular}
\label{tab:user_study}
\vspace{-1.0em}
\end{wraptable}

\begin{figure}[t]
 \centering
 \includegraphics[width=.8\linewidth]{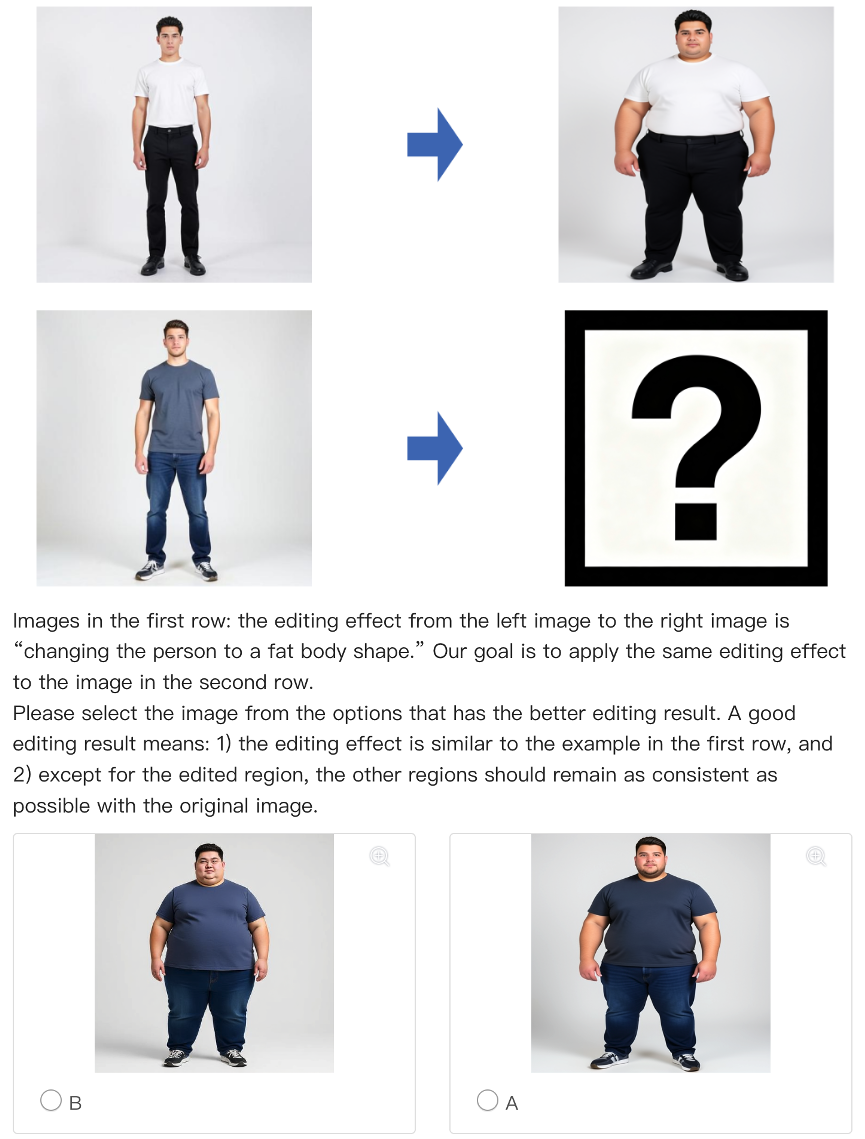}
\caption{\textbf{An example question from the user study.} Each question follows a two-alternative forced-choice format: participants view an exemplar pair $(a, a')$, a query image $b$, and two anonymized candidate edits produced by Delta-Adapter and a baseline, randomly assigned to positions A and B. Participants select the candidate that better reflects the transformation demonstrated by the exemplar pair while preserving the unedited regions of the query image.}
\label{fig:user_study_interface}

\end{figure}

To further validate the perceptual quality of our results, we conduct a pairwise preference study comparing Delta-Adapter against each baseline. In each trial, participants are presented with an exemplar pair and a query image, alongside two candidate edits: one produced by our method and one by a baseline. They are asked to select the edit that better captures the transformation demonstrated by the exemplar pair while preserving the unedited regions of the query image. We collect 768 judgments from 32 participants, with responses evenly distributed across seen and unseen editing tasks. As reported in Table~\ref{tab:user_study}, participants consistently preferred Delta-Adapter over all baselines by a substantial margin. A representative interface screenshot is shown in Figure~\ref{fig:user_study_interface}.

\section{Societal Impact}
\label{sec:appendix_impact}
Delta-Adapter enables image edits specified entirely through visual exemplar pairs, without textual instructions, lowering the barrier to exemplar-based editing and supporting applications such as content creation, artistic exploration, and visual prototyping. Like other generative editing methods, it may be misused to create misleading imagery. Mitigating such risks calls for reliable detection of synthetic or edited images~\cite{wang2020cnngenerated,corvi2022detection}, along with responsible deployment practices such as provenance tracking and user-facing disclosures.

\section{Licenses for Pre-trained Models and Datasets}
\label{sec:appendix_license}
Our implementation builds upon the publicly available FLUX.2-klein-4B model, released under the FLUX.2 Community License, and the SigLIP-2 image encoder, released under the Apache 2.0 License. For training, we use three publicly available image-pair datasets: Relation~\cite{gong2025relationadapter}, Pico-Banana~\cite{picobanana}, and NHR-Edit~\cite{nhr}, and adhere to the licenses specified by their respective authors. All evaluation images are drawn from these datasets or sourced from~\cite{vmdiff,contradictory,controlnet,lu2025pairedit}.

\begin{figure*}[t]
    \centering
    \renewcommand{\arraystretch}{0.3}
    \setlength{\tabcolsep}{0.6pt}

    {\footnotesize
    \begin{tabular}{c c c @{\hspace{0.04cm}} | @{\hspace{0.04cm}} c @{\hspace{0.12cm}} c c c @{\hspace{0.04cm}} | @{\hspace{0.04cm}} c}

        \multicolumn{1}{c}{\normalsize Source ($a$)} &
        \multicolumn{1}{c}{\normalsize Target ($a'$)} &
        \multicolumn{1}{c}{\normalsize Query ($b$)} &
        \multicolumn{1}{c}{\normalsize Ours} &
        \multicolumn{1}{c}{\normalsize Source ($a$)} &
        \multicolumn{1}{c}{\normalsize Target ($a'$)} &
        \multicolumn{1}{c}{\normalsize Query ($b$)} &
        \multicolumn{1}{c}{\normalsize Ours} \\

        \includegraphics[width=0.12\textwidth]{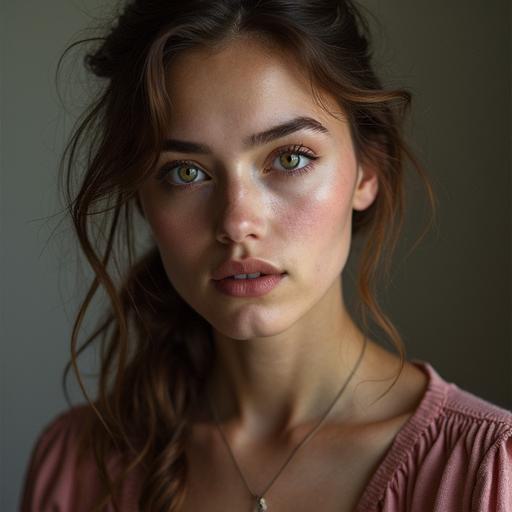} &
        \includegraphics[width=0.12\textwidth]{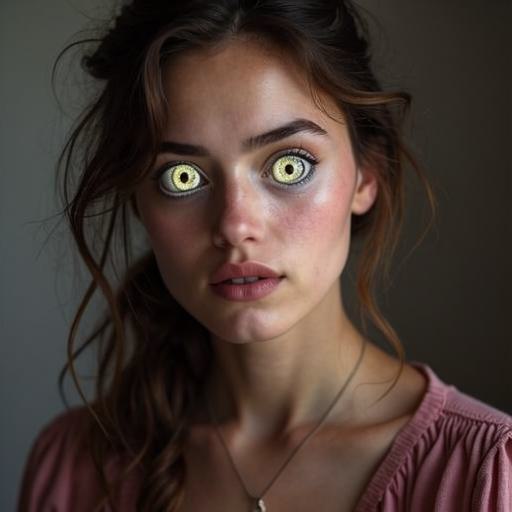} &
        \includegraphics[width=0.12\textwidth]{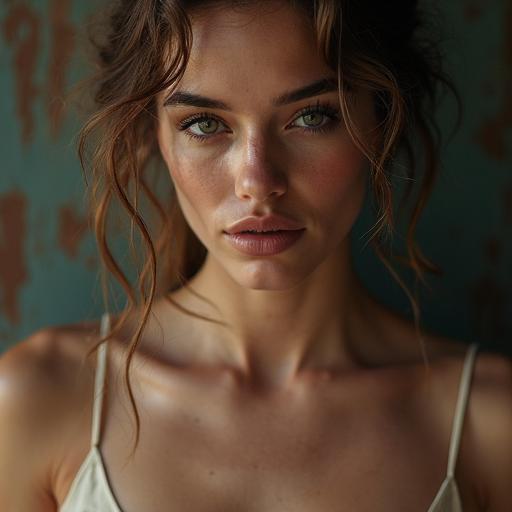} &
        \includegraphics[width=0.12\textwidth]{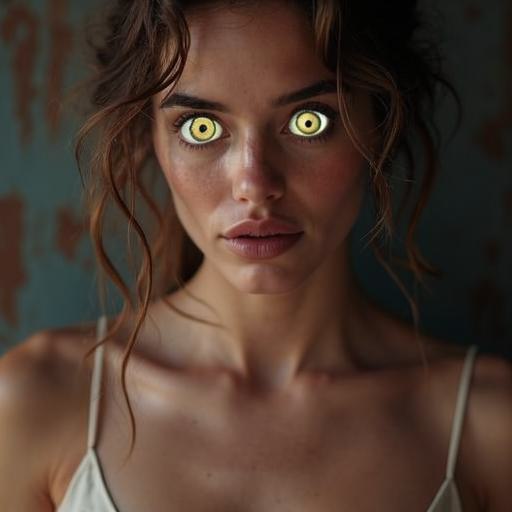} &
        \includegraphics[width=0.12\textwidth]{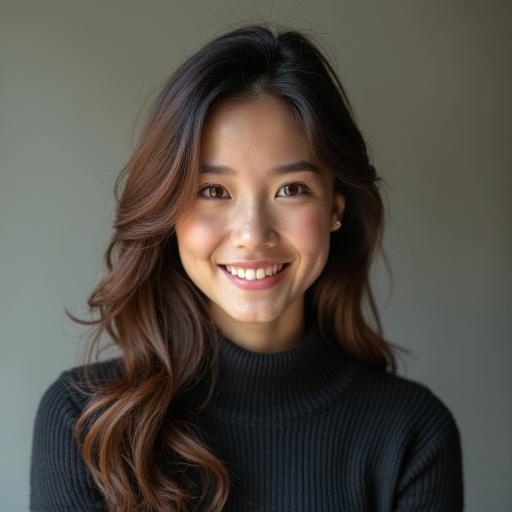} &
        \includegraphics[width=0.12\textwidth]{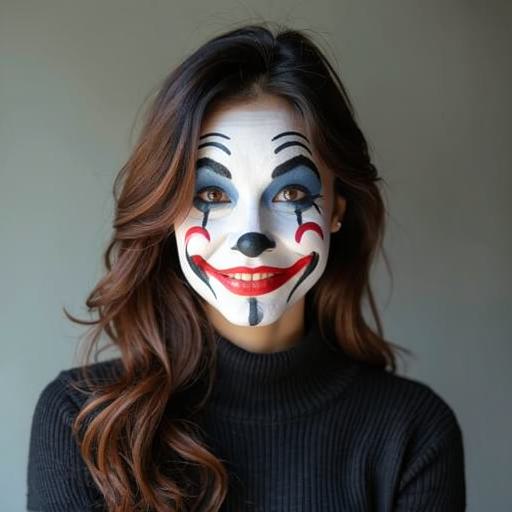} &
        \includegraphics[width=0.12\textwidth]{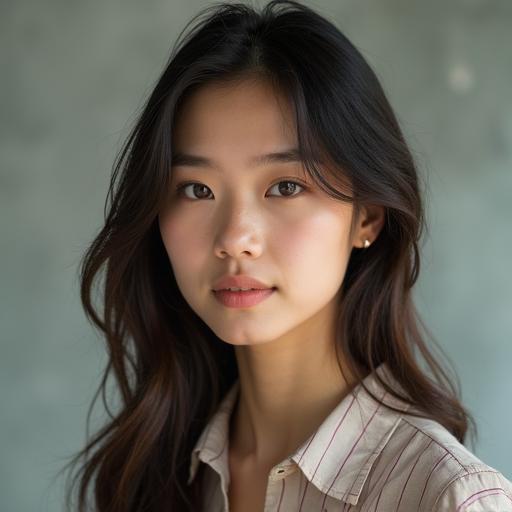} &
        \includegraphics[width=0.12\textwidth]{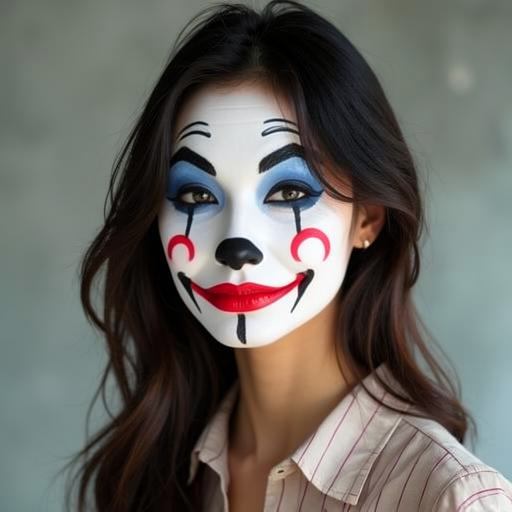} \\

        \includegraphics[width=0.12\textwidth]{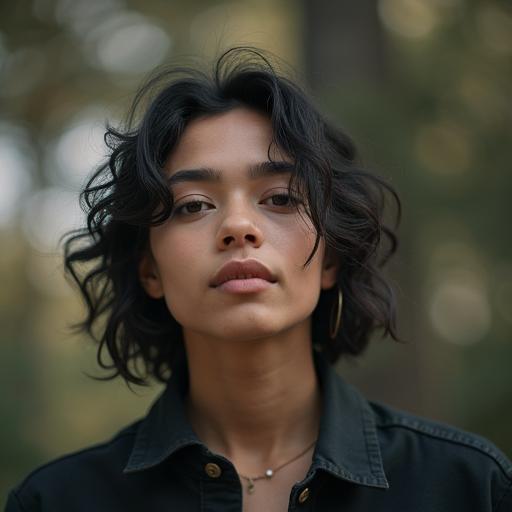} &
        \includegraphics[width=0.12\textwidth]{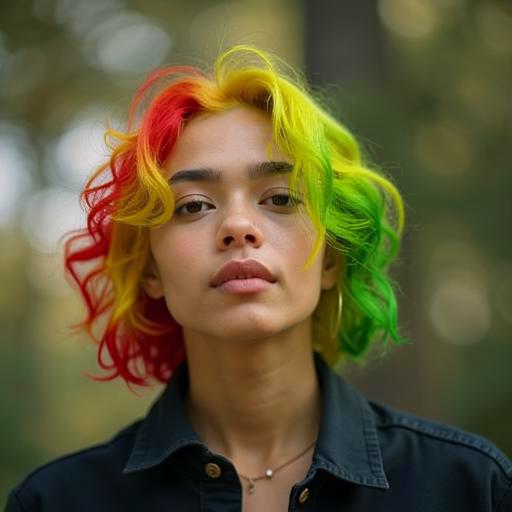} &
        \includegraphics[width=0.12\textwidth]{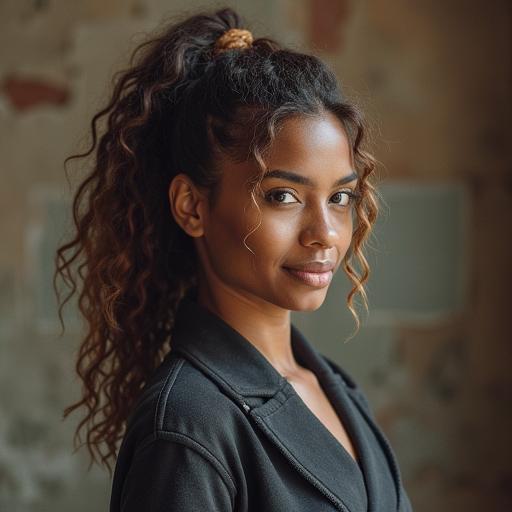} &
        \includegraphics[width=0.12\textwidth]{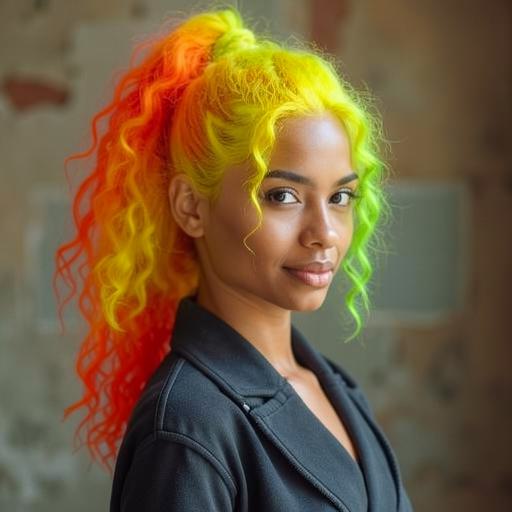} &
        \includegraphics[width=0.12\textwidth]{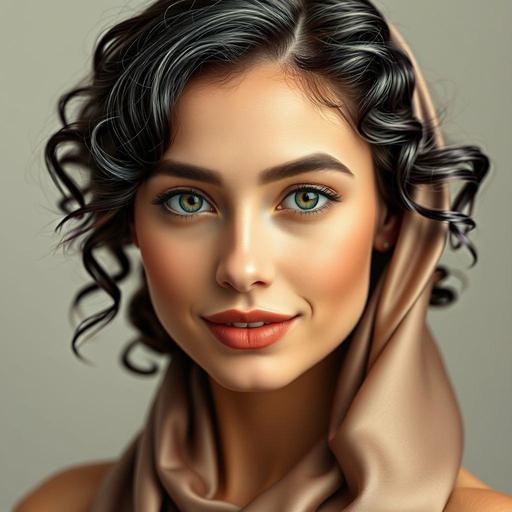} &
        \includegraphics[width=0.12\textwidth]{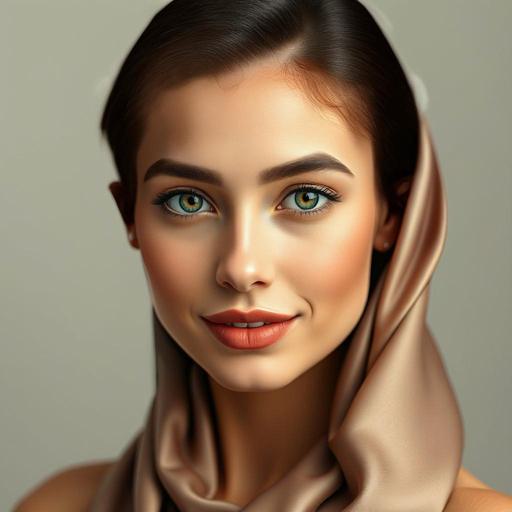} &
        \includegraphics[width=0.12\textwidth]{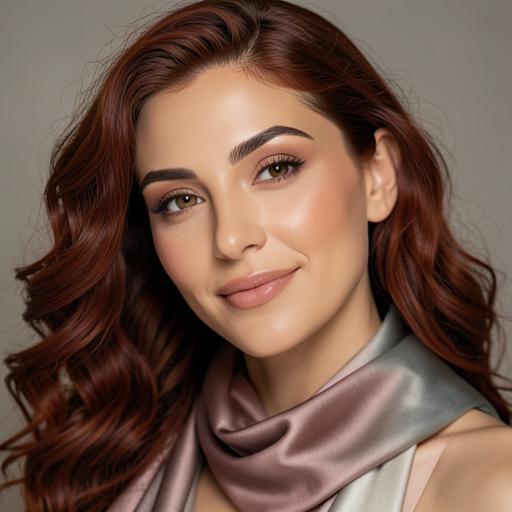} &
        \includegraphics[width=0.12\textwidth]{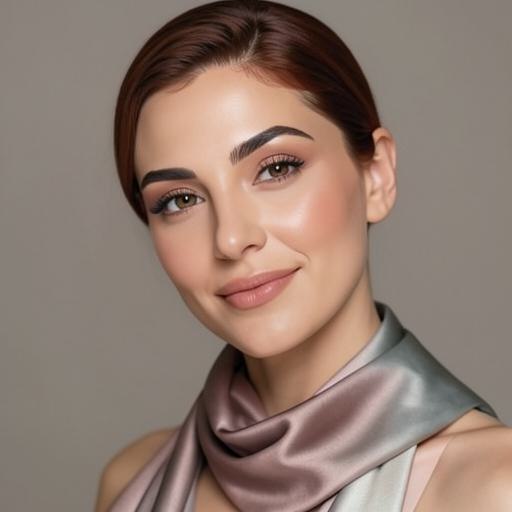} \\

        \includegraphics[width=0.12\textwidth]{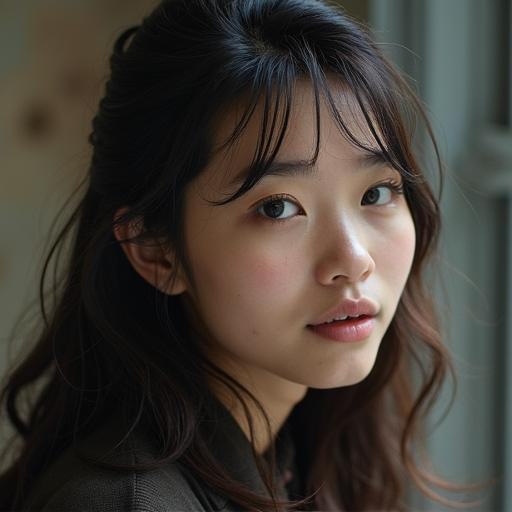} &
        \includegraphics[width=0.12\textwidth]{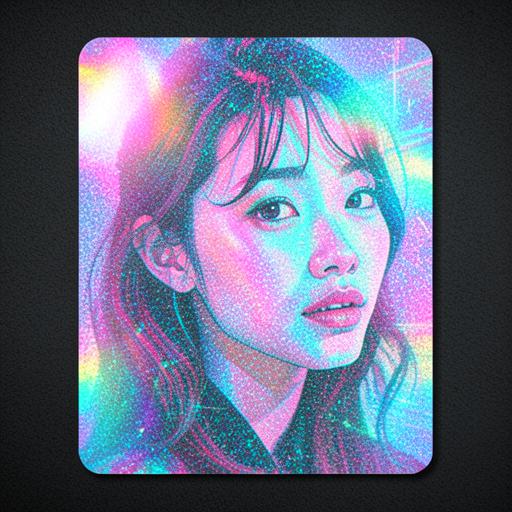} &
        \includegraphics[width=0.12\textwidth]{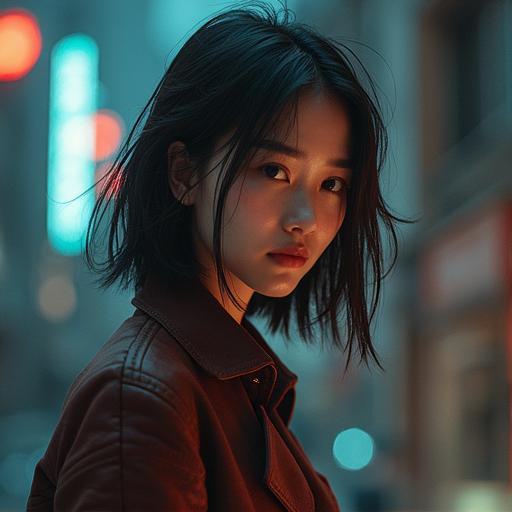} &
        \includegraphics[width=0.12\textwidth]{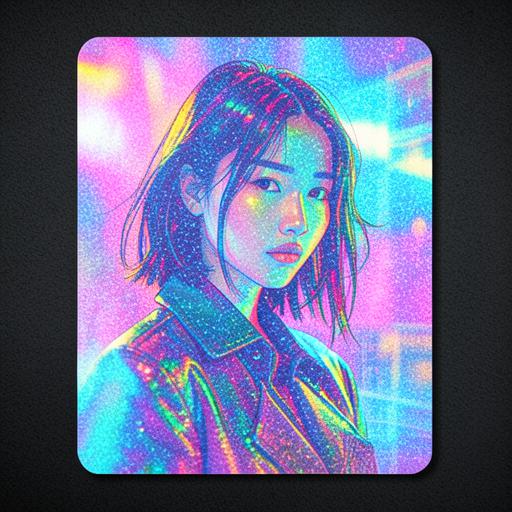} &
        \includegraphics[width=0.12\textwidth]{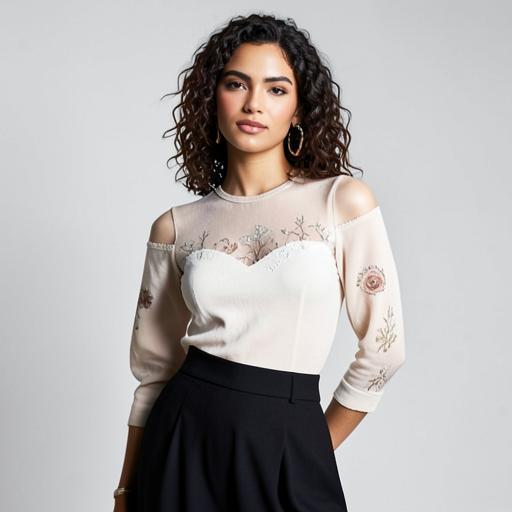} &
        \includegraphics[width=0.12\textwidth]{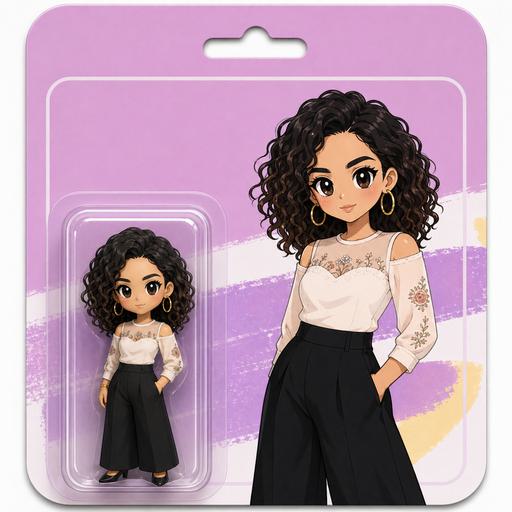} &
        \includegraphics[width=0.12\textwidth]{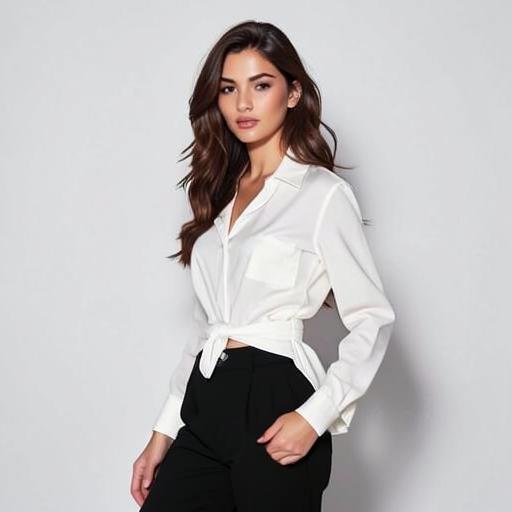} &
        \includegraphics[width=0.12\textwidth]{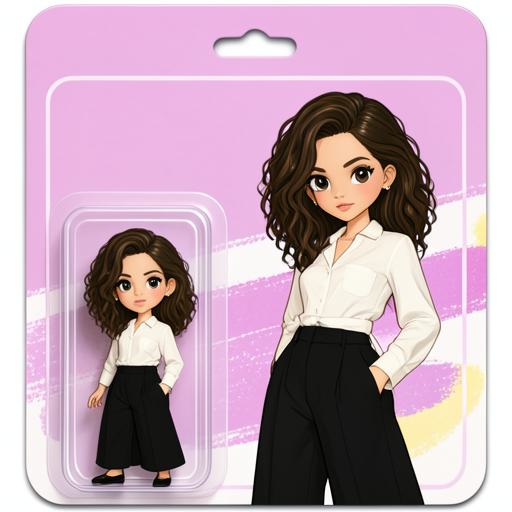} \\

        \includegraphics[width=0.12\textwidth]{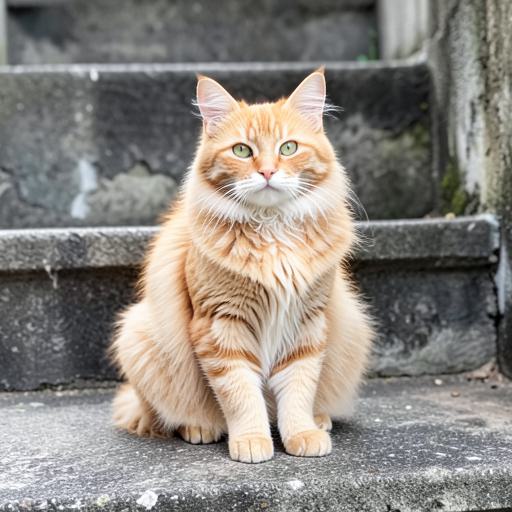} &
        \includegraphics[width=0.12\textwidth]{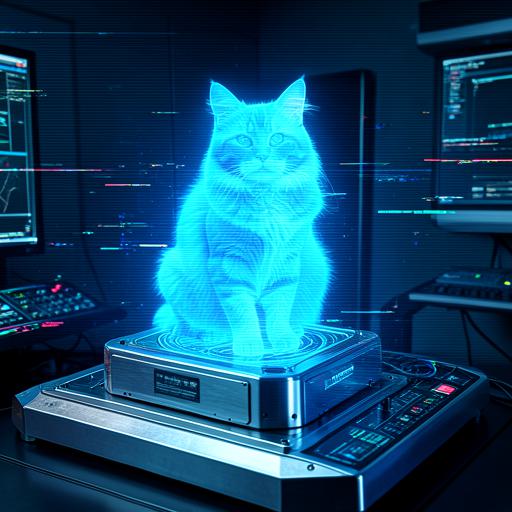} &
        \includegraphics[width=0.12\textwidth]{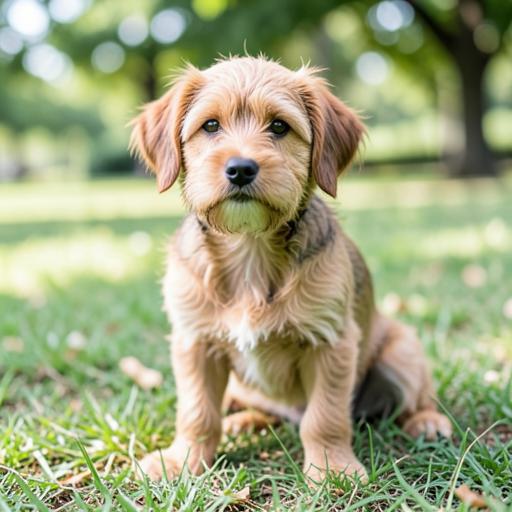} &
        \includegraphics[width=0.12\textwidth]{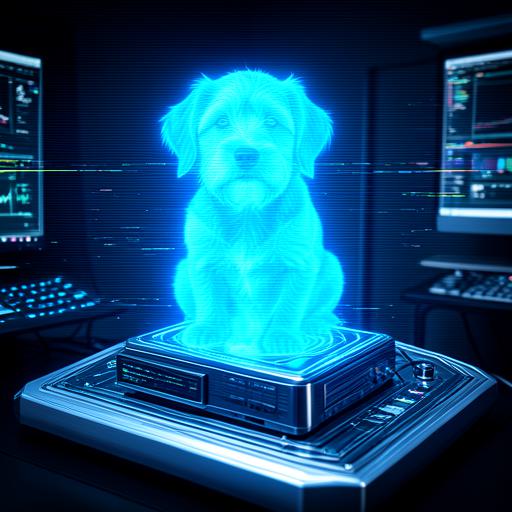} &
        \includegraphics[width=0.12\textwidth]{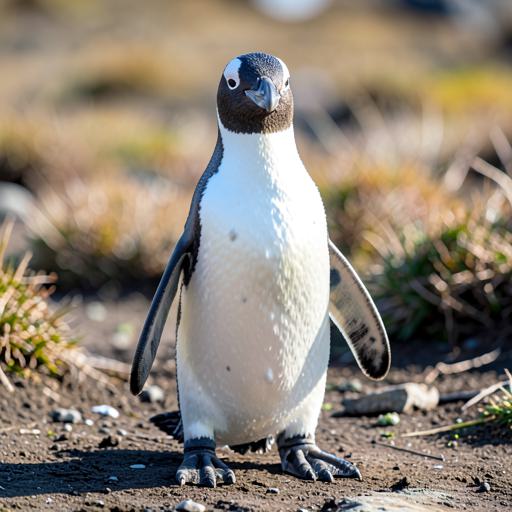} &
        \includegraphics[width=0.12\textwidth]{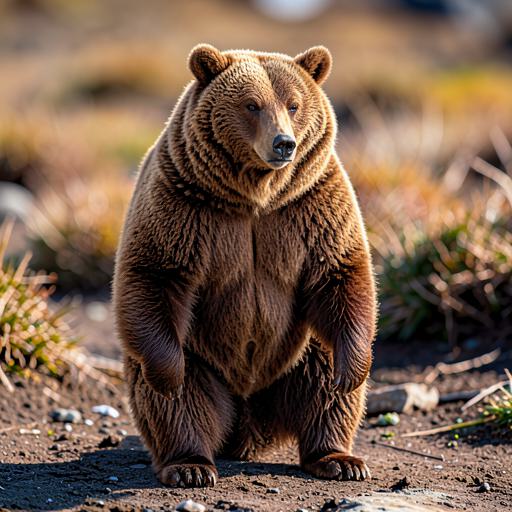} &
        \includegraphics[width=0.12\textwidth]{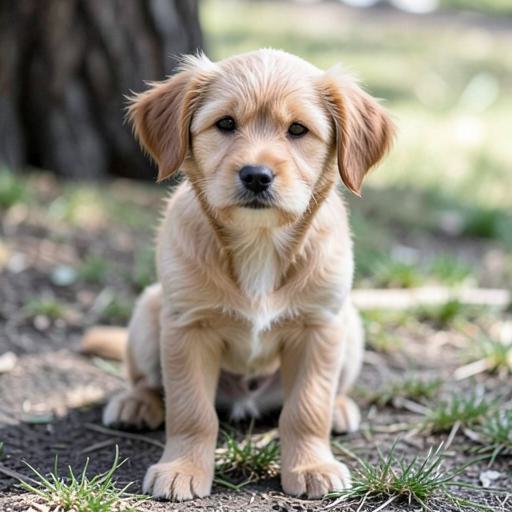} &
        \includegraphics[width=0.12\textwidth]{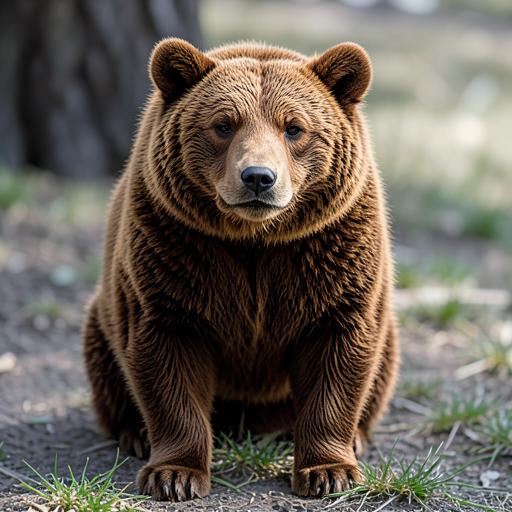} \\

        \includegraphics[width=0.12\textwidth]{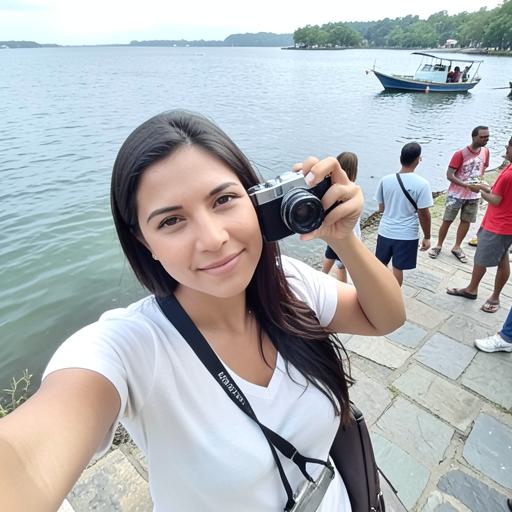} &
        \includegraphics[width=0.12\textwidth]{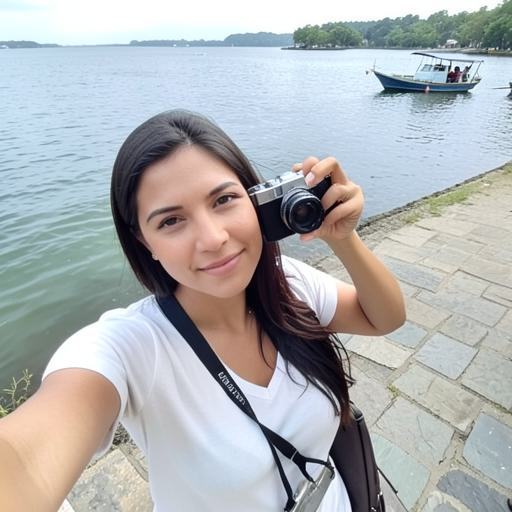} &
        \includegraphics[width=0.12\textwidth]{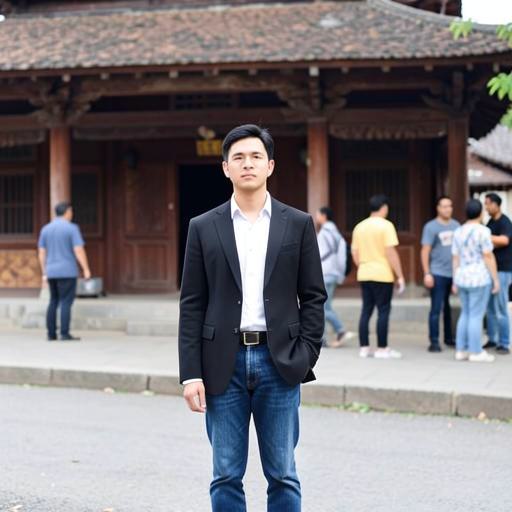} &
        \includegraphics[width=0.12\textwidth]{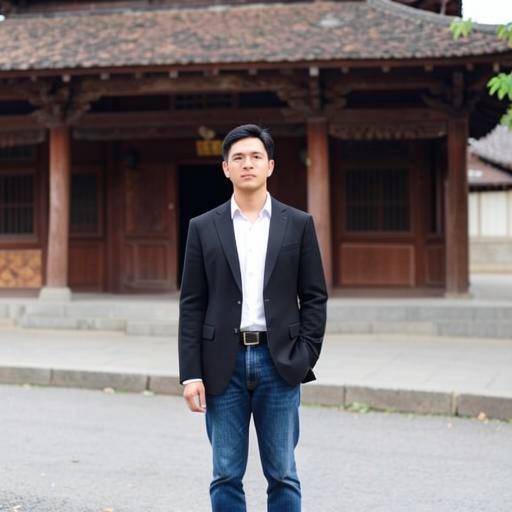} &
        \includegraphics[width=0.12\textwidth]{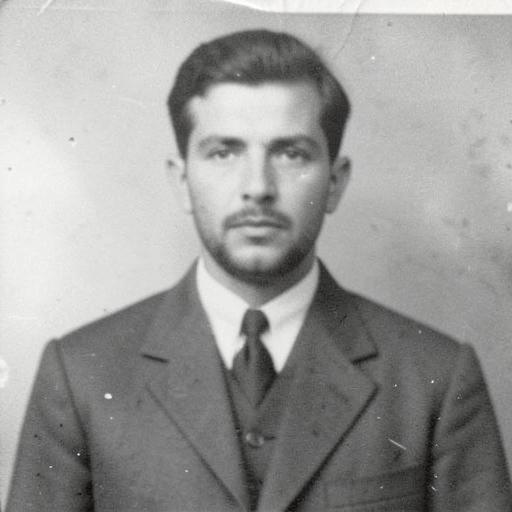} &
        \includegraphics[width=0.12\textwidth]{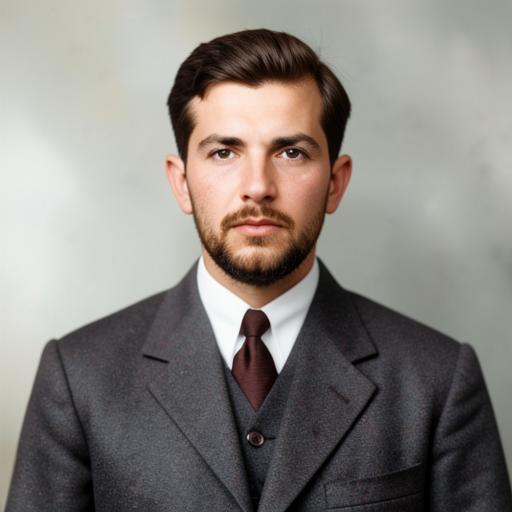} &
        \includegraphics[width=0.12\textwidth]{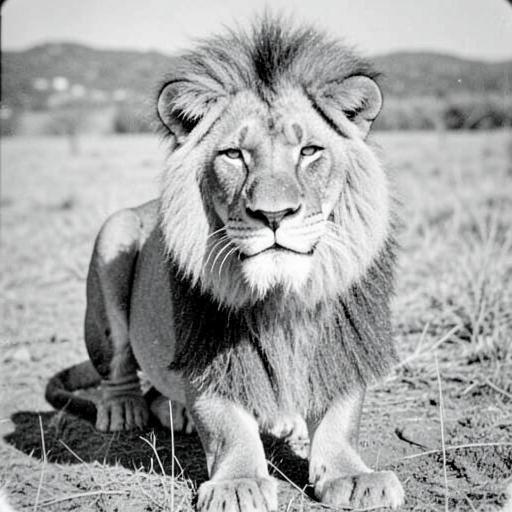} &
        \includegraphics[width=0.12\textwidth]{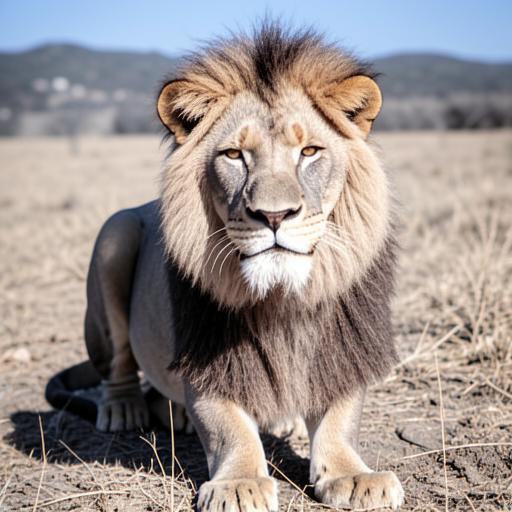} \\

        \includegraphics[width=0.12\textwidth]{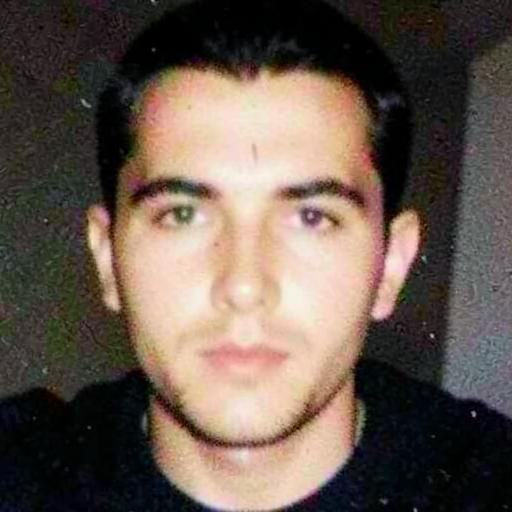} &
        \includegraphics[width=0.12\textwidth]{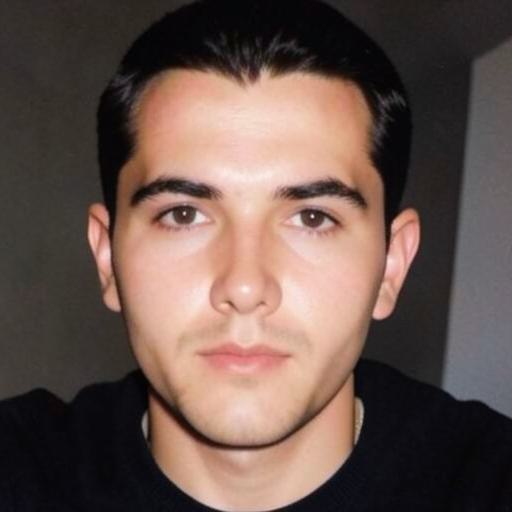} &
        \includegraphics[width=0.12\textwidth]{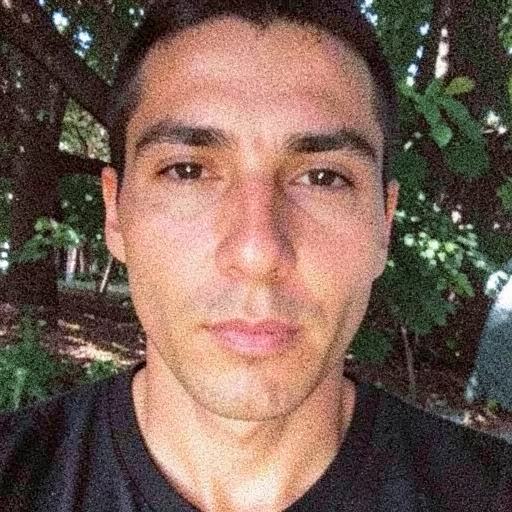} &
        \includegraphics[width=0.12\textwidth]{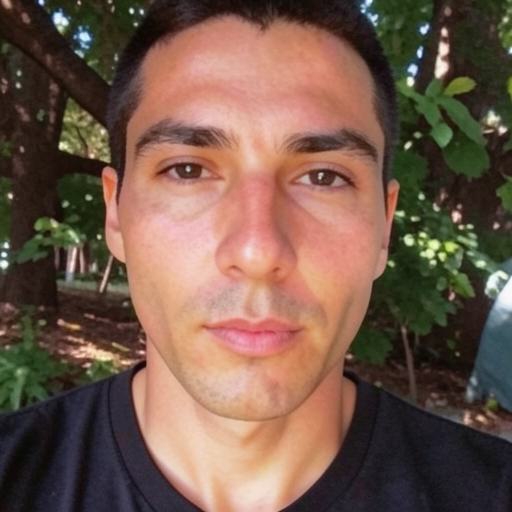} &
        \includegraphics[width=0.12\textwidth]{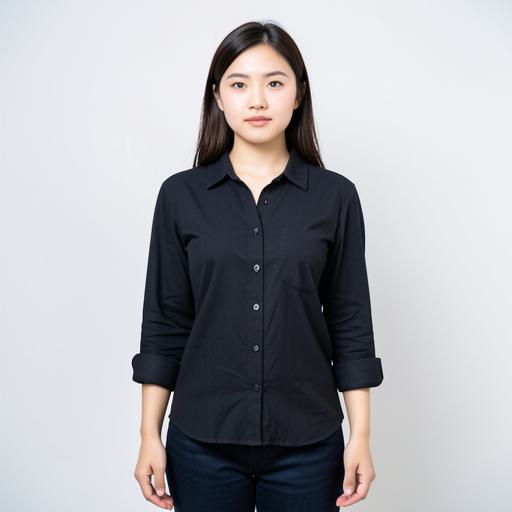} &
        \includegraphics[width=0.12\textwidth]{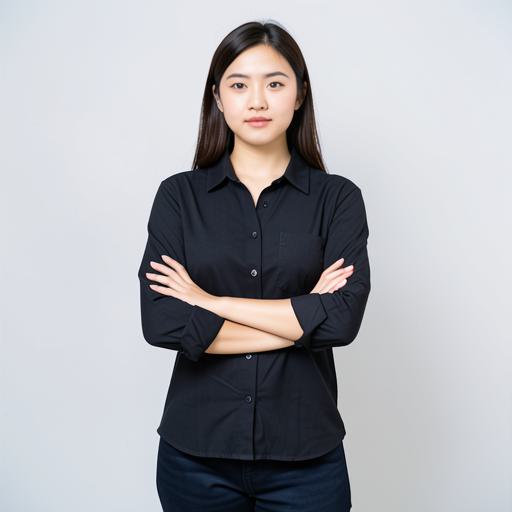} &
        \includegraphics[width=0.12\textwidth]{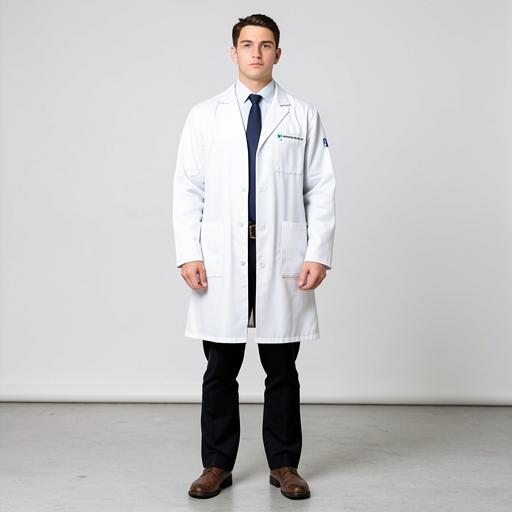} &
        \includegraphics[width=0.12\textwidth]{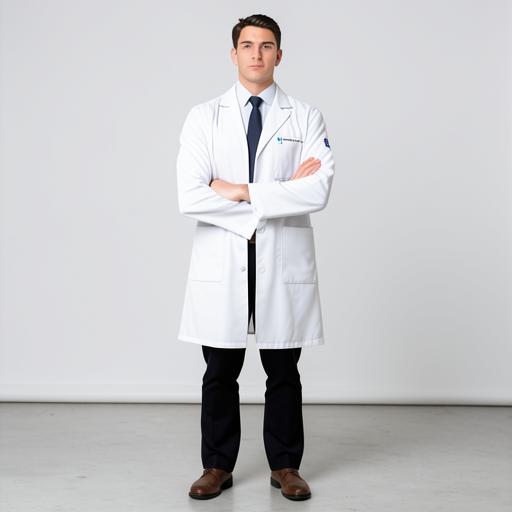} \\

        \includegraphics[width=0.12\textwidth]{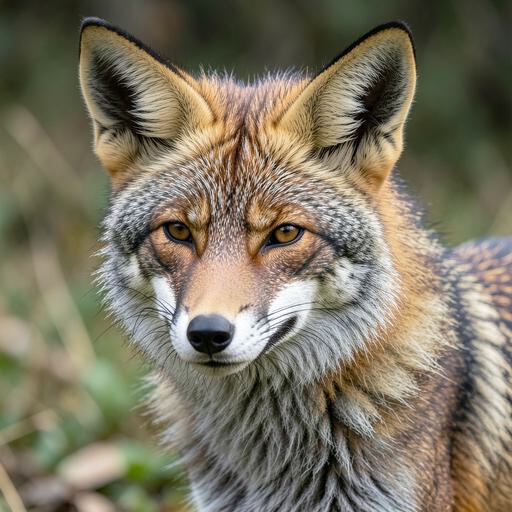} &
        \includegraphics[width=0.12\textwidth]{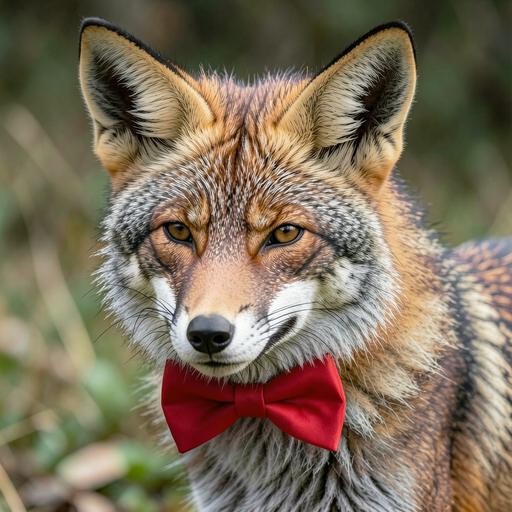} &
        \includegraphics[width=0.12\textwidth]{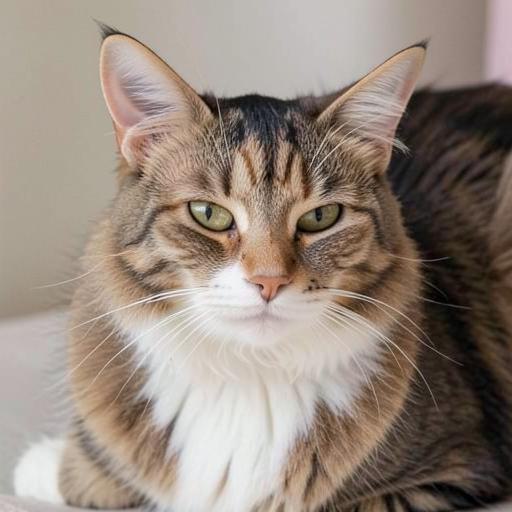} &
        \includegraphics[width=0.12\textwidth]{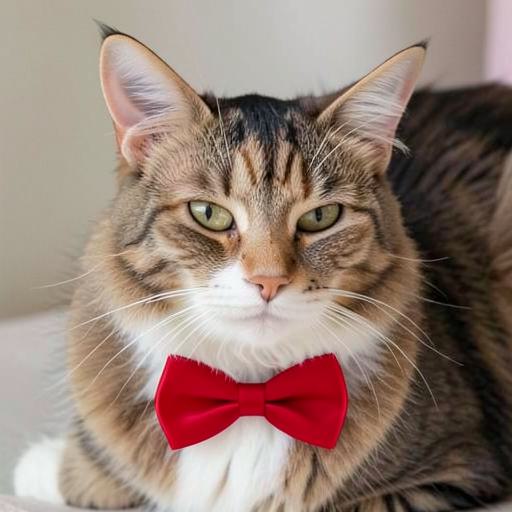} &
        \includegraphics[width=0.12\textwidth]{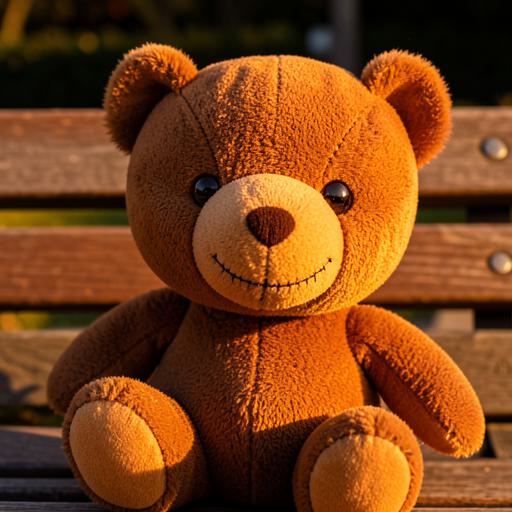} &
        \includegraphics[width=0.12\textwidth]{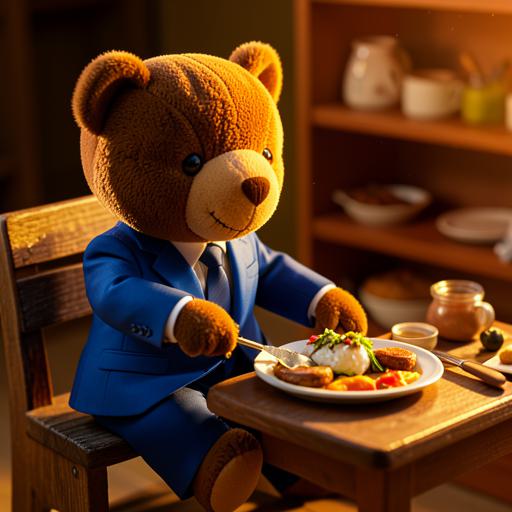} &
        \includegraphics[width=0.12\textwidth]{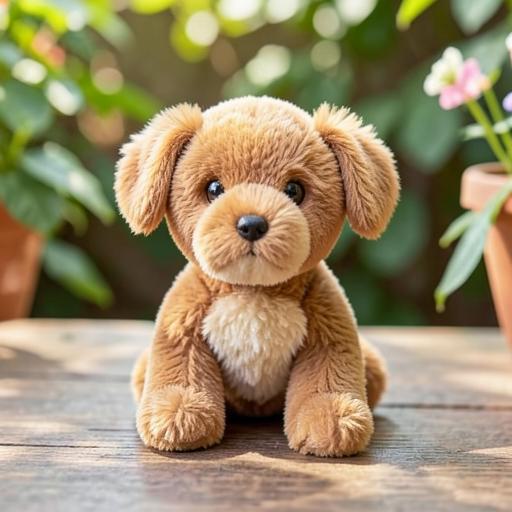} &
        \includegraphics[width=0.12\textwidth]{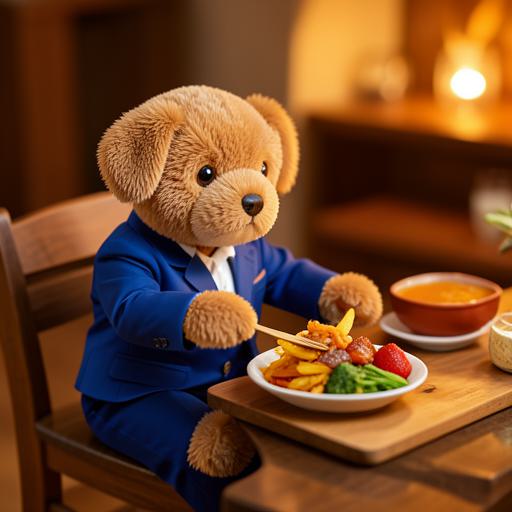} \\

        \includegraphics[width=0.12\textwidth]{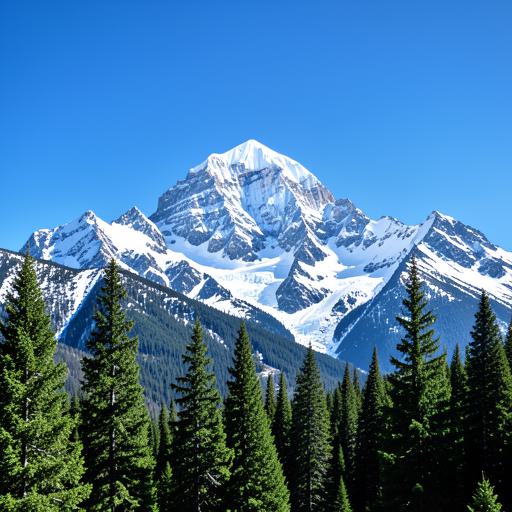} &
        \includegraphics[width=0.12\textwidth]{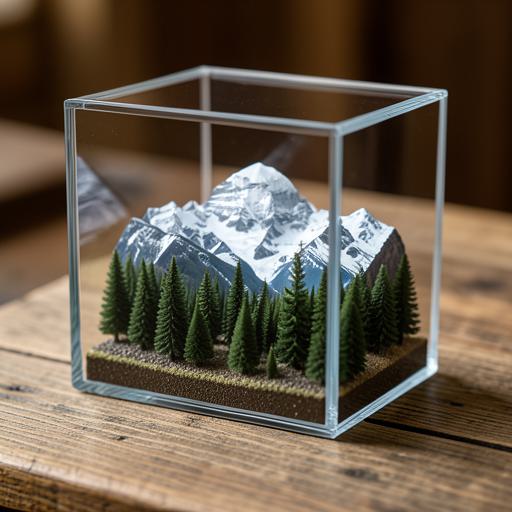} &
        \includegraphics[width=0.12\textwidth]{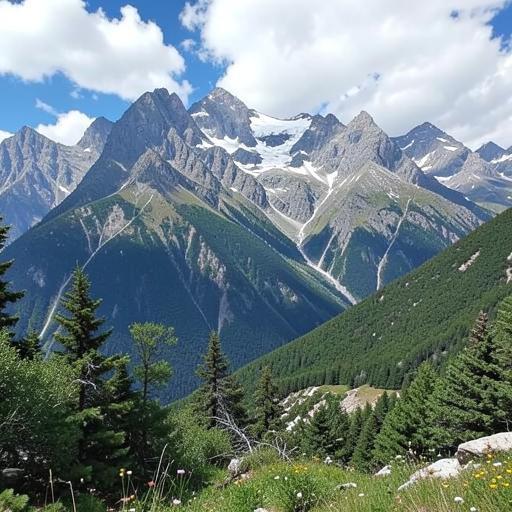} &
        \includegraphics[width=0.12\textwidth]{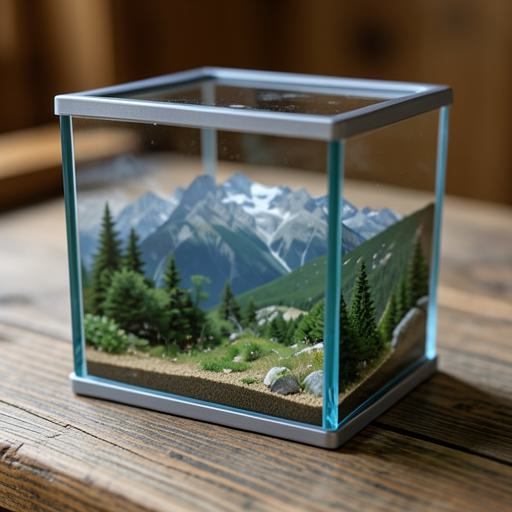} &
        \includegraphics[width=0.12\textwidth]{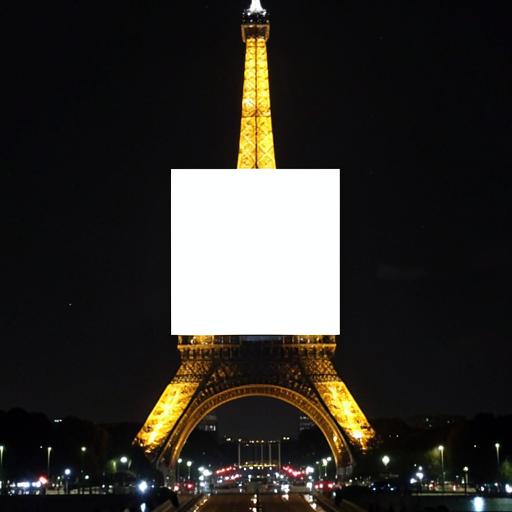} &
        \includegraphics[width=0.12\textwidth]{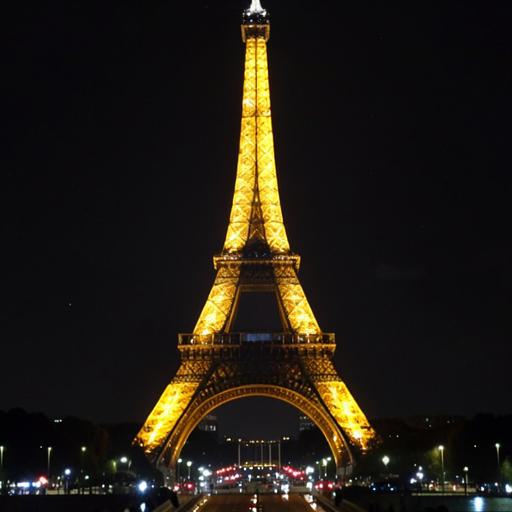} &
        \includegraphics[width=0.12\textwidth]{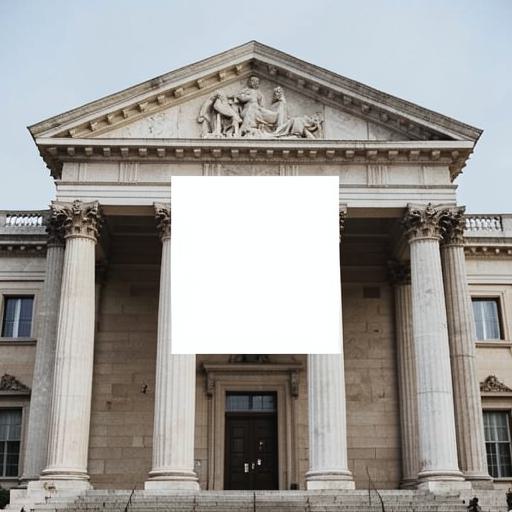} &
        \includegraphics[width=0.12\textwidth]{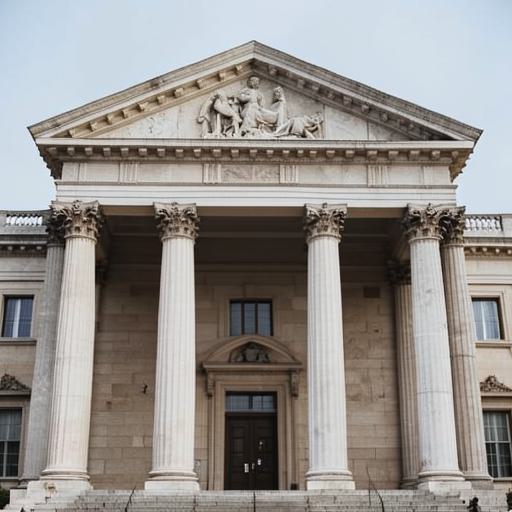} \\

        \includegraphics[width=0.12\textwidth]{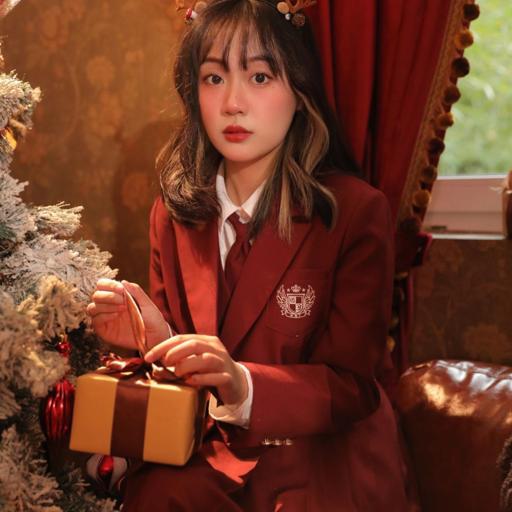} &
        \includegraphics[width=0.12\textwidth]{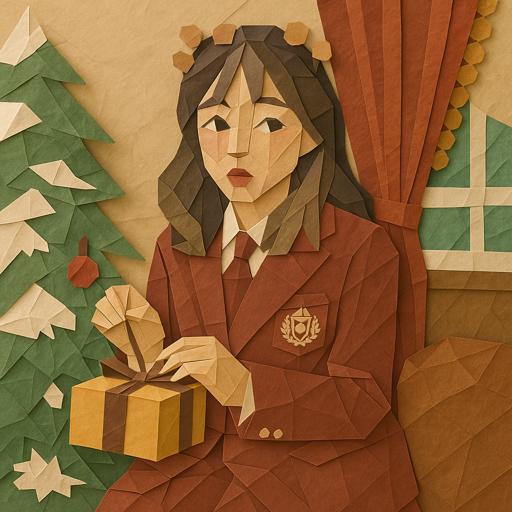} &
        \includegraphics[width=0.12\textwidth]{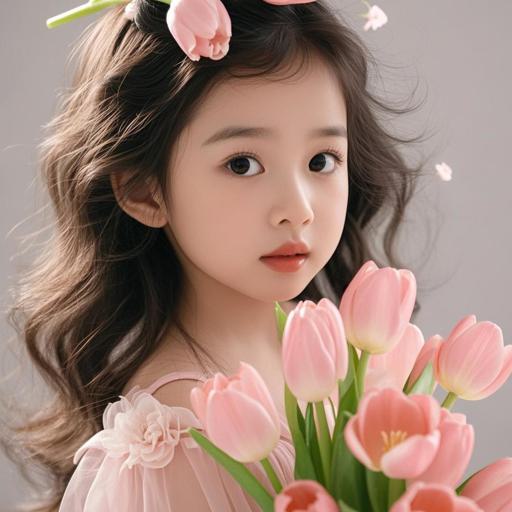} &
        \includegraphics[width=0.12\textwidth]{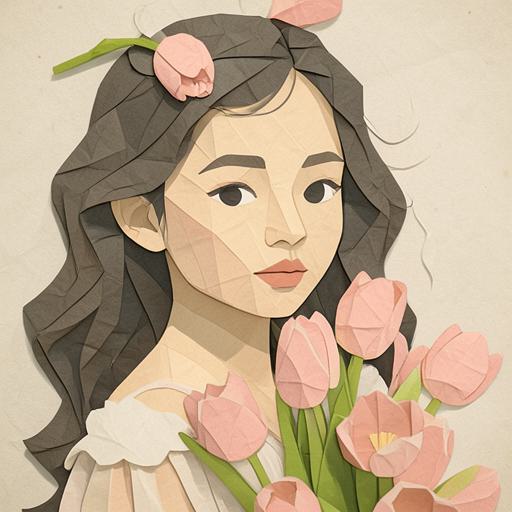} &
        \includegraphics[width=0.12\textwidth]{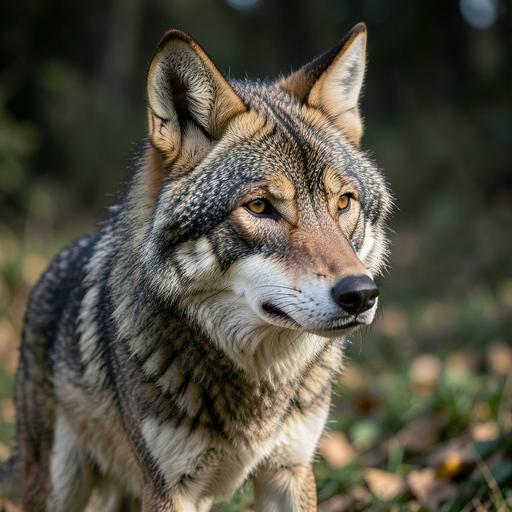} &
        \includegraphics[width=0.12\textwidth]{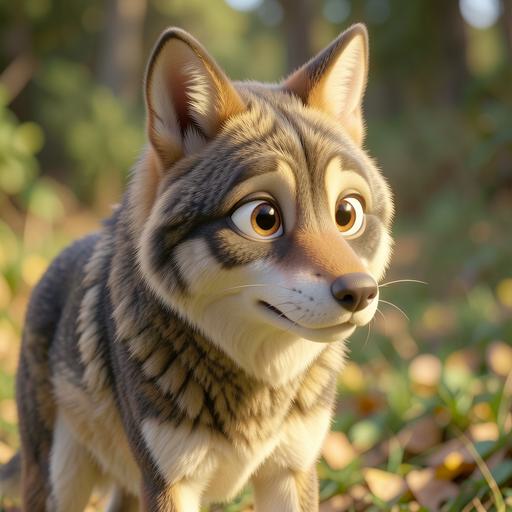} &
        \includegraphics[width=0.12\textwidth]{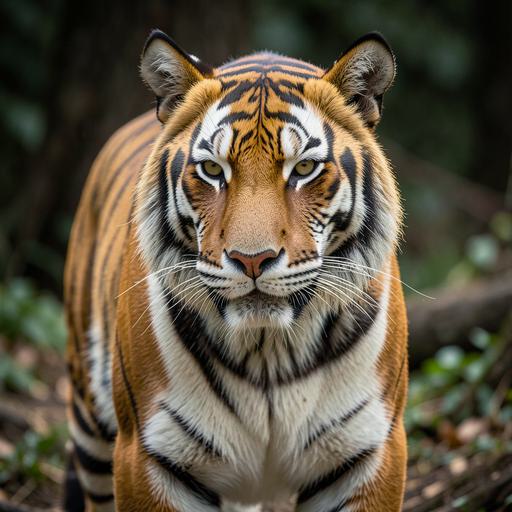} &
        \includegraphics[width=0.12\textwidth]{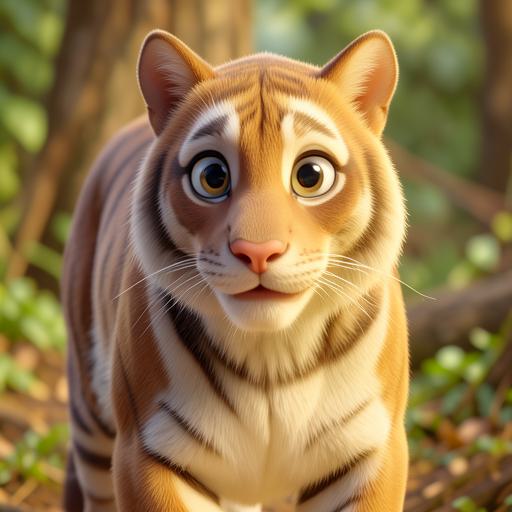} \\

        \includegraphics[width=0.12\textwidth]{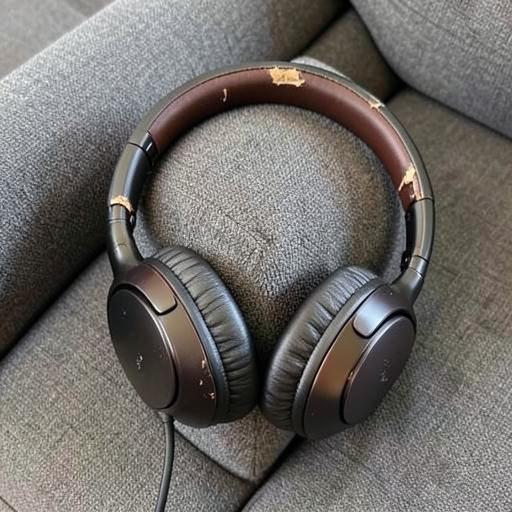} &
        \includegraphics[width=0.12\textwidth]{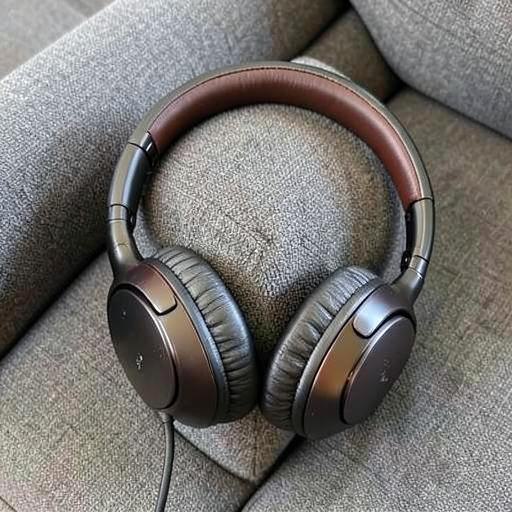} &
        \includegraphics[width=0.12\textwidth]{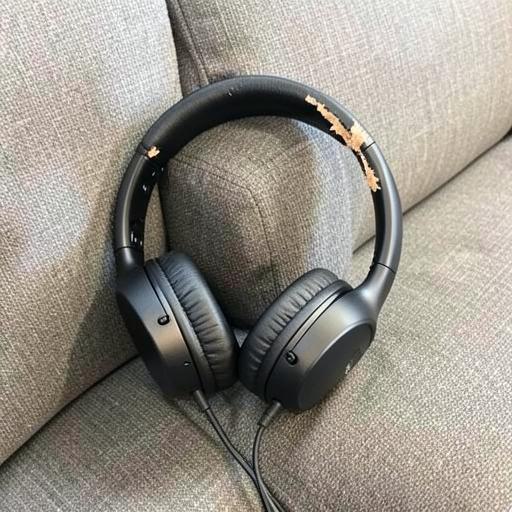} &
        \includegraphics[width=0.12\textwidth]{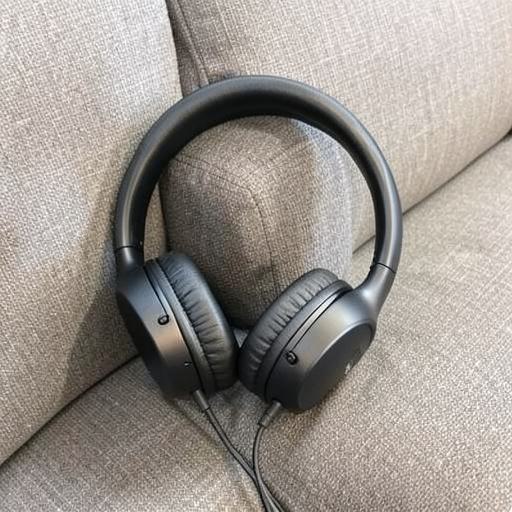} &
        \includegraphics[width=0.12\textwidth]{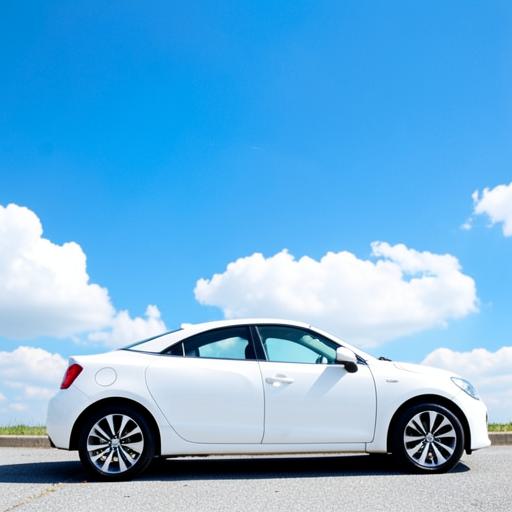} &
        \includegraphics[width=0.12\textwidth]{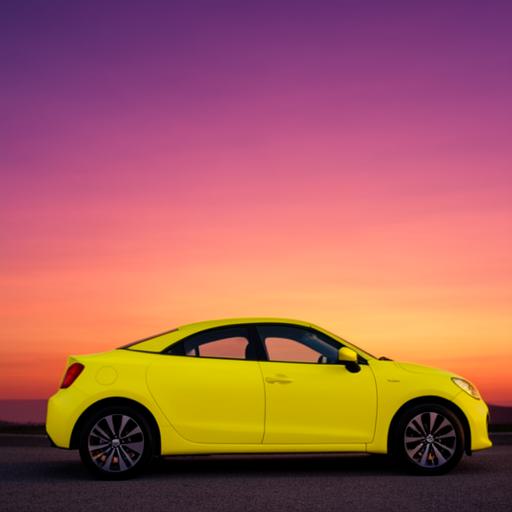} &
        \includegraphics[width=0.12\textwidth]{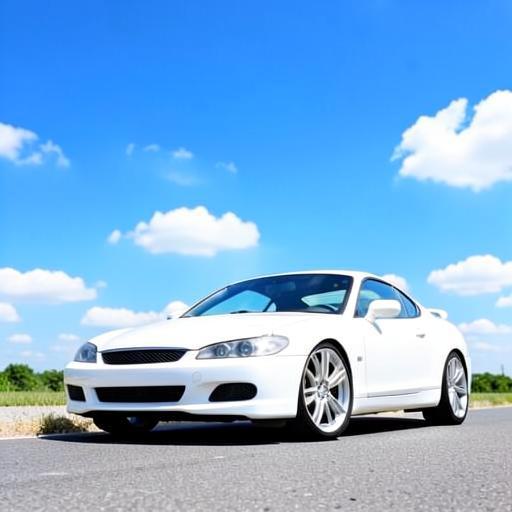} &
        \includegraphics[width=0.12\textwidth]{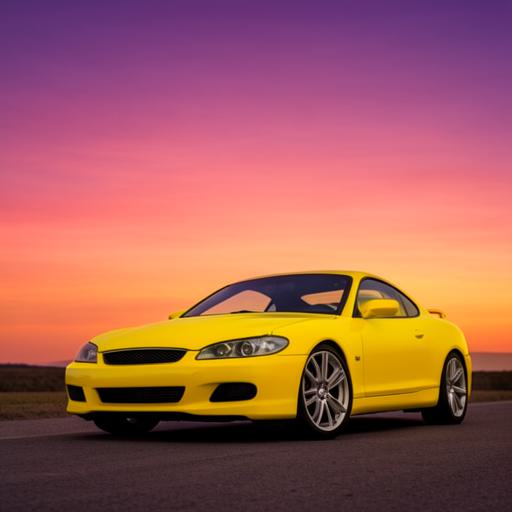} \\

        \includegraphics[width=0.12\textwidth]{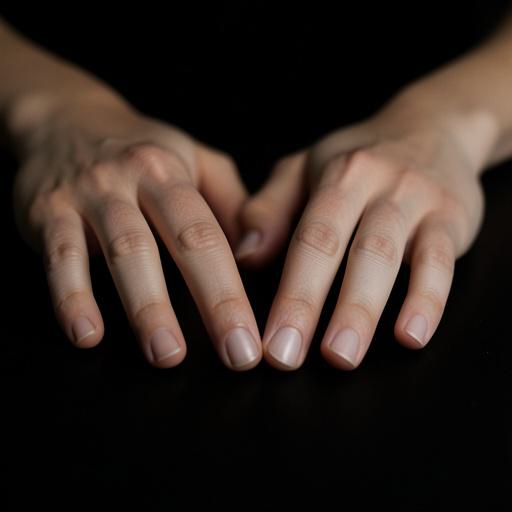} &
        \includegraphics[width=0.12\textwidth]{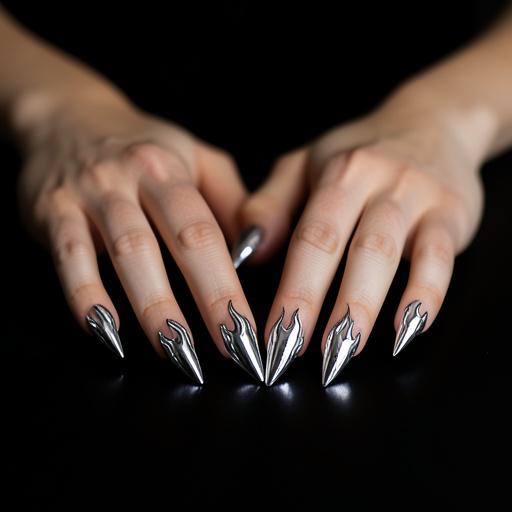} &
        \includegraphics[width=0.12\textwidth]{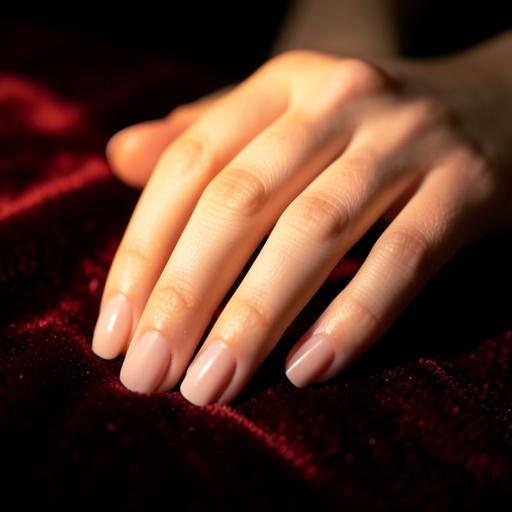} &
        \includegraphics[width=0.12\textwidth]{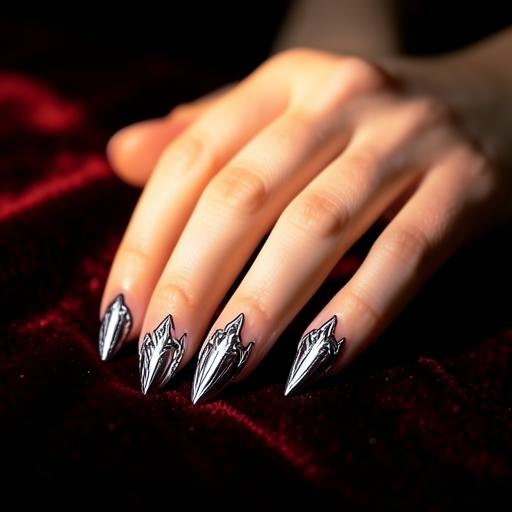} &
        \includegraphics[width=0.12\textwidth]{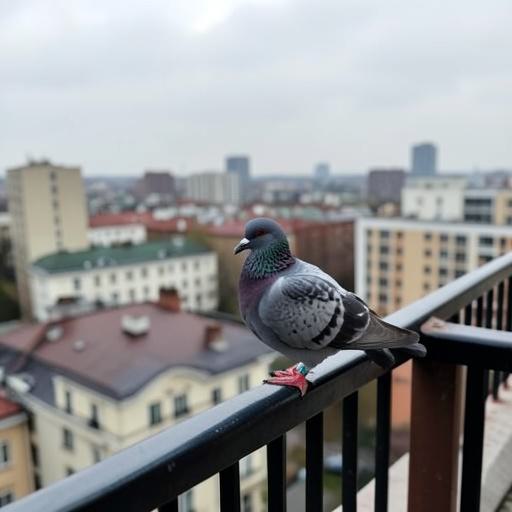} &
        \includegraphics[width=0.12\textwidth]{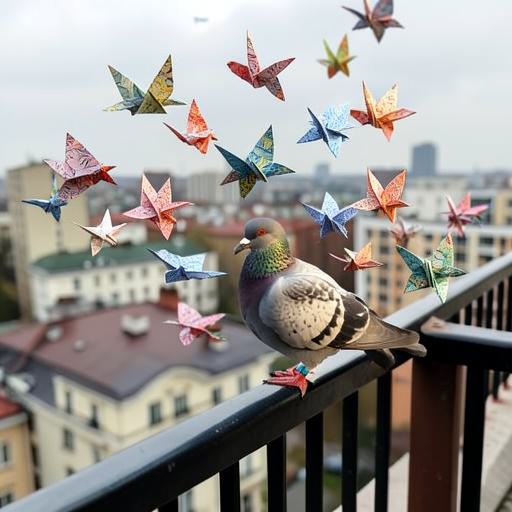} &
        \includegraphics[width=0.12\textwidth]{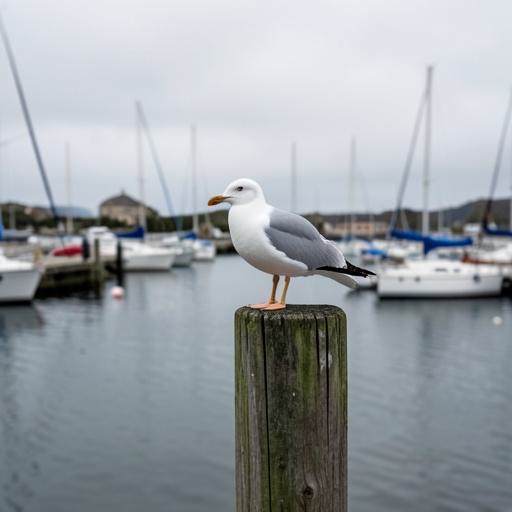} &
        \includegraphics[width=0.12\textwidth]{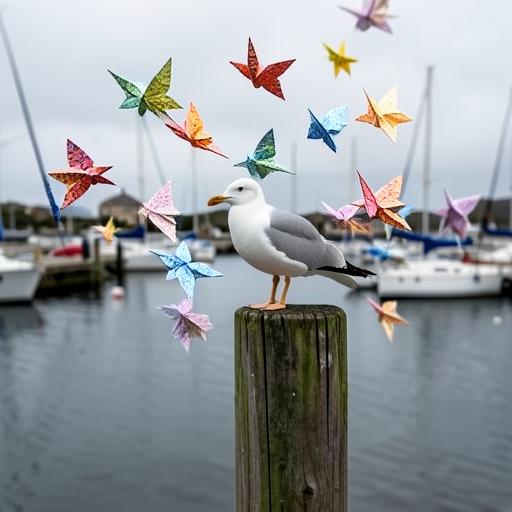} \\

        \includegraphics[width=0.12\textwidth]{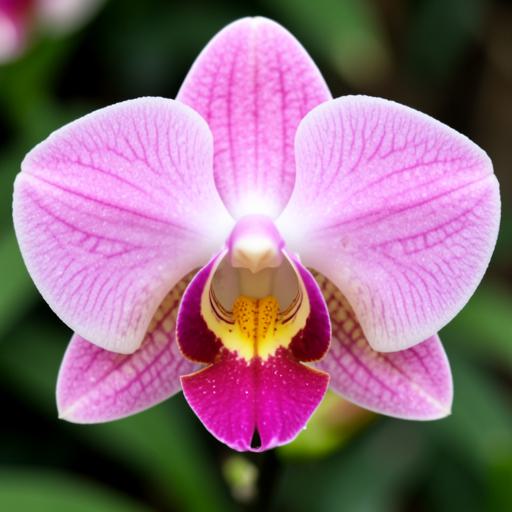} &
        \includegraphics[width=0.12\textwidth]{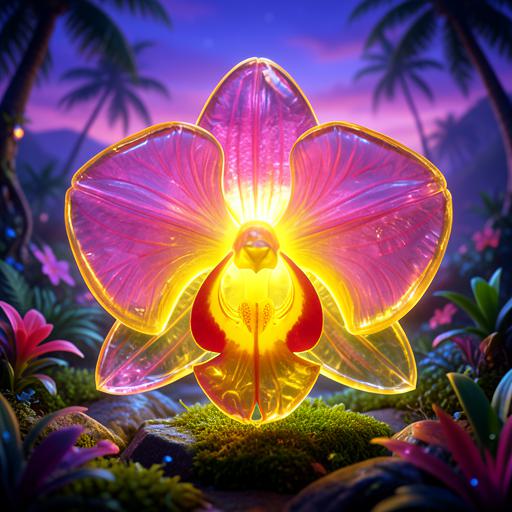} &
        \includegraphics[width=0.12\textwidth]{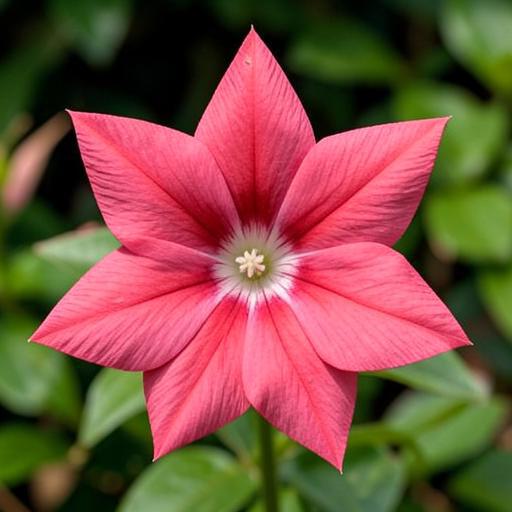} &
        \includegraphics[width=0.12\textwidth]{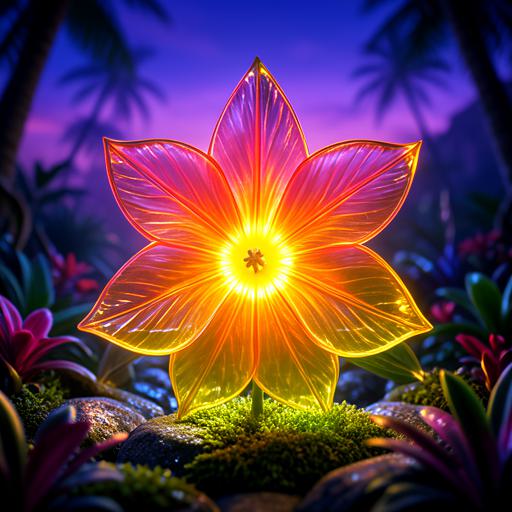} &
        \includegraphics[width=0.12\textwidth]{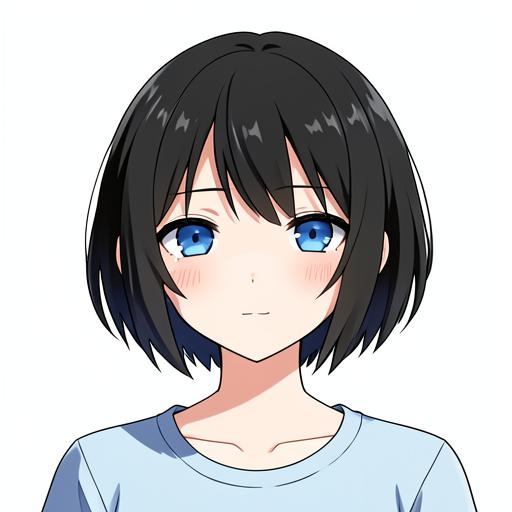} &
        \includegraphics[width=0.12\textwidth]{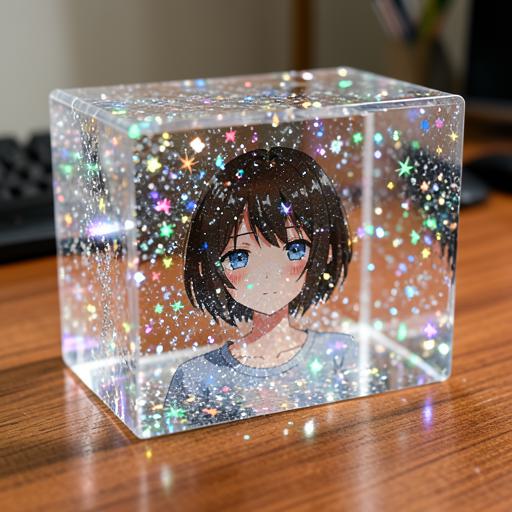} &
        \includegraphics[width=0.12\textwidth]{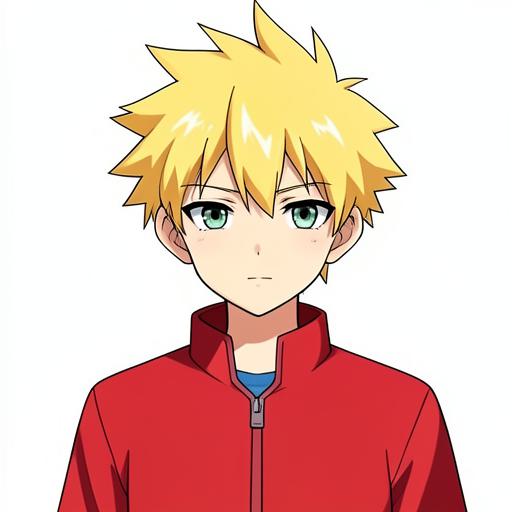} &
        \includegraphics[width=0.12\textwidth]{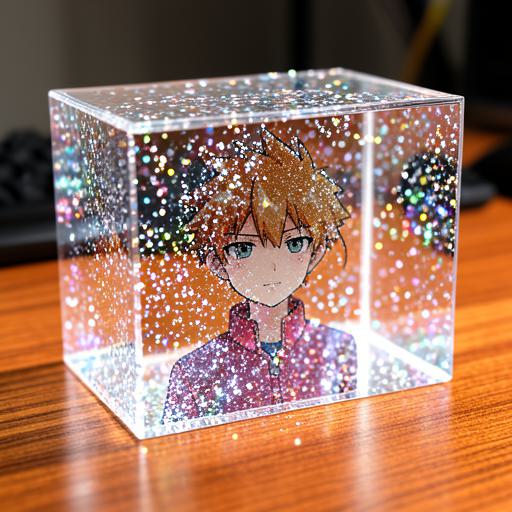} \\

    \end{tabular}
    }
    \caption{\textbf{Additional editing results produced by Delta-Adapter.} Given an exemplar pair, our method infers the underlying transformation and faithfully applies it to unseen query images.}
\label{fig:appendix_more_ours}
\end{figure*}

\begin{figure*}[t]
    \centering
    \renewcommand{\arraystretch}{0.3}
    \setlength{\tabcolsep}{0.6pt}

    {\footnotesize
    \begin{tabular}{c c c @{\hspace{0.05cm}} | @{\hspace{0.05cm}}  c c c c }

        \multicolumn{1}{c}{\normalsize Source ($a$)} &
        \multicolumn{1}{c}{\normalsize Target ($a'$)} &
        \multicolumn{1}{c}{\normalsize Query ($b$)} &
        \multicolumn{1}{c}{\normalsize Edit Transfer} &
        \multicolumn{1}{c}{\normalsize LoRWeB} &
        \multicolumn{1}{c}{\begin{tabular}{c}\normalsize Relation\\\normalsize Adapter\end{tabular}} &
        \multicolumn{1}{c}{\normalsize Ours} \\

        \includegraphics[width=0.14\textwidth]{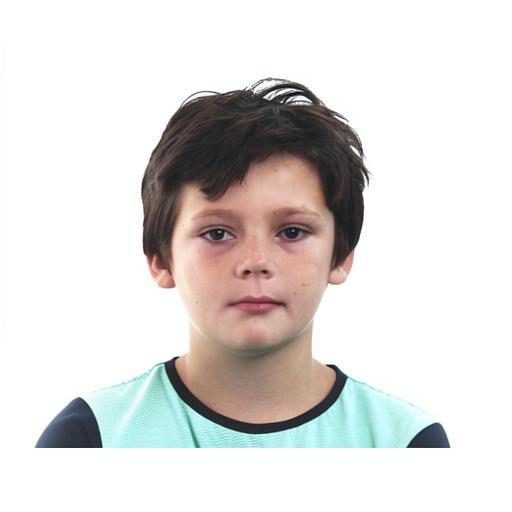} &
        \includegraphics[width=0.14\textwidth]{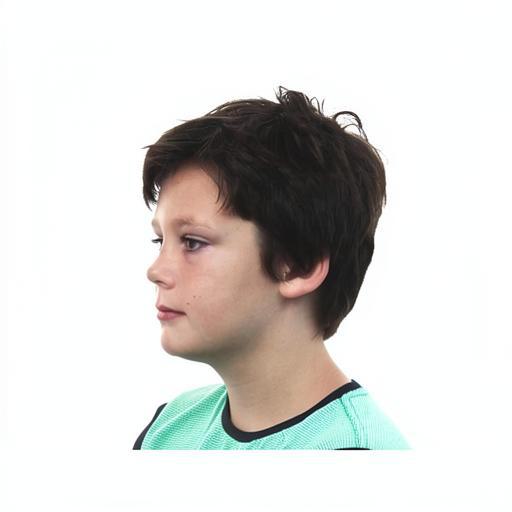} &
        \includegraphics[width=0.14\textwidth]{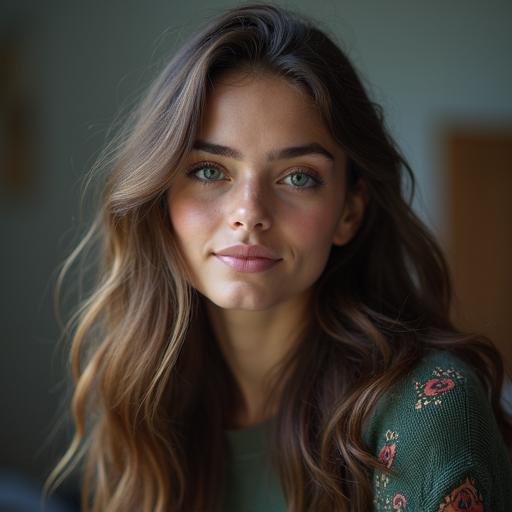} &
        \includegraphics[width=0.14\textwidth]{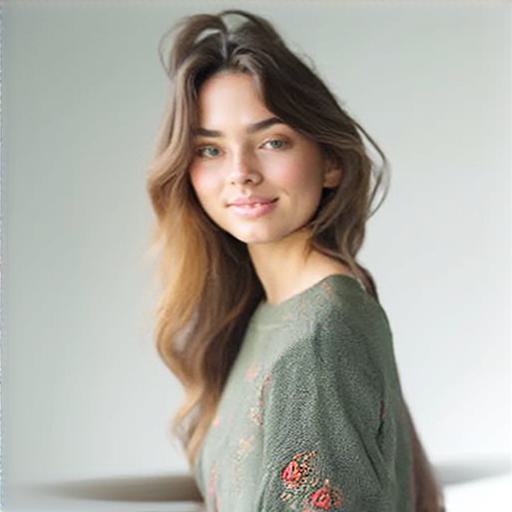} &
        \includegraphics[width=0.14\textwidth]{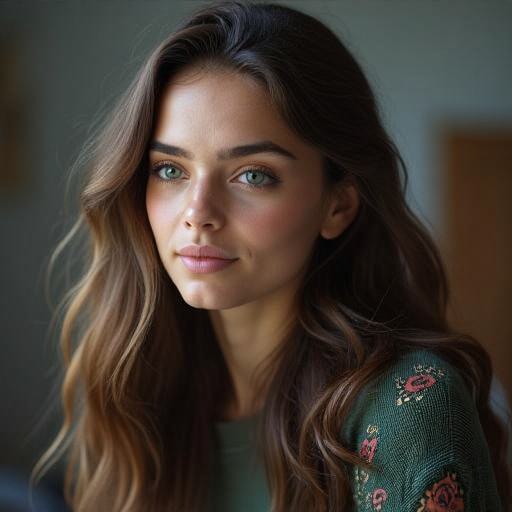} &
        \includegraphics[width=0.14\textwidth]{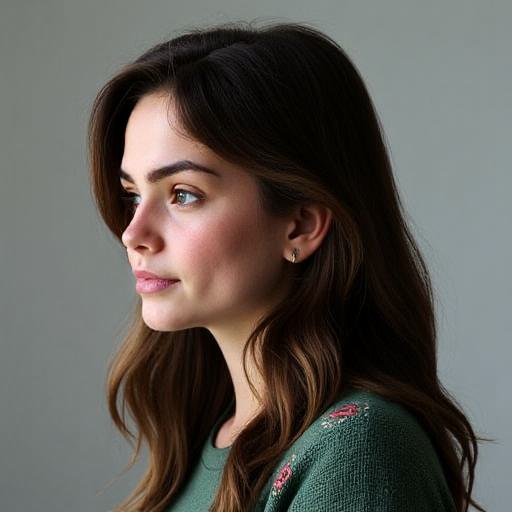} &
        \includegraphics[width=0.14\textwidth]{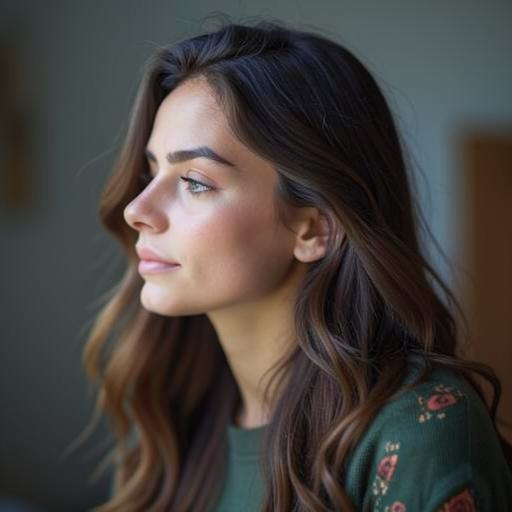} \\

        \includegraphics[width=0.14\textwidth]{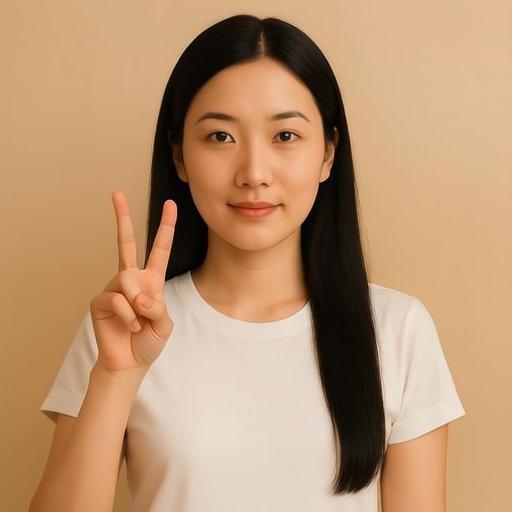} &
        \includegraphics[width=0.14\textwidth]{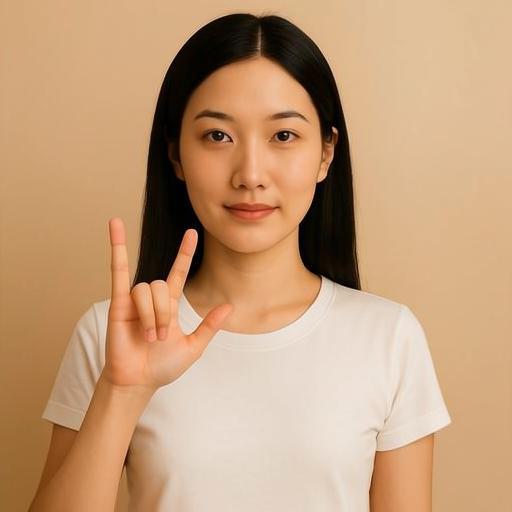} &
        \includegraphics[width=0.14\textwidth]{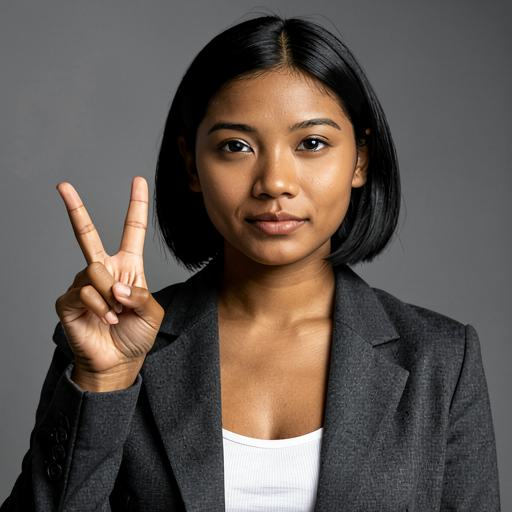} &
        \includegraphics[width=0.14\textwidth]{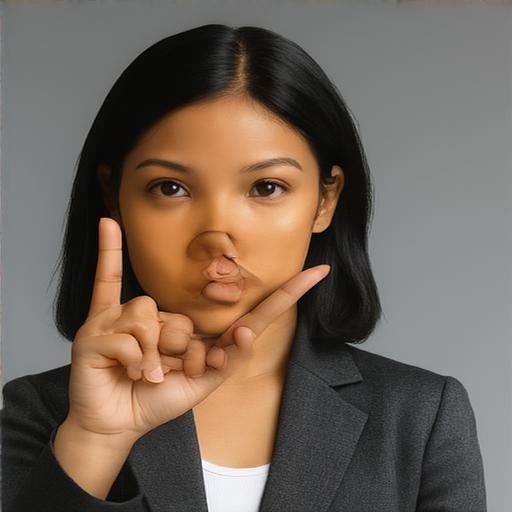} &
        \includegraphics[width=0.14\textwidth]{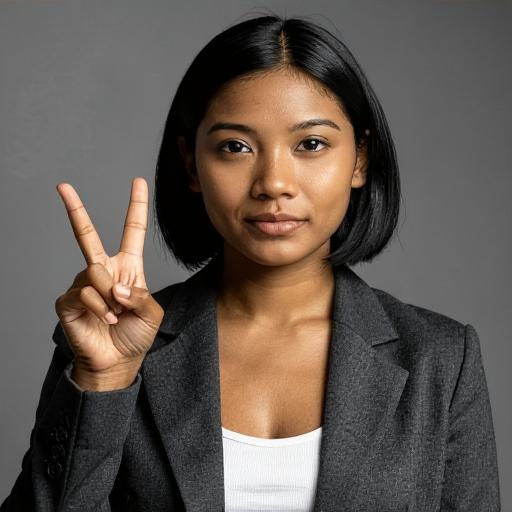} &
        \includegraphics[width=0.14\textwidth]{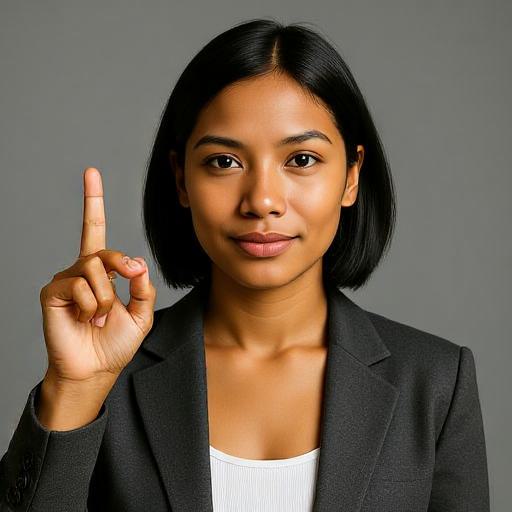} &
        \includegraphics[width=0.14\textwidth]{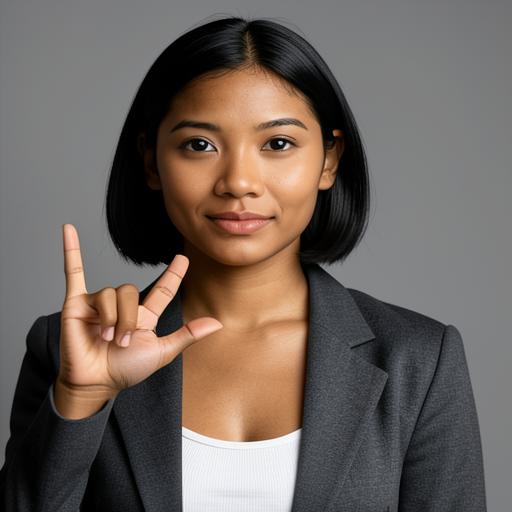} \\

        \includegraphics[width=0.14\textwidth]{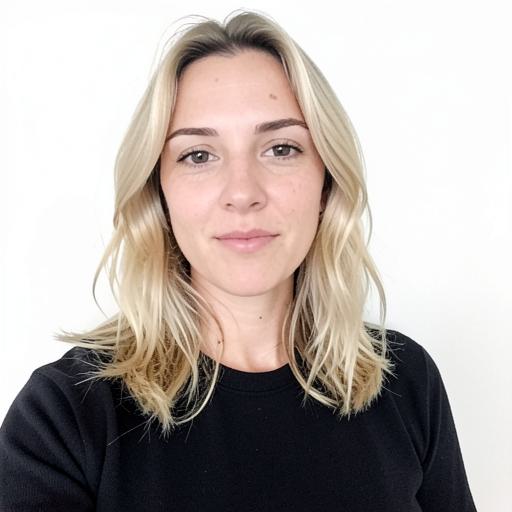} &
        \includegraphics[width=0.14\textwidth]{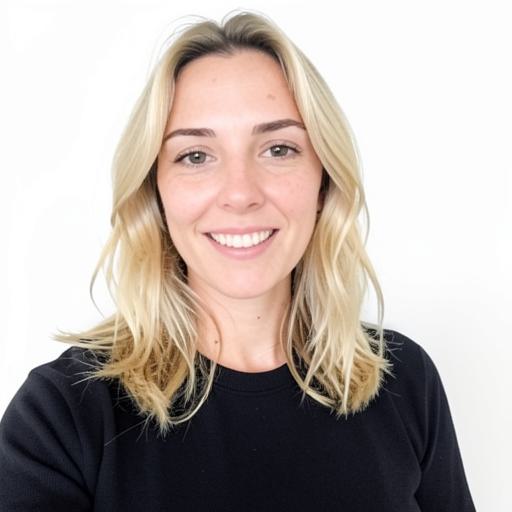} &
        \includegraphics[width=0.14\textwidth]{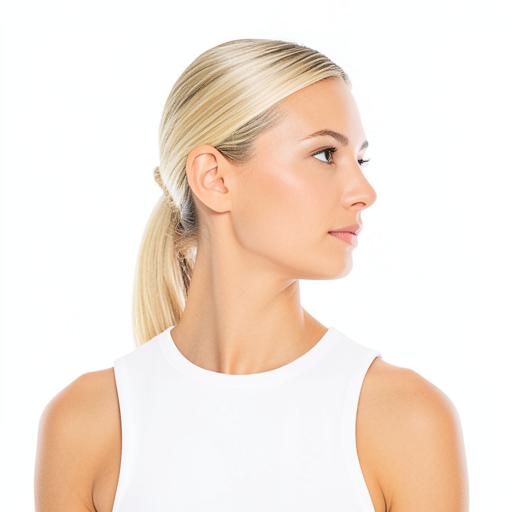} &
        \includegraphics[width=0.14\textwidth]{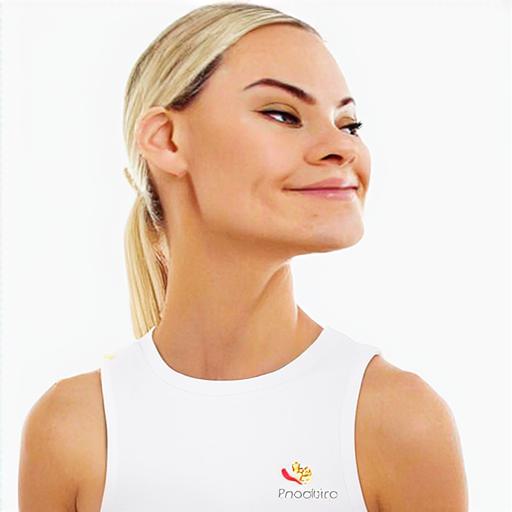} &
        \includegraphics[width=0.14\textwidth]{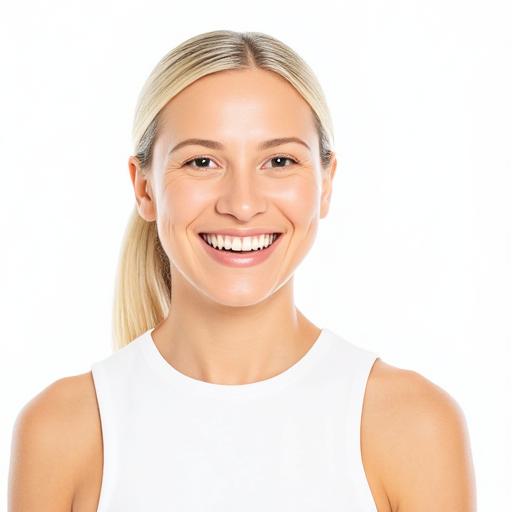} &
        \includegraphics[width=0.14\textwidth]{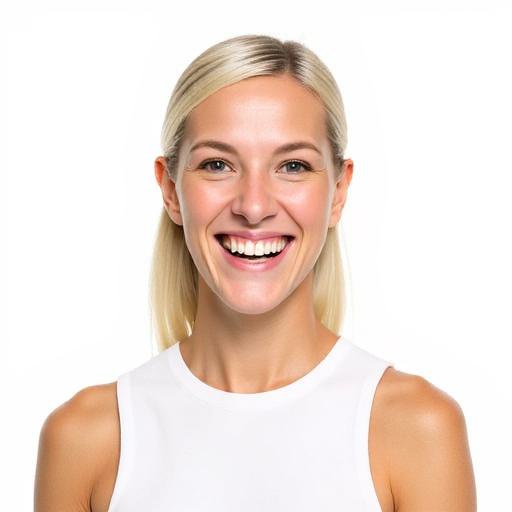} &
        \includegraphics[width=0.14\textwidth]{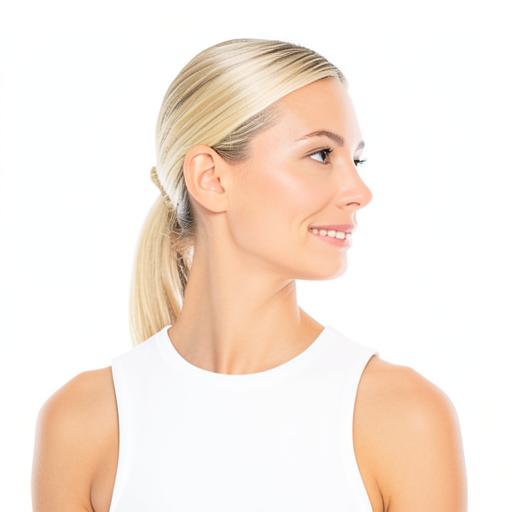} \\

        \includegraphics[width=0.14\textwidth]{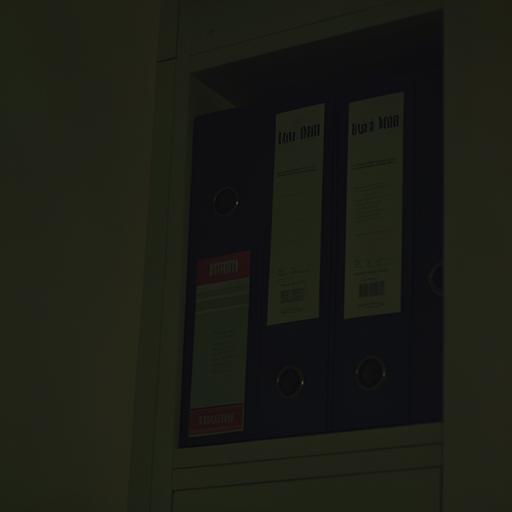} &
        \includegraphics[width=0.14\textwidth]{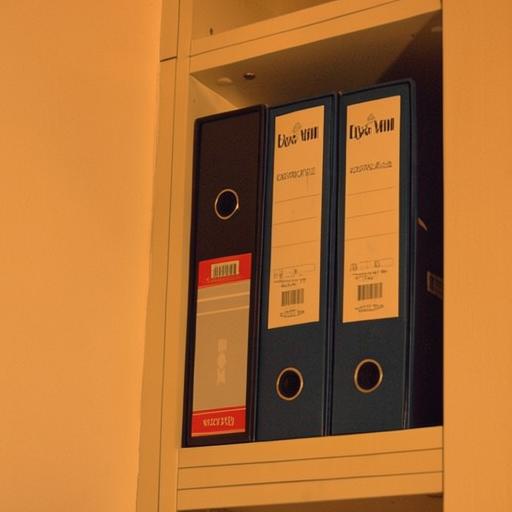} &
        \includegraphics[width=0.14\textwidth]{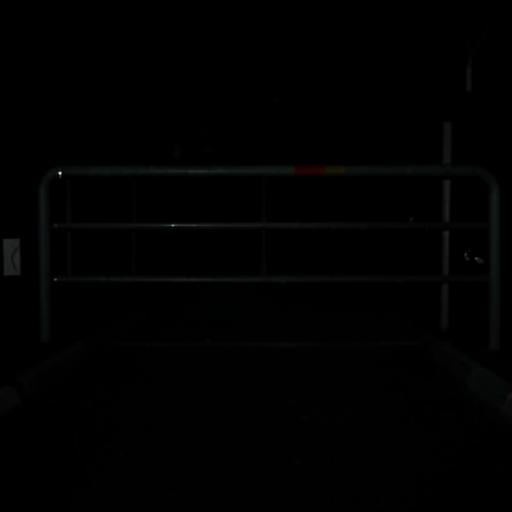} &
        \includegraphics[width=0.14\textwidth]{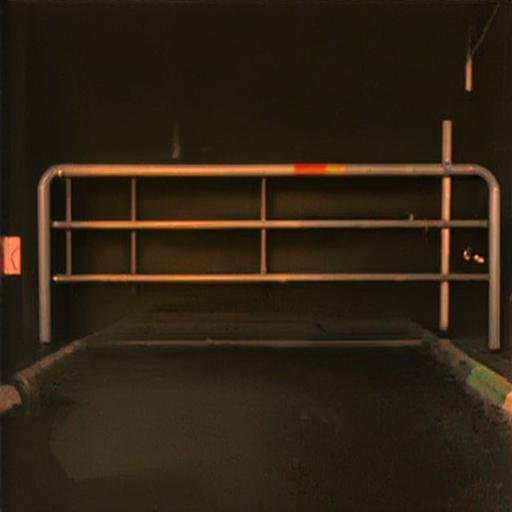} &
        \includegraphics[width=0.14\textwidth]{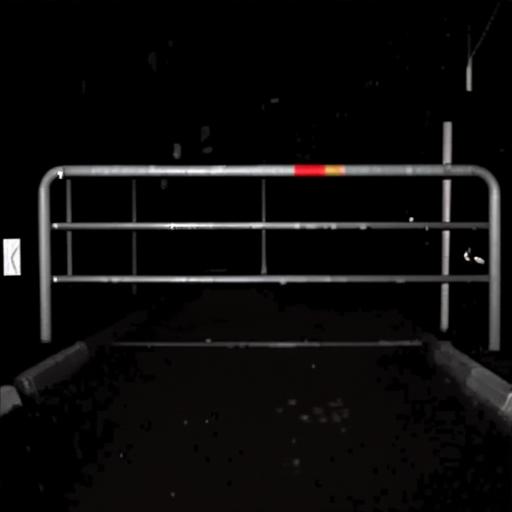} &
        \includegraphics[width=0.14\textwidth]{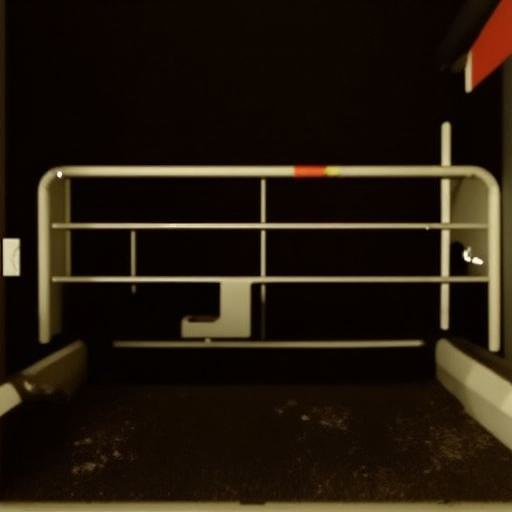} &
        \includegraphics[width=0.14\textwidth]{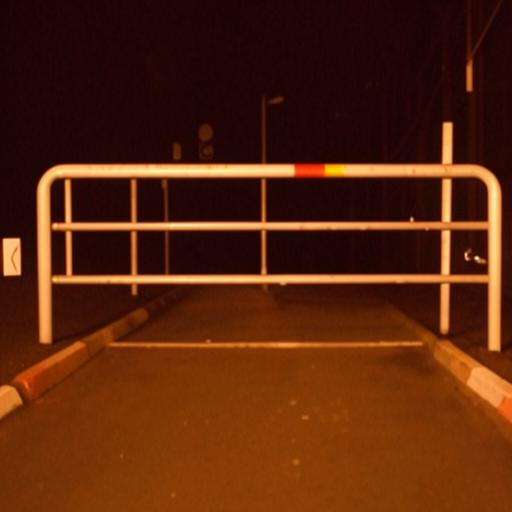} \\

        \includegraphics[width=0.14\textwidth]{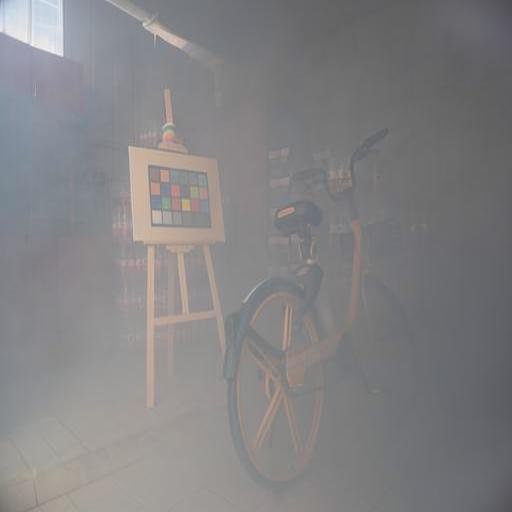} &
        \includegraphics[width=0.14\textwidth]{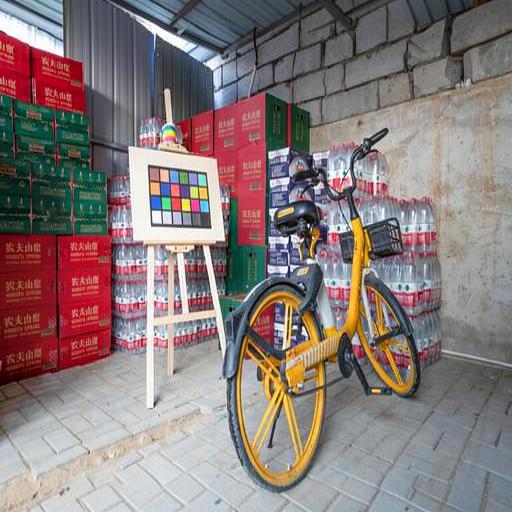} &
        \includegraphics[width=0.14\textwidth]{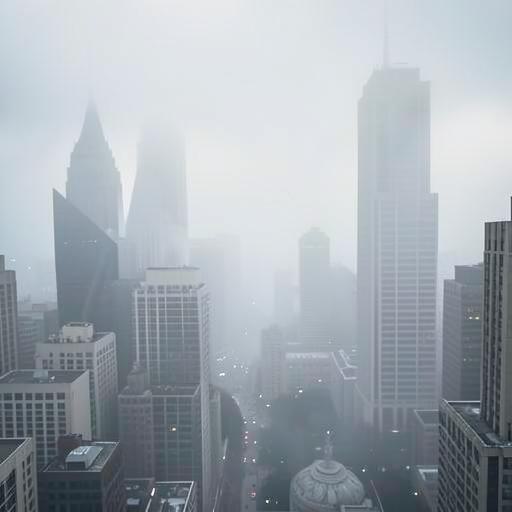} &
        \includegraphics[width=0.14\textwidth]{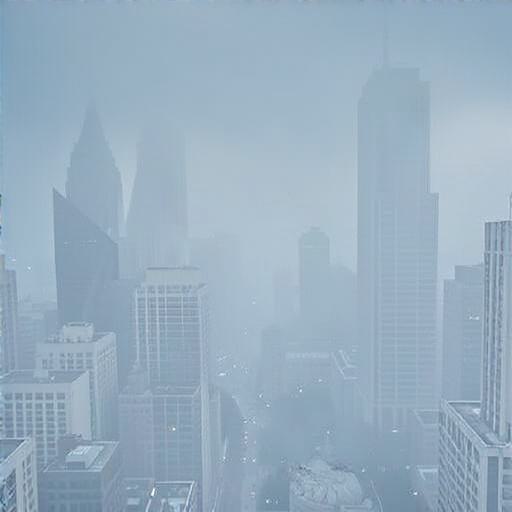} &
        \includegraphics[width=0.14\textwidth]{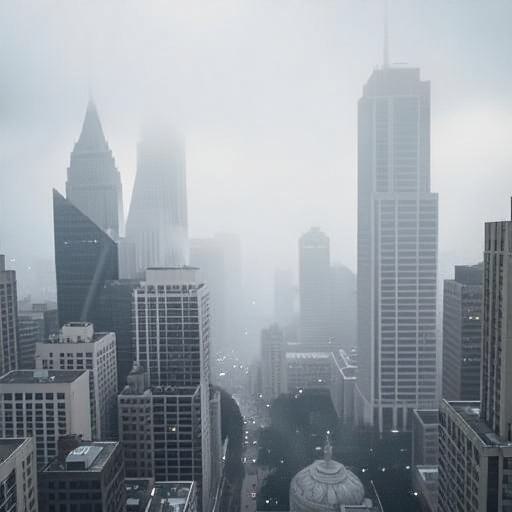} &
        \includegraphics[width=0.14\textwidth]{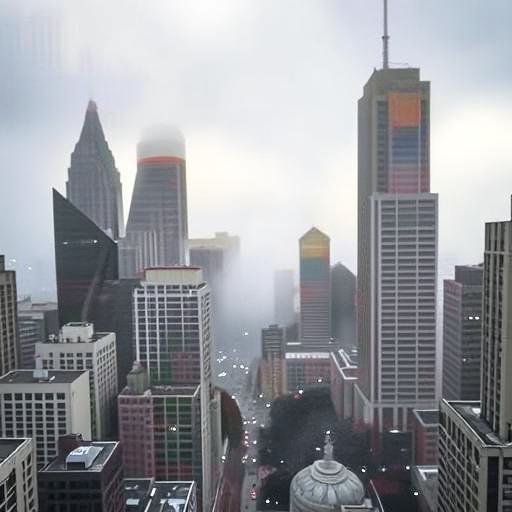} &
        \includegraphics[width=0.14\textwidth]{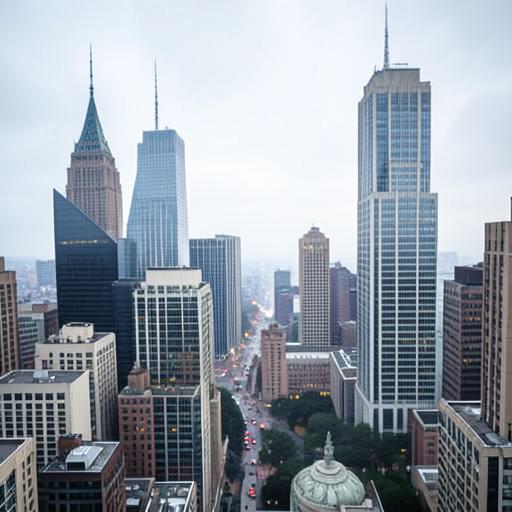} \\

        \includegraphics[width=0.14\textwidth]{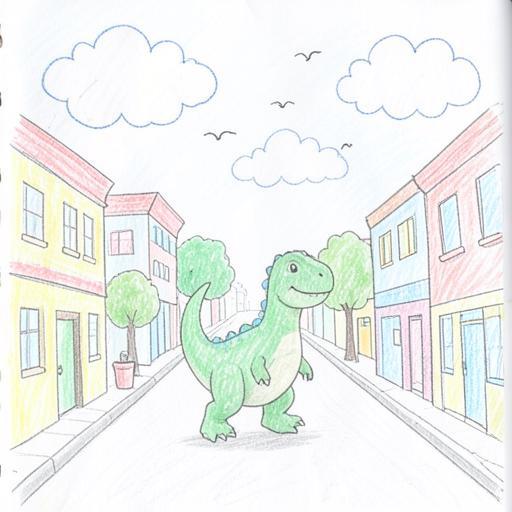} &
        \includegraphics[width=0.14\textwidth]{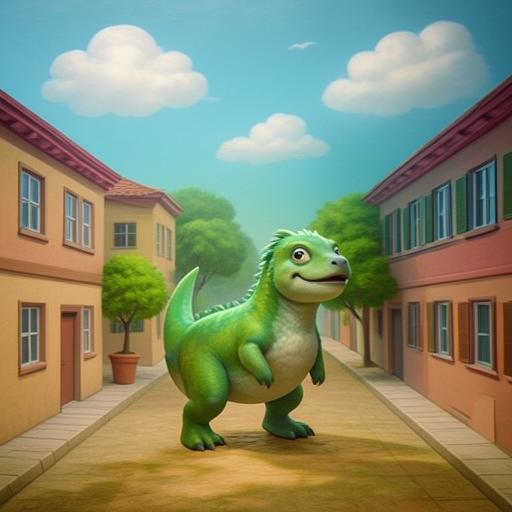} &
        \includegraphics[width=0.14\textwidth]{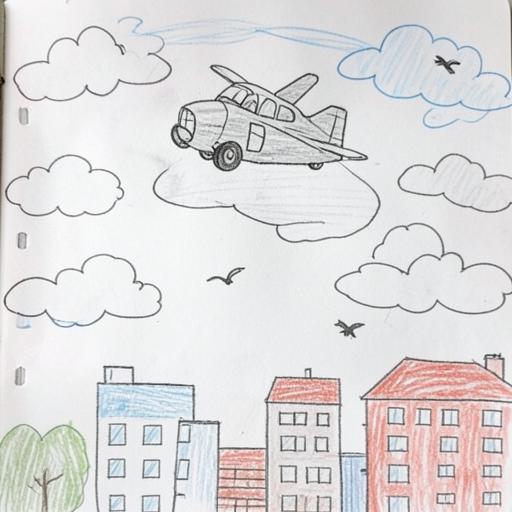} &
        \includegraphics[width=0.14\textwidth]{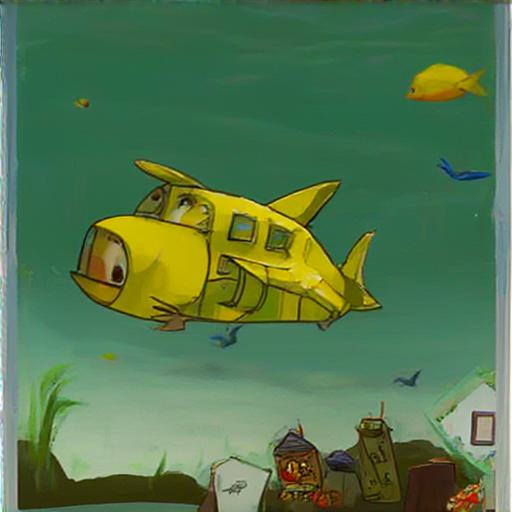} &
        \includegraphics[width=0.14\textwidth]{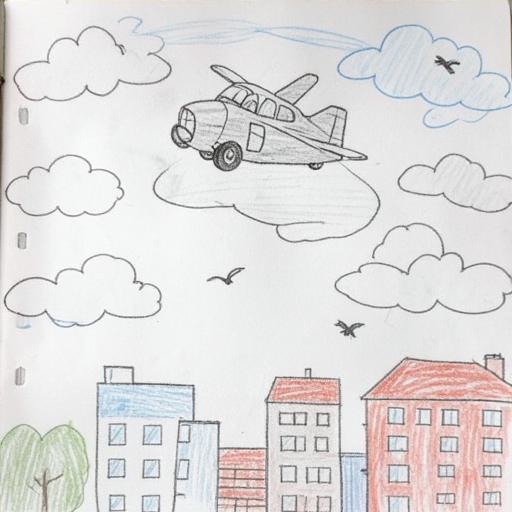} &
        \includegraphics[width=0.14\textwidth]{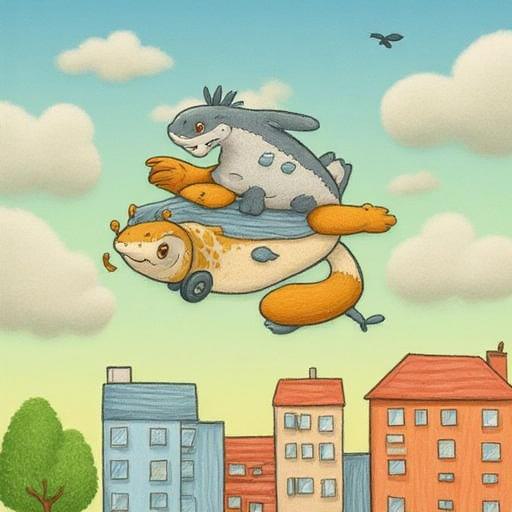} &
        \includegraphics[width=0.14\textwidth]{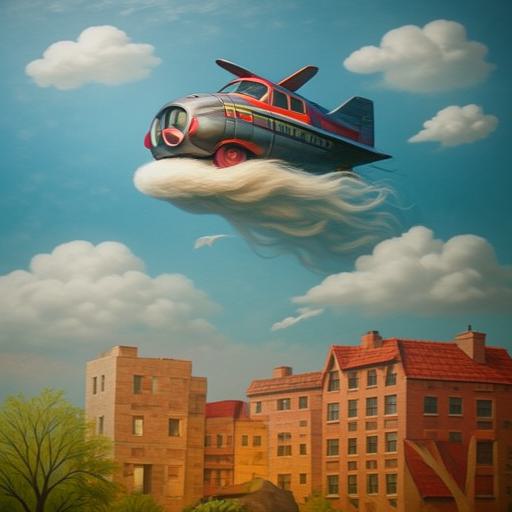} \\

        \includegraphics[width=0.14\textwidth]{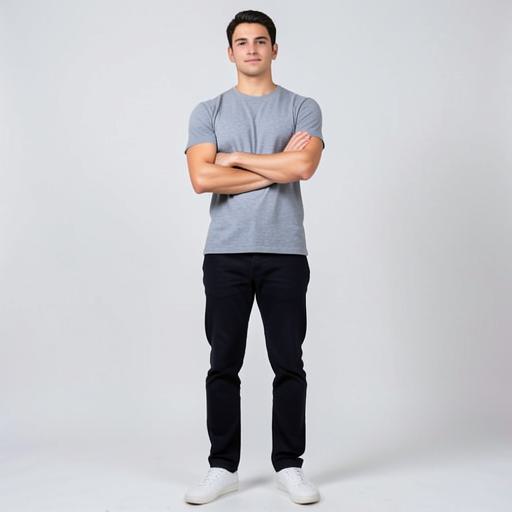} &
        \includegraphics[width=0.14\textwidth]{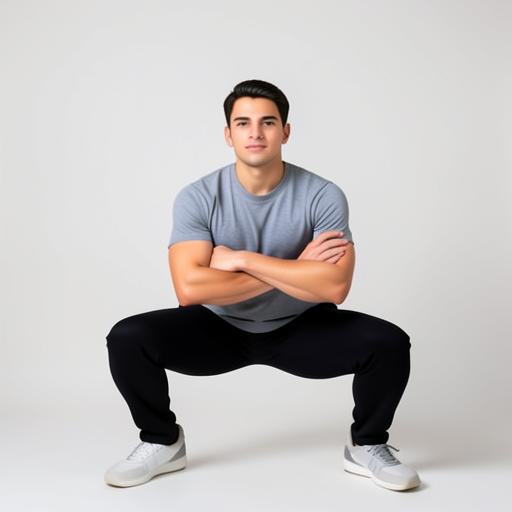} &
        \includegraphics[width=0.14\textwidth]{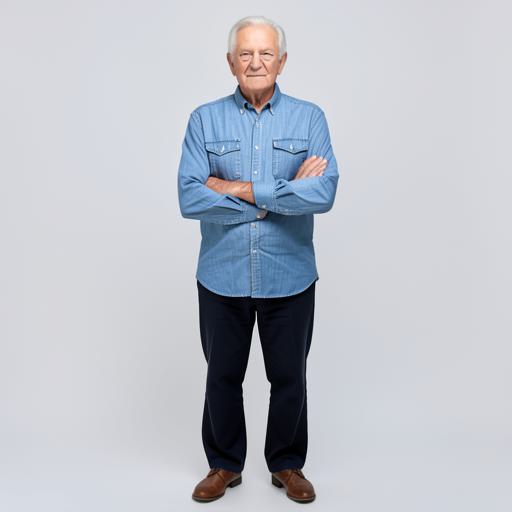} &
        \includegraphics[width=0.14\textwidth]{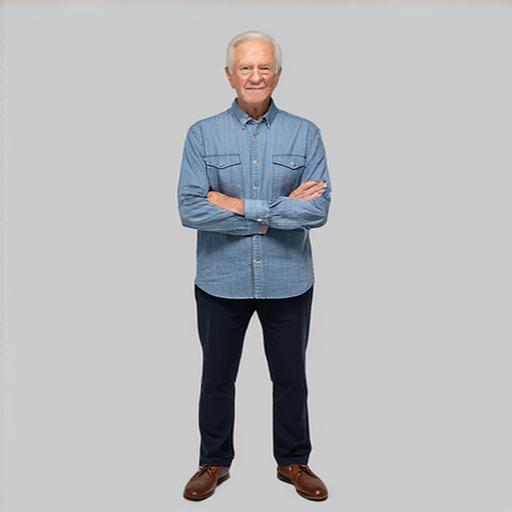} &
        \includegraphics[width=0.14\textwidth]{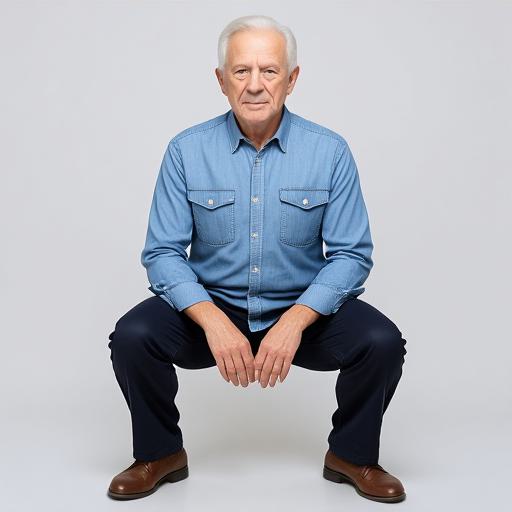} &
        \includegraphics[width=0.14\textwidth]{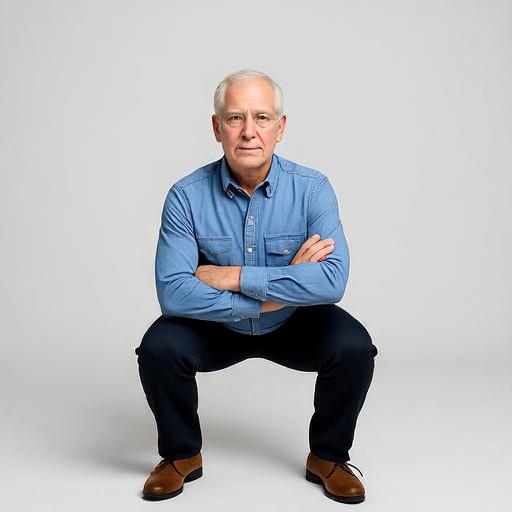} &
        \includegraphics[width=0.14\textwidth]{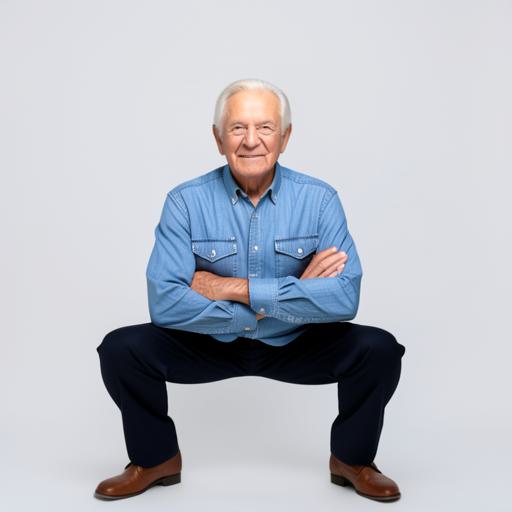} \\

        \includegraphics[width=0.14\textwidth]{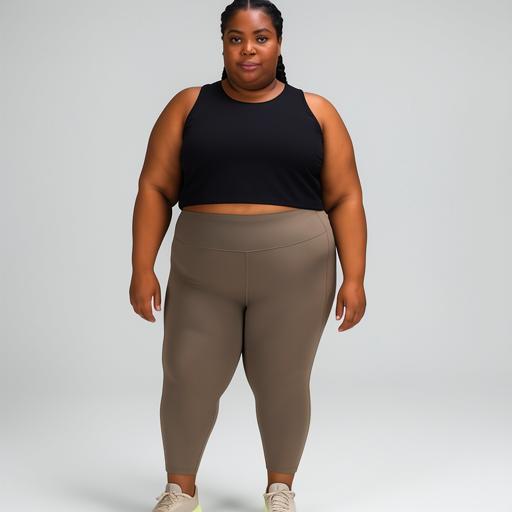} &
        \includegraphics[width=0.14\textwidth]{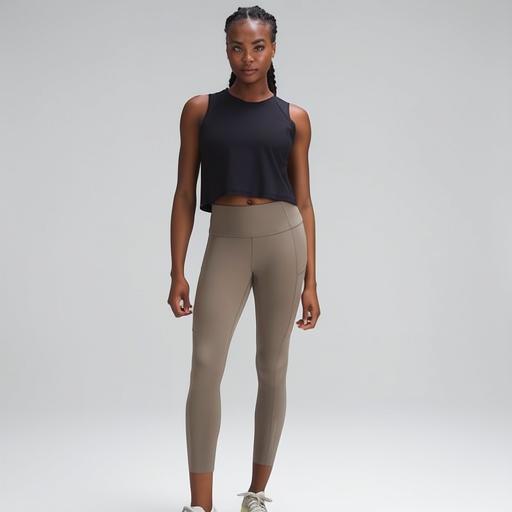} &
        \includegraphics[width=0.14\textwidth]{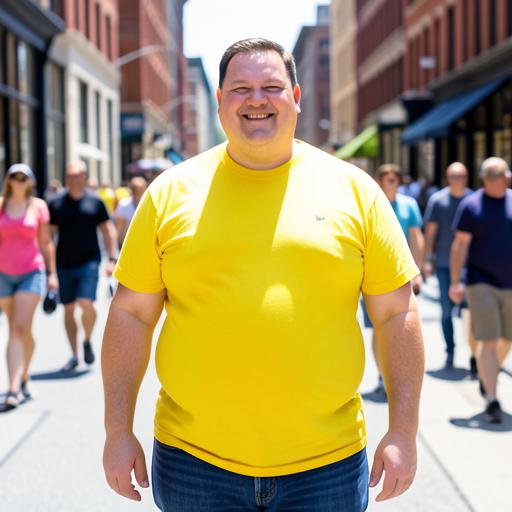} &
        \includegraphics[width=0.14\textwidth]{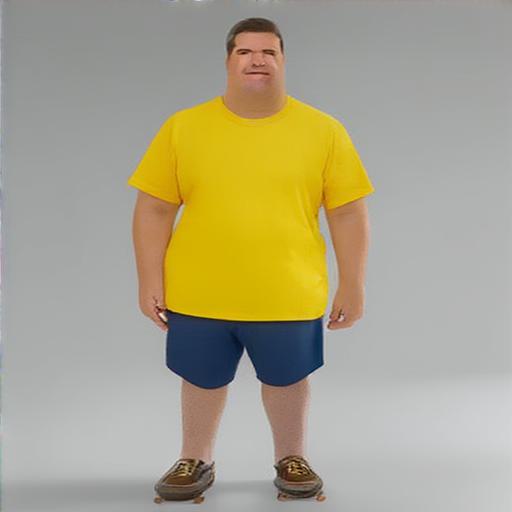} &
        \includegraphics[width=0.14\textwidth]{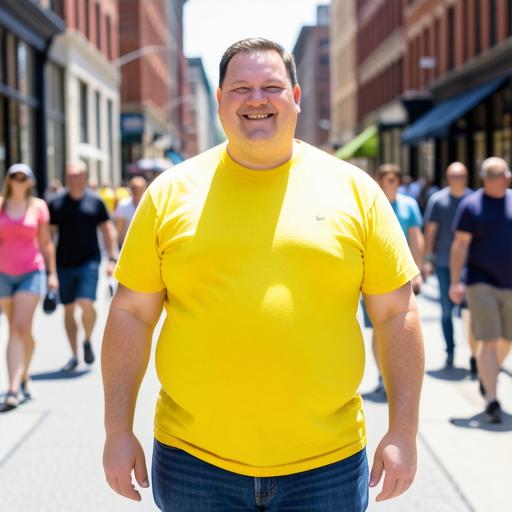} &
        \includegraphics[width=0.14\textwidth]{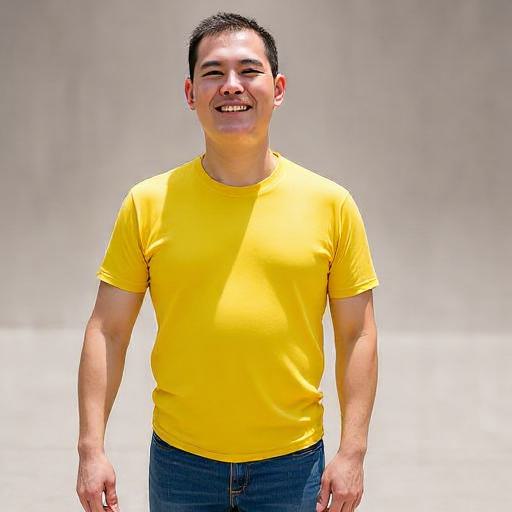} &
        \includegraphics[width=0.14\textwidth]{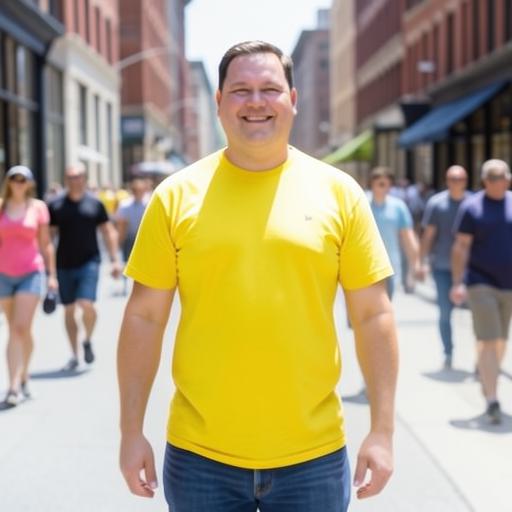} \\

        \includegraphics[width=0.14\textwidth]{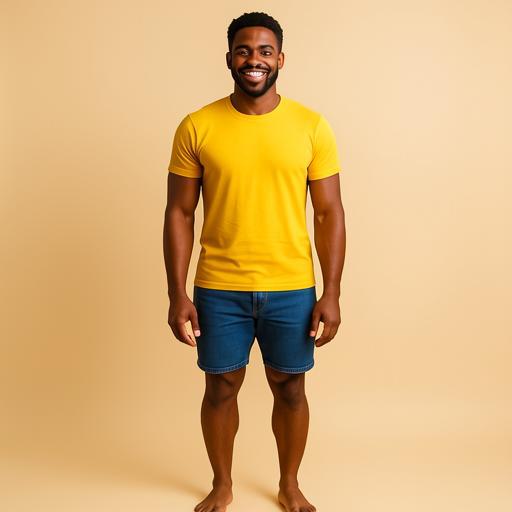} &
        \includegraphics[width=0.14\textwidth]{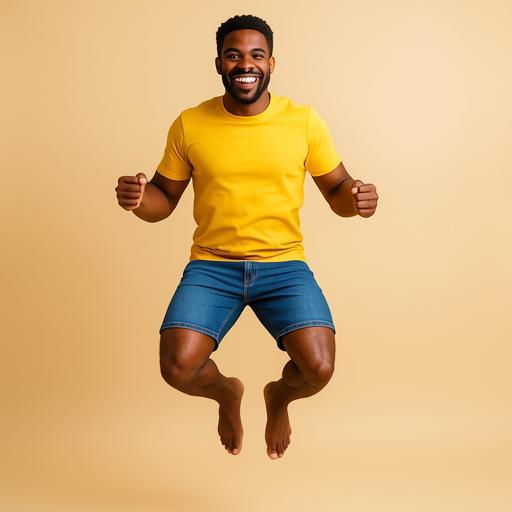} &
        \includegraphics[width=0.14\textwidth]{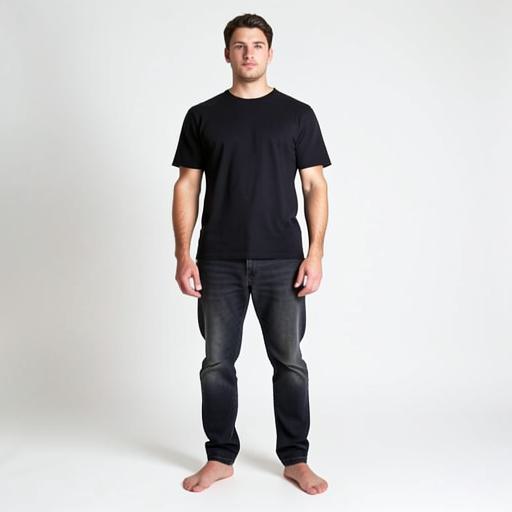} &
        \includegraphics[width=0.14\textwidth]{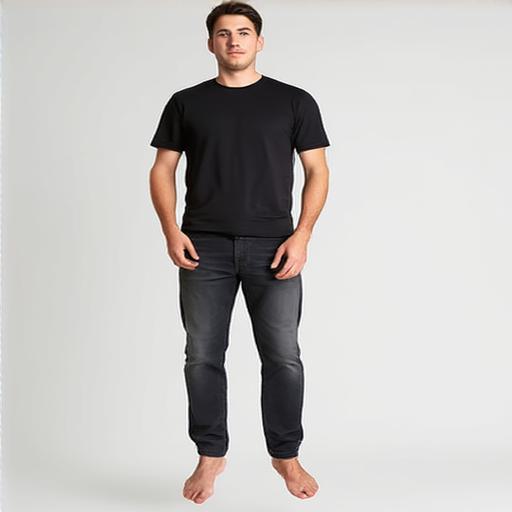} &
        \includegraphics[width=0.14\textwidth]{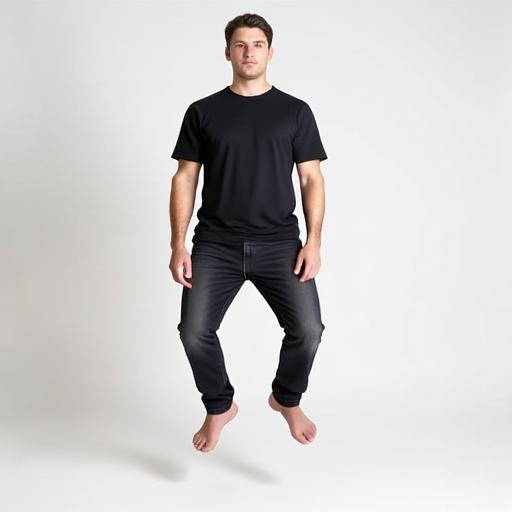} &
        \includegraphics[width=0.14\textwidth]{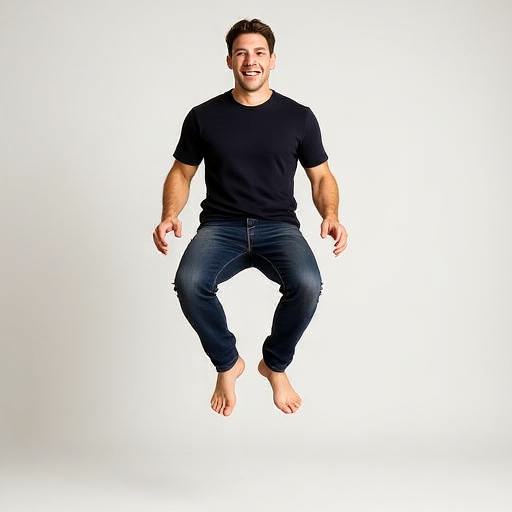} &
        \includegraphics[width=0.14\textwidth]{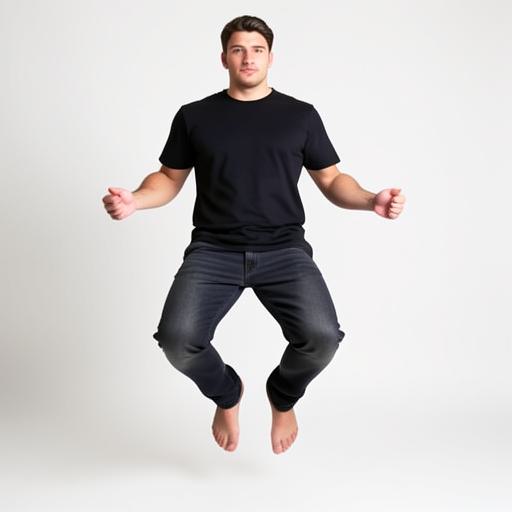} \\

    \end{tabular}
    }
    \caption{\textbf{Additional qualitative comparison on seen editing tasks.} We compare Delta-Adapter with RelationAdapter~\cite{gong2025relationadapter}, LoRWeB~\cite{lorweb}, and Edit Transfer~\cite{edit_transfer}. Delta-Adapter more faithfully captures the edit semantics implied by the exemplar pair and applies them to the query image, while better preserving its underlying structure and identity.}

\label{fig:additional_qualitative_seen}
\end{figure*}
\begin{figure*}[t]
    \centering
    \renewcommand{\arraystretch}{0.3}
    \setlength{\tabcolsep}{0.6pt}

    {\footnotesize
    \begin{tabular}{c c  c @{\hspace{0.05cm}} | @{\hspace{0.05cm}} c c c c }

        \multicolumn{1}{c}{\normalsize Source ($a$)} &
        \multicolumn{1}{c}{\normalsize Target ($a'$)} &
        \multicolumn{1}{c}{\normalsize Query ($b$)} &
        \multicolumn{1}{c}{\normalsize VisualCloze} &
        \multicolumn{1}{c}{\normalsize LoRWeB} &
        \multicolumn{1}{c}{\begin{tabular}{c}\normalsize Relation\\\normalsize Adapter\end{tabular}} &
        \multicolumn{1}{c}{\normalsize Ours} \\

        \includegraphics[width=0.14\textwidth]{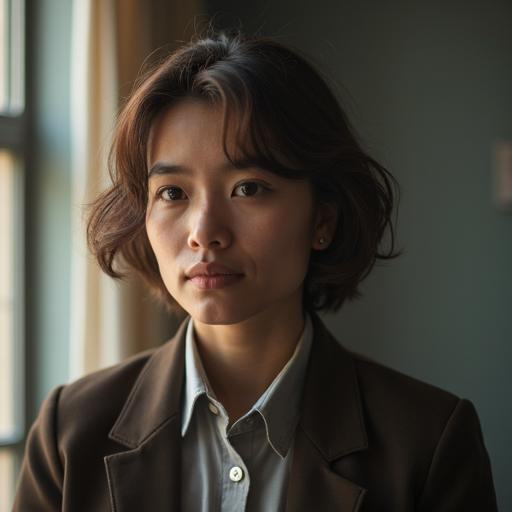} &
        \includegraphics[width=0.14\textwidth]{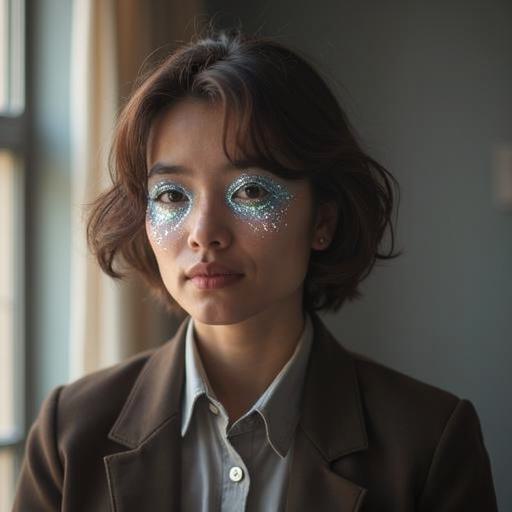} &
        \includegraphics[width=0.14\textwidth]{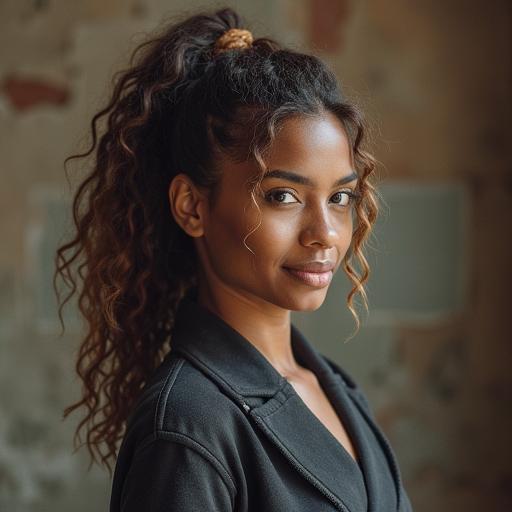} &
        \includegraphics[width=0.14\textwidth]{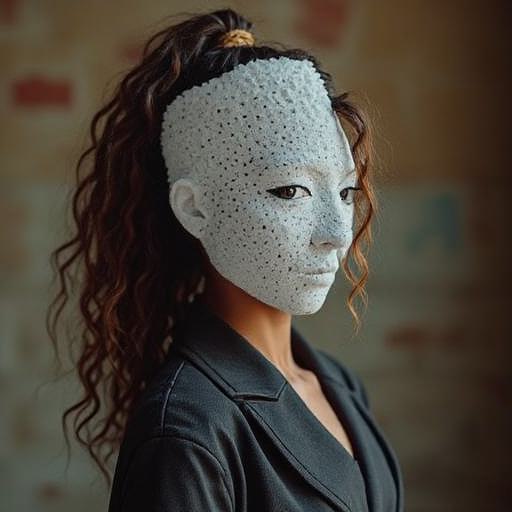} &
        \includegraphics[width=0.14\textwidth]{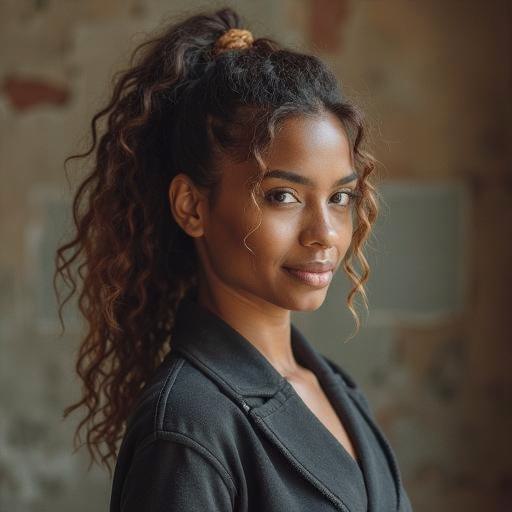} &
        \includegraphics[width=0.14\textwidth]{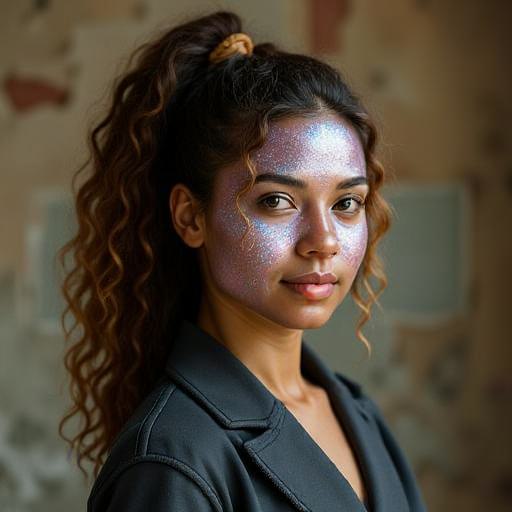} &
        \includegraphics[width=0.14\textwidth]{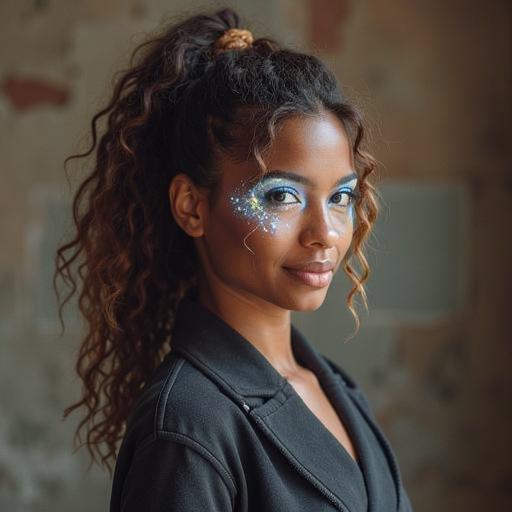} \\

        \includegraphics[width=0.14\textwidth]{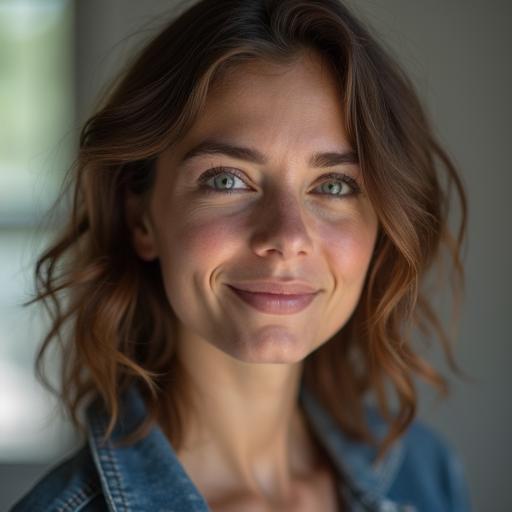} &
        \includegraphics[width=0.14\textwidth]{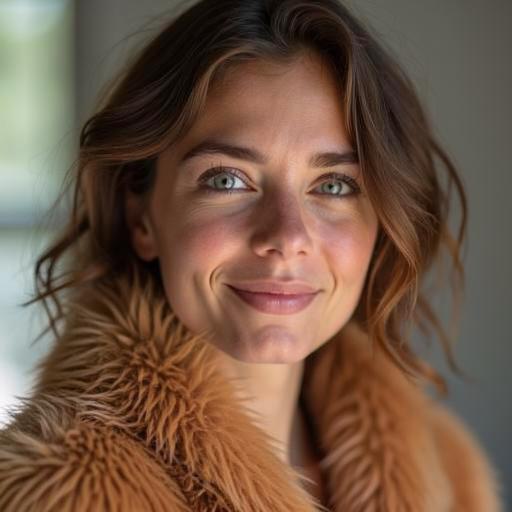} &
        \includegraphics[width=0.14\textwidth]{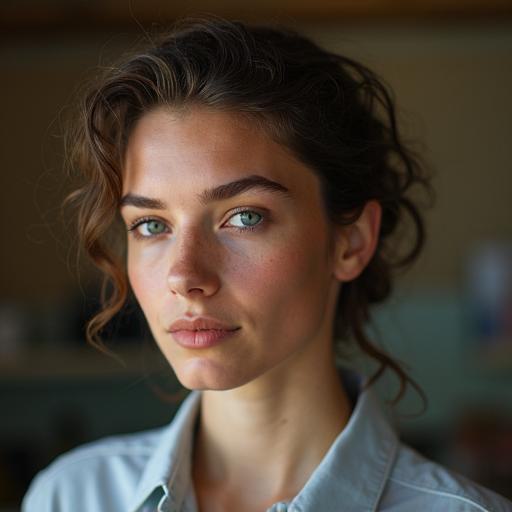} &
        \includegraphics[width=0.14\textwidth]{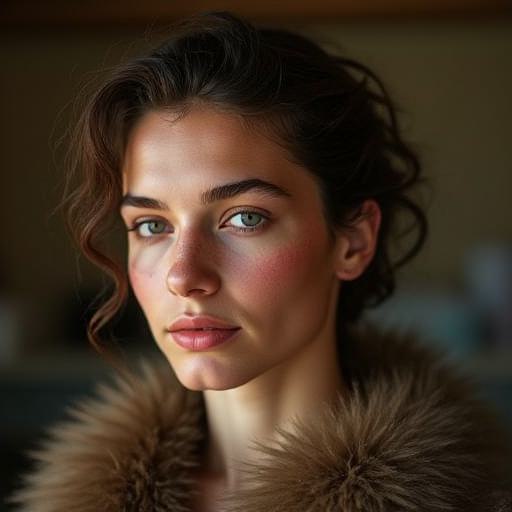} &
        \includegraphics[width=0.14\textwidth]{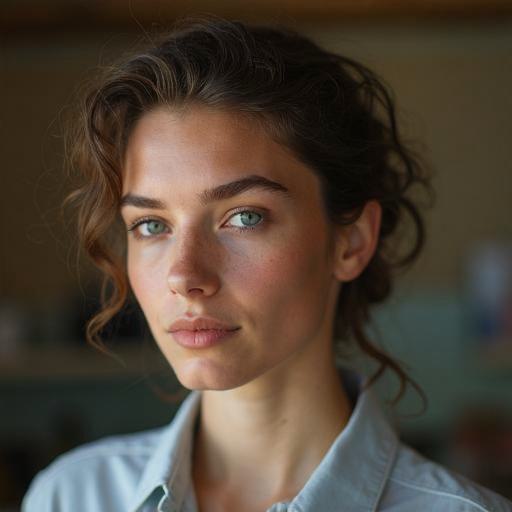} &
        \includegraphics[width=0.14\textwidth]{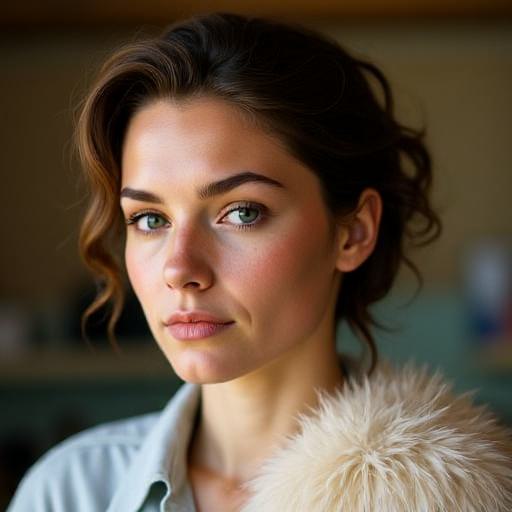} &
        \includegraphics[width=0.14\textwidth]{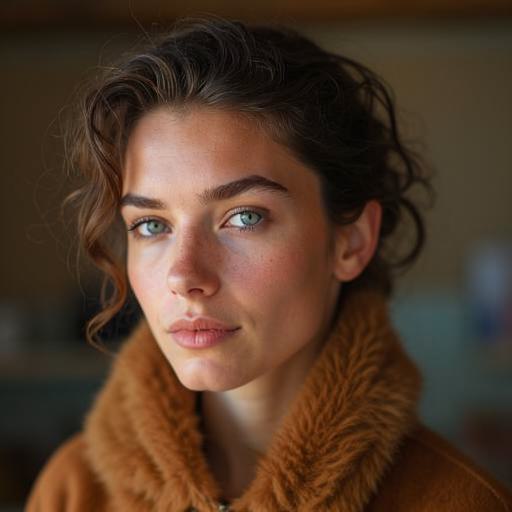} \\
        
        \includegraphics[width=0.14\textwidth]{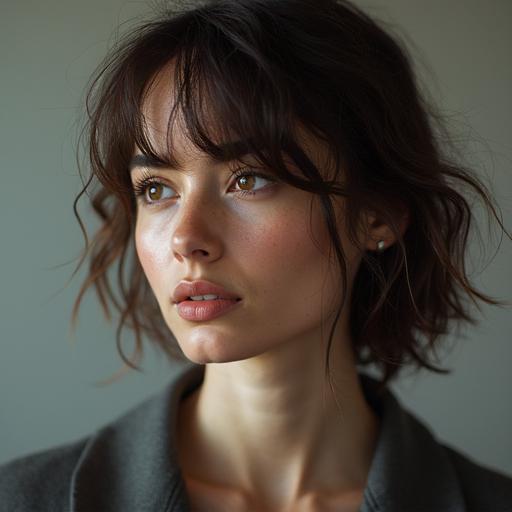} &
        \includegraphics[width=0.14\textwidth]{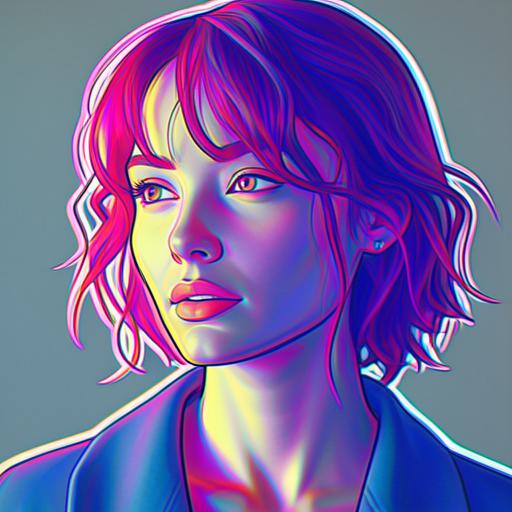} &
        \includegraphics[width=0.14\textwidth]{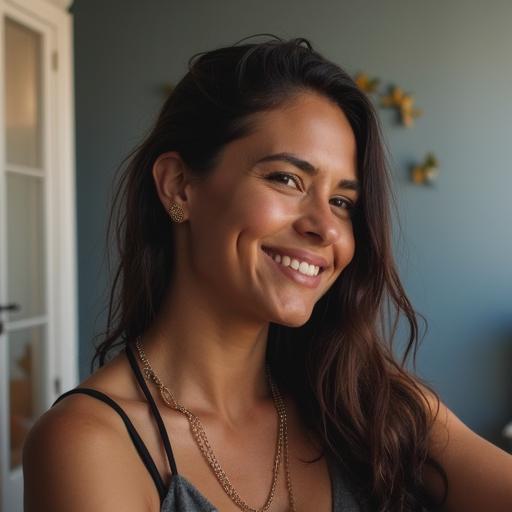} &
        \includegraphics[width=0.14\textwidth]{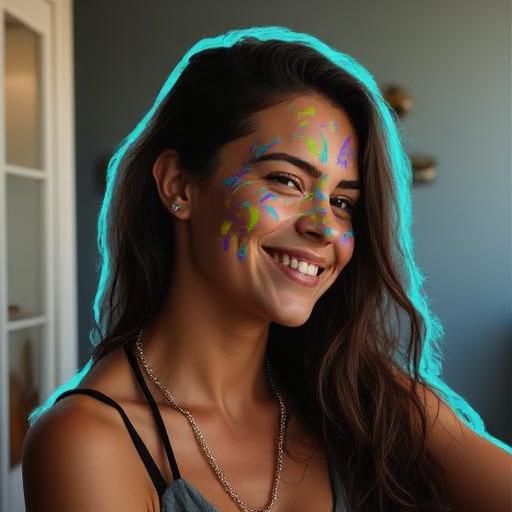} &
        \includegraphics[width=0.14\textwidth]{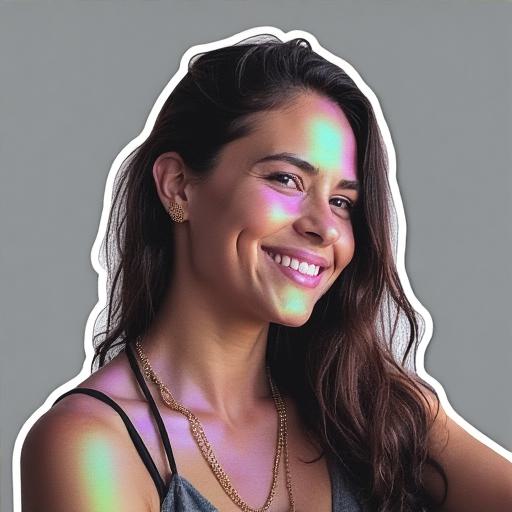} &
        \includegraphics[width=0.14\textwidth]{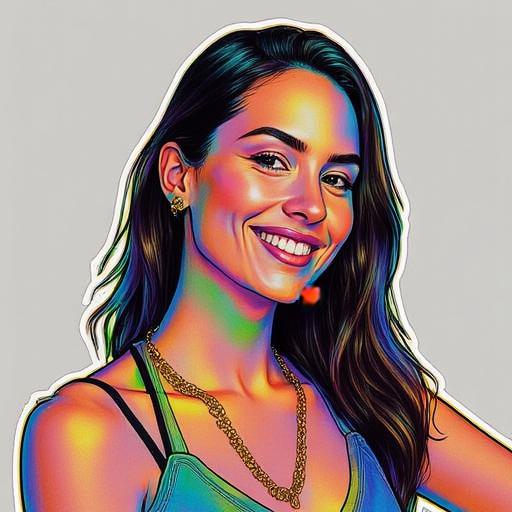} &
        \includegraphics[width=0.14\textwidth]{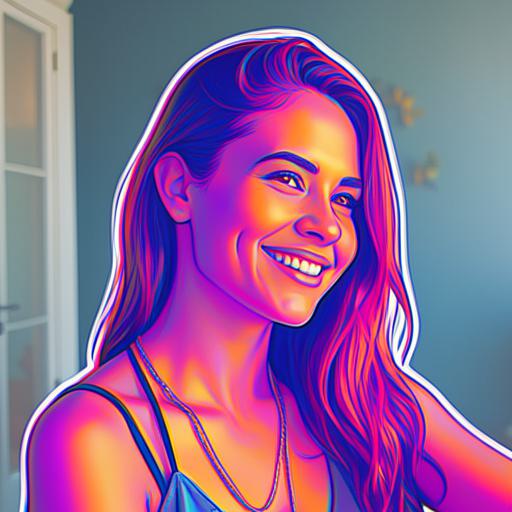} \\

        \includegraphics[width=0.14\textwidth]{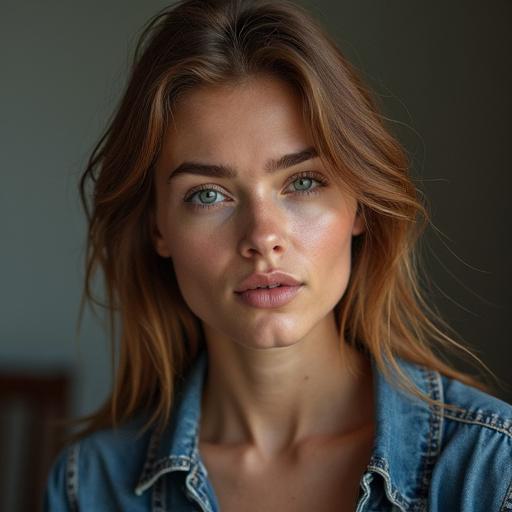} &
        \includegraphics[width=0.14\textwidth]{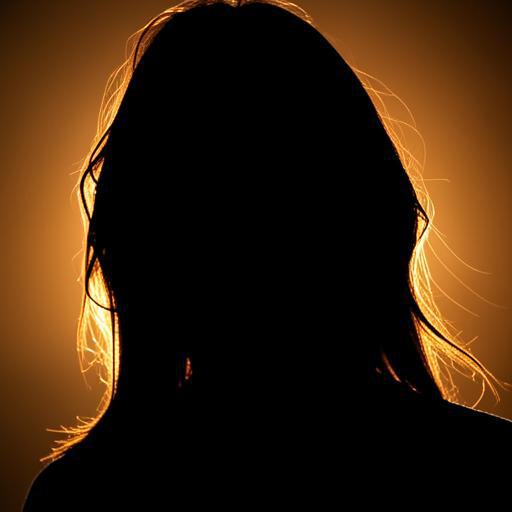} &
        \includegraphics[width=0.14\textwidth]{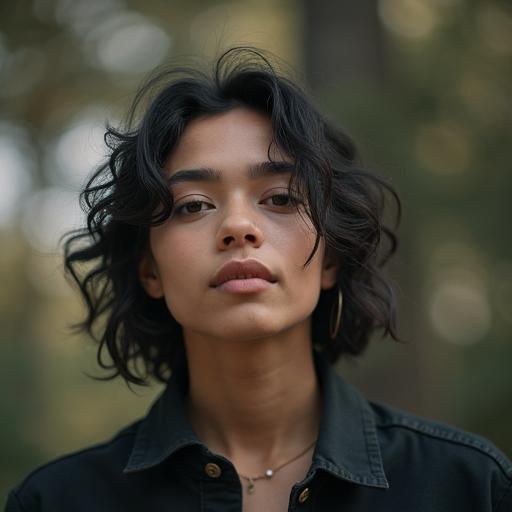} &
        \includegraphics[width=0.14\textwidth]{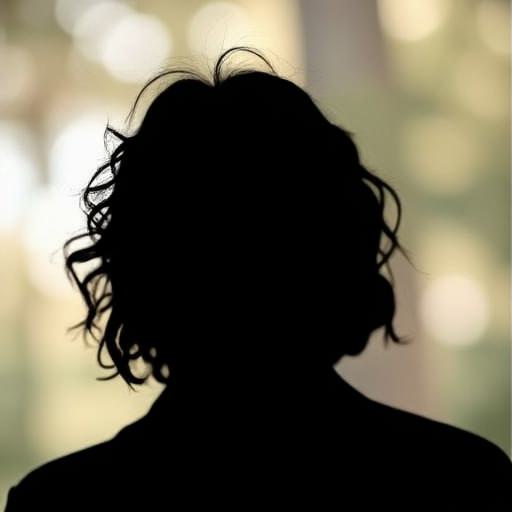} &
        \includegraphics[width=0.14\textwidth]{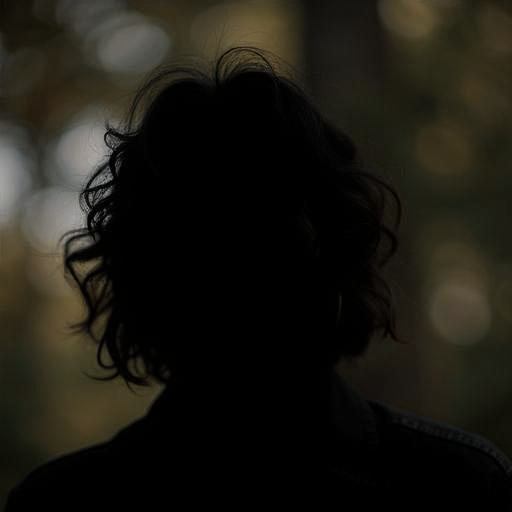} &
        \includegraphics[width=0.14\textwidth]{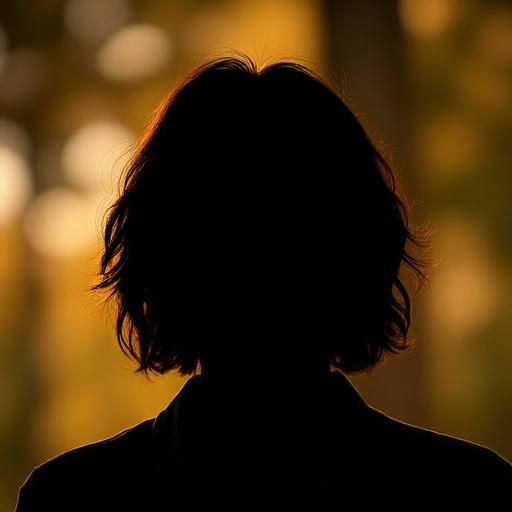} &
        \includegraphics[width=0.14\textwidth]{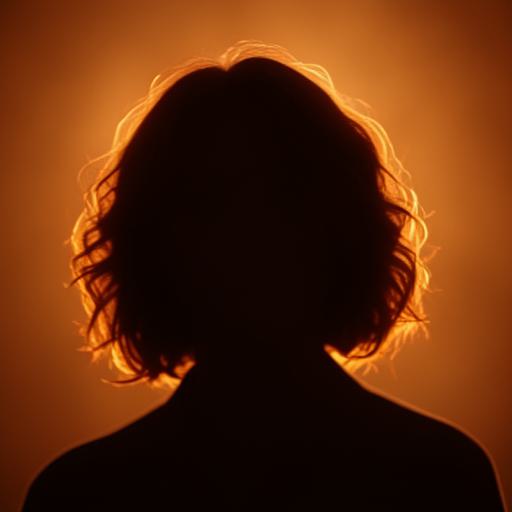} \\

        \includegraphics[width=0.14\textwidth]{images/Test_Useen_baseline/pearlescent_coating/ref/left.jpg} &
        \includegraphics[width=0.14\textwidth]{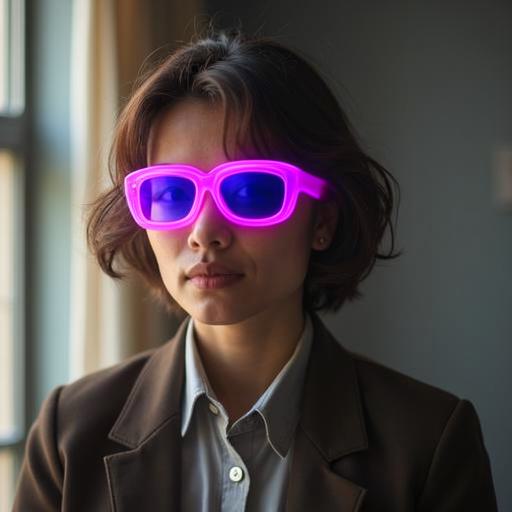} &
        \includegraphics[width=0.14\textwidth]{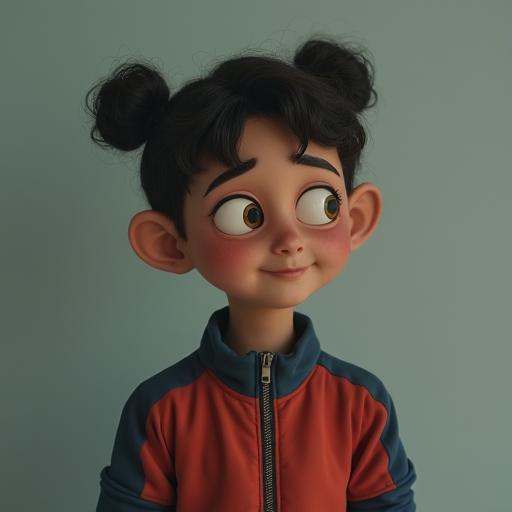} &
        \includegraphics[width=0.14\textwidth]{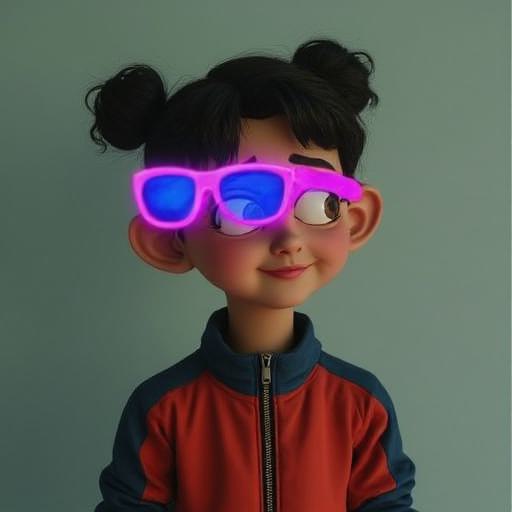} &
        \includegraphics[width=0.14\textwidth]{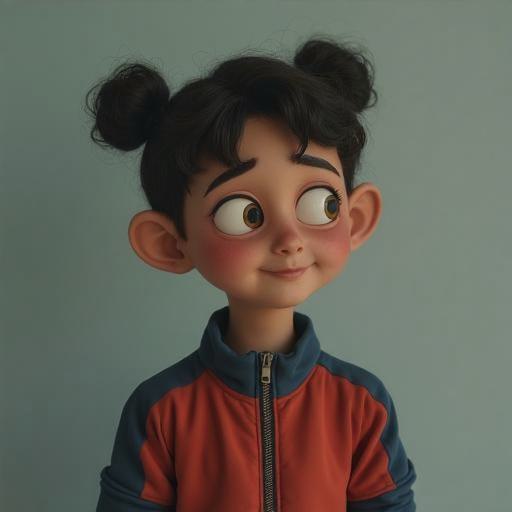} &
        \includegraphics[width=0.14\textwidth]{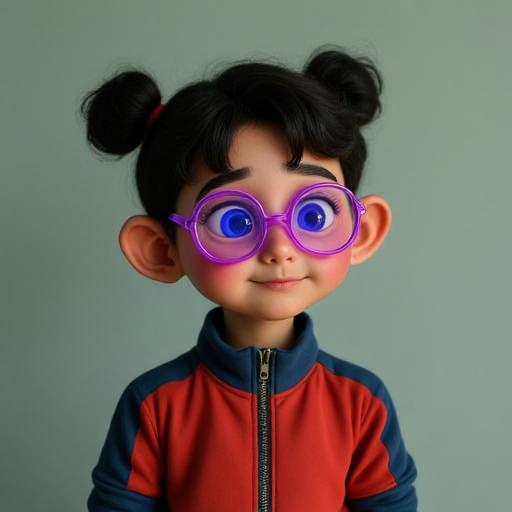} &
        \includegraphics[width=0.14\textwidth]{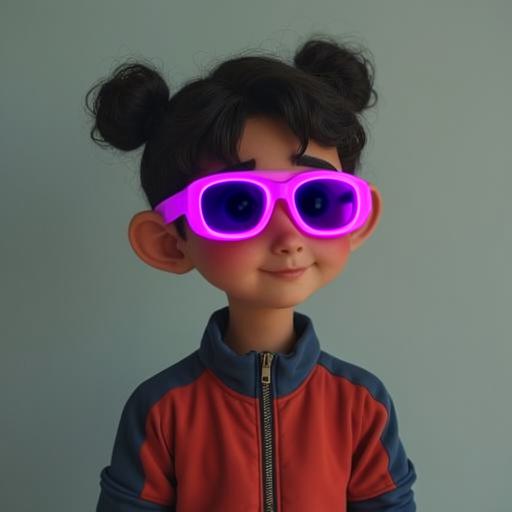} \\
        
        \includegraphics[width=0.14\textwidth]{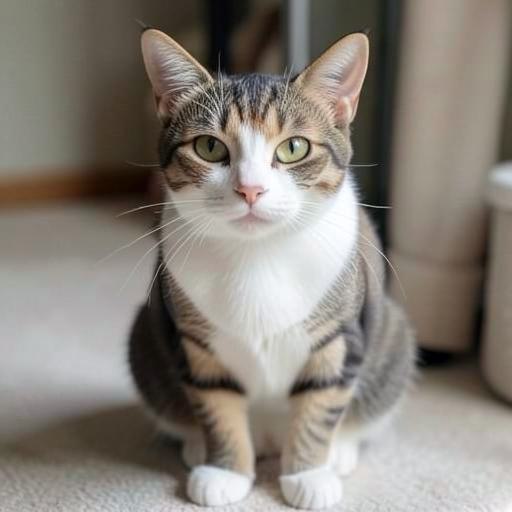} &
        \includegraphics[width=0.14\textwidth]{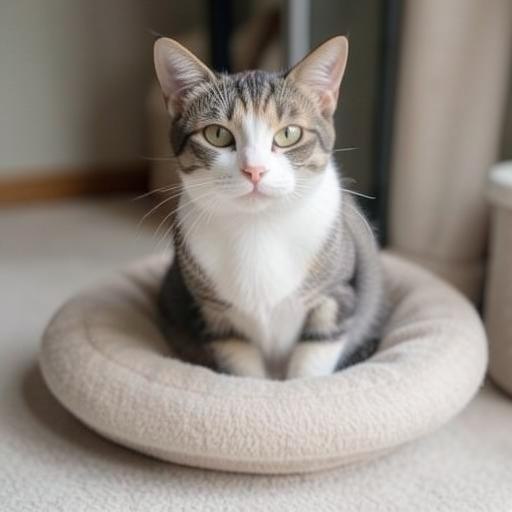} &
        \includegraphics[width=0.14\textwidth]{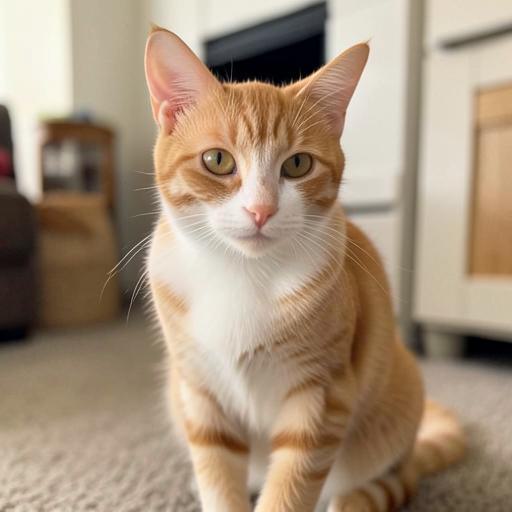} &
        \includegraphics[width=0.14\textwidth]{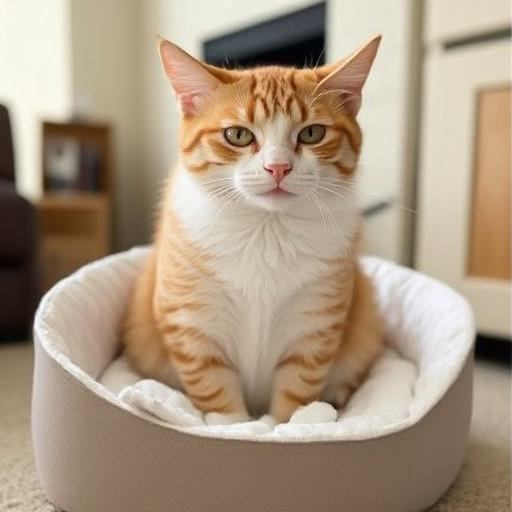} &
        \includegraphics[width=0.14\textwidth]{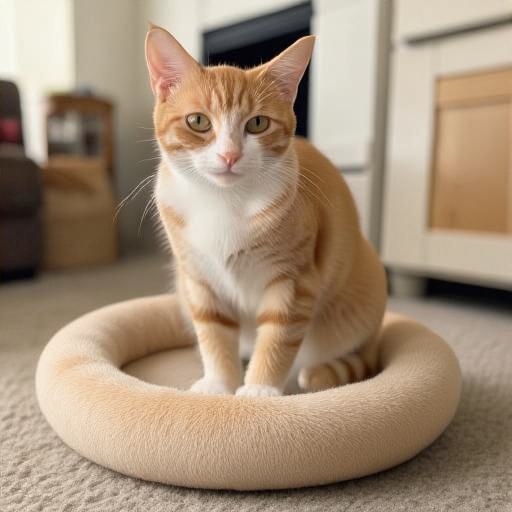} &
        \includegraphics[width=0.14\textwidth]{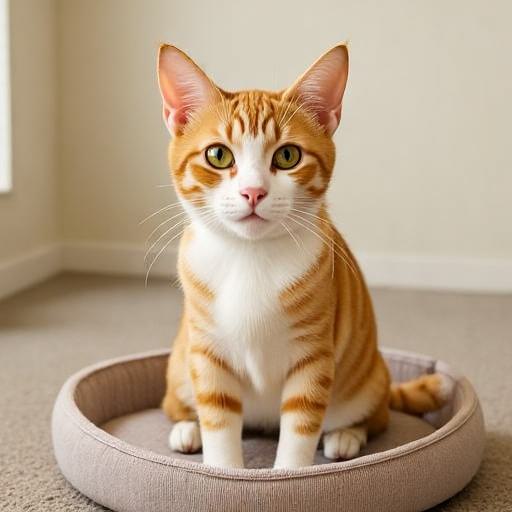} &
        \includegraphics[width=0.14\textwidth]{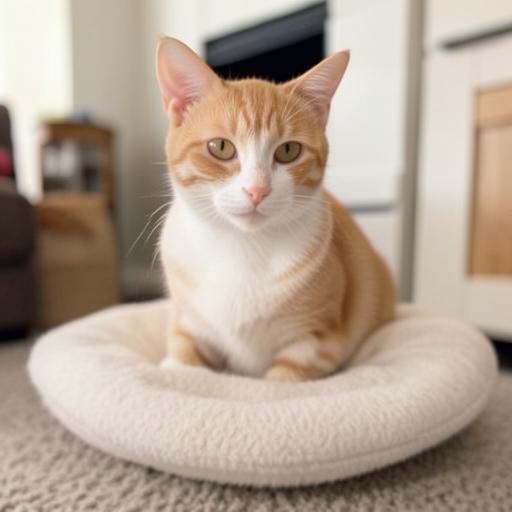} \\

        \includegraphics[width=0.14\textwidth]{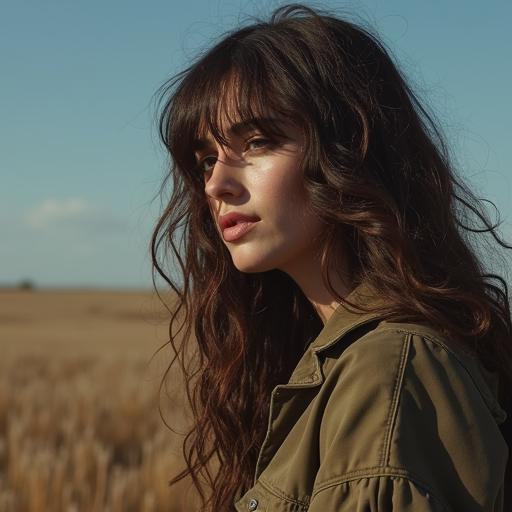} &
        \includegraphics[width=0.14\textwidth]{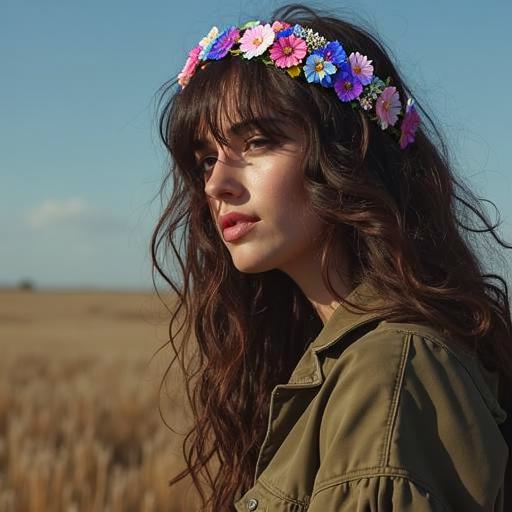} &
        \includegraphics[width=0.14\textwidth]{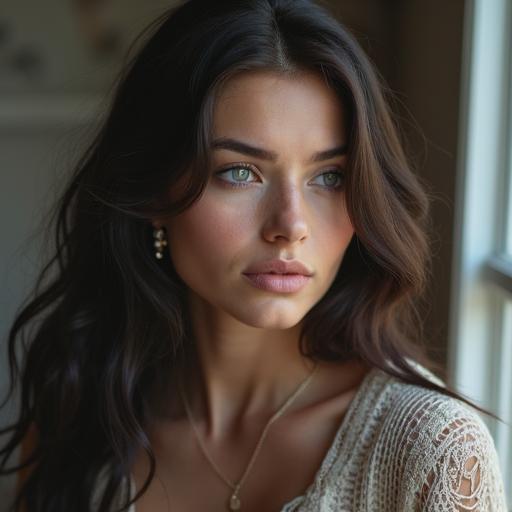} &
        \includegraphics[width=0.14\textwidth]{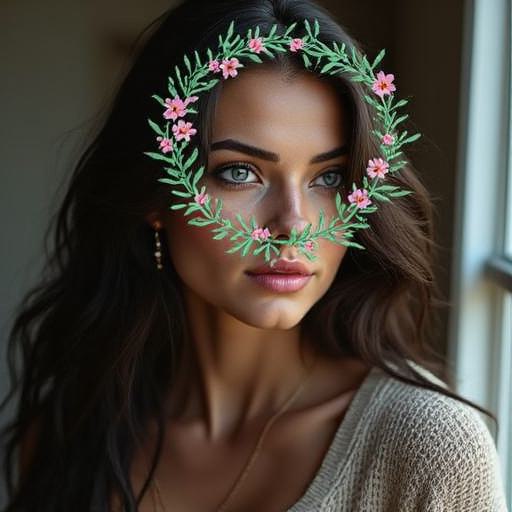} &
        \includegraphics[width=0.14\textwidth]{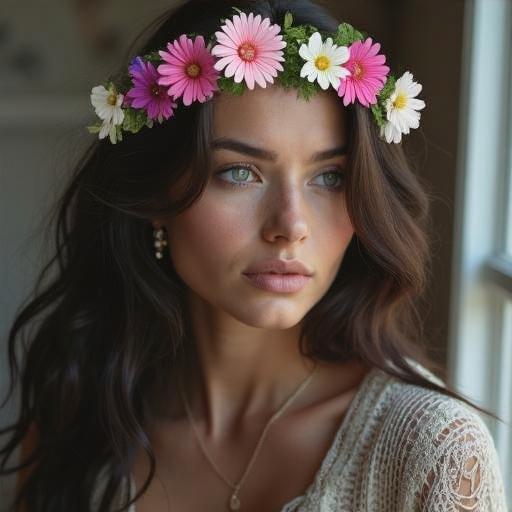} &
        \includegraphics[width=0.14\textwidth]{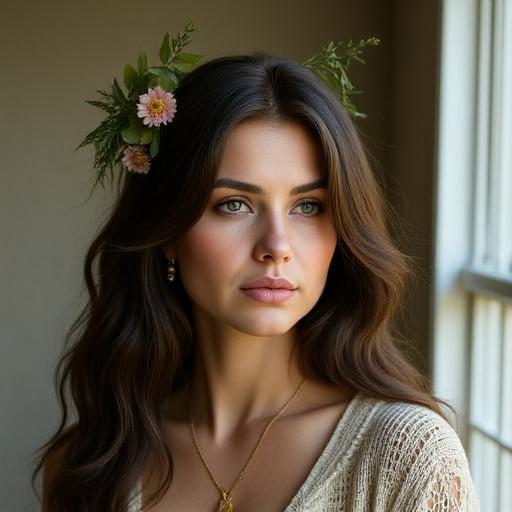} &
        \includegraphics[width=0.14\textwidth]{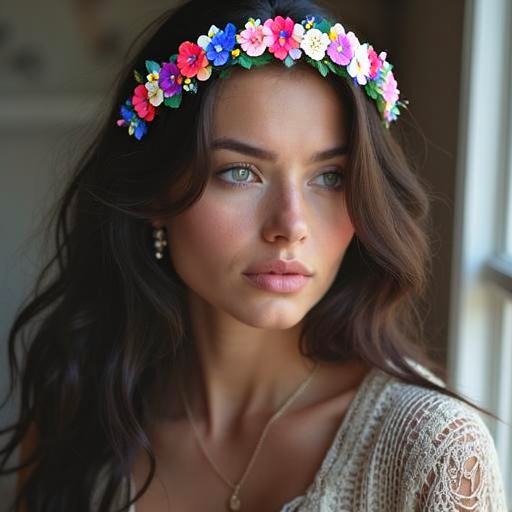} \\

        \includegraphics[width=0.14\textwidth]{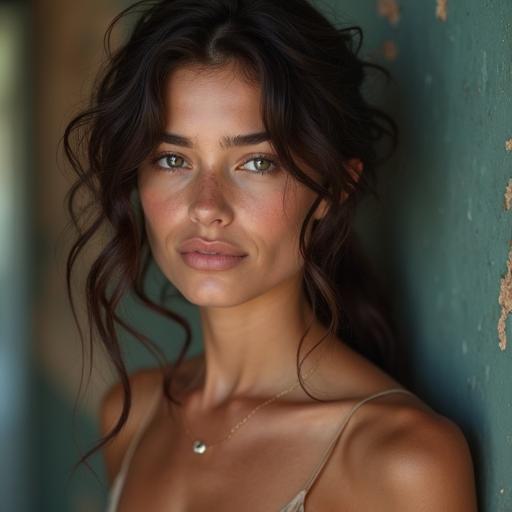} &
        \includegraphics[width=0.14\textwidth]{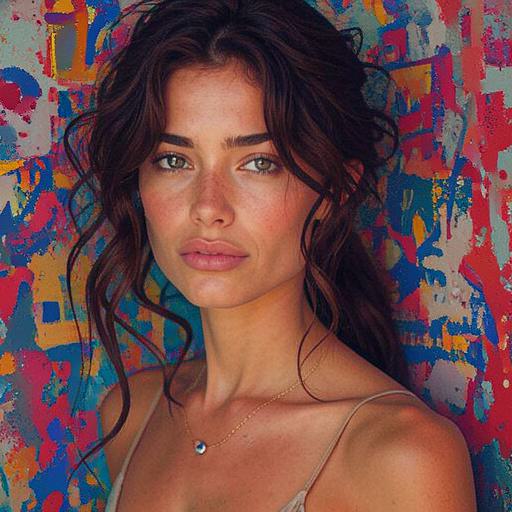} &
        \includegraphics[width=0.14\textwidth]{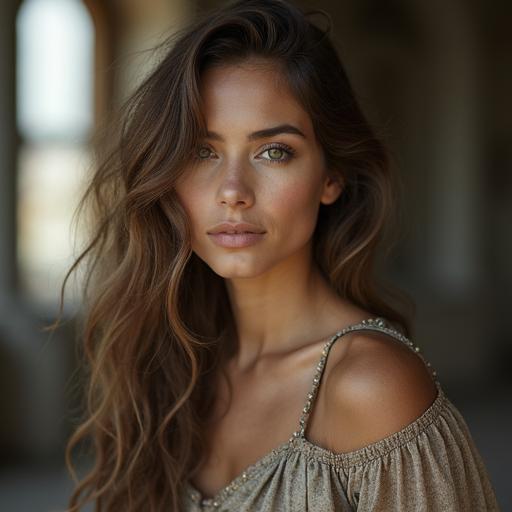} &
        \includegraphics[width=0.14\textwidth] {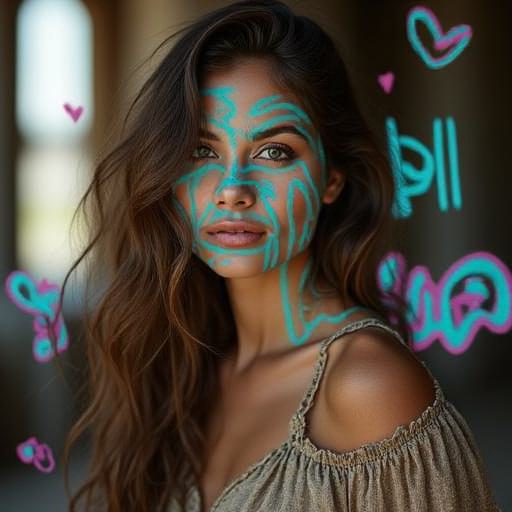} &
        \includegraphics[width=0.14\textwidth]{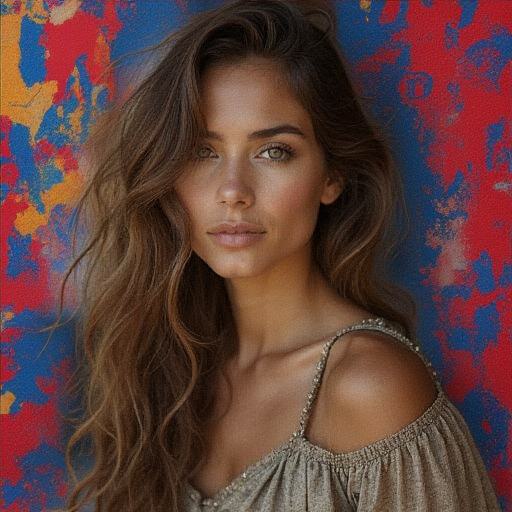} &
        \includegraphics[width=0.14\textwidth]{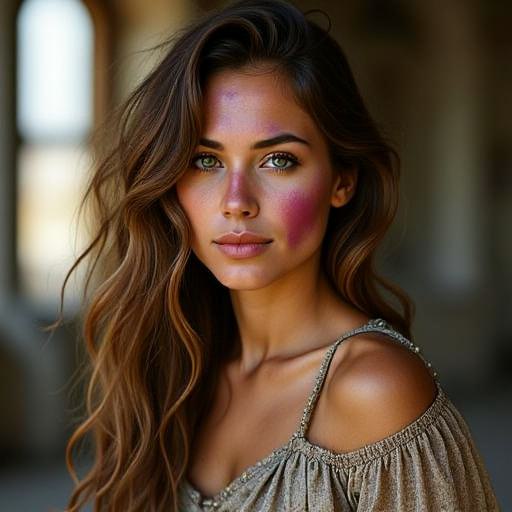} &
        \includegraphics[width=0.14\textwidth]{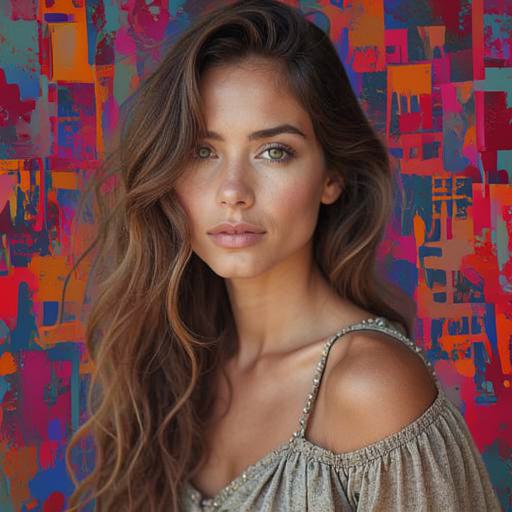} \\

        \includegraphics[width=0.14\textwidth]{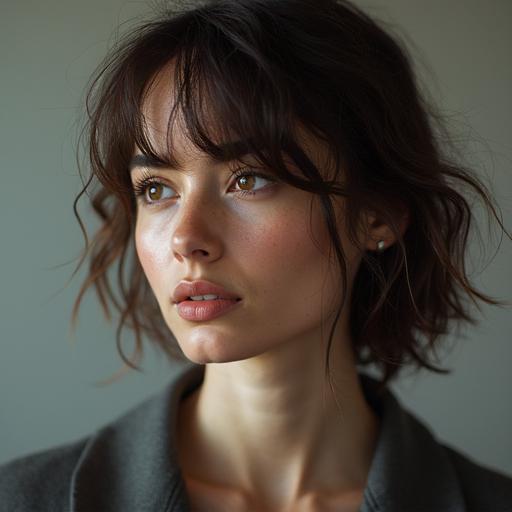} &
        \includegraphics[width=0.14\textwidth]{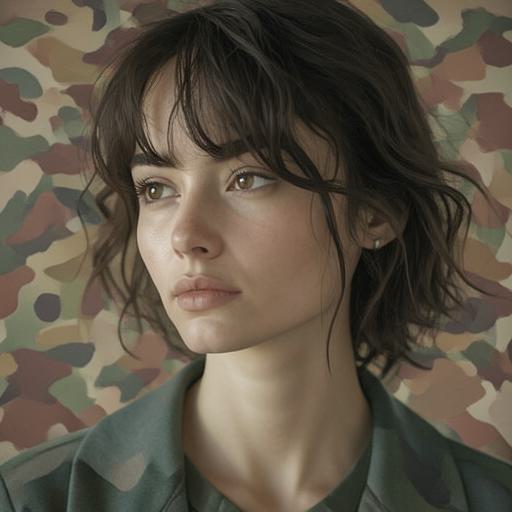} &
        \includegraphics[width=0.14\textwidth]{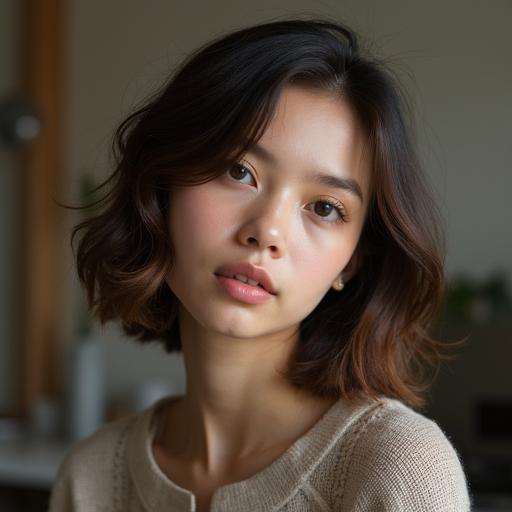} &
        \includegraphics[width=0.14\textwidth]{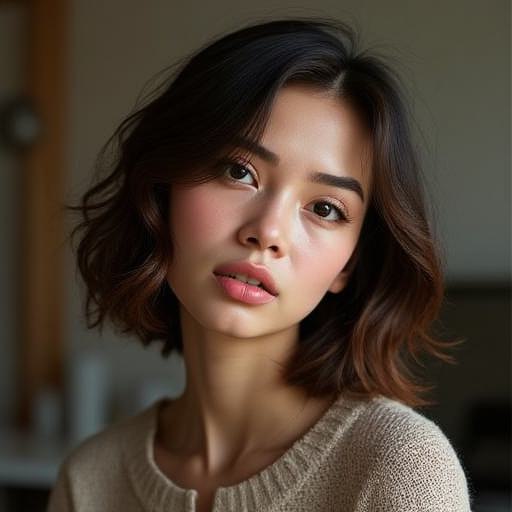} &
        \includegraphics[width=0.14\textwidth]{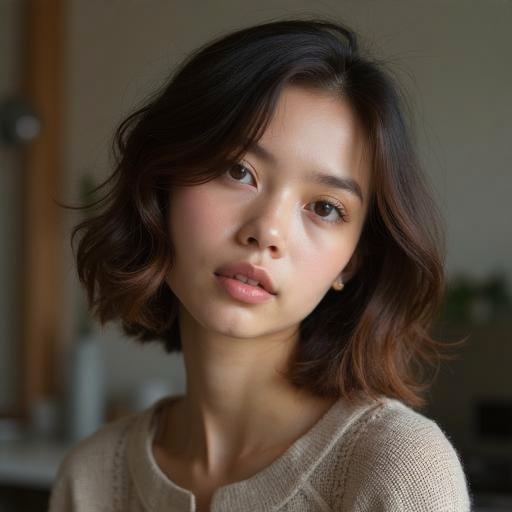} &
        \includegraphics[width=0.14\textwidth]{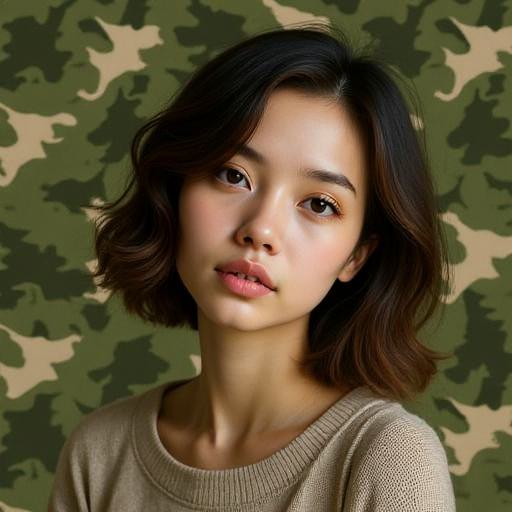} &
        \includegraphics[width=0.14\textwidth]{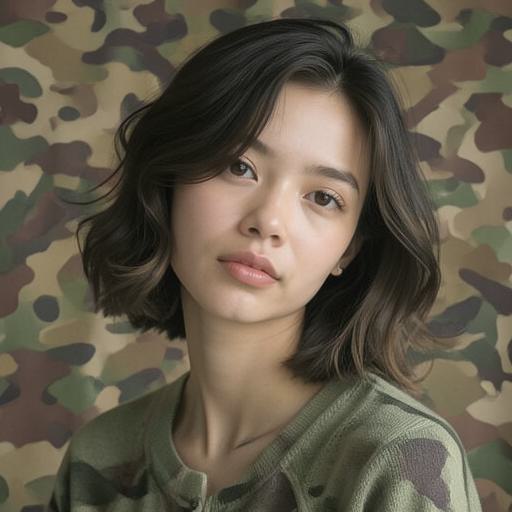} \\

    \end{tabular}
    }
    \caption{\textbf{Additional qualitative comparison on unseen editing tasks.} We compare Delta-Adapter with RelationAdapter~\cite{gong2025relationadapter}, LoRWeB~\cite{lorweb}, and VisualCloze~\cite{li2025visualcloze}. Across diverse unseen transformations, Delta-Adapter produces outputs that are semantically aligned with the exemplar edit, demonstrating superior generalization over the baselines.}

\label{fig:additional_qualitative_unseen}
\end{figure*}

\begin{figure*}[t]
    \centering
    \renewcommand{\arraystretch}{0.3}
    \setlength{\tabcolsep}{1pt}

    {\footnotesize
    \begin{tabular}{c c c @{\hspace{0.07cm}} | @{\hspace{0.07cm}} c c c}

        \multicolumn{1}{c}{\normalsize Source ($a$)} &
        \multicolumn{1}{c}{\normalsize Target ($a'$)} &
        \multicolumn{1}{c}{\normalsize Query ($b$)} &
        \multicolumn{1}{c}{\normalsize Nano Banana 2} &
        \multicolumn{1}{c}{\normalsize GPT-Image-2} &
        \multicolumn{1}{c}{\normalsize Ours} \\

        \includegraphics[width=0.16\textwidth]{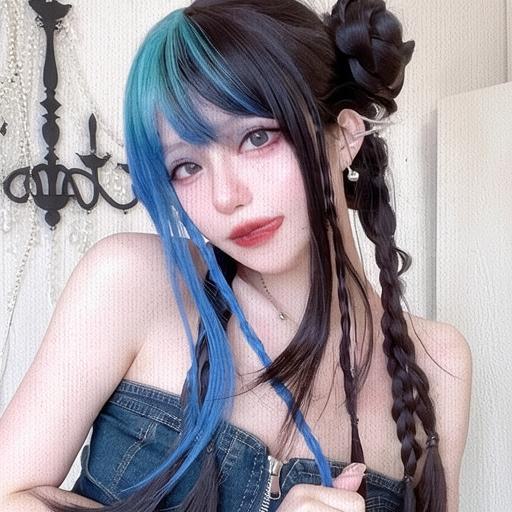} &
        \includegraphics[width=0.16\textwidth]{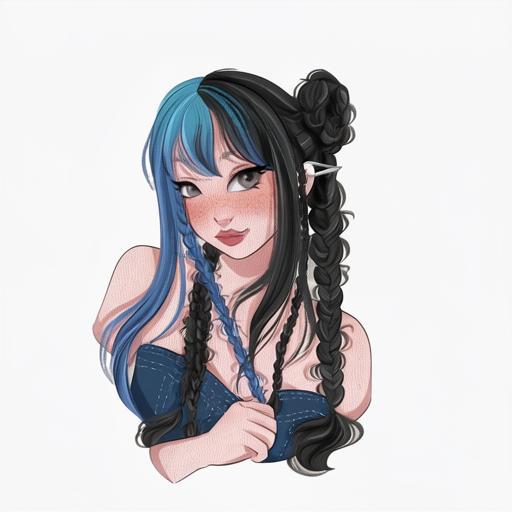} &
        \includegraphics[width=0.16\textwidth]{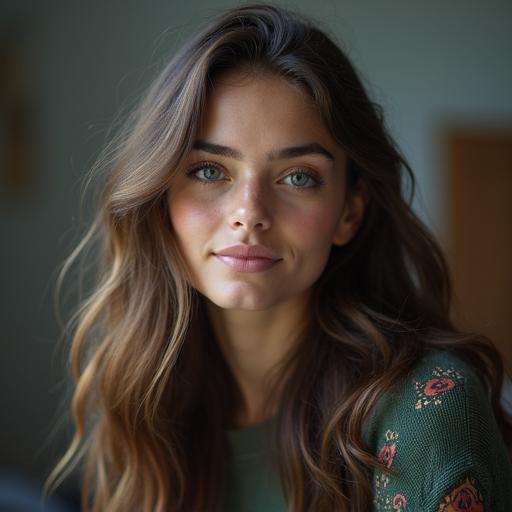} &
        \includegraphics[width=0.16\textwidth]{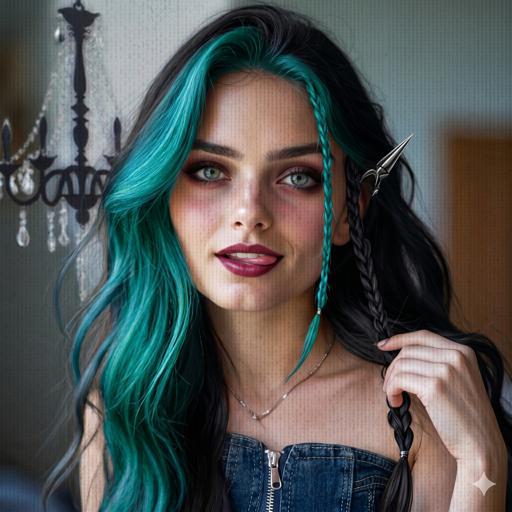} &
        \includegraphics[width=0.16\textwidth]{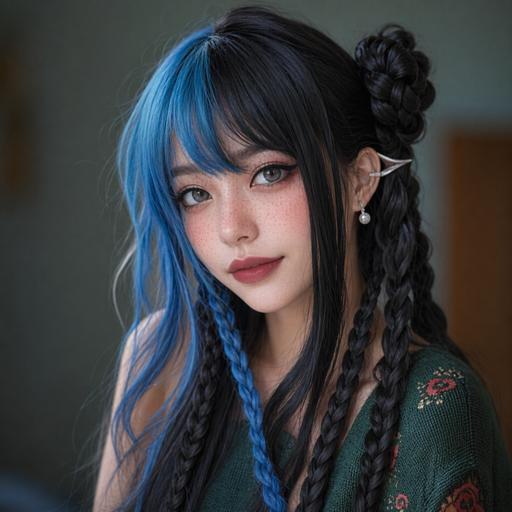} &
        \includegraphics[width=0.16\textwidth]{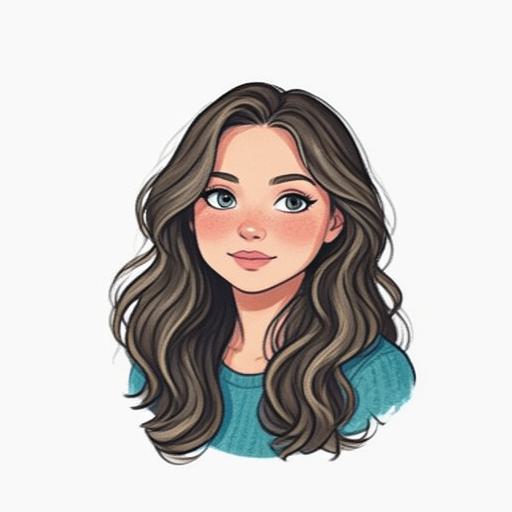} \\

        \includegraphics[width=0.16\textwidth]{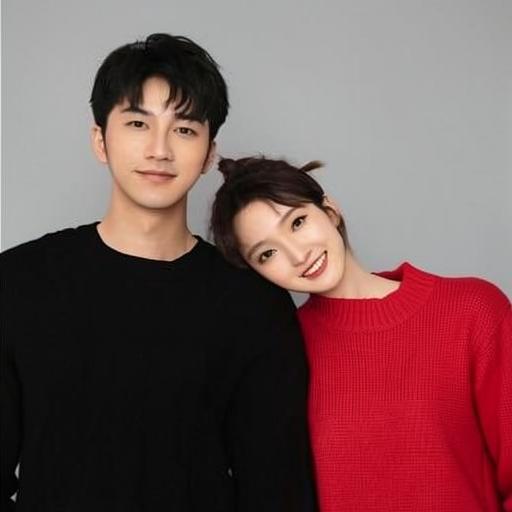} &
        \includegraphics[width=0.16\textwidth]{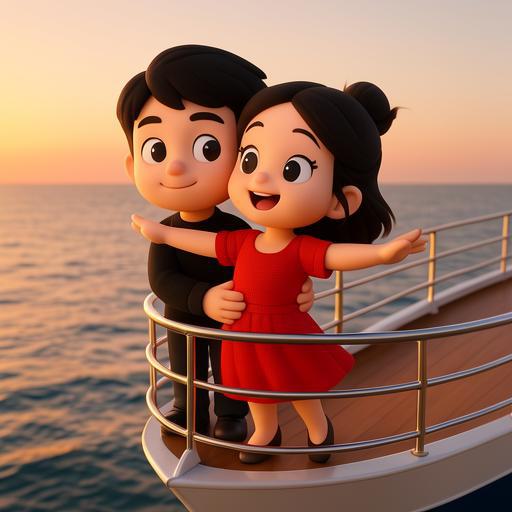} &
        \includegraphics[width=0.16\textwidth]{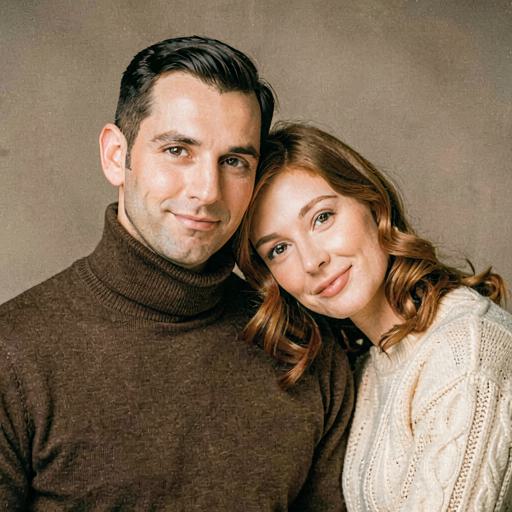} &
        \includegraphics[width=0.16\textwidth]{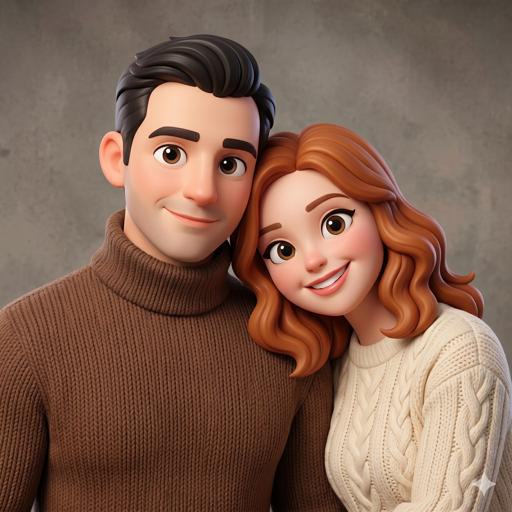} &
        \includegraphics[width=0.16\textwidth]{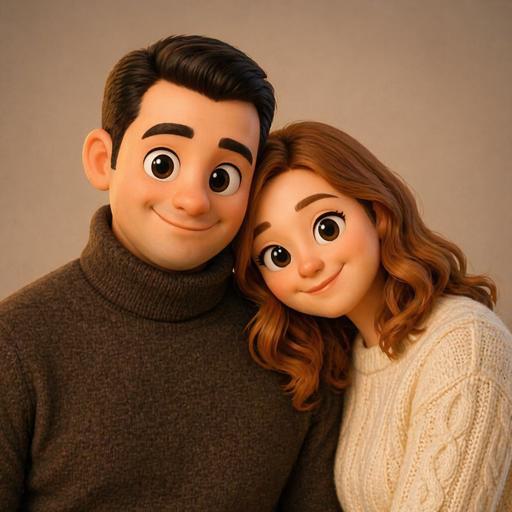} &
        \includegraphics[width=0.16\textwidth]{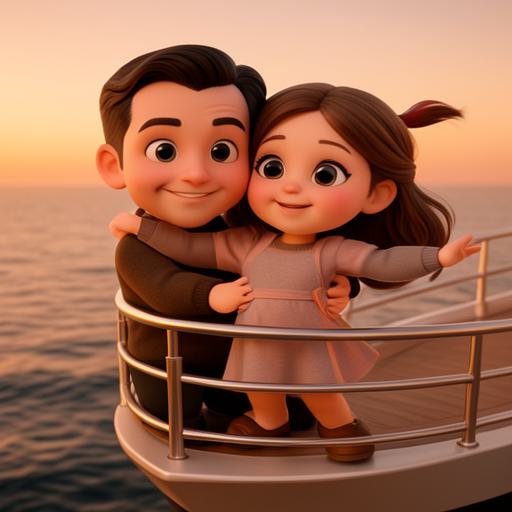} \\

        \includegraphics[width=0.16\textwidth]{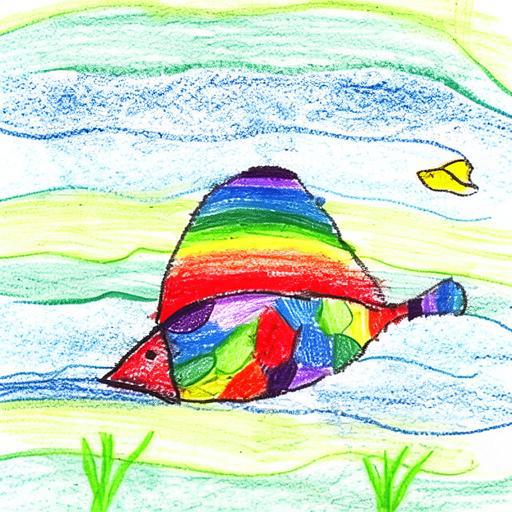} &
        \includegraphics[width=0.16\textwidth]{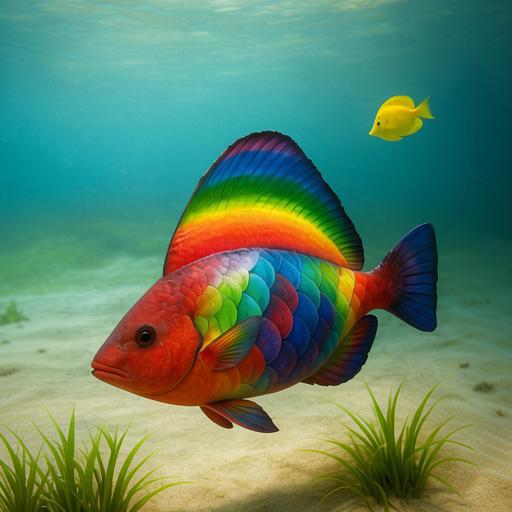} &
        \includegraphics[width=0.16\textwidth]{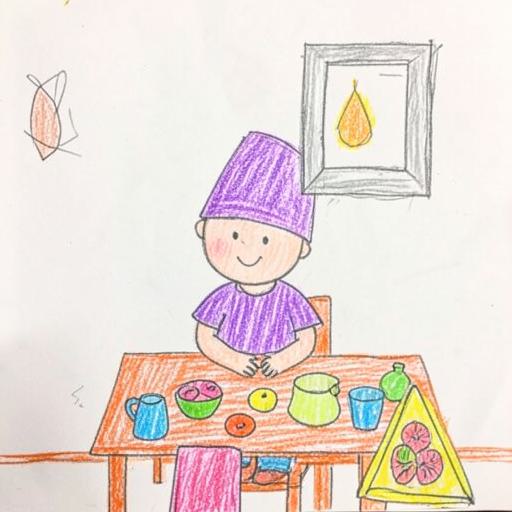} &
        \includegraphics[width=0.16\textwidth]{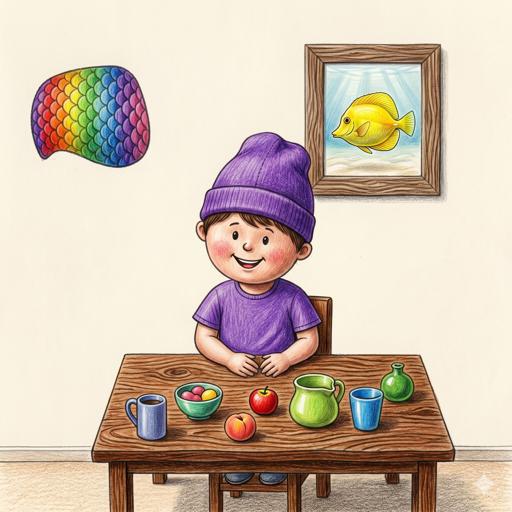} &
        \includegraphics[width=0.16\textwidth]{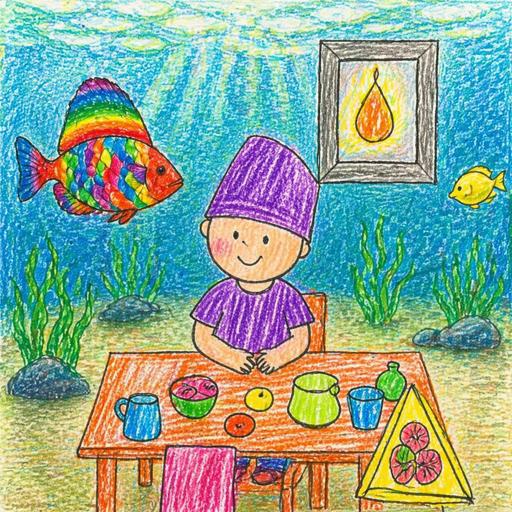} &
        \includegraphics[width=0.16\textwidth]{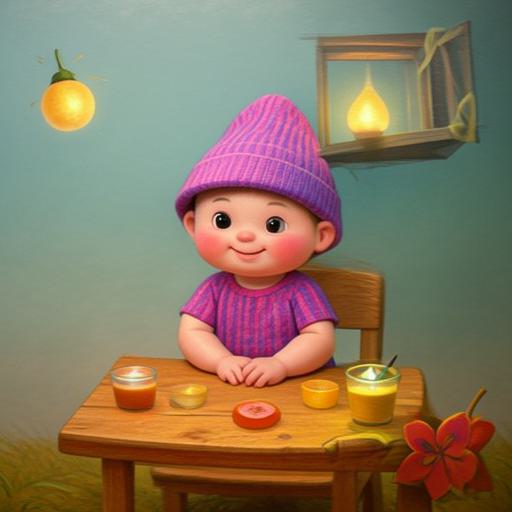} \\

        \includegraphics[width=0.16\textwidth]{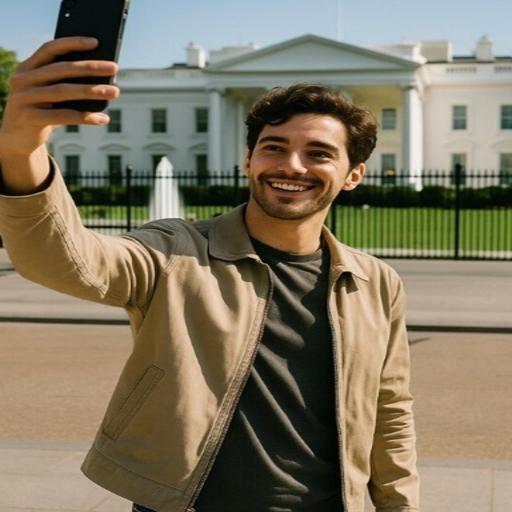} &
        \includegraphics[width=0.16\textwidth]{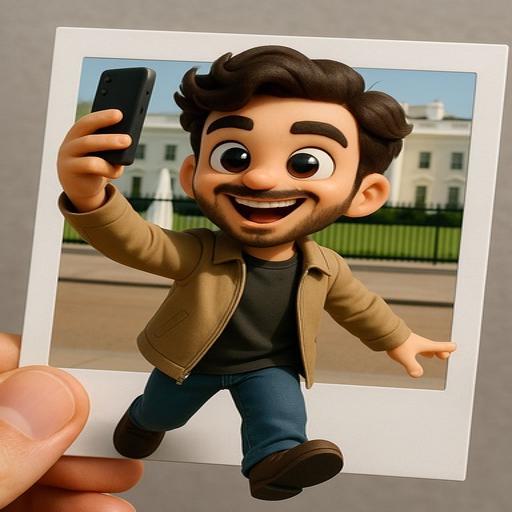} &
        \includegraphics[width=0.16\textwidth]{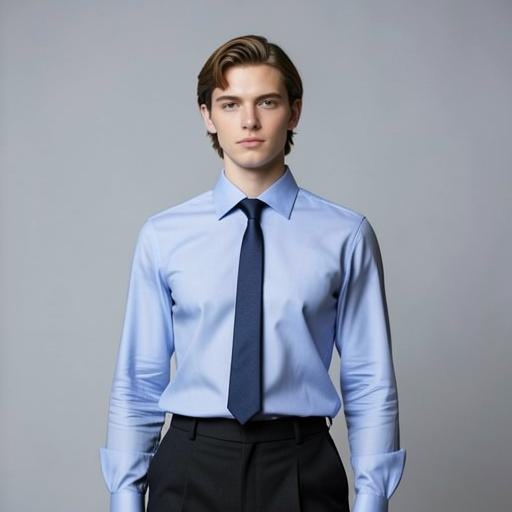} &
        \includegraphics[width=0.16\textwidth]{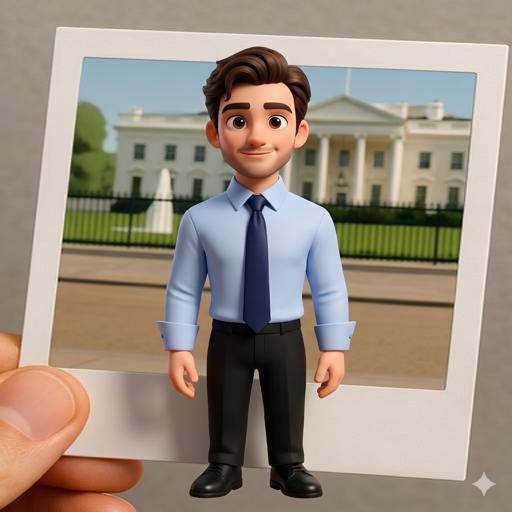} &
        \includegraphics[width=0.16\textwidth]{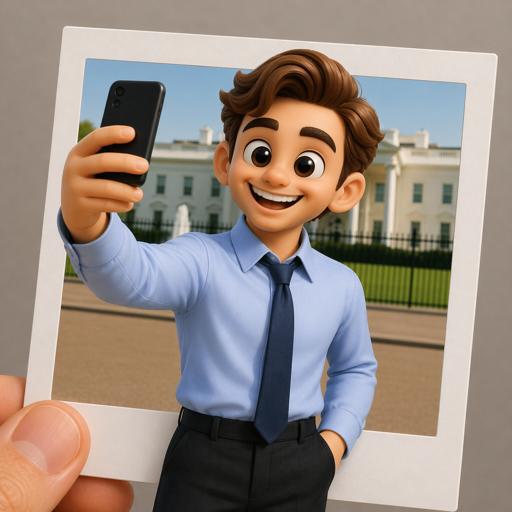} &
        \includegraphics[width=0.16\textwidth]{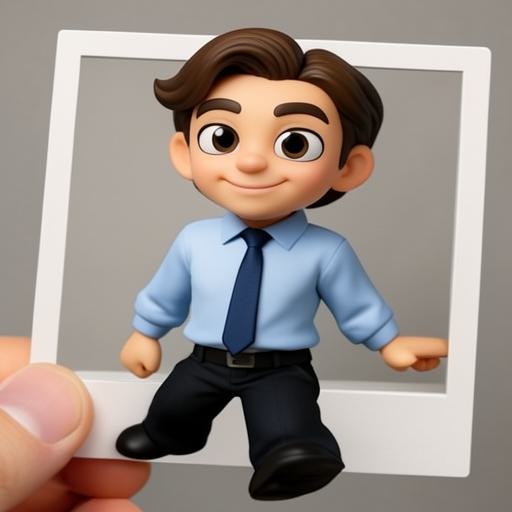} \\

        \includegraphics[width=0.16\textwidth]{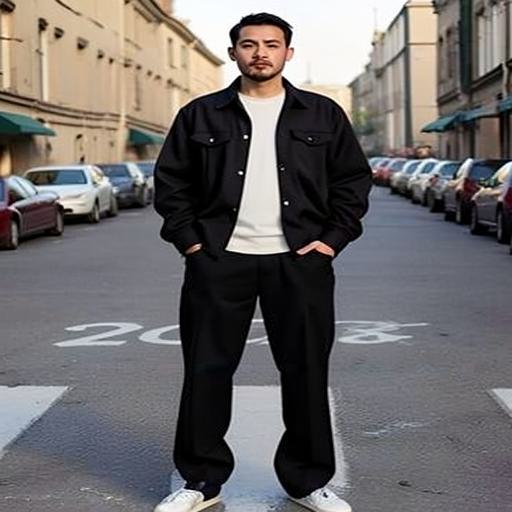} &
        \includegraphics[width=0.16\textwidth]{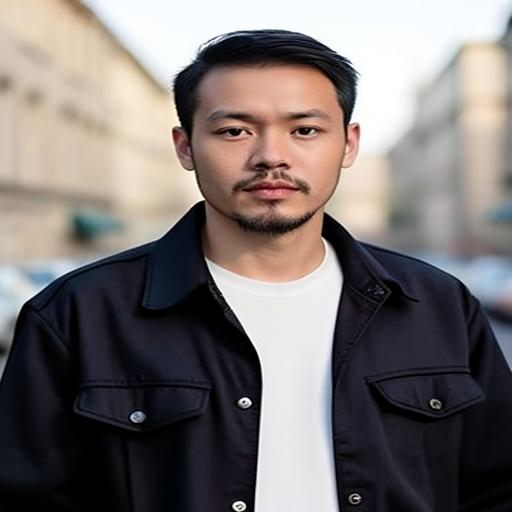} &
        \includegraphics[width=0.16\textwidth]{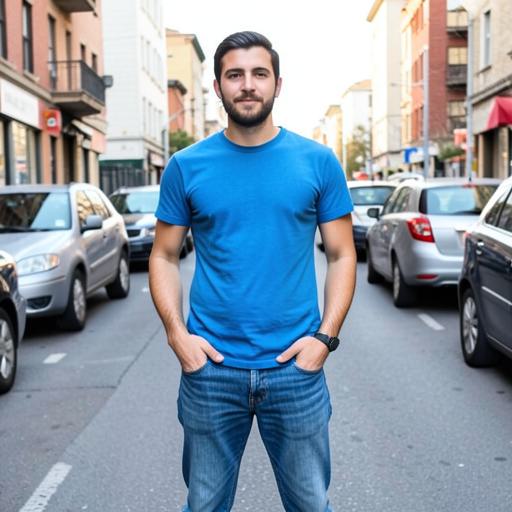} &
        \includegraphics[width=0.16\textwidth]{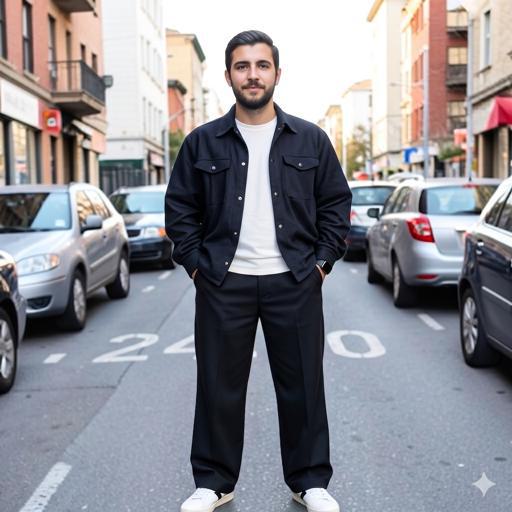} &
        \includegraphics[width=0.16\textwidth]{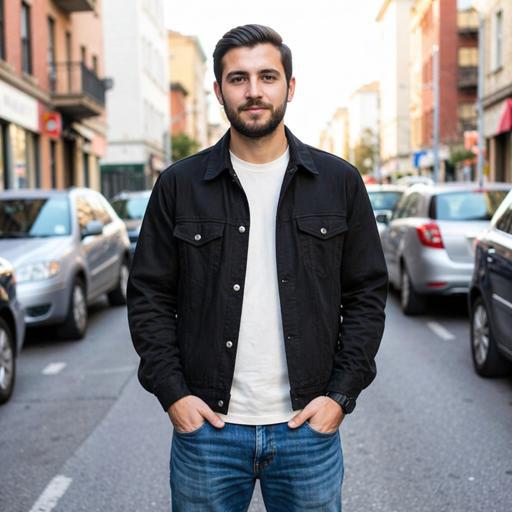} &
        \includegraphics[width=0.16\textwidth]{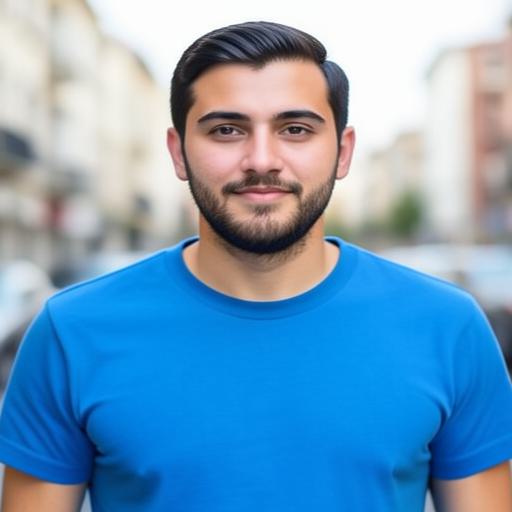} \\

        \includegraphics[width=0.16\textwidth]{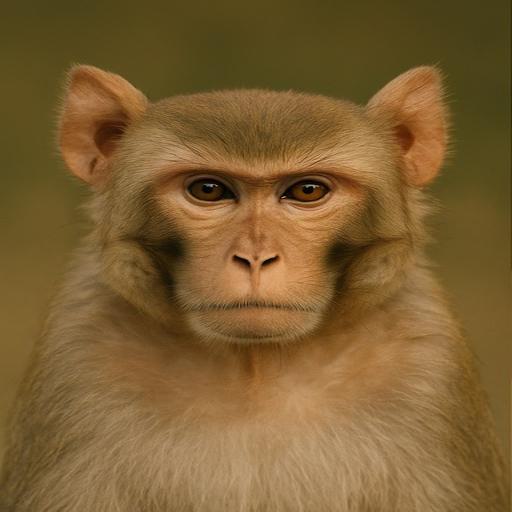} &
        \includegraphics[width=0.16\textwidth]{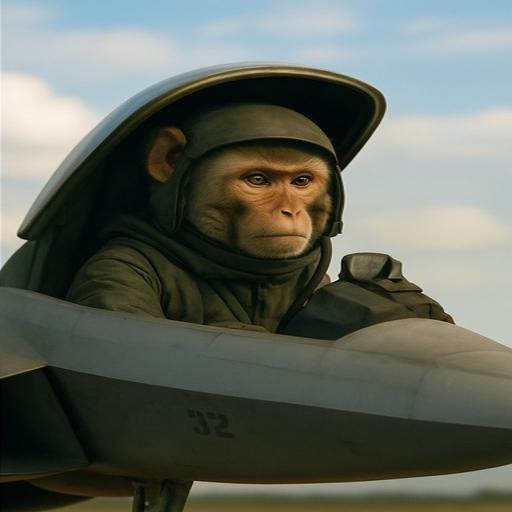} &
        \includegraphics[width=0.16\textwidth]{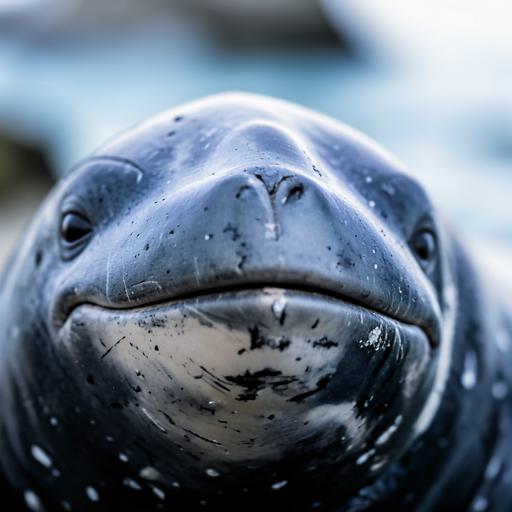} &
        \includegraphics[width=0.16\textwidth]{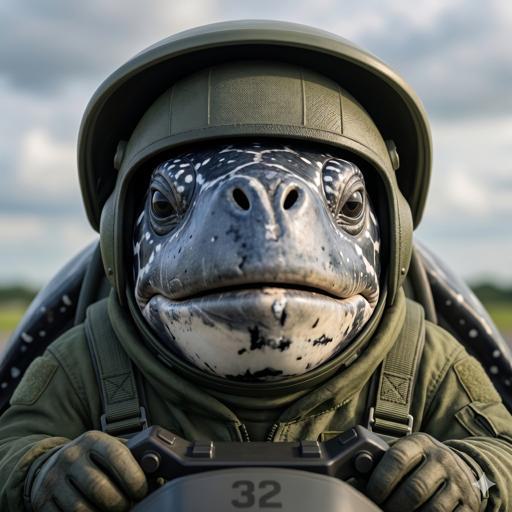} &
        \includegraphics[width=0.16\textwidth]{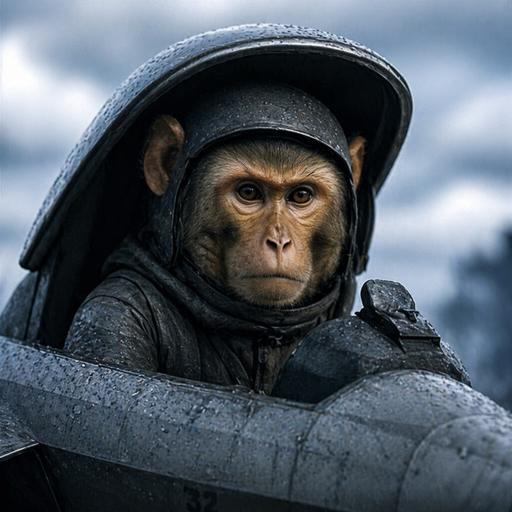} &
        \includegraphics[width=0.16\textwidth]{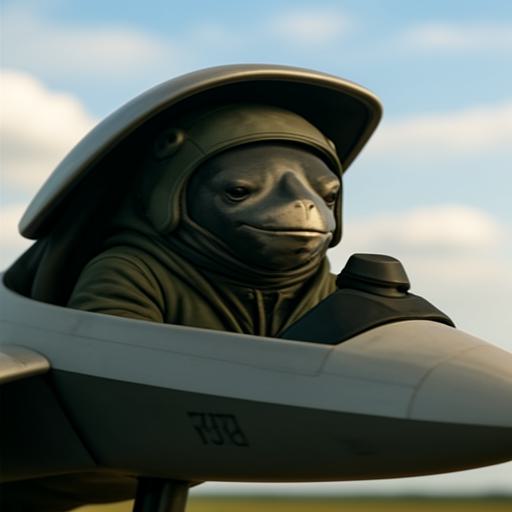} \\

       \includegraphics[width=0.16\textwidth]{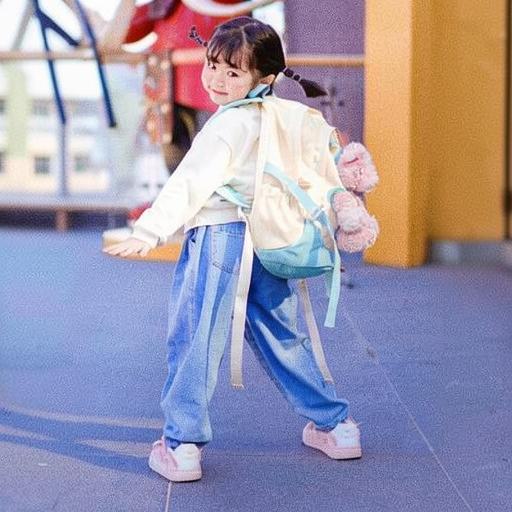} &
        \includegraphics[width=0.16\textwidth]{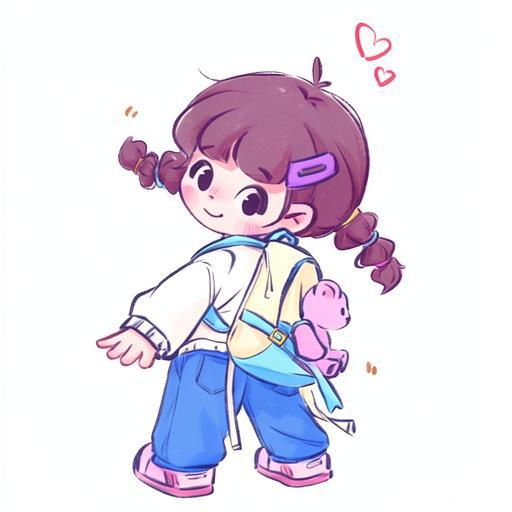} &
        \includegraphics[width=0.16\textwidth]{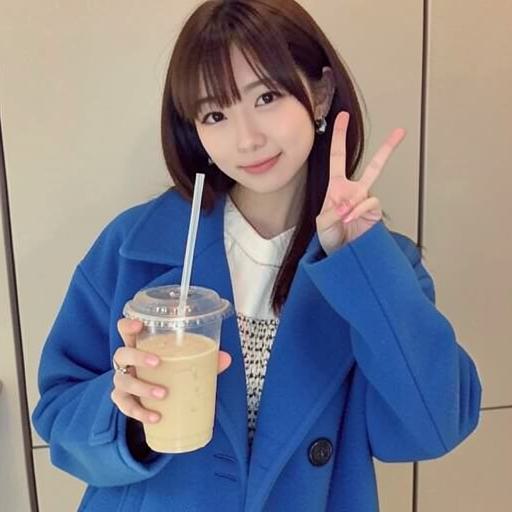} &
        \includegraphics[width=0.16\textwidth]{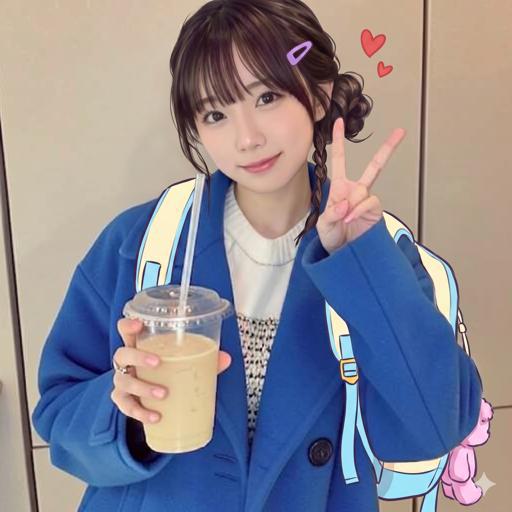} &
        \includegraphics[width=0.16\textwidth]{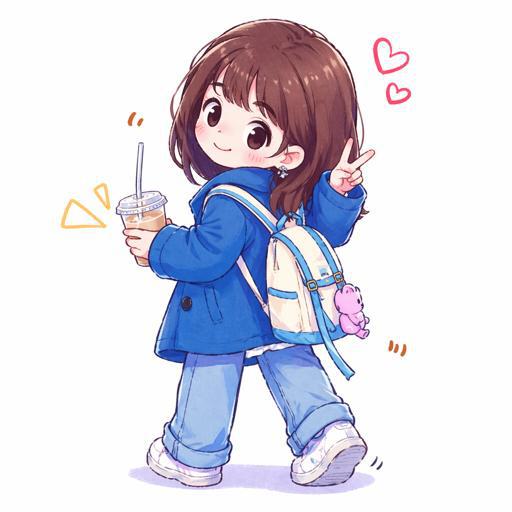} &
        \includegraphics[width=0.16\textwidth]{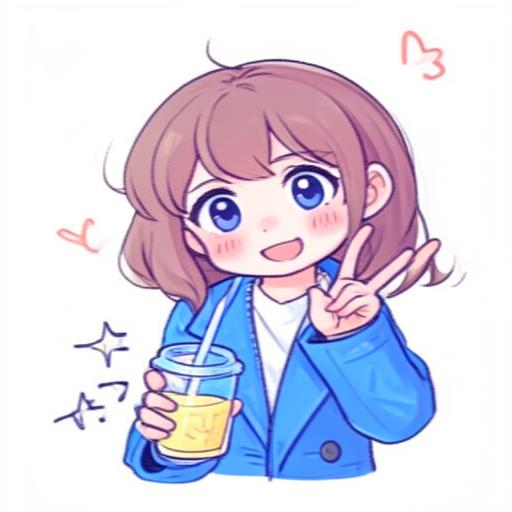} \\

        \includegraphics[width=0.16\textwidth]{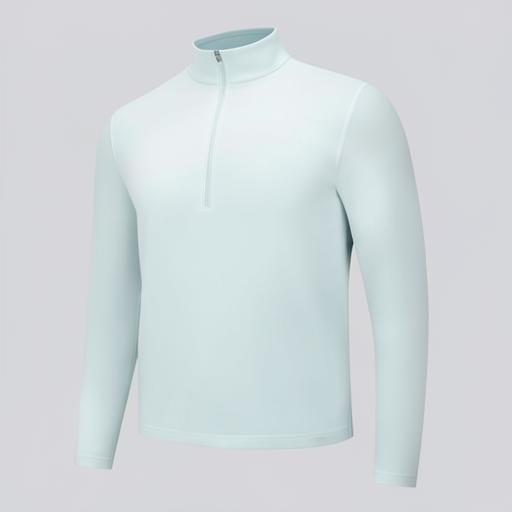} &
        \includegraphics[width=0.16\textwidth]{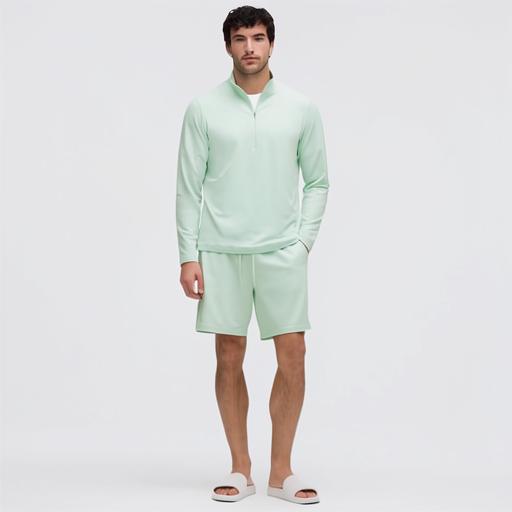} &
        \includegraphics[width=0.16\textwidth]{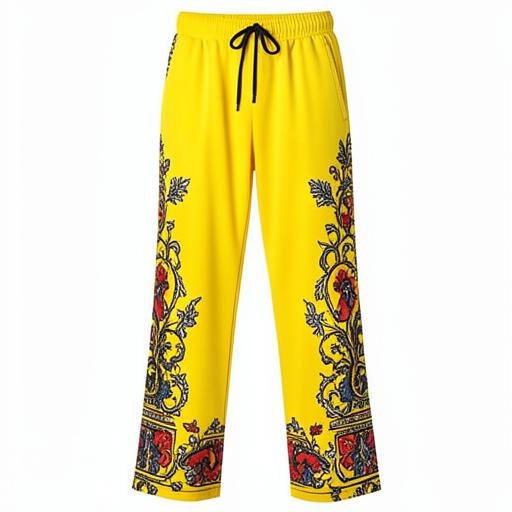} &
        \includegraphics[width=0.16\textwidth]{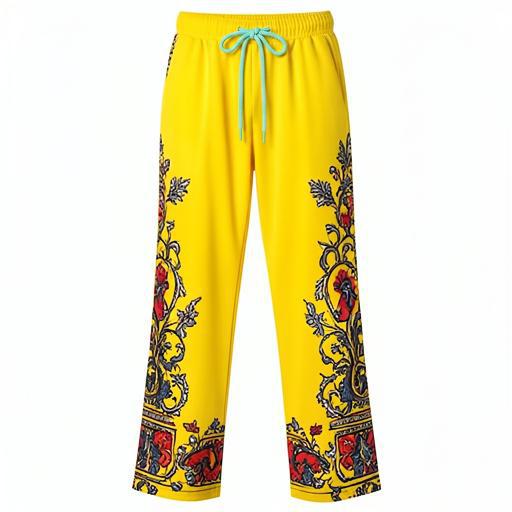} &
        \includegraphics[width=0.16\textwidth]{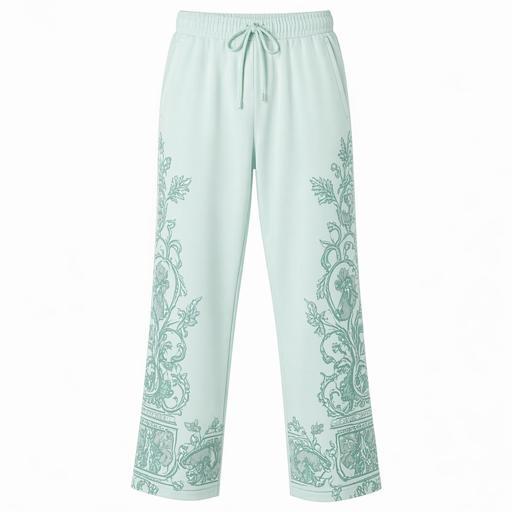} &
        \includegraphics[width=0.16\textwidth]{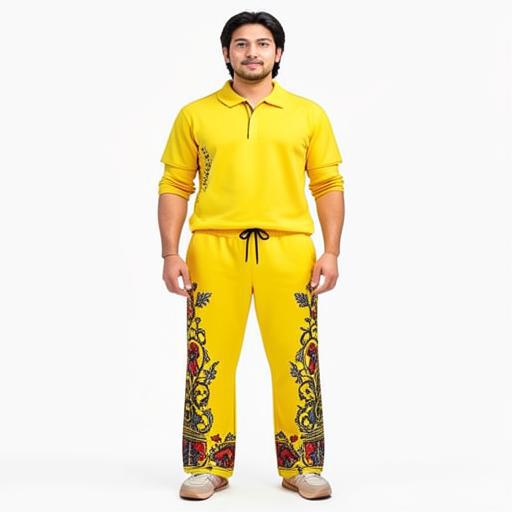} \\

    \end{tabular}
    }
    \caption{\textbf{Qualitative comparison with Nano Banana 2~\cite{banana} and GPT-Image-2~\cite{gpt_image}.} Both models frequently fail to capture the intended transformation (rows 1–3) and tend to leak appearance cues from the exemplar images into the output (rows 4–6).}
    \label{fig:qualitative_gpt_nano}
\end{figure*}
\begin{figure*}[!t]
    \centering
    \renewcommand{\arraystretch}{0.3}
    \setlength{\tabcolsep}{1pt}

    {\footnotesize
    \begin{tabular}{c c c @{\hspace{0.07cm}} | @{\hspace{0.07cm}} c c}

        \multicolumn{1}{c}{\normalsize Source ($a$)} &
        \multicolumn{1}{c}{\normalsize Target ($a'$)} &
        \multicolumn{1}{c}{\normalsize Query ($b$)} &
        \multicolumn{1}{c}{\normalsize PairEdit} &
        \multicolumn{1}{c}{\normalsize Ours} \\

        \includegraphics[width=0.16\textwidth]{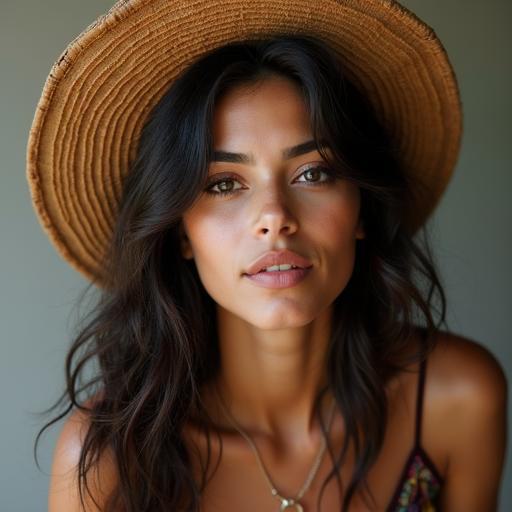} &
        \includegraphics[width=0.16\textwidth]{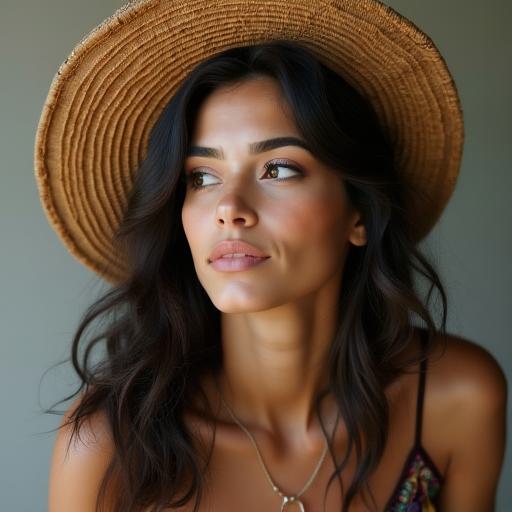} &
        \includegraphics[width=0.16\textwidth]{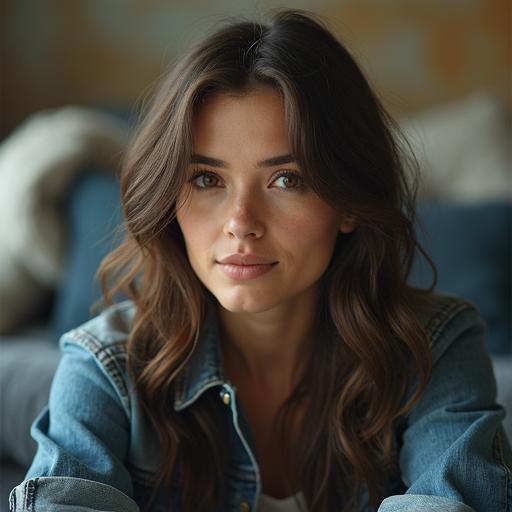} &
        \includegraphics[width=0.16\textwidth]{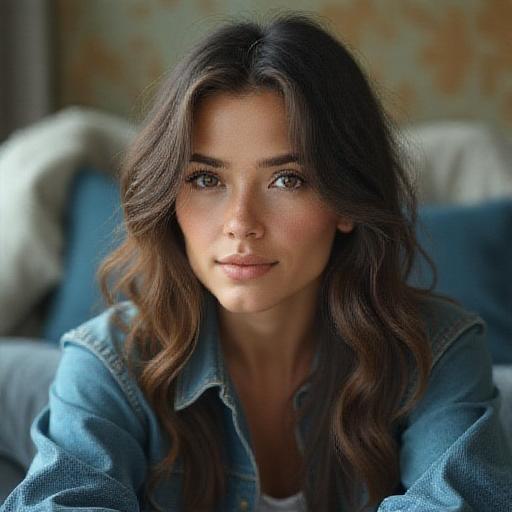} &
        \includegraphics[width=0.16\textwidth]{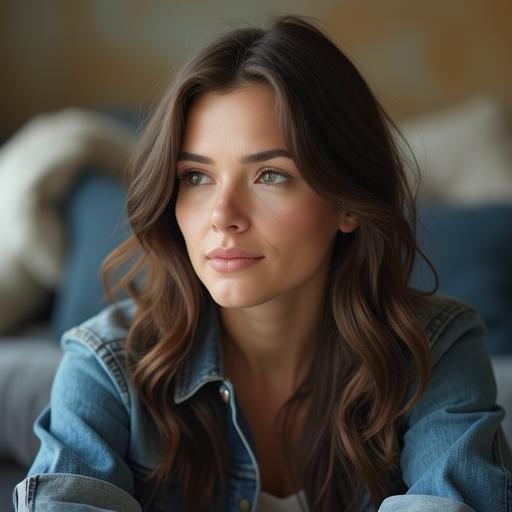} \\
        
        \includegraphics[width=0.16\textwidth]{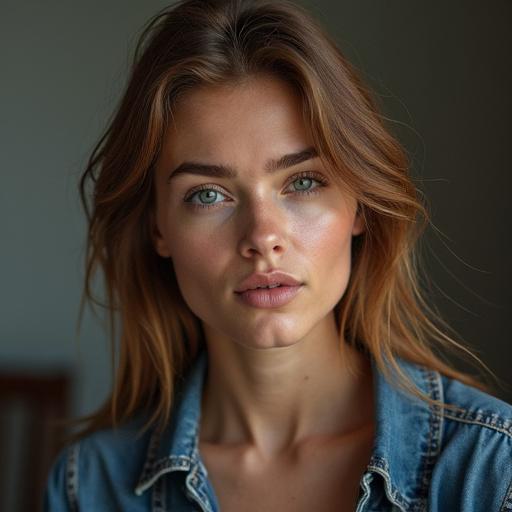} &
        \includegraphics[width=0.16\textwidth]{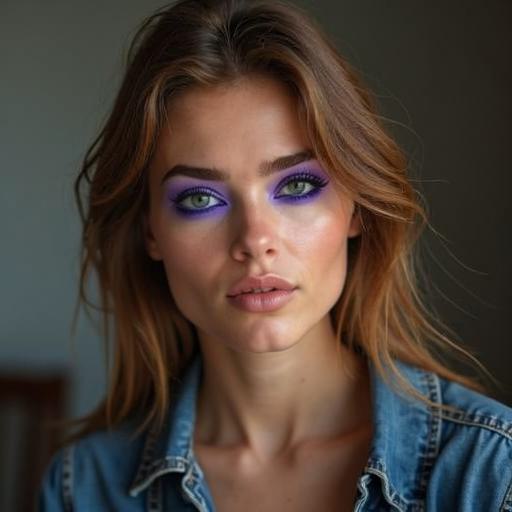} &
        \includegraphics[width=0.16\textwidth]{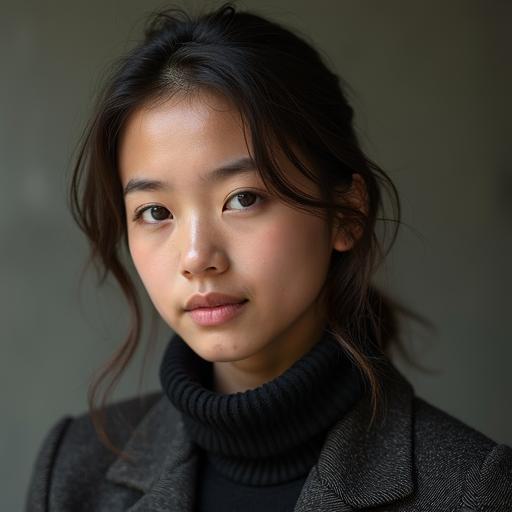} &
        \includegraphics[width=0.16\textwidth]{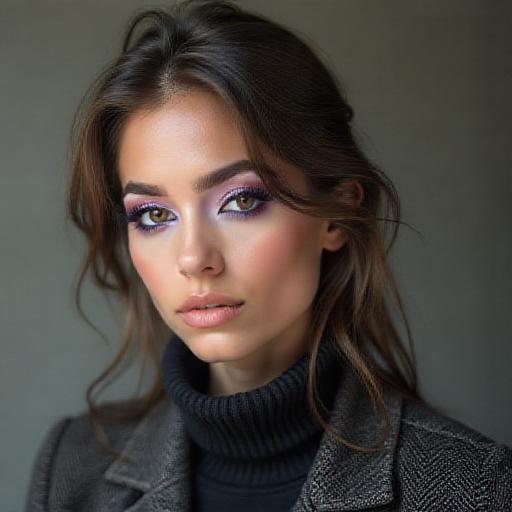} &
        \includegraphics[width=0.16\textwidth]{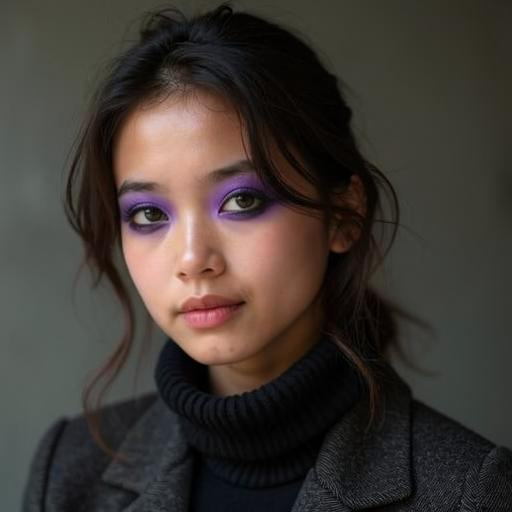} \\

        \includegraphics[width=0.16\textwidth]{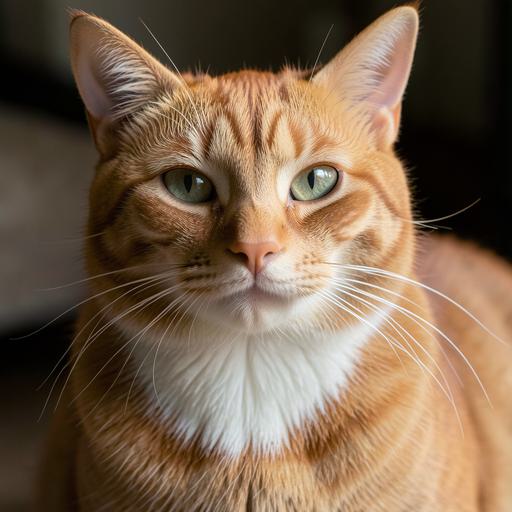} &
        \includegraphics[width=0.16\textwidth]{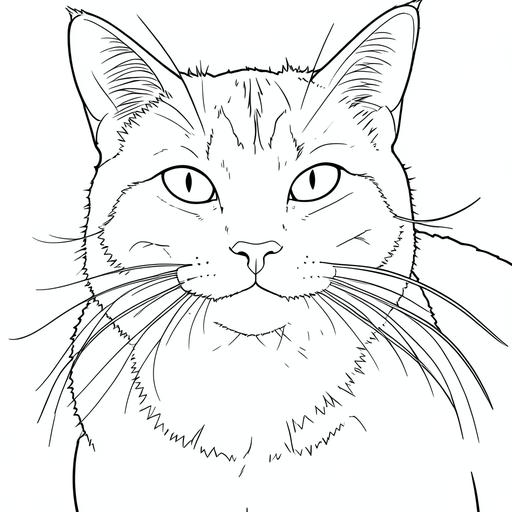} &
        \includegraphics[width=0.16\textwidth]{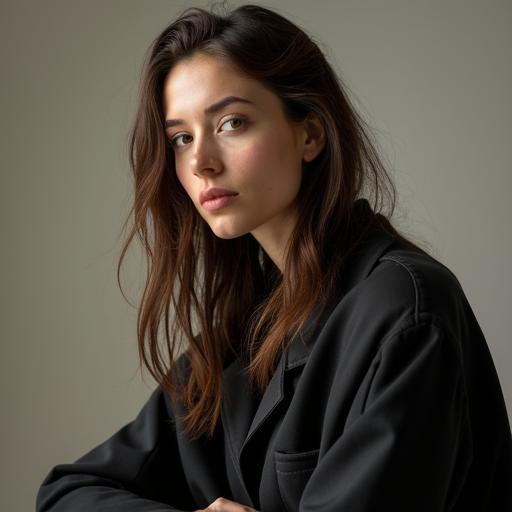} &
        \includegraphics[width=0.16\textwidth]{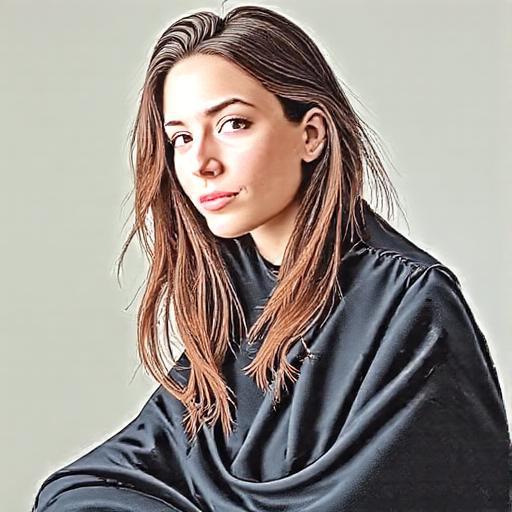} &
        \includegraphics[width=0.16\textwidth]{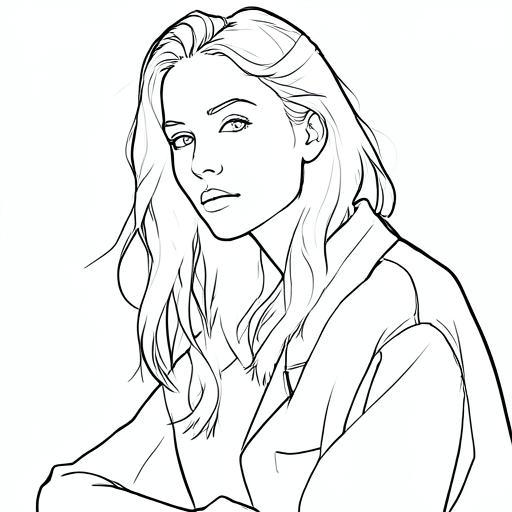} \\

        \includegraphics[width=0.16\textwidth]{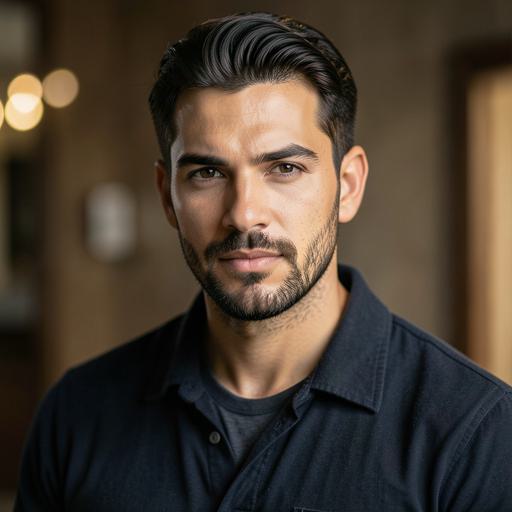} &
        \includegraphics[width=0.16\textwidth]{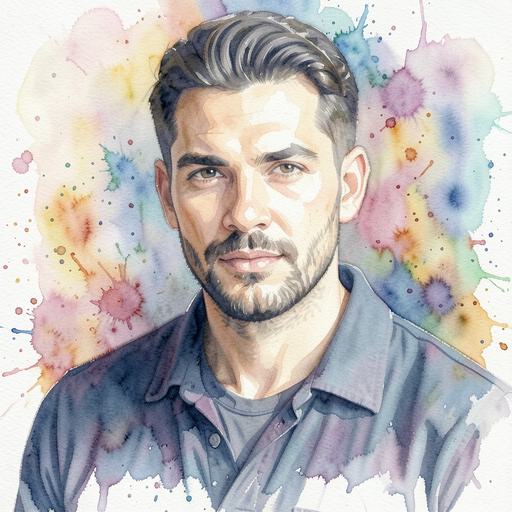} &
        \includegraphics[width=0.16\textwidth]{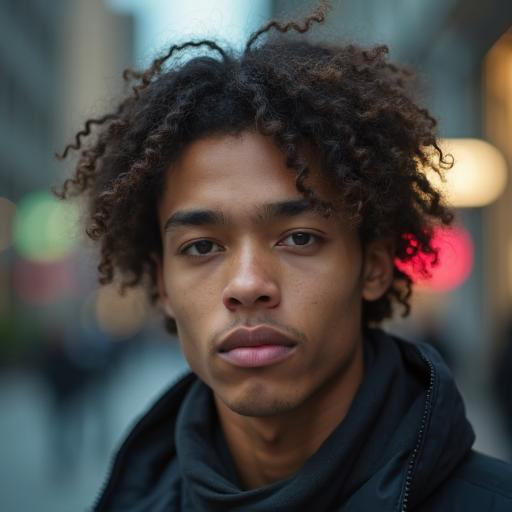} &
        \includegraphics[width=0.16\textwidth]{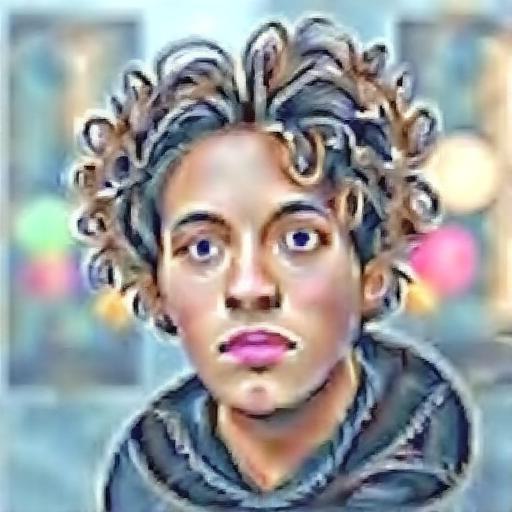} &
        \includegraphics[width=0.16\textwidth]{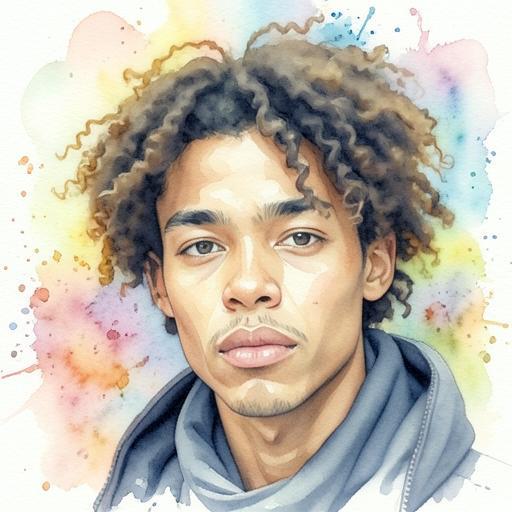} \\

        \includegraphics[width=0.16\textwidth]{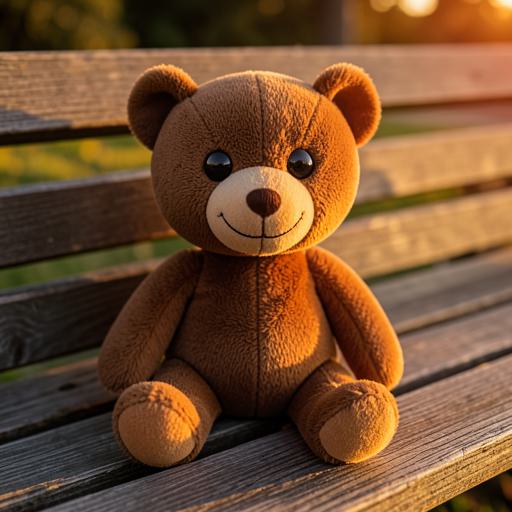} &
        \includegraphics[width=0.16\textwidth]{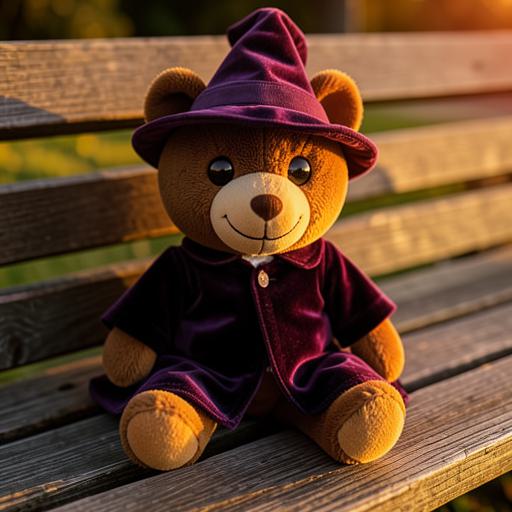} &
        \includegraphics[width=0.16\textwidth]{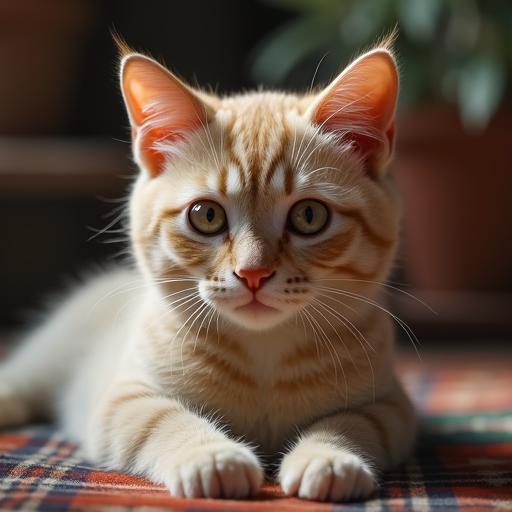} &
        \includegraphics[width=0.16\textwidth]{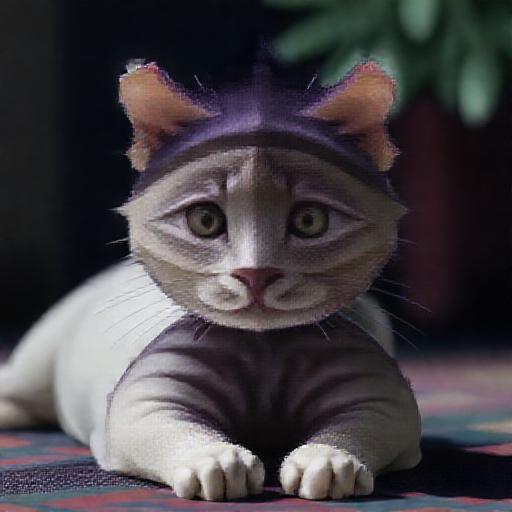} &
        \includegraphics[width=0.16\textwidth]{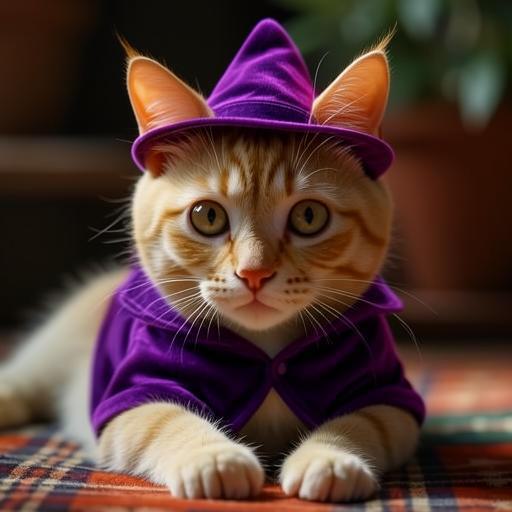} \\

    \end{tabular}
    }
    \caption{\textbf{Qualitative comparison with PairEdit~\cite{lu2025pairedit}.} PairEdit struggles to capture complex edits from the exemplar pair.}
    \label{fig:qualitative_pairedit}
\end{figure*}
\begin{figure*}[t]
    \centering
    \renewcommand{\arraystretch}{0.3}
    \setlength{\tabcolsep}{0.6pt}

    {\footnotesize
    \begin{tabular}{c c c c}

        \multicolumn{1}{c}{\normalsize Source ($a$)} &
        \multicolumn{1}{c}{\normalsize Target ($a'$)} &
        \multicolumn{1}{c}{\normalsize Query ($b$)} &
        \multicolumn{1}{c}{\normalsize Result} \\

        \includegraphics[width=0.16\textwidth]{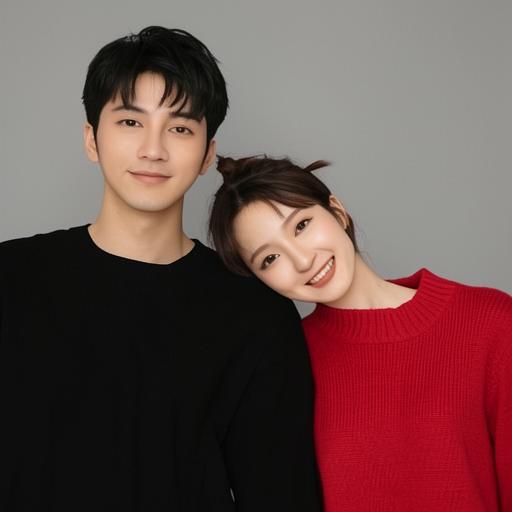} &
        \includegraphics[width=0.16\textwidth]{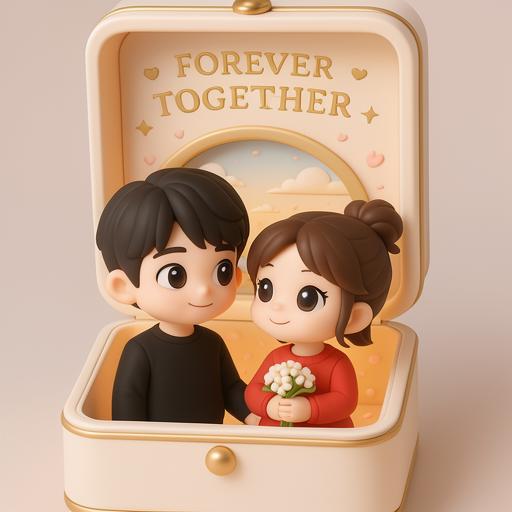} &
        \includegraphics[width=0.16\textwidth]{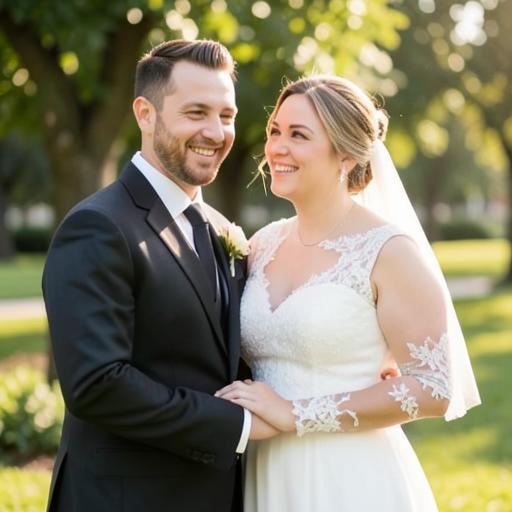} &
        \includegraphics[width=0.16\textwidth]{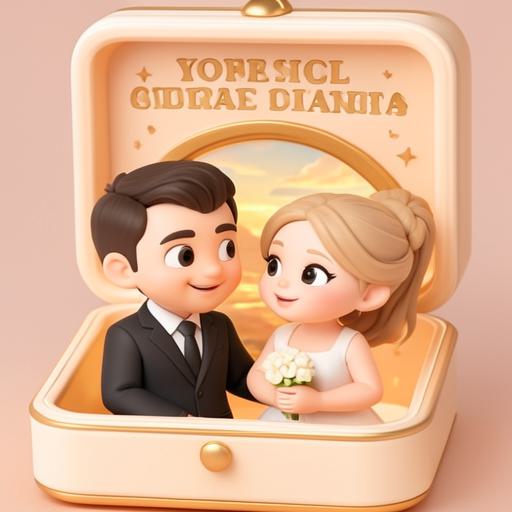} \\

        \includegraphics[width=0.16\textwidth]{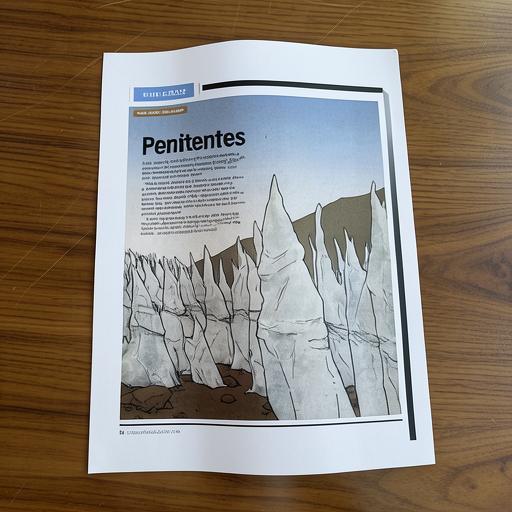} &
        \includegraphics[width=0.16\textwidth]{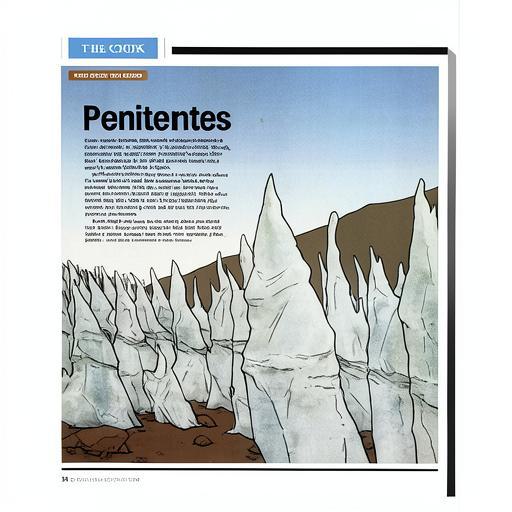} &
        \includegraphics[width=0.16\textwidth]{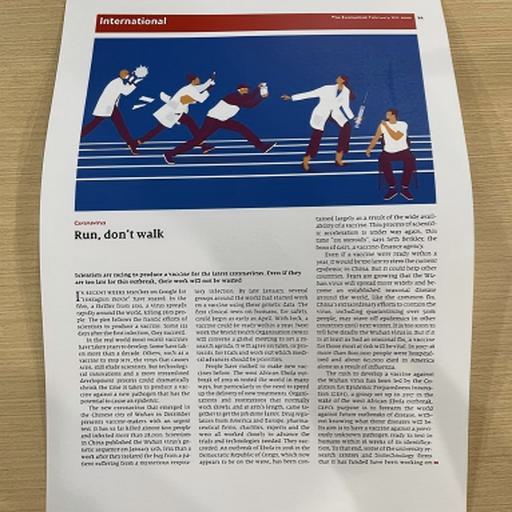} &
        \includegraphics[width=0.16\textwidth]{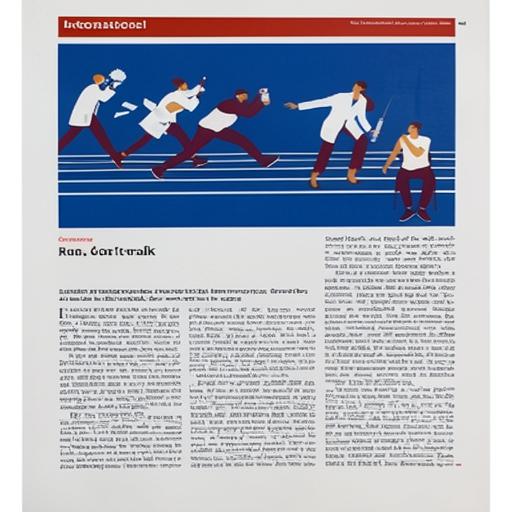} \\

    \end{tabular}
    }
    \caption{\textbf{Failure cases of Delta-Adapter.} Our method struggles with editing tasks that require precise text rendering. When the exemplar pair contains textual content, the model produces characters that are inconsistent with those in the exemplar.}
    \label{fig:failure_cases}
\end{figure*}

\begin{figure}[t]
\centering
\begin{tcolorbox}[
    colback=gray!10,
    colframe=black!75,
    coltitle=white,
    colbacktitle=black!85,
    title=\textbf{Text Prompts},
    fonttitle=\bfseries,
    arc=2mm,
    boxrule=0.4pt,
    left=3mm, right=3mm, top=2mm, bottom=2mm,
    width=0.95\linewidth,
    fontupper=\small,
]
You are a strict visual evaluation assistant for reference-pair-guided image editing.

\medskip
You will receive four images:
\begin{itemize}[leftmargin=*,nosep]
    \item A: reference source image
    \item B: reference target image
    \item C: source image to be edited
    \item D: candidate edited image
\end{itemize}

\medskip
\textbf{Your job:}
\begin{enumerate}[leftmargin=*,nosep]
    \item Infer the intended edit from the actual visual transformation from $A \rightarrow B$.
    \item Judge whether the same edit is correctly transferred from $C \rightarrow D$.
    \item Score $D$ on two dimensions:
\end{enumerate}

\medskip
\textbf{GPT-A (Editing Accuracy):} Whether the candidate correctly applies the intended edit or transformation shown by $A \rightarrow B$.

\medskip
\textbf{GPT-C (Content Consistency):} Whether the candidate preserves the source image $C$ everywhere outside the intended edit. Focus on non-target content preservation: identity, structure, geometry, layout, background, pose/camera, object count, and spatial relations should remain unchanged unless a change is clearly required by the intended edit shown in $A \rightarrow B$.

\medskip
\textbf{Core principle:} The final judgment must rely only on the visual evidence in $A$, $B$, $C$, and $D$. Do not assume an edit category from filenames, directory names, or metadata.

\medskip
\textbf{General rules:} Be strict. Do not reward unrelated changes. Score and explanation must be fully consistent. Judge transfer quality by comparing $C$ with $D$ under the transformation inferred from $A \rightarrow B$. Penalize GPT-C only for unwanted changes outside the target edit region.

\medskip
\textbf{Output format.} Return ONLY valid JSON with exactly this schema:
\begin{verbatim}
{
  "inferred_edit": "<one short sentence>",
  "D": {
    "scores": {
      "gpt_a": <integer 1-5>,
      "gpt_c": <integer 1-5>
    }
  }
}
\end{verbatim}
\end{tcolorbox}
\caption{System prompt used for GPT-based automated evaluation of editing accuracy (GPT-A) and content consistency (GPT-C).}
\label{fig:gpt_prompt}
\end{figure}
\begin{figure*}[t]
    \centering
    \renewcommand{\arraystretch}{0.3}
    \setlength{\tabcolsep}{0.6pt}

    {\footnotesize
    \begin{tabular}{c c c c c c c }

        \multicolumn{1}{c}{\normalsize Source ($a$)} &
        \multicolumn{1}{c}{\normalsize Target ($a'$)} &
        \multicolumn{1}{c}{\normalsize Query ($b$)} &
        \multicolumn{1}{c}{\normalsize Step 1} &
        \multicolumn{1}{c}{\normalsize Step 2} &
        \multicolumn{1}{c}{\normalsize Step 3} &
        \multicolumn{1}{c}{\normalsize Step 4} \\

        \includegraphics[width=0.14\textwidth]{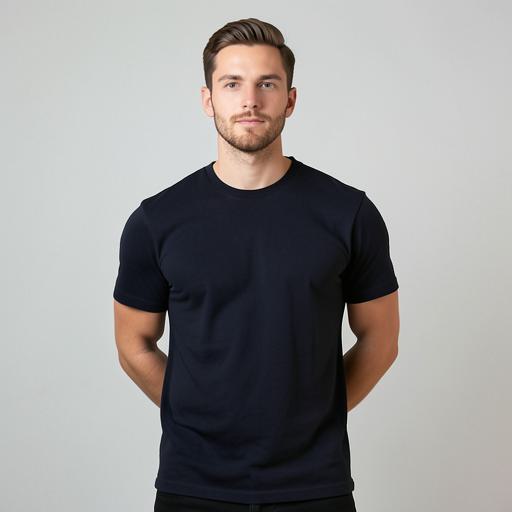} &
        \includegraphics[width=0.14\textwidth]{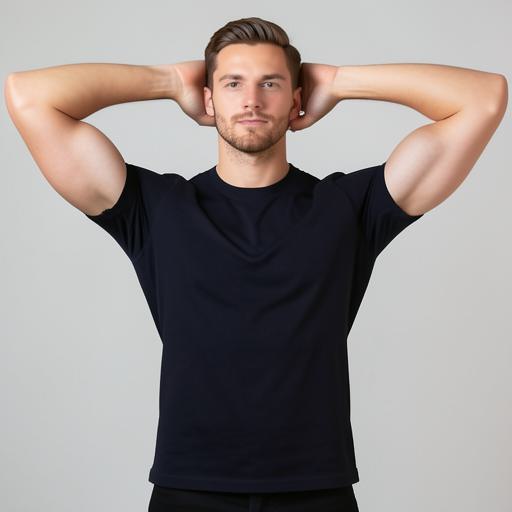} &
        \includegraphics[width=0.14\textwidth]{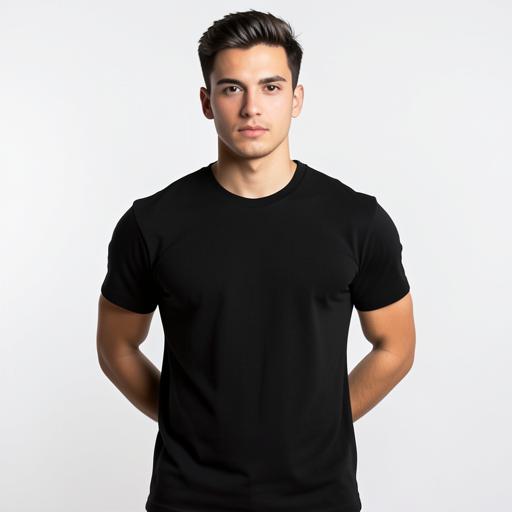} &
        \includegraphics[width=0.14\textwidth]{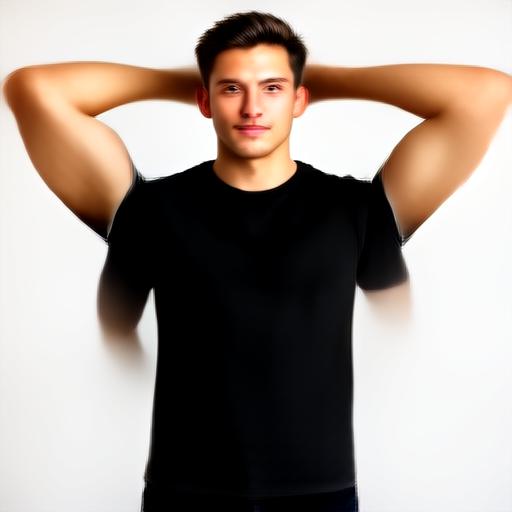} &
        \includegraphics[width=0.14\textwidth]{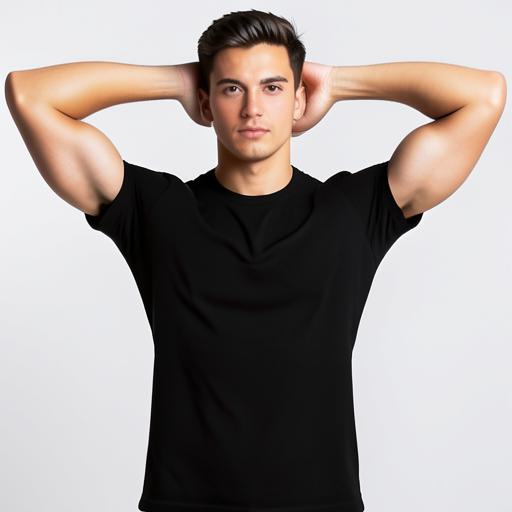} &
        \includegraphics[width=0.14\textwidth]{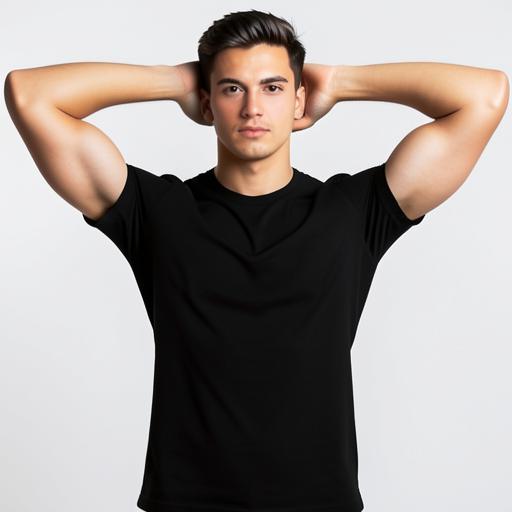} &
        \includegraphics[width=0.14\textwidth]{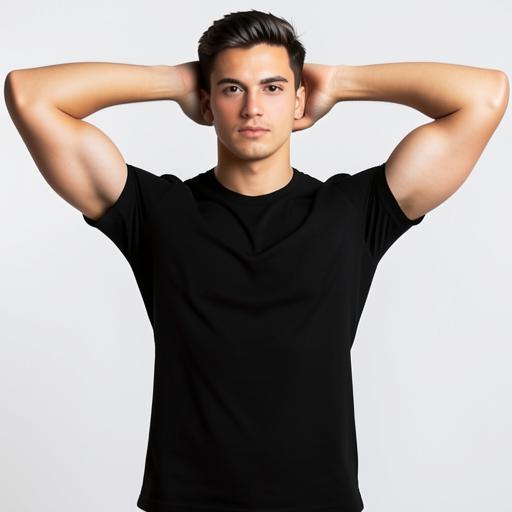} \\

        \includegraphics[width=0.14\textwidth]{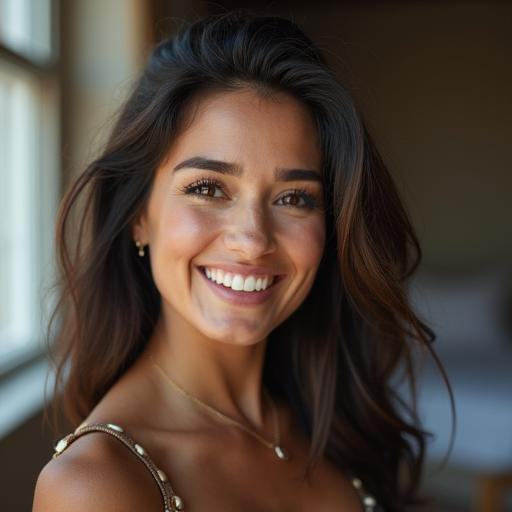} &
        \includegraphics[width=0.14\textwidth]{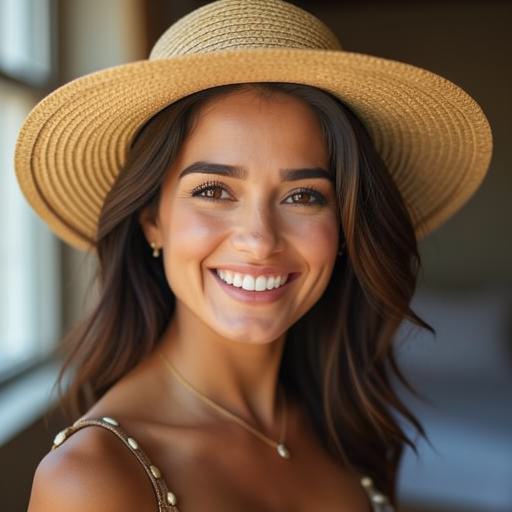} &
        \includegraphics[width=0.14\textwidth]{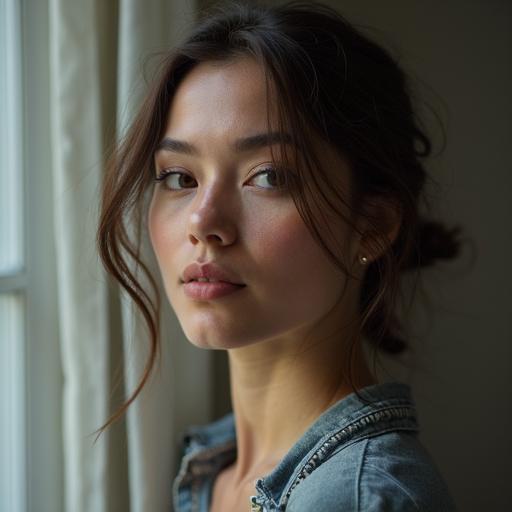} &
        \includegraphics[width=0.14\textwidth]{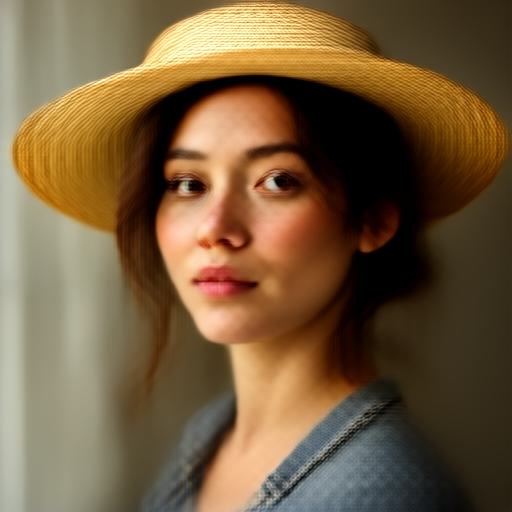} &
        \includegraphics[width=0.14\textwidth]{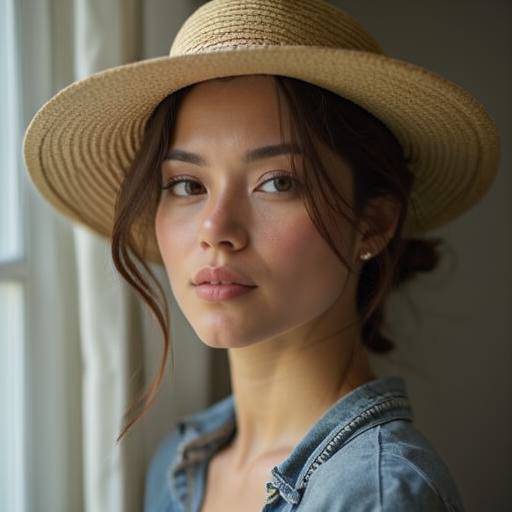} &
        \includegraphics[width=0.14\textwidth]{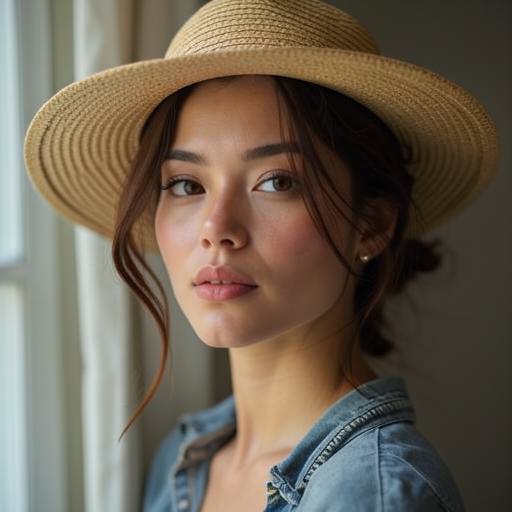} &
        \includegraphics[width=0.14\textwidth]{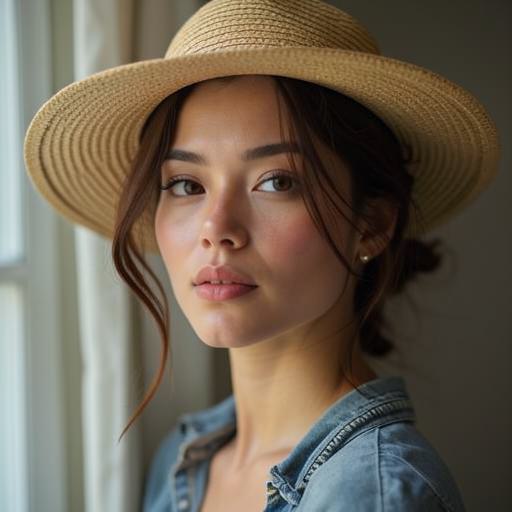} \\

    \end{tabular}
    }
    \caption{\textbf{Visualization of recovered clean latents during four-step denoising.} Given a single exemplar pair $(a, a')$ and a query image $b$, we visualize the decoded clean latent estimate $\hat{z}_0$ predicted at each denoising step of our four-step backbone. Even at the early denoising stages, the recovered $\hat{z}_0$ already forms coherent image structures and reflects the intended edit semantics, providing sufficiently reliable visual features for our dense semantic supervision.}

\label{fig:klein_steps}
\end{figure*}

\end{document}